\documentclass[11pt]{article}
\usepackage[utf8]{inputenc}
\usepackage[margin=1in]{geometry}
\usepackage{graphicx} 
\usepackage{enumitem}
\usepackage{comment}
\usepackage{amsmath} 
\usepackage{amsfonts}
\usepackage{amsthm, amssymb}
\usepackage{mathrsfs}
\usepackage{thm-restate}
\usepackage{booktabs}
\usepackage[title]{appendix}
\usepackage[]{hyperref}
\hypersetup{
    colorlinks = true,
    linkcolor = blue,
    citecolor = blue
    }
    
\usepackage[
        noabbrev,
        capitalise,
        nameinlink,
    ]
    {cleveref}
\usepackage[dvipsnames]{xcolor}
\usepackage{bbm}
\usepackage{dirtytalk}
\usepackage{subcaption}
\usepackage{float}
\usepackage{tikz}
\usepackage{tikz-cd}
\usepackage{algorithm}
\usepackage[noend]{algpseudocode}
\usepackage{makecell}
\usetikzlibrary{arrows.meta,decorations.markings, calc, positioning, matrix}
\usepackage{array} 
\usepackage{booktabs} 

\newcolumntype{C}[1]{>{\centering\arraybackslash}p{#1}}

\usepackage{pgfplots}
\pgfplotsset{compat=1.18}
\declaretheorem[name=Theorem, numberwithin=section]{thm} 

\declaretheorem[name=Lemma, numberwithin=section]{lemma}

\declaretheorem[name=Assumption, sibling=lemma]{asmpt} 
\declaretheorem[name=Proposition, sibling=lemma]{prop} 
\declaretheorem[name=Corollary, sibling=lemma]{corollary} 

\declaretheorem[name=Remark, sibling=lemma, style=definition]{rmk} 
\declaretheorem[name=Definition, sibling=lemma]{defn} 
\declaretheorem[name=Example, sibling=lemma]{example} 

\newcommand{\man}{\mathcal{M}}

\definecolor{cyan}{cmyk}{1, 0.4, 0, 0}

\newcommand{\new}[1]{{\color{black} #1}}

\newcommand{\PT}{\operatorname{PT}}

\newcommand{\OT}{\operatorname{OT}}

\newcommand{\KL}{\operatorname{KL}}

\usepackage[round]{natbib} 
\title{Second-order Sinkhorn Geometry and $\sqrt{n}$-Inference}

\author{
\begin{tabular}{ccc} 
\multicolumn{1}{c}{\makecell[c]{Tristan Luca Saidi\textsuperscript{\(\dagger\)}\\ \href{mailto:tsaidi@andrew.cmu.edu}{\nolinkurl{tsaidi@andrew.cmu.edu}}}} & 
\multicolumn{1}{c}{\makecell[c]{Florian Gunsilius\textsuperscript{\(*\)}\\ \href{mailto:florian.felix.gunsilius@emory.edu}{\nolinkurl{fgunsil@emory.edu}}}} & 
\multicolumn{1}{c}{\makecell[c]{Gonzalo Mena\textsuperscript{\(\dagger\)}\\ \href{mailto:gmena@andrew.cmu.edu}{\nolinkurl{gmena@andrew.cmu.edu}}}} 
\end{tabular}
\vspace{3mm}\\ 
\textsuperscript{\(\dagger\)}Department of Statistics and Data Science, Carnegie Mellon University\\
\textsuperscript{\(*\)}Department of Economics, Emory University
}

\begin{document}

\maketitle

\begin{abstract}
    The Sinkhorn divergence, a debiased form of entropic optimal transport, has been widely used in various scientific disciplines due to its computational and statistical tractability. Unlike the quadratic Wasserstein distance, however, the Sinkhorn divergence is not a metric and it does not naturally admit a formal Riemannian structure. These properties are central to many applications of optimal transport methods that utilize interpolation, barycentric averaging, parallel transport or other geometric operations. Recent work addresses these shortcomings by constructing a formal Riemannian tangent space and a metric tensor on the space of probability measures that quadratically approximates the Sinkhorn divergence. In this work, we extend this framework with two major threads of contributions. For a suitable class of finite-smoothness cost functions, we prove Hadamard differentiability of the Sinkhorn metric tensor and we derive the Levi-Civita covariant derivative in this geometry; we then use the covariant derivative to obtain the geodesic and parallel transport equations, each of which amount to an ODE on a fixed Reproducing Kernel Hilbert Space (RKHS) with smooth coefficients. Our second contribution is to linearize the solutions to the geodesic and parallel transport equations using standard tools from Banach space ODE theory. We show that under standard regularity conditions these linearizations imply $\sqrt{n}$-limiting Gaussian processes centered at population values for the plugin estimators of the exponential map, the logarithmic map, parallel transport and the induced Sinkhorn distance. Moreover, we show that the nonparametric bootstrap is consistent for all objects mentioned. To the best of our knowledge, these are the first $\sqrt{n}$--limit theorems and bootstrap consistency results for canonical geometric operations in a transport-based geometry on the space of probability measures.
\end{abstract}


\section{Introduction}

The Entropic Optimal Transport (EOT) cost was proposed by \citet{cuturi2013sinkhorn} as a method to regularize the Wasserstein distance arising from the classical Kantorovich optimal transport problem \citep{santambrogio2015optimal}. Since then, it has been used extensively in the biological sciences, machine learning and statistics communities due to its favorable computational and statistical properties. Despite this widespread adoption, EOT suffers from some drawbacks; for example, it does not endow the space of probability measures with metric or Riemannian structure
and its associated cost does not even satisfy the criteria of a statistical divergence \citep{nishiyama2018generalized}. To rectify the latter issue, the Sinkhorn divergence was proposed by \citet{genevay2017learninggenerativemodelssinkhorn} to debias the EOT cost, and it has become a standard tool amongst scientists and statisticians who use optimal transport.

Despite this rectification, the Sinkhorn divergence still does not endow the space of probability measures with metric or Riemannian structure. \citet{lavenant2024riemannian} propose a solution to this by constructing a formal Riemannian tangent space and metric tensor that quadratically approximates the Sinkhorn divergence. Their construction has laid the groundwork for the study of gradient flows on entropic geometries \citep{hardion2025gradient} and the convergence of EOT to unregularized Optimal Transport (OT) \citep{mordant2026convergence}; with that said, many important geometric objects in the construction have remained underived, and many open questions about the statistical tractability of the construction remain unanswered.

In this work we address these open questions by extending the geometric constructions of \citet{lavenant2024riemannian} and deriving the statistical properties of associated geometric objects. The principal statistical consequence of our geometric construction can be summarized informally as follows. Given independent samples from the underlying distributions, and under the regularity conditions developed below, the plug-in estimators of the Sinkhorn geodesic interpolant, logarithmic map, parallel-imputation trajectory, and squared geodesic distance admit $\sqrt{n}$-Gaussian limits. In particular, for each such operation $\Theta$, we establish
\[
\sqrt{n}(\widehat\Theta_n-\Theta)\rightsquigarrow\mathbb Z_\Theta,
\]
where $\mathbb Z_\Theta$ is a centered Gaussian element in the appropriate function space, and the nonparametric bootstrap consistently approximates its limiting distribution. These results are especially pertinent to single-cell genomics: there, optimal transport is used routinely to infer cell-state trajectories from snapshot RNA-sequencing measurements \citet{schiebinger2019optimal}. More recently \citet{saidi2026wassersteinparalleltransportpredicting} employed parallel imputation to transfer gene-expression dynamics across biological systems. However, statistical estimation for the associated functionals is both computationally and statistically prohibitive. Our results provide a tractable statistical foundation for uncertainty quantification for these estimands when using the Sinkhorn geometry as a surrogate of the Wasserstein geometry. 

\medskip

\noindent{\emph{Contributions}}. We briefly remark on our contributions in this work more formally. Specifically, our results fall into two categories: 

\begin{enumerate}
    \item \emph{Analytical:} Our first contribution is to further the construction of \citet{lavenant2024riemannian} by deriving the Levi-Civita covariant derivative using the technique employed by \citet{gigli2012second}. This derivation requires that we prove a sufficiently strong notion of differentiability -- in particular, we prove full Hadamard differentiability --  for the metric tensor $\tilde{\mathbf{g}}_\mu$ along suitable curves of measures. We note that our results restrict to a class of finite smoothness costs; despite this restriction, the setting is widely applicable, as it subsumes costs that are both widely used in practice and approximate the squared Euclidean cost arbitrarily well. As an immediate corollary, the Levi-Civita covariant derivative yields the geodesic equations and the parallel transport equations, the first of which further yields the exponential and logarithmic maps in this geometry. Our analysis reveals that both the exponential map and parallel transport each amount to an ODE on a fixed Reproducing Kernel Hilbert Space (RKHS) with smooth coefficients. This finding has immediate practical implications, as it suggests a path to a tractable computational procedure for geodesic interpolation and parallel transport.

    \item \emph{Statistical:} Leveraging the differentiation results described above, we derive linearizations of the solutions to the geodesic and parallel transport equations. We then use the functional delta method to obtain root-$n$ Gaussian process limits for the plugin estimator of the exponential map, the logarithmic map, parallel transport and the Sinkhorn distance. Moreover, we show that our linearizations are sufficiently smooth for validity of the nonparametric bootstrap, circumventing the need for a derivation of an asymptotic variance estimator. To the best of our knowledge, the statistical results presented here are the first $\sqrt{n}$--limit theorems and bootstrap consistency results for canonical geometric operations in a transport-based geometry on the space of probability measures.

\end{enumerate}

\noindent \emph{Outline of the paper.} In \Cref{notation} we introduce important notation, while in \Cref{related work} and \Cref{preliminaries} we review prior work and recall preceding results on the Sinkhorn Riemannian geometry. In \Cref{sec: parameterizing action} we reparameterize the Sinkhorn metric tensor $\tilde{\mathbf{g}}_\mu$ through a family of operators on a fixed RKHS. This provides the starting point for the first major technical contribution of the paper: establishing the Hadamard differentiability of the metric tensor in \Cref{sec: hadamard diff main text} (\Cref{metric hadamard diff}). This result does not follow immediately from existing differentiability results for the Schr\"odinger potentials. Indeed, its proof requires developing a series of operator-level linearizations in the relevant negative Sobolev and RKHS topologies. These analytical arguments constitute a substantial portion of the mathematical contribution of the first part of the work. With this regularity theory established, in \Cref{sec: deriving the connection} we derive the Levi-Civita covariant derivative, and in \Cref{sec: geodesic equations and pt} we instantiate covariant differentiation to obtain the geodesic and parallel transport equations. These constructions, comprising \Cref{levi civita}, \Cref{geodesic equations} and \Cref{PT IVP}, follow from standard Riemannian-geometric principles, albeit through extensive calculations. We note that our interest in Hadamard differentiability is motivated by statistical inference, since it provides the foundation for applying the functional delta method. However, obtaining limit theorems for the remaining geometric objects requires additional linearizations that do not follow directly from the differentiability of the metric tensor.

In \Cref{sec: linearizations} we establish linearizations of the Sinkhorn Riemannian distance (\Cref{sinkhorn distance hadamard diff}), the Riemannian exponential map (\Cref{sec: expmap linearization}), the Riemannian logarithmic map (\Cref{sec: logmap linearization}) and the solutions to the parallel transport equations (\Cref{sec: pt linearization}). The second major technical contribution of the paper concerns the regularity theory needed to obtain the latter three linearizations. Although the differentiability of solutions to sufficiently smooth Banach space ODEs is well understood, the equations derived in \Cref{sec: geodesic equations and pt} are not immediately amenable to these results, as they do not directly define ODEs on an open subset of a Banach space. Consequently, in \Cref{sec: extensions and smoothness} we construct smooth extensions and reparameterizations of these equations on suitable open neighborhoods in the ambient function spaces. Establishing the existence and smoothness of these extensions, particularly the nonlinear coefficients of the geodesic equations, requires additional functional analytic arguments and constitutes the principal technical challenge of the second part of the work. The central regularity result is \Cref{smoothness of geodesic coefficients}. Once this regularity has been established, the remaining ODE linearizations can be obtained using standard Banach space ODE theory, together with the additional analysis required for the logarithmic map. Finally, in \Cref{sec: clts} we apply the functional delta method \citep{van2000asymptotic} to these linearizations to establish Gaussian process limits for the plugin estimators of the aforementioned geometric objects.

\subsection{Related Work} \label{related work}

Our work relates to many different areas of optimal transport and statistical inference. We discuss each area below.

\emph{Optimal Transport Geometry.}
The geometric interpretation of quadratic optimal transport originates with displacement interpolation \citep{mccann1997convexity}, the dynamic formulation of \citet{benamou2000computational}, and the formal Riemannian calculus of \citet{otto2001geometry}. This structure was subsequently developed to include second-order geometric operations. In particular, \citet{lott2008geometric} computed the formal Levi--Civita connection and curvature, \citet{ambrosio2008parallel} constructed parallel transport along regular Wasserstein curves, and \citet{gigli2012second} developed a comprehensive second-order calculus encompassing covariant differentiation, parallel transport, the exponential map, and Jacobi fields. These constructions underpin a broad algorithmic literature that uses the Wasserstein tangent-space, geodesic interpolation, and parallel transport for procedures on the space of probability measures.

\emph{Statistical Inference for Optimal Transport.} Statistical inference for such geometric objects has, however, largely been developed in special settings or for individual functionals. \citet{sommerfeld2018inference} derive distributional limits for empirical Wasserstein distances on finite spaces, while \citet{delbarrio2019central} establish central limit theorems and bootstrap procedures for one-dimensional Wasserstein barycentric functionals. In Bures--Wasserstein geometry, \citet{kroshnin2021statistical} prove asymptotic normality for empirical barycenters and \citet{kroshnin2024generalized} develop corresponding bootstrap guarantees. More recently, \citet{sadhu2024stability} obtain limit theorems and nonparametric bootstrap consistency for Brenier maps in the semidiscrete setting. Tangent space comparison and parallel transport have also been incorporated into statistical methodology for univariate distributions \citep{chen2023wasserstein}, and multivariate distributional dynamics modeling \citep{saidi2026wassersteinparalleltransportpredicting}. These results motivate the study of inference on transport-geometric quantities, though they typically concern restricted geometries or particular operations. 

The construction directly underlying our work in this paper is due to \citet{lavenant2024riemannian}; they define a Riemannian metric tensor from the Hessian of the Sinkhorn divergence, identify the associated RKHS-based tangent space, and construct the induced geodesic distance. Subsequent work has studied gradient flows in this geometry \citep{hardion2025gradient} and its relation to Wasserstein geometry as the regularization vanishes \citep{mordant2026convergence}. More recently, \citet{lavenant2026approximation} developed spatial and temporal discretization schemes for approximating the geodesic Sinkhorn distance, establishing convergence of both the discrete distances and associated geodesics. We instead develop its second-order differential geometry: following the strategy of \citet{gigli2012second}, we derive the Levi--Civita covariant derivative and the corresponding geodesic and parallel transport equations, which in turn define the exponential and logarithmic maps studied below.

\medskip

\emph{Statistical Inference for EOT.}
Entropic regularization is well known to substantially improve the statistical behavior of optimal transport at fixed regularization. \citet{genevay2019sample} established parametric-rate sample-complexity bounds for Sinkhorn divergences, while \citet{mena2019statistical} proved a central limit theorem for the empirical EOT cost and \citet{rigollet2025sample} obtained dimension-free parametric rates for several EOT quantities. In discrete settings, \citet{bigot2019central} and \citet{klatt2020empirical} developed limit theory and bootstrap procedures for regularized transport costs, plans, and Sinkhorn divergences. \new{\citet{gonzalezsanz2022weak} established functional central limit theorems for empirical Schr\"odinger potentials and weak limits for empirical entropic transport plans evaluated against test functions, together with asymptotic distributions for the Sinkhorn divergence under both the null and alternative hypotheses.} \citet{goldfeld2024limit} establish first and second-order Hadamard differentiability of the Schr\"odinger potentials and derive Gaussian limits and bootstrap consistency for empirical EOT potentials, barycentric EOT maps, and the Sinkhorn divergence. We note, however, that these EOT quantities are distinct from the Riemannian operations studied here and, unlike our exponential map, do not generally yield an endpoint-interpolating analogue of the Brenier map.

\subsection{Notation} \label{notation}

\emph{Statistics and Analysis.} Suppose $\mathbb{B}_1,\mathbb{B}_2$ are Banach spaces. We write $\mathbb{B}^*$ for the continuous dual of a Banach space $\mathbb{B}$ and $\mathcal{L}(\mathbb{B}_1,\mathbb{B}_2)$ for the space of bounded linear operators from $\mathbb{B}_1$ to $\mathbb{B}_2$, endowed with the operator norm $\|A\|_{\operatorname{op}}:=\sup_{\|x\|_{\mathbb{B}_1}\leq 1}\|Ax\|_{\mathbb{B}_2}$. For normed spaces $\mathbb{E}$ and $\mathbb{F}$, we write $\mathbb{E}\hookrightarrow\mathbb{F}$ when $\mathbb{E}$ injects continuously into $\mathbb{F}$. For nonnegative quantities $a$ and $b$, we write $a\lesssim b$ if $a\leq Cb$ for a constant $C>0$ independent of the quantities under consideration, and $a\asymp b$ if both $a\lesssim b$ and $b\lesssim a$; dependence of the implicit constant may be indicated by a subscript for emphasis. We write $C^m(X)$ for the space of functions from $X$ to $\mathbb{R}$ with $m$ continuous derivatives, equipped with the norm $\|f\|_{C^m}:=\sum_{|\alpha|\leq m}\|\partial^\alpha f\|_{L^\infty(X)}$. Similarly, $C^{(m,m)}(X\times X)$ denotes the space of functions whose mixed derivatives of the $(\alpha, \beta)$ order are continuous for every $|\alpha|,|\beta|\leq m$. We denote the constant-one function on $X$ by $\mathbbm{1}_X$. Suppose $(\mathcal{F},\|\cdot\|_{\mathcal{F}})$ is a normed vector space of functions from $X$ to $\mathbb{R}$. We denote by $\mathcal{F}/\mathbb{R}$ the space $\mathcal{F}$ modulo constant functions, with elements denoted by $[f]$, and equip it with the quotient norm $\|[f]\|_{\mathcal{F}/\mathbb{R}}:=\inf_{\lambda\in\mathbb{R}}\|f-\lambda\mathbbm{1}_X\|_{\mathcal{F}}$. For any Hilbert space $H$ and closed subspace $M$, we equip $H/M$ with the canonical Hilbert-space structure induced by the identification $H/M\simeq M^\perp$, namely $\langle[f],[g]\rangle_{H/M}:=\langle P_{M^\perp}f,P_{M^\perp}g\rangle_H$ \citep[Equation 16.6]{fricain2016theoryvol2}. In particular, since $\mu$ is a probability measure, we identify $L^2(\mu)/\mathbb{R}$ with the subspace of mean-zero functions and write
\[
    \langle[f],[g]\rangle_{L^2(\mu)/\mathbb{R}}
    :=
    \left\langle
        f-\int_X f\,d\mu,\,
        g-\int_X g\,d\mu
    \right\rangle_{L^2(\mu)}.
\]
\medskip

\noindent\emph{Differentiation and Convergence.} For a map $F:\mathbb{E}\to\mathbb{F}$ between normed spaces, $\operatorname{D}_hF_x$ denotes the derivative of $F$ at $x$ in the direction $h$, while $\operatorname{D}F_x$ denotes the corresponding derivative operator. When the base point is clear from context, we may suppress the subscript $x$. If $F$ has several arguments, $\operatorname{D}_{i,h}F$ denotes differentiation with respect to its $i$th argument in the direction $h$, and $\operatorname{D}^k_{(h_1,\ldots,h_k)}F_x$ denotes the $k$th derivative evaluated in the directions $(h_1,\ldots,h_k)$. The notion of differentiability---Fr\'echet or Hadamard---will be clear from context. We write $\rightsquigarrow$ for weak convergence of random elements and $\rightsquigarrow_B$ for conditional weak convergence of bootstrap random elements, in outer probability. We write $\mu_n\rightharpoonup\mu$ for narrow convergence of probability measures, $x_n\xrightarrow{w}x$ for weak convergence in a Banach space (i.e., convergence against every continuous linear functional), and $x_n\to x$ for convergence in norm.

\medskip

\noindent\emph{Differential and Riemannian Geometry.} For a manifold $\mathcal{M}$ and $p\in\mathcal{M}$, $T_p\mathcal{M}$ denotes the tangent space to $\mathcal{M}$ at $p$, and we denote by $T\mathcal{M}:=\{(p,v):p\in\mathcal{M},\,v\in T_p\mathcal{M}\}$ its \textit{tangent bundle}. We write $\exp_p$ and $\log_p$ for the Riemannian exponential and logarithmic maps, respectively, and $\operatorname{PT}_{p\to q}$ for parallel transport from $p$ to $q$ along the curve under consideration.

\subsection{Preliminaries} \label{preliminaries}

In this section we review important background material relevant to our results. We will begin by reviewing the EOT cost and its debiased counterpart, the Sinkhorn divergence. We will then introduce the formal Riemannian geometry proposed by  \citet{lavenant2024riemannian} that quadratically approximates the Sinkhorn divergence. 

\subsubsection{The Sinkhorn Divergence}

In this work we will consider the following setting. Suppose that $X$ is a convex and compact subset of $\mathbb{R}^d$ such that $X = \overline{\operatorname{int}(X)}$, the ground cost $c: X\times X \to [0,\infty)$ is some symmetric and nonnegative cost function to be specified in more detail later, and $\mathcal{P}(X)$ is the set of Borel probability measures on $X.$ Fix an $\varepsilon > 0$ as a regularization parameter, which we regard as fixed throughout this work. The entropic optimal transport (EOT) cost is defined by 
\begin{equation}
    \OT_\varepsilon(\mu, \nu) = \inf_{\pi \in \Pi(\mu, \nu)} \left\{\int_{X \times X} c\, d\pi + \varepsilon \KL(\pi \,| \, \mu \otimes \nu)\right\}
\end{equation}
where $\Pi(\mu, \nu)$ is the set of probability measures on $X \times X$ with marginals $(\mu, \nu)$ and KL is the Kullback-Leibler divergence,
\[\KL(\pi\,|\,\rho) = \begin{cases}
    \int_{X\times X} \log\left(\frac{d\pi}{d\rho}\right)\, d\pi & \text{if $\pi \ll \rho$} \\
    +\infty & \text{else}
\end{cases}.\]
The entropic optimal transport problem also admits a dual formulation, given by 
\begin{equation}
    \OT_\varepsilon(\mu, \nu) = \sup_{f, g \in C_b(X)} \left\{ \int_X f\, d\mu + \int_X g \,d\nu  - \varepsilon\int_{X \times X}\left(\exp\left(\frac{1}{\varepsilon}\left(f \oplus g - c\right)\right) -1\right)\, d(\mu \otimes \nu)\, \right\}
\end{equation}
where $f \oplus g$ is the map $(x, y) \mapsto f(x) + g(y)$ and $C_b(X)$ is the set of bounded and continuous functions over $X$. The dual problem admits maximizers $f_{\mu, \nu}, g_{\mu, \nu}$, called \textit{Schr\"odinger potentials}, that satisfy $\OT_\varepsilon(\mu, \nu) = \int_X f_{\mu, \nu}\, d\mu + \int_X g_{\mu, \nu}\, d\nu$ and are solutions to the so-called Schr\"odinger system
\begin{align}
    f_{\mu, \nu}(y) &= -\varepsilon \log \int_X\exp\left(\frac{1}{\varepsilon}\left(g_{\mu, \nu}(x) - c(y,x)\right)\right)\, d\nu(x) \\
    g_{\mu, \nu}(y) &= -\varepsilon \log \int_X\exp\left(\frac{1}{\varepsilon}\left(f_{\mu, \nu}(x) - c(x,y)\right)\right)\, d\mu(x)
\end{align}
for every $y \in X$. With optimal dual potentials, the optimal transport plan $\pi$ is simply given by the reweighted product measure, $d\pi = \exp((f_{\mu, \nu} \oplus g_{\mu, \nu} - c)/\varepsilon)\,d(\mu \otimes \nu).$ 

Note that the Schr\"odinger potentials are unique modulo the following equivalence: $(f, g) \sim (f + \lambda \mathbbm{1}_X, g - \lambda \mathbbm{1}_X)$ for any $\lambda \in \mathbb{R}. $ A canonical selection principle is obtained by setting $f_{\mu, \nu}(x_0) = g_{\mu, \nu}(x_0)$ for some fixed $x_0 \in X$, which yields the property $f_{\mu, \nu} = g_{\nu, \mu}$, and therefore $f_{\mu, \mu} = g_{\mu, \mu}$. 
\begin{lemma}\label{uniqueness lemma}
    If the cost $c$ satisfies $c(x,y) = c(y,x)$ and $(f_{\mu, \nu}, g_{\mu, \nu})$ are optimal Schr\"odinger potentials for $(\mu, \nu)$, then $(g_{\mu, \nu}, f_{\mu, \nu})$ are optimal Schr\"odinger potentials for the pair $(\nu, \mu).$ Moreover, if $f_{\mu, \nu}(x_0) = g_{\mu, \nu}(x_0)$ for some fixed $x_0 \in X$, then $f_{\mu, \nu} = g_{\nu, \mu}.$
\end{lemma}

The entropic cost $\OT_\varepsilon$ is not a valid distance metric \emph{nor} a statistical divergence, as $\OT_\varepsilon(\mu, \mu)$ is nonzero in general. The Sinkhorn divergence was proposed by \citet{genevay2017learninggenerativemodelssinkhorn} to rectify this bias, and is defined by
\begin{equation}
    S_\varepsilon(\mu, \nu) = \OT_\varepsilon(\mu, \nu) - \frac{1}{2}\OT_\varepsilon(\mu, \mu) - \frac{1}{2}\OT_\varepsilon(\nu, \nu).
\end{equation}

Observe that $S_\varepsilon(\mu,\mu)=0$ and $S_\varepsilon(\mu,\nu)\geq 0$, with equality if and only if $\mu=\nu$. Nevertheless, $\sqrt{S_\varepsilon}$ fails to satisfy the triangle inequality, as shown by \citet{lavenant2024riemannian}, and hence does not itself define a metric on $\mathcal{P}(X)$. From a statistical perspective, however, the Sinkhorn divergence retains the favorable $\sqrt{n}$-rate estimability of EOT \citep{goldfeld2024limit,del2023improved,mena2019statistical,gonzalez2022weak}, making it an attractive transport-based divergence for statistical applications. This combination of transport structure and statistical regularity motivates the Riemannian approximation proposed by \citet{lavenant2024riemannian}. Their construction equips $\mathcal{P}(X)$ with a formal Riemannian structure that locally approximates the Sinkhorn divergence and, much like the Wasserstein geometry of OT, lifts the underlying geometry of $X$ to the space of probability measures, while seeking to preserve the smoothness and favorable statistical properties induced by entropic regularization.

\paragraph{Riemannian Geometry.}  
In this section we will review key aspects of the Riemannian construction proposed by \citet{lavenant2024riemannian}. Their construction formalizes the Riemannian tangent space at a measure, and then obtains the Hessian of the Sinkhorn divergence with respect to perturbations within this tangent space. We will assume that $X$ is a compact and convex subset of $\mathbb{R}^d$ and $c \in C^{(m, m)}(X \times X)$, and we recall the following definition of $C^{-m}$ perturbations of a measure from \citet{lavenant2024riemannian}.



\begin{defn}[$C^{-m}$ perturbations, \citet{lavenant2024riemannian}] \label{Cm perturbations}
    Let $m > 0$ and $\tau > 0$. We call a path $(\mu_t)_{t\in (-\tau, \tau)}$ valued in $\mathcal{P}(X)$ a $C^{-m}$-perturbation of $\mu = \mu_0$ if it is continuously differentiable in $C^m(X)^*$ endowed with the weak-$*$ topology. In particular, for every $\phi \in C^m(X)$ it must hold that
    \[\frac{d}{dt}\int \phi \, d\mu_t = \dot \mu_t(\phi)\]
    for some $\dot \mu_t \in C^m(X)^*$, where $ \dot \mu_t$ is weakly continuous. We write $\dot \mu \in C^m(X)^*$ for the derivative of the curve at time $t = 0$. 
\end{defn}

Note that $m$ does \textit{not} denote the order to which the perturbation approximates $\mu_t$. Instead, it controls the regularity of the test function class, and therefore determines the topology in which the curve is differentiated. Since $C^0(X)^* \subset C^1(X)^* \subset \dots$, a larger $m$ yields a larger dual space and a weaker topology. Therefore, a $C^{-k}$ perturbation is automatically a $C^{-m}$ perturbation for any $k \leq m$. Thus, the space of perturbations gets \say{rougher} as $m$ gets larger, while it gets smoother as $m$ approaches $0$.

Now we'll introduce important objects that will feature prominently in the tangent space and metric tensor. Define $k_c(x,y) = \exp( - c(x,y)/\varepsilon)$, and define the so-called \emph{self-transport kernel},
\[k_\mu(x,y) := \exp\left(\frac{1}{\varepsilon}\left(f_{\mu, \mu}(x) + f_{\mu, \mu}(y) - c(x,y)\right)\right).\]
Recall that our selection principle for the Schr\"odinger potentials implies $f_{\mu, \nu} = g_{\nu, \mu}$, and thus $f_{\mu, \mu} = g_{\mu, \mu}.$  By our assumptions on $c$, the kernels $k_\mu$ and $k_c$ are both non-negative, continuous, positive-definite and universal. A kernel $k$ satisfying such properties induces a unique Reproducing Kernel Hilbert Space (RKHS) $\mathcal{H}_k$; thus, we will write $\mathcal{H}_\mu, \mathcal{H}_c$ to be the RKHSs associated with  $k_{\mu}$ and $k_c$ respectively. For more information on the theory of RKHSs, we direct readers to \Cref{preliminaries on RKHS}.   
\begin{defn}[Integral operators]\label{kernel integral operators}
    For $\mu \in \mathcal{P}(X)$, define the following operators $K_\mu, H_\mu$, $H_c$ by 
    \begin{gather*}
        K_\mu: \mathcal{H}_{\mu} \to \mathcal{H}_{\mu}, \quad K_\mu(\phi)(y) = \int_X \phi(x)k_\mu(x,y) \, d\mu(x) \\
        H_\mu: \mathcal{H}_{\mu}^*  \to \mathcal{H}_{\mu},  \quad H_\mu(\nu)(y) =  \nu(k_\mu(\cdot, y)) \quad \text{and} \quad  H_c: \mathcal{H}_{c}^* \to \mathcal{H}_{c}, \quad H_c(\nu)(y) =  \nu(k_c(\cdot, y)) .
    \end{gather*}
\end{defn}
For a bounded linear functional $\sigma \in \mathcal{H}_\mu^*$, the element $H_\mu(\sigma)$ is its Riesz representative in $\mathcal{H}_\mu$ \citep{lavenant2024riemannian}, thus satisfying $ \sigma(\phi) = \langle H_{\mu}(\sigma), \phi\rangle _{\mathcal{H}_{\mu}}$ for any $\phi \in \mathcal{H}_\mu$. Also observe that $K_{\mu}(\phi) = H_{\mu}(\phi\mu)$ can therefore be interpreted as the Riesz representative of the functional $\psi \mapsto \int_X \phi \psi \,d\mu$ induced by the signed measure $\phi\mu$. Note that if $\nu$ is a signed Borel measure on $X$, then $H_\mu$ and $H_c$ can be written in integral form as
    \begin{align*}
        H_\mu(\nu) = \int_X k_\mu(x,\cdot)\, d\nu(x) \quad \text{and} \quad H_c(\nu) = \int_X k_c(x,\cdot)\, d\nu(x).
    \end{align*}
When $\nu$ is a Borel probability measure, the operators $H_\mu$ and $H_c$ are often referred to as the kernel mean embeddings of $\nu$ into the RKHSs $\mathcal{H}_\mu$, $\mathcal H_c$ respectively \citep{muandet2017kernel}.

\medskip

\emph{The Metric and Tangent Space.} Write $\mathcal{H}_\mu/\mathbb{R}$ as the space $\mathcal{H}_{\mu}$ modulo constant functions, equipped with the canonical quotient inner product. This space $\mathcal{H}_\mu/\mathbb{R}$ is dual to $\mathcal{H}_{\mu, 0}^*$, the space of bounded linear functionals $\sigma$ on $\mathcal{H}_\mu$ such that $\sigma (\mathbbm{1}_X) = 0$ \citep[Lemma 4.5]{lavenant2024riemannian}.
Therefore, $H_\mu$ can be restricted to an operator $H_\mu: \mathcal{H}_{\mu, 0}^* \to \mathcal{H}_{\mu}/\mathbb{R}$ that also maps a $\sigma \in \mathcal{H}_{\mu, 0}^*$ to its Riesz representative $H_\mu(\sigma) \in \mathcal{H}_\mu/\mathbb{R}$. \citet{lavenant2024riemannian} prove that the Sinkhorn divergence has the following second order behavior
\[\lim_{t \to 0}\frac{S_\varepsilon(\mu, \mu_t)}{t^2} = \frac{\varepsilon}{2}\left\langle H_\mu(\dot \mu),   \left(\operatorname{id}_{\mathcal{H}_\mu /\mathbb{R}} - K_\mu^2\right)^{-1}H_{\mu}(\dot \mu)\right\rangle_{\mathcal{H}_\mu/\mathbb{R}}\]
for a $C^{-m}$ perturbation $(\mu_t)_{t}$ of $\mu = \mu_0$. This characterization readily leads to a definition of the metric tensor and the tangent space. 

\begin{defn}[The metric on the tangent space]\label{original tangent space definition}
    Suppose $c \in C^{(m, m)}(X \times X)$ is symmetric and nonnegative, and $k_c = \exp(-c/\varepsilon)$ is positive definite and universal. Define the Sinkhorn tangent space to a measure $\mu\in \mathcal{P}(X)$ as \[\text{$T_{\mu}\mathcal{P}(X) := \overline{C^m(X)^*_0}^{\|\cdot\|_{\mathbf{g}_{\mu}}}$ where $\|\sigma\|_{\mathbf{g}_{\mu}} := \sqrt{\mathbf{g}_\mu(\sigma, \sigma)}$}.\]
    and for $\dot \mu_1, \dot \mu_2$ in $C^m(X)_0^*$ define the metric tensor $\mathbf{g}_\mu$ as
    \begin{equation} \label{sinkhorn metric tensor}
        \mathbf{g}_\mu(\dot \mu_1, \dot \mu_2) = \frac{\varepsilon}{2}\dot \mu_1\left (  (\operatorname{id} - K_\mu^2)^{-1}H_{\mu}(\dot \mu_2)\right) = \frac{\varepsilon}{2}\left\langle H_\mu(\dot \mu_1),  (\operatorname{id}_{\mathcal{H}_\mu/\mathbb{R}} - K_\mu^2)^{-1}H_{\mu}(\dot \mu_2)\right \rangle_{\mathcal{H}_\mu/\mathbb{R}}.
    \end{equation}
\end{defn}

\begin{rmk}[\citet{lavenant2024riemannian}, Theorem 4.6]
    The Sinkhorn tangent space satisfies $T_{\mu}\mathcal{P}(X) \cong \mathcal{H}_{\mu, 0}^*$, thereby justifying the evaluation of $H_\mu$ at an element in $C^m(X)_0^*$ (since $C^m(X)^* \hookrightarrow \mathcal{H}_\mu^*$ by \Cref{continuous embedding}).
\end{rmk}

Observe that these definitions and the result above together imply that the quadratic approximation to the Sinkhorn divergence is only sensitive to a perturbation $\dot \mu$ through its Riesz representer in $\mathcal{H}_\mu$. 


\medskip

\emph{Change of variables.}  \citet{lavenant2024riemannian} propose using the change of variables $\mu \mapsto \exp(-f_{\mu,\mu}/\varepsilon)$ employed previously in the EOT literature by \citet{feydy2019interpolating}. This change of variables maps $\mathcal{P}(X)$ to a subset of the unit sphere of the RKHS $\mathcal{H}_c$ induced by the kernel $k_c = \exp(-c/\varepsilon)$, allowing all geometry to operate in the fixed ambient Hilbert space $\mathcal{H}_c$. Write $a_\mu = \exp(f_{\mu, \mu}/\varepsilon)$, and define the change of variable map $B: \mathcal{P}(X) \to \mathcal{H}_c$ given by
\[\quad B_{\mu} = H_c\left(a_\mu\mu\right) = \exp\left(-\frac{1}{\varepsilon}f_{\mu, \mu}\right).\]
As alluded to, one can readily verify that for all $\mu \in \mathcal{P}(X)$, $\|B_\mu\|_{\mathcal{H}_c} = 1$; thus, $B[\mathcal{P}(X)] \subset \mathbb{S}(\mathcal{H}_c)$ where $\mathbb{S}(\mathcal{H}_c)$ is the unit-sphere in the fixed Hilbert space $\mathcal{H}_c$. 
\begin{thm}[The metric under $\mu \mapsto B_\mu$, \citep{lavenant2024riemannian}, Theorem 4.12 ] \label{metric under change of vars thm}
    Let $(\mu_t)_t$ be a curve of measures such that $\beta_t  = B_{\mu_t}$ is weakly differentiable in $\mathcal{H}_c$ at $t = 0$. Then $\langle \dot \beta, B_\mu\rangle_{\mathcal{H}_c} = 0$, the path $(\mu_t)_t$ is weakly differentiable in $\mathcal{H}_\mu^*$ at $t = 0$, and $\mathbf{g}_\mu(\dot \mu, \dot \mu) = \tilde{\mathbf{g}}_\mu(\dot \beta, \dot \beta)$ where 
    \[\tilde{\mathbf{g}}_\mu(\dot \beta, \dot \beta) = \frac{\varepsilon}{2}\left(\|\dot \beta\|^2_{\mathcal{H}_c} + 2\left\langle a_\mu \dot \beta,(\operatorname{id}_{L^2(\mu)/\mathbb{R}} - K_\mu)^{-1}(a_\mu\dot \beta) \right \rangle_{L^2(\mu)}\right).\]
    Moreover, if $\dot \beta_1, \dot \beta_2 \in \mathcal{H}_c$ with $\langle \dot \beta_1, B_\mu\rangle_{\mathcal{H}_c} = \langle \dot \beta_2, B_\mu\rangle_{\mathcal{H}_c} = 0$, then we have
    \[\tilde{\mathbf{g}}_{\mu}(\dot \beta_1, \dot \beta_2) = \frac{\varepsilon}{2}\left(\langle \dot \beta_1, \dot \beta_2\rangle_{\mathcal{H}_c} + 2\left\langle  a_\mu\dot \beta_1, (\operatorname{id}_{L^2(\mu)/\mathbb{R}} - K_\mu)^{-1}(a_\mu \dot \beta_2) \right\rangle_{L^2(\mu)} \right).\]
\end{thm}
Since $\tilde{\mathbf{g}}_{\mu}(a, a)$ is a quadratic form, the symmetric bilinear form  $\tilde{\mathbf{g}}_{\mu}(a, b)$ can be recovered from the quadratic form via the polarization identities \citep{thomas2014polarization} and an application of the fact that $(\operatorname{id}_{L^2(\mu)/\mathbb{R}} - K_\mu)$ is self-adjoint on $L^2(\mu)/\mathbb{R}$ \citep[Theorem 4.3]{lavenant2024riemannian}. The result shows that the Sinkhorn Riemannian geometry can be described as an augmented inner product on an appropriate subset of $\mathbb{S}(\mathcal{H}_c).$ We'll define the set
    \[\mathfrak{T}_\mu \mathcal{P}(X) = \left\{\dot \beta \in \mathcal{H}_c: \langle \dot \beta, B_{\mu}\rangle_{\mathcal{H}_c} = 0\right\}.\]
Throughout the paper, we will denote $T_\mu \mathcal{P}(X)$ as the original tangent space at $\mu$, and $\mathfrak{T}_\mu \mathcal{P}(X)$ will denote the tangent space to $\mathbb{S}(\mathcal{H}_c)$ at $B_\mu$. 


    

\subsection{Preliminaries on Reproducing Kernel Hilbert Spaces (RKHS)} \label{preliminaries on RKHS}

For a thorough treatment of the theory of RKHSs, we recommend \citet{paulsen2016introduction}. We will provide a brief overview of the theory and some relevant results in this section.

For any non-negative, continuous, positive-definite and universal kernel $k: X\times X\to [0, \infty)$, there exists a unique so-called Reproducing Kernel Hilbert Space $\mathcal{H}$ associated to it. The RKHS $\mathcal{H}$ consists of the completion of the linear span of kernel sections $\{k(x,\cdot)\,|\, x\in X\}$ under the metric induced by inner product
\[\left\langle \sum_{i = 1}^n \alpha_i k(x_i, \cdot), \sum_{j = 1}^m \beta_j k(x_j, \cdot)\right\rangle_{\mathcal{H}} = \sum_{i = 1}^n \sum_{j = 1}^m\alpha_i \beta_j k(x_i, x_j). \]
Moreover, function evaluation in this RKHS can be written $f(x) = \langle f, k(x, \cdot)\rangle_{\mathcal{H}}$, rendering it a bounded linear functional on the space.

\begin{prop} \label{continuous embedding}
    If $c \in C^{(m,m)}(X \times X)$, then $\mathcal{H}_c \hookrightarrow C^m(X)$ and $\mathcal{H}_\mu \hookrightarrow C^m(X)$. 
\end{prop}

We provide the proof in \Cref{proof of continuous embedding}. For our results in subsequent sections, we will restrict our analysis to specific cost functions that induce a finite, but sufficiently high, smoothness RKHS $\mathcal H_c$. We distinguish the smoothness order $m$, which governs the $C^{-m}$ perturbation and statistical theory, from the Sobolev order $s$ of the RKHS.

\begin{asmpt}[Spectral decay] \label{spectral decay assumption}
    Fix an integer $m>d/2$ and let $s>m+d/2$ with $s \in \mathbb N$. Let $c(x,y)=-\varepsilon\log \kappa(x-y)$, where $\kappa:\mathbb R^d\to(0,\infty)$ is symmetric, continuous and positive definite, and suppose that its Fourier transform satisfies $\widehat{\kappa}(\omega)\asymp (1+\|\omega\|^2)^{-s}$.
\end{asmpt}

Under \Cref{spectral decay assumption}, the kernel $k_c=\exp(-c/\varepsilon)=\kappa$ induces an RKHS $\mathcal H_c$ that is norm-equivalent to $H^s(X)$. Indeed, \citet[Corollary 10.48]{wendland2004scattered} implies $\mathcal H_c\simeq H^s(X)$. Moreover, the condition $s>m+d/2$ guarantees sufficient classical regularity of the kernel for the $C^{-m}$ perturbation theory. In particular,
\[
    \int_{\mathbb R^d}\|\omega\|^{2m}\widehat{\kappa}(\omega)\,d\omega<\infty,
\]
so $\widehat{\kappa}$ is nonnegative and integrable. The Fourier
differentiation theorem \cite[Theorem~8.22(d)]{folland1999real}
implies that $\kappa\in C^{2m}(\mathbb R^d)$, and hence
$k_c(x,y)=\kappa(x-y)$ belongs to $C^{(m,m)}(X\times X)$. Since $\kappa$ is strictly positive and $X-X$ is compact, $\inf_{z\in X-X}\kappa(z)>0$, and therefore $c(x,y)=-\varepsilon\log\kappa(x-y)$ also belongs to $C^{(m,m)}(X\times X)$.

\begin{lemma}\label{RKHS norm equivalence}
    Let $k_c=\exp(-c/\varepsilon)$. Under \Cref{spectral decay assumption}, the RKHS $\mathcal H_c$ induced by $k_c$ is norm-equivalent to $H^s(X)$.
\end{lemma}

Since $s>d/2$, $H^s(X)$ is a Banach algebra under pointwise multiplication \citep[Theorem 4.39]{adams2003sobolev}. In particular, there exists $C>0$ such that $\|fg\|_{H^s(X)}\leq C\|f\|_{H^s(X)}\|g\|_{H^s(X)}$ for all $f,g\in H^s(X)$. By the norm equivalence $\mathcal H_c\simeq H^s(X)$, the same property holds in $\mathcal H_c$. In particular, for any $f,g\in\mathcal H_c$, the pointwise product $fg$ is also an element of $\mathcal H_c$. As a preview of the utility of this property, it will guarantee that the Levi--Civita covariant derivative for the Sinkhorn Riemannian geometry is a bounded operator on the tangent space $\mathfrak T_\mu\mathcal P(X)$.

\begin{rmk}[Smoothness of the tangent space]
    Combining \Cref{RKHS norm equivalence}, \Cref{original tangent space definition} and the change of variables leads one to conclude somewhat surprisingly that 
    \[\overline{C^m(X)_0^*}^{\|\cdot\|_{\mathbf{g}_\mu}} \cong H^s(X)_0^* \quad \text{with $s > m + d/2$.}\]
    Thus, although the Sinkhorn tangent space is constructed initially from $m$-th order distributional perturbations, its metric completion is governed by the Sobolev regularity of the RKHS induced by the transport kernel. 
\end{rmk}

It should also be made explicit that the squared Euclidean cost $c(x,y)\propto\|x-y\|_2^2$ is not an admissible cost under the present finite-smoothness assumptions, as it induces an RKHS that is not closed under pointwise multiplication. For $c(x,y)=\|x-y\|_2^2$, the induced kernel is the Gaussian kernel $k_c(x,y)=\exp(-\|x-y\|_2^2/\varepsilon)$. Its Fourier transform satisfies $\widehat{k}_c(\omega)\propto\exp(-\varepsilon\|\omega\|_2^2/4)$, and the associated RKHS is characterized by
\[
    \int_{\mathbb R^d}|\widehat f(\omega)|^2\exp\left(\frac{\varepsilon}{4}\|\omega\|_2^2\right)d\omega<\infty
\]
as shown in \citet[Section 3.1, Lemma 10]{vert2006consistency}. This space is not closed under pointwise multiplication. By contrast, under \Cref{spectral decay assumption} we have $\mathcal H_c\simeq H^s(X)$ with $s>d/2$, and hence $\mathcal H_c$ is a Banach algebra.

\begin{example}[The Mat\'ern Kernel]
    A canonical family of kernels $\kappa$ satisfying both \Cref{spectral decay assumption} and the hypotheses of the previous sections (i.e. positive definiteness and universality) is the Mat\'ern family,
    \begin{equation}
        \kappa_{\vartheta, \ell}(r) = \frac{2^{1-\vartheta}}{\Gamma(\vartheta)}\left(\frac{\sqrt{2\vartheta}\,r}{\ell}\right)^{\vartheta}K_\vartheta\left(\frac{\sqrt{2\vartheta}\,r}{\ell}\right)
    \end{equation}
    where $K_\vartheta$ is the modified Bessel function of the second kind \citep[Section 9.6]{abramowitz1964handbook}. 
\end{example}
The Mat\'ern kernel $\kappa_{\vartheta,\ell}$ has spectral decay $\widehat{\kappa}_{\vartheta,\ell}(\omega)\asymp_{\vartheta,\ell,d}(1+\|\omega\|_2^2)^{-(\vartheta+d/2)}$ \citep[Equation 4.15]{rasmussen2006gaussian}, and therefore has a Sobolev order of $s=\vartheta+d/2$. Consequently, the condition $\vartheta>m, \vartheta + d/2 \in \mathbb N$ implies $s>m+d/2$, and hence the Mat\'ern kernel satisfies \Cref{spectral decay assumption}. Importantly, it induces the cost $c_{\vartheta,\ell}(x,y)=-\varepsilon\log\kappa_{\vartheta,\ell}(x-y)$, which approximates the squared Euclidean cost in the sense that
\[
    -\varepsilon\log\kappa_{\vartheta,\ell}(x-y)\overset{\vartheta\to\infty}{\longrightarrow}\frac{\varepsilon}{2\ell^2}\|x-y\|_2^2.
\]





\section{Sinkhorn Riemannian Geometry} \label{sec: Sinkhorn riemannian geometry}

\new{In this section, we further the construction of the Sinkhorn Riemannian geometry by deriving its Levi-Civita covariant derivative \`a la \citet{gigli2012second}, and subsequently obtaining the corresponding geodesic and parallel transport equations. Recall that the Levi-Civita covariant derivative is the unique metric-compatible and torsion-free covariant derivative, and underpins many of the fundamental constructions of Riemannian geometry that have found algorithmic applications. For a review of covariant differentiation and Riemannian geometry, we recommend \citet{lee2018introduction}. Unlike in finite dimensions, however, the existence of a Levi-Civita connection is not automatic in infinite dimensions, and its construction requires establishing sufficient regularity of the metric tensor.

Our approach proceeds in four steps. First, in \Cref{sec: parameterizing action}, we show that the Sinkhorn metric tensor can be characterized as the inner product induced by a bounded linear operator $S_\mu$ on $\mathcal H_c$. Next, in \Cref{sec: operator differentiability}, we establish Hadamard differentiability of $\mu\mapsto S_\mu$ and derive its time derivative along sufficiently regular curves of measures. These results provide the necessary ingredients for verifying metric compatibility. In \Cref{sec: deriving the connection}, we use metric compatibility and the torsion-free requirement to identify a Levi-Civita covariant derivative. Finally, in \Cref{sec: geodesic equations and pt}, we derive the associated geodesic and parallel transport equations and establish their smooth ODE formulations on $\mathcal H_c$.}

\subsection{Parameterizing the Action of $\tilde{\mathbf{g}}_{\mu}$} \label{sec: parameterizing action}

In this subsection we re-express the metric tensor evaluated at two tangent directions $v,w \in \mathfrak{T}_\mu \mathcal{P}(X)$ as an inner product induced by a bounded linear operator $S_\mu$, $\tilde{\mathbf{g}}_\mu(v,w) = \frac{\varepsilon}{2}\langle S_\mu v, w\rangle_{\mathcal{H}_c}$. \new{We note that the non-bolded operators $R_\mu$ and $J_\mu$ introduced below act on $\mu$-dependent spaces, whereas their boldface counterparts $\mathbf{R}_\mu$ and $\mathbf{J}_\mu$ are obtained through a change of variables and belong to $\mathcal{L}(\mathcal{H}_c,\mathcal{H}_c)$. This reformulation places the operators on a common, $\mu$-independent Hilbert space, facilitating the subsequent analysis of their differentiability with respect to $\mu$.} 
\begin{corollary}\label{metric tensor bounded}
    Let $\mu \in \mathcal{P}(X)$, $\beta = B_\mu$ and let $v \in \mathfrak{T}_\mu \mathcal{P}(X)$. Then the operator $w \mapsto \tilde{\mathbf{g}}_{\mu}(v,w)$ is a bounded linear operator on $\mathcal{H}_c$.
\end{corollary}
One can readily verify that \citet[Proposition 4.15]{lavenant2024riemannian} and Cauchy--Schwarz imply \Cref{metric tensor bounded}. By the Riesz representation theorem, for a fixed $v\in \mathfrak T_\mu \mathcal{P}(X)$, there exists a unique element $G_v \in \mathfrak T_\mu \mathcal{P}(X)$ such that $\tilde{\mathbf{g}}_\mu(v,w) = \frac{\varepsilon}{2}\langle G_v, w\rangle_{\mathcal{H}_c}$. By the symmetry of $\tilde{\mathbf{g}}_{\mu}(v,w)$, this can be extended to the statement that there exist unique elements $G_v,G_w \in \mathfrak T_\mu \mathcal{P}(X)$ such that $\tilde{\mathbf{g}}_\mu(v,w) = \frac{\varepsilon}{2}\langle G_v, w\rangle_{\mathcal{H}_c} = \frac{\varepsilon}{2}\langle v, G_w\rangle_{\mathcal{H}_c}$. We will now show that $v \mapsto G_v$ can be written as a self-adjoint and bounded linear operator $S_\mu.$ To see this, define $\mathcal{Q}_{\mu} = \mathcal{H}_{\mu}/ \mathbb{R}$ and recall the isometric isomorphism $\iota_{\mu} : \mathcal{H}_{c} \to \mathcal{H}_{\mu}$ defined in \citet[Proposition 4.7]{lavenant2024riemannian} given by $\iota_{\mu}u = a_{\mu}u$.

Using $\iota_{\mu}^{-1}$, we can pull back constant functions in $\mathcal{H}_{\mu}$ to $\mathcal{H}_c$ with $\iota_{\mu}^{-1}\lambda = a_{\mu}^{-1}\lambda$, where $a_{\mu}^{-1}$ is the pointwise reciprocal of $a_{\mu}$. Set $E_\mu:=\operatorname{span}\{B_\mu\}$. Writing $\mathcal{Q}_\mu = \mathcal{H}_\mu/\mathbb{R}$ and noting that $a_\mu^{-1} = B_\mu$, we can deduce that constant functions pull back to $ \operatorname{span}\{a_{\mu}^{-1}\} = \mathfrak{T}_\mu \mathcal{P}(X)^\perp$, and $\mathcal{Q}_\mu$ pulls back to $\mathfrak{T}_\mu \mathcal{P}(X)$ under $\iota$. Now define $\Pi_{\mu}: \mathcal{H}_c \to \mathfrak{T}_\mu \mathcal{P}(X)$ as the $\mathcal{H}_c$-orthogonal projection onto $\mathfrak{T}_\mu \mathcal{P}(X)$ and consider the operator,
\begin{equation}\label{Smu original definition}
\begin{aligned}
    S_{\mu} = \operatorname{id}_{\mathcal{H}_c} + 2J_{\mu}R_{\mu}A_{\mu}, \quad \text{where }&A_{\mu}: \mathcal{H}_c \to \mathcal{Q}_{\mu} \text{ is given by $A_{\mu} z = [a_{\mu}z]$} \\
    &R_{\mu}: \mathcal{Q}_{\mu}  \to \mathcal{Q}_{\mu} \text{ is given by $R_{\mu} [z] = (\operatorname{id}_{\mathcal{Q}_{\mu}} - K_{\mu})^{-1}[z]$}, \\
    \text{and }&J_{\mu}:\mathcal{Q}_{\mu} \to \mathcal{H}_c \text{ is given by }J_{\mu}[z] = H_c(a_{\mu}^2 \Pi_{\mu}(a_{\mu}^{-1}[z])\mu).
\end{aligned}
\end{equation}
Note that $\Pi_\mu$ is also the $\tilde{\mathbf{g}}_\mu$-orthogonal projection onto $\mathfrak{T}_\mu\mathcal{P}(X)$ where $\tilde{\mathbf{g}}_\mu$ is regarded as an inner product on $\mathcal{H}_c$ instead of solely $\mathfrak{T}_\mu\mathcal{P}(X)$; this fact is due to \Cref{Pimu g orthogonal projection}. Indeed, the operators $R_{\mu}$ and $J_{\mu}$ are well-defined as maps on the quotient $\mathcal{Q}_{\mu}$, as proved in \Cref{well defined r j}. The operator $S_{\mu}$ will be the mechanism through which we will study the metric tensor $\tilde{\mathbf{g}}_{\mu}$. Under this definition, we show that $\tilde{\mathbf{g}}_{\mu}(v,w) = \frac{\varepsilon}{2}\langle S_{\mu}v, w\rangle_{\mathcal{H}_c}$ in \Cref{sec: operator metric equivalence}. 

\begin{thm}\label{operator metric equivalence}
    For every $v,w\in \mathfrak{T}_{\mu}\mathcal P(X)$
    the metric tensor satisfies $\tilde{\mathbf g}_{\mu}(v,w)
    =
    \frac{\varepsilon}{2}
    \left\langle
    S_{\mu}v,w
    \right\rangle_{\mathcal H_c}.$ Moreover, $S_\mu:\mathcal{H}_c \to \mathcal{H}_c$ is self-adjoint, and the restriction $S_{\mu}|_{\mathfrak{T}_\mu \mathcal{P}(X)}:\mathfrak T_{\mu}\mathcal P(X)\to \mathfrak T_{\mu}\mathcal P(X)$ is self-adjoint.
\end{thm}
As we will discuss in detail in \Cref{sec: deriving the connection}, to obtain the Levi-Civita covariant derivative one needs to show metric compatibility \citep{lee2018introduction}. This entails finding a curvewise derivative operator $\nabla_{\dot \mu_t}$ satisfying the identity $\frac{d}{dt}\tilde{\mathbf{g}}_{\mu_t}(v_t, w_t) =  \tilde{\mathbf{g}}_{\mu_t}(\nabla_{\dot \mu_t}v_t, w_t) + \tilde{\mathbf{g}}_{\mu_t}(v_t, \nabla_{\dot \mu_t}w_t)$ for a sufficiently regular curve of measures $(\mu_t)_t$ and tangents $(v_t)_t, (w_t)_t$.  To make sense of the left-hand-side of this identity, we need to differentiate $S_\mu$ along $(\mu_t)_t.$

\medskip

\emph{Reparameterizing $S_\mu$.} To facilitate differentiation of the map $\mu \mapsto S_\mu$, we will reparameterize $S_\mu$ so intermediate (co)domains in \Cref{Smu original definition} are independent of $\mu$. To do this, observe that we can associate with each $f \in \mathcal{H}_{\mu}/\mathbb{R}$ a function in $\mathfrak{T}_\mu \mathcal{P}(X) \subset \mathcal{H}_c$ through a map $\Theta_{\mu}: \mathcal{H}_{\mu}/\mathbb{R} \to \mathfrak{T}_\mu \mathcal{P}(X)$ given by $ [f ]\mapsto  \Pi_{\mu}\iota_{\mu}^{-1}f.$ With it, we can write
\begin{equation}\label{reparameterization}
    \begin{aligned}
    S_{\mu} = \operatorname{id}_{\mathcal{H}_c} + 2\mathbf{J}_{\mu} \mathbf{R}_{\mu}, \quad \text{where }&\mathbf{J}_{\mu}:  \mathcal{H}_c \to \mathcal{H}_c \text{ is given by $\mathbf{J}_{\mu} = J_{\mu}\Theta_{\mu}^{-1}\Pi_{\mu}$} \\
    \text{and }&\mathbf{R}_{\mu}: \mathcal{H}_c  \to \mathcal{H}_c \text{ is given by $\mathbf{R}_{\mu}  = \Theta_{\mu} R_{\mu} \Theta_{\mu}^{-1}\Pi_{\mu}$}
    \end{aligned}
\end{equation}
since $\Theta_\mu A_\mu =\Pi_\mu$. Note that the added composition with $\Pi_{\mu}$ serves to extend the domains of the operators to all of $\mathcal{H}_c$ where needed. One can readily verify that $\Theta_{\mu}$ is indeed a well-defined map from the quotient $\mathcal{H}_{\mu}/\mathbb{R}$ with a well-defined inverse (\Cref{theta inverse}).


\subsection{Operator Differentiability} \label{sec: operator differentiability}


Now we will use the parameterization described in \Cref{sec: parameterizing action} to differentiate the metric tensor. Since the ambient RKHS is fixed, differentiating $\tilde{\mathbf{g}}_\mu(v, w) = \frac{\varepsilon}{2}\langle S_\mu v, w\rangle_{\mathcal{H}_c}$ reduces to differentiating the operator valued map $\mu \mapsto S_\mu$.  Our strategy is chosen with statistical inference in mind, as in statistical theory, full Hadamard differentiability is the standard regularity condition underlying the functional delta method and related asymptotic results for plug-in estimators \citep{van2000asymptotic}. It therefore provides a common differentiability framework for both the geometric and statistical results that follow. Along curves $(\mu_t)_t$ that are differentiable in a suitable topology, the Hadamard derivative yields the corresponding time derivative of $S_{\mu_t}$ by the chain rule, while at the same time yielding a sufficiently strong notion of differentiability for the limit theorems developed in \Cref{sec: linearizations}.

\subsubsection{Hadamard Differentiability}\label{sec: hadamard diff main text}

We will proceed with proving Hadamard differentiability of the metric tensor $\mu \mapsto \tilde{\mathbf{g}}_{\mu}$ using the operator $S_\mu$. For a primer on notions of differentiability of maps between normed spaces, we refer the reader to \Cref{normed space differentiability} and \citet[Chapter 20]{van2000asymptotic}. We have deferred many proofs required for the results in this section to \Cref{sec: proofs for Sinkhorn riemannian geometry}. 

\medskip

\emph{The Statistical Tangent Space.} To establish Hadamard differentiability, we will first choose a suitable functional space and norm; in particular, we will follow the construction of \citet{goldfeld2024limit}. Fix an $m \in \mathbb{N}$, define $B^m = \{f \in C^m(X) : \|f\|_{C^m(X)} \leq 1\}$ and denote $\ell^\infty(B^m)$ as the space of bounded real-valued functions on $B^m$, equipped with the induced supremum norm
\[\ell^\infty(B^m) = \left\{z : \sup_{f \in B^m}|z(f)| < \infty\right\}.\]
The fact that $\ell^\infty(B^m) \neq C^m(X)^*$ is noteworthy, and follows from the requirement that $C^m(X)^*$ functionals are linear. For any signed measure $\gamma$, we can associate to it the functional $(f \mapsto \int_X f\,d\gamma)$ equipped with the same supremum norm. This construction therefore identifies $\mathcal{P}(X)$ with a subset of the normed space $(\ell^\infty(B^m), \|\cdot\|_{C^m(X)^*})$. Now fix a $\mu \in \mathcal{P}(X)$ and write \[\mathcal{P}_\mu = \{\nu \in \mathcal{P}(X): \operatorname{supp}(\nu) \subseteq \operatorname{supp}(\mu)\} \quad \text{and}\quad \mathfrak{M}_\mu = \{t(\nu - \mu) : \nu \in \mathcal{P}_\mu, t > 0\}.\]
Note that when $m > d/2$ the Sobolev embedding theorem \citep[Chapter 4]{adams2003sobolev} implies $H^m(X) \hookrightarrow C^0(X);$ thus, for every $\sigma\in \mathfrak{M}_\mu$ we have $|\sigma(f)| \leq \|\sigma\|_{\operatorname{TV}}\|f\|_\infty \lesssim \|\sigma\|_{\operatorname{TV}} \|f\|_{H^m(X)}$ for all $f \in H^m(X)$, therefore implying $\mathfrak{M}_\mu \subset H^m(X)^*$. With this in mind, we will define our statistical tangent space at $\mu \in \mathcal{P}(X)$, denoted with $\mathbb{T}_\mu^m$, to be the $H^m(X)^*$--closure of $\mathfrak{M}_\mu.$ This statistical tangent space is generally smaller than the full Riemannian tangent space under our assumptions on the ground cost $c$ (\Cref{spectral decay assumption}). Under a further positive-density assumption, however, it embeds continuously and densely into the Riemannian tangent space.

\begin{prop}[Statistical and Riemannian tangent spaces]
\label{statistical riemannian tangent embedding}
Suppose $X=\overline{\operatorname{int}(X)}$ is convex, $c\in C^{(m,m)}(X\times X)$, and $\mathcal H_c$ is norm-equivalent to $H^s(X)$ with $s>m$. Let $\mu=\rho\,dx\in\mathcal P(X)$ satisfy $\rho\geq\rho_{\min}>0$. Then $\mathbb T_\mu^m=H^m(X)^*_0$, while $T_\mu\mathcal P(X)\cong H^s(X)^*_0$ with equivalent norms. In particular, the inclusion $\mathbb T_\mu^m\hookrightarrow T_\mu\mathcal P(X)$ is continuous and has dense range.
\end{prop}

\begin{rmk}[On the distinction of the statistical and Riemannian tangent spaces]
    The distinction between the statistical perturbation space and the Riemannian tangent space arises naturally from the required regularity. In particular, the derivation of the Levi--Civita covariant derivative requires $\mathcal H_c$ to be closed under pointwise multiplication to ensure boundedness (see \Cref{sec: deriving the connection} and discussion therein). We therefore restrict to kernels whose RKHS is norm-equivalent to a finite-order Sobolev space $H^s(X)$ with $s>d/2$, so that $\mathcal H_c$ is a Banach algebra. At the same time, in order for a translation-invariant kernel with spectral decay $\widehat{\kappa}(\omega)\asymp(1+\|\omega\|^2)^{-s}$ to induce a cost $c=-\varepsilon\log\kappa$ of class $C^{(m,m)}$, we require $s>m+d/2$. Consequently, the classical smoothness order $m$ governing the $C^{-m}$ perturbation and statistical theory is necessarily lower than the Sobolev order $s$ governing the RKHS. This yields two corresponding tangent topologies: the statistical perturbation space is $\mathbb T_\mu^m=H^m(X)^*_0$, while the Riemannian tangent space is naturally identified with $H^s(X)^*_0$. 
\end{rmk}



\medskip

\emph{Differentiating $\mu \mapsto S_\mu$.} We will now prove that the map $S_{(\cdot)}: \mathcal{P}(X)\subset \ell^\infty(B^m) \to \mathcal{L}(\mathcal{H}_c, \mathcal{H}_c)$ given by $\mu \mapsto S_\mu$ is Hadamard differentiable at $\mu$ tangentially to $\mathbb{T}_\mu^m$. Note that when dealing with the product space $\ell^\infty(B^m) \times \ell^\infty(B^m)$ we equip it with the norm $\|\mu\|_{C^m(X)^*} \vee \|\nu\|_{C^m(X)^*}$ for $(\mu, \nu) \in \mathcal{P}(X) \times \mathcal{P}(X)$. With this tangent space definition in place, we are now ready to invoke the Hadamard differentiability of the Schr\"odinger potentials.


\begin{prop}[Hadamard differentiability of self-Schr\"odinger potentials] \label{self schrodinger hadamard diff}
    Fix a $\mu \in \mathcal{P}(X)$. The self-Schr\"odinger potential $f_{\mu, \mu}$, regarded as a map $\mathcal{P}(X) \subset \ell^\infty(B^m) \to C^m(X)$, is Hadamard differentiable at $\mu$ tangentially to $\mathbb{T}_\mu^m$ with derivative $ \gamma \mapsto \operatorname{D}_{\gamma}f_{\mu, \mu}$. 
\end{prop}

We provide the proof of \Cref{self schrodinger hadamard diff} in \Cref{proof of self schrodinger hadamard diff}. Because the operator $S_\mu$ depends upon the self-Schr\"odinger potential (among other objects), the above result is directly applicable to the differentiability of the constituent operators in \Cref{Smu original definition}, the operator $S_\mu$ and therefore the metric tensor. As a result, we prove Hadamard differentiability of $S_\mu$ and the metric tensor $\tilde{\mathbf{g}}_\mu$ in \Cref{metric hadamard diff}.


\begin{thm}[Hadamard differentiability of the metric] \label{metric hadamard diff}
    The map $S_{(\cdot)}: \mathcal{P}(X) \subset \ell^\infty(B^m) \to \mathcal{L}(\mathcal{H}_c, \mathcal{H}_c)$ given by $S_\mu = \operatorname{id}_{\mathcal{H}_c} + 2\mathbf{J}_{\mu}\mathbf{R}_\mu$ is Hadamard differentiable at $\mu$ tangentially to $\mathbb{T}_\mu^m$, with a Hadamard derivative given by 
    \[\operatorname{D}_{\gamma}S_{\mu} = 2 (\operatorname{D}_{\gamma }\mathbf{J}_{\mu}) \mathbf{R}_\mu + 2 \mathbf{J}_\mu (\operatorname{D}_{\gamma} \mathbf{R}_\mu)\]
    where $\gamma \mapsto \operatorname{D}_{\gamma}\mathbf{J}_\mu$ and $\gamma \mapsto \operatorname{D}_{\gamma}\mathbf{R}_\mu$ are the Hadamard derivatives of $\mathbf{J}_\mu$, $\mathbf{R}_\mu$ respectively. Moreover, the map $(\mu, v, w) \mapsto \tilde{\mathbf{g}}_\mu(v,w)$ regarded as a map from $\mathcal{P}(X) \times \mathcal{H}_c \times \mathcal{H}_c \to \mathbb{R}$ is Hadamard differentiable at $(\mu, v, w)$ tangentially to $(\mathbb{T}_\mu^m \times \mathcal{H}_c \times \mathcal{H}_c)$ with Hadamard derivative,
    \begin{align*}
        \operatorname{D}_{(\gamma, \dot v, \dot w)}\tilde{\mathbf{g}}_\mu(v,w) = \frac{\varepsilon}{2}\bigg[\big\langle (\operatorname{D}_\gamma S_\mu)v, w\big\rangle_{\mathcal{H}_c} + \big\langle S_\mu \dot v, w\big\rangle_{\mathcal{H}_c} + \big\langle S_\mu v, \dot w\big\rangle_{\mathcal{H}_c}\bigg].
    \end{align*}
\end{thm}
We provide the proof of \Cref{metric hadamard diff} in \Cref{proof of metric hadamard deriv}. We note that the proof of this result utilizes many auxiliary results. In particular, it requires differentiation of $\mu \mapsto \mathbf{J}_\mu$ and $\mu \mapsto \mathbf{R}_\mu$, each of which are obtained by compositions of several operators (see \Cref{reparameterization}). The proofs of all auxiliary results involved here can be found in \Cref{sec: operator regularity}, and follow from standard arguments in conjunction with the Hadamard differentiability of the self-Schr\"odinger potentials (\Cref{self schrodinger hadamard diff}).


Using the Hadamard differentiability of the map $(\mu, v, w) \mapsto \tilde{\mathbf{g}}_\mu(v,w)$, we can now consider the time derivative of the metric tensor along a curve of measures $(\mu_t)_{t \in [0,1]}.$ In particular, we will show that the needed regularity for time-differentiation of these operators follows from mild assumptions on the curve of measures $\mu_t$. From there, the time derivative of the metric tensor follows from another application of the Hadamard chain rule. We provide the proof in \Cref{proof of metric time deriv}.



\begin{thm}[Time derivative of the metric] \label{metric time deriv}
    For a curve of measures $\mu_t\in H^{1}([0,1];H^m(X)^*)$ satisfying the positive density criterion, it holds that for almost every $t\in[0,1]$,
    \[
        \frac{d}{dt}S_{\mu_t}
        =
        \dot S_{\mu_t}
        =
        \operatorname D_{\dot\mu_t}S_{\mu_t},
    \]
    where the time derivative is taken with respect to the
    $\mathcal L(\mathcal H_c,\mathcal H_c)$ topology. Moreover, if we let $v_t, w_t \in H^{1}([0,1]; \mathcal{H}_c)$ be two $\mu_t$-tangent curves (i.e. $v_t,w_t \in \mathfrak{T}_{\mu_t}\mathcal{P}(X)$ for all $t \in [0,1]$) then the metric tensor  $\tilde{\mathbf{g}}_{\mu_t}$ evaluated at $(v_t,w_t)$ is time differentiable for a.e. $t \in [0,1]$ with the derivative
    \[\frac{d}{dt} \tilde{\mathbf{g}}_{\mu_t}(v_t,w_t) = \frac{\varepsilon}{2}\bigg[\big\langle \dot{S}_{\mu_t}v_t, w_t\big\rangle_{\mathcal{H}_c} + \big\langle S_{\mu_t} \dot v_t, w_t\big\rangle_{\mathcal{H}_c} + \big\langle S_{\mu_t} v_t, \dot w_t\big\rangle_{\mathcal{H}_c}\bigg].\]
\end{thm}

\Cref{metric time deriv} establishes the needed differentiability to verify the metric compatibility of a candidate covariant differentiation operator $\nabla$. In the following section we will use this result extensively in constructing the Levi-Civita covariant derivative. But before we move on, we remark on a key aspect of our proof strategy. 


\new{We remark that the preceding time-differentiation result does not require Hadamard differentiability of $\mu\mapsto S_\mu$; suitable differentiability along curves of measures would suffice. Nevertheless, as previously noted, our decision to establish Hadamard differentiability is motivated by statistical inference, as it enables an application of the functional delta method to metric-based plugin estimators, particularly the logarithmic-map-based estimator of the squared Sinkhorn distance (\Cref{geometric clt}). The limit theorems for the remaining geometric operations require additional linearizations, developed in \Cref{sec: linearizations}.}

\subsection{Deriving the Covariant Derivative}\label{sec: deriving the connection}

Now we'll apply an analogous technique to that of \citet{gigli2012second} to derive a family of metric compatible covariant derivatives. Once we have said family, we will then identify the torsion-free element of the family. Just as a reminder, we restate all accumulated assumptions thus far.

\begin{asmpt}[Standing assumptions] \label{standing sinkhorn assumptions}
    Fix integers $m>d/2$ and $s>m+d/2$.
    Throughout the remainder of the manuscript,
    we impose the following assumptions on the
    domain and kernel, and on population measures
    whenever their positive-density condition is
    invoked.
    \begin{enumerate}
        \item The domain $X\subset\mathbb R^d$ is a closed Euclidean ball of positive radius.
        \item The cost $c:X\times X\to[0,\infty)$ is symmetric and belongs to $C^{(m,m)}(X\times X)$. Moreover, $k_c=\exp(-c/\varepsilon)$ is symmetric, positive definite and universal, and is the restriction to $X\times X$ of a strictly positive translation-invariant kernel $\kappa(x-y)$ satisfying $\widehat{\kappa}(\omega)\asymp(1+\|\omega\|_2^2)^{-s}$.
        \item We restrict attention to measures $\mu\in\mathcal P(X)$ that admit a Lebesgue density $\mu=\rho\,dx$ satisfying $\rho(x)\geq\rho_{\min}>0$ for every $x\in X$.
    \end{enumerate}
\end{asmpt}

Under \Cref{standing sinkhorn assumptions}, $\mathcal H_c$ is norm-equivalent to $H^s(X)$ and is therefore a Banach algebra under pointwise multiplication (see \Cref{preliminaries on RKHS}). Moreover, \Cref{statistical riemannian tangent embedding} gives
\[
    \mathbb T_\mu^m=H^m(X)^*_0\hookrightarrow T_\mu\mathcal P(X)\cong H^s(X)^*_0,
\]
where the inclusion is continuous and dense. Thus the Riemannian geometry uses the weaker negative-Sobolev topology $H^s(X)^*$, whereas the statistical differentiability theory is carried out on the dense subspace $H^m(X)^*$.

\begin{rmk}[On the standing assumptions]
The conditions in \Cref{standing sinkhorn assumptions} are intended as convenient sufficient conditions for the regularity theory developed below, rather than as necessary conditions for the existence of a Sinkhorn Riemannian structure. Assumptions 1--2 provide a particularly transparent setting in which the RKHS $\mathcal H_c$ can be identified, with equivalent norms, with the Sobolev space $H^s(X)$ and consequently inherits the algebra and embedding properties used throughout our arguments. Similar results should therefore extend to other domains and kernels whenever the corresponding RKHS enjoys the same properties. 

Assumption 3 is probably the strongest restriction. The uniform lower bound $\rho\geq\rho_{\min}>0$ ensures that sufficiently small two-sided zero-mass density perturbations remain probability measures, which facilitates the dense embedding $\mathbb T_\mu^m \hookrightarrow T_\mu\mathcal P(X)$ and the differentiation arguments used below. We do not expect such a lower bound, or even absolute continuity, to be intrinsically necessary for a more general Sinkhorn Riemannian regularity theory. Rather, singular or vanishing-density measures introduce boundary phenomena for which the admissible perturbation space may depend more delicately on $\mu$, and treating these cases would require additional analysis beyond the scope of the present work.
\end{rmk}

\subsubsection{Regular Curves}\label{sec: regular curves}

Defining a candidate covariant derivative $\nabla$ along a curve of measures $(\mu_t)$ first requires a suitable notion of regularity for $(\mu_t)$ and any tangent vector fields along it. 

\begin{lemma}[Bijectivity of $\operatorname{D}_{(\cdot)}B_{\mu}$] \label{change of vars bijectivity}
    The map $\operatorname{D}_{(\cdot)}B_{\mu}: T_\mu\mathcal{P}(X) \to \mathfrak{T}_\mu\mathcal{P}(X)$ is a bijection and admits an inverse \[\left(\operatorname{D}_{(\cdot)}B_{\mu}\right)^{-1}(w) = H_{\mu}^{-1}\left(\left(\operatorname{id}_{\mathcal{H}_\mu} + K_\mu\right)(a_\mu w)\right).\]
\end{lemma}

We provide the proof of \Cref{change of vars bijectivity} in \Cref{sec: proof of change of vars bijectivity}. Note that we identify the corresponding $\mathcal{H}_c$-level tangent velocity under the change of variables $\mu_t \mapsto B_{\mu_t} := \beta_t$ with $\dot \beta_t = \operatorname{D}_{\dot\mu_t}B_{\mu_t} \in \mathfrak{T}_{\mu_t}\mathcal{P}(X).$ Indeed, one could readily verify that under \Cref{standing sinkhorn assumptions}, the following commutative diagram holds,
\[
\begin{tikzcd}[
    column sep=large,
    row sep=large,
    ampersand replacement=\&
]
C^m(X)^*_0
    \arrow[r, "\operatorname{D}_{(\cdot)}B_\mu"]
    \arrow[d]
\&
(\operatorname{D}_{(\cdot)}B_\mu)\left[C^m(X)^*_0\right]
    \arrow[d]
\\
T_\mu\mathcal{P}(X)
    \arrow[r, "\operatorname{D}_{(\cdot)}B_\mu"', "\cong"]
\&
\mathfrak{T}_\mu\mathcal{P}(X)
\end{tikzcd}
\]
where the vertical arrows denote the inclusions into the respective completions. In particular,
\[
T_\mu\mathcal{P}(X)
=
\overline{C^m(X)_0^*}^{\|\cdot\|_{\mathbf g_\mu}},
\quad \text{and}\quad
\mathfrak{T}_\mu\mathcal{P}(X)
=
\overline{
\operatorname{D}_{(\cdot)}B_\mu[C^m(X)^*_0]
}^{\|\cdot\|_{\tilde{\mathbf g}_\mu}}.\]
Indeed, these two parameterizations of the tangent space are isomorphic as $T_\mu \mathcal{P}(X) \cong H^s(X)_0^*$ (due to \Cref{statistical riemannian tangent embedding}) while $\mathfrak{T}_\mu \mathcal{P}(X) = B_\mu^\perp \cong H^s(X)_0^*$ (see \Cref{preliminaries on RKHS}).

\begin{defn}[First-order Sinkhorn regular curves] \label{sinkhorn regular curves}
A curve $\mu:[0,1]\to\mathcal P(X)$ is called \emph{first-order Sinkhorn regular} if
$\mu\in H^1([0,1];H^m(X)^*)$.
\end{defn}

Under the positive-density condition in \Cref{standing sinkhorn assumptions}, the velocity of a first-order Sinkhorn regular curve satisfies
$\dot\mu_t\in H^m(X)^*_0=\mathbb T_{\mu_t}^m$ for almost every $t\in[0,1]$. Since $s>m$, the continuous embedding
$H^m(X)^*\hookrightarrow H^s(X)^*$ implies that every first-order Sinkhorn regular curve is also an $H^1([0,1];H^s(X)^*)$-valued curve. Consequently, its image under the change of variables satisfies
$B_{\mu}\in H^1([0,1];\mathcal H_c)$. We prove this fact rigorously in \Cref{sinkhorn regular implies admissible}. We warn, however, that the converse implication need not hold. Indeed, admissibility only requires
$B_{\mu}\in H^1([0,1];\mathcal H_c)$, which under the identification $\mathcal H_c\simeq H^s(X)$ corresponds to regularity in the weaker $H^s(X)^*$ topology. Since $s>m$, our first-order Sinkhorn regularity condition therefore restricts attention to a stronger class of curves.





To define covariant differentiation, we also need the tangent velocity fields along a curve of measures $(\mu_t)$ to be sufficiently regular. This motivates the following definition.

\begin{defn}[Sinkhorn regular tangent field]
    Let $(\mu_t)_{t\in [0,1]}, (\beta_t)_{t \in [0,1]}$ be a Sinkhorn regular curve (\Cref{sinkhorn regular curves}). A map $v:[0,1]\to\mathcal{H}_c$ is called a \emph{Sinkhorn regular tangent field} along
    $(\mu_t)_{t\in [0,1]}$ if $v_t\in H^1([0,1];\mathcal{H}_c)$ and $v_t\in\mathfrak{T}_{\mu_t}\mathcal{P}(X)$ for a.e. $t\in [0,1]$.
\end{defn}

We will see that Sinkhorn regular tangent fields have enough regularity for differentiation along sufficiently regular curves. Finally, to reason about \emph{geodesic paths} on the space of probability measures we will need a second order notion of differentiability as well. We therefore define second order Sinkhorn regular curves to be those Sinkhorn regular curves with tangent velocities that are differentiable in this sense.

\begin{defn}[Second-order Sinkhorn regular curves]
    A curve
    $(\mu_t)_{t\in [0,1]}\subset\mathcal{P}(X)$ is called a 
    \emph{second-order Sinkhorn regular curve} if
    \begin{enumerate}
        \item it is first-order Sinkhorn regular;
        \item its $\mathcal{H}_c$-level velocity $\dot \beta_t = \operatorname{D}_{\dot \mu_t}B_{\mu_t}$ is a Sinkhorn regular tangent field. 
    \end{enumerate}
    Unless otherwise specified, we take \say{Sinkhorn regular} to implicitly mean \say{first-order Sinkhorn regular.}
\end{defn}

With these rigorous notions of curvewise and tangent field regularity, we can now formalize the construction of covariant differentiation. 

\subsubsection{The Covariant Derivative}

Following the curvewise formulation of \citet[Chapter 5]{gigli2012second},
we regard a covariant derivative along a Sinkhorn-regular curve as an
$\mathbb{R}$-linear operator on regular tangent fields satisfying the
Leibniz rule with respect to scalar functions of time. In the classical
smooth manifold setting, such an operator is induced by an affine
connection and additionally satisfies the compatibility property of
\citet[Theorem 4.24]{lee2018introduction}; since we do not assume an
a priori smooth affine connection on $\mathcal{P}(X)$, we instead construct
the curvewise derivative directly.

\begin{thm}[Metric-compatible covariant derivative family]
\label{metric compatible family}
Let $v_t \in H^{1}([0,1]; \mathcal{H}_c)$ be a Sinkhorn regular tangent vector field along a curve of measures $(\mu_t)_{t \in [0,1]}$, and suppose \Cref{standing sinkhorn assumptions} holds. If $\Omega_{\mu_t} \in \mathcal{L}(\mathfrak{T}_{\mu_t}\mathcal{P}(X), \mathfrak{T}_{\mu_t}\mathcal{P}(X))$ satisfies $\langle S_{\mu_t}\Omega_{\mu_t}v, w\rangle_{\mathcal{H}_c}+\langle S_{\mu_t}v, \Omega_{\mu_t}w\rangle_{\mathcal{H}_c}=0$ for all $v_t,w_t \in \mathfrak{T}_{\mu_t}\mathcal{P}(X)$ then 
\[\nabla^{\Omega}_{\dot\beta_t}v_t
=
\Pi_{\mu_t}\left(
\dot v_t
+
\left(\frac{1}{2}S_{\mu_t}^{-1}\dot S_{\mu_t}
+
\Omega_{\mu_t} 
\right)v_t\right),
\]
defines a $\tilde{\mathbf{g}}_{\mu_t}$-metric compatible covariant derivative.
\end{thm}

Note that we identify objects indexed by $\dot \beta_t$ and $\dot \mu_t$ since $\dot \beta_t$ and $\dot \mu_t$ refer to two parameterizations of the same tangent, as discussed in \Cref{sec: regular curves}. We provide the proof of \Cref{metric compatible family} in \Cref{sec: metric compatible family}, wherein we verify that all curvewise covariant derivatives of the form stated above indeed satisfy the metric compatibility criterion,
\[\frac{d}{dt}\tilde{\mathbf{g}}_{\mu_t}(v_t, w_t) =  \tilde{\mathbf{g}}_{\mu_t}(\nabla_{\dot \mu_t}^{\Omega}v_t, w_t) + \tilde{\mathbf{g}}_{\mu_t}(v_t, \nabla_{\dot \mu_t}^{\Omega}w_t).\]
Now that we have verified metric compatibility for a family of covariant derivatives, it remains to identify a torsion-free one. By \Cref{agreement of base derivatives}, the derivative $\gamma\mapsto\operatorname D_\gamma S_\mu$ extends uniquely and continuously from the statistical tangent space $\mathbb T_\mu^m=H^m(X)^*_0$ to the full Riemannian tangent space $T_\mu\mathcal P(X)\cong H^s(X)^*_0$. Accordingly, for $w\in\mathfrak T_\mu\mathcal P(X)$ and $\sigma_w=(\operatorname D_{(\cdot)}B_\mu)^{-1}(w)\in T_\mu\mathcal P(X)$, we interpret $\operatorname D_{\sigma_w}S_\mu$ as this extended derivative, since $\sigma_w$ need not belong to the smaller statistical tangent space $\mathbb T_\mu^m$. Fix a $\mu\in\mathcal P(X)$, and for every $u,v\in \mathfrak{T}_\mu\mathcal P(X)$, define the functional $C_\mu(u,v):\mathfrak{T}_\mu\mathcal P(X)\to\mathbb R$
given by
\[
    C_\mu(u,v)(w)
    :=
    \left\langle
    (\operatorname D_{\sigma_w}S_\mu)u,v
    \right\rangle_{\mathcal H_c},
    \qquad
    \sigma_w
    :=
    \left(
    \operatorname D_{(\cdot)}B_\mu
    \right)^{-1}(w).
\]
Note that \Cref{Cmu riesz rep} shows that $C_\mu(u,v)$ admits a Riesz representative $C_\mu(u,v)^\sharp$ in $\mathfrak{T}_\mu\mathcal{P}(X)$ for every $(\mu, u,v) \in \mathcal{P}(X) \times \mathfrak{T}_\mu \mathcal{P}(X)\times \mathfrak{T}_\mu \mathcal{P}(X)$,
\[C_\mu(u,v)^\sharp = 2\left( a_\mu (\mathbf{R}_\mu u)(\mathbf{R}_\mu v) - H_c\!\left[ a_\mu^3 (\mathbf{R}_\mu u)(\mathbf{R}_\mu v)\mu \right] \right) \in \mathcal{H}_c.\]
Herein lies the need for the finite smoothness of the ground cost $c$. As we discuss in \Cref{preliminaries on RKHS}, infinitely smooth costs like the squared Euclidean distance induce an RKHS that is not closed under pointwise multiplication. In such a setting, the functional $C_\mu(u,v)$ would not be bounded, and therefore it would not admit a Riesz representative. This can also be seen in the expression for $C_\mu(u,v)^\sharp$ above, as it contains the pointwise product of many $\mathcal{H}_c$ elements. A lack of closure under pointwise multiplication would therefore violate the membership in $\mathcal{H}_c$. In this case, one would have to define the Levi-Civita covariant derivative weakly; in particular one would have to define it as a functional and characterize it by testing against every element of $\mathfrak{T}_\mu \mathcal{P}(X).$

\begin{rmk}[Trading off smoothness and geometry]
    One interpretation of this phenomenon is as follows: when one chooses an infinitely smooth cost, the ambient RKHS $\mathcal{H}_c$ becomes so smooth and restrictive that it may no longer be closed under the action of second-order objects. In this sense, the ambient space is sufficiently rich to support the first-order geometry, but may be too small to realize the second-order geometry classically. Choosing a cost $c$ according to \Cref{standing sinkhorn assumptions} remedies this by ensuring that the induced RKHS is sufficiently large to retain the required second order objects. We emphasize that the cost induced by the widely used Mat\'ern kernel both has this property, and approximates the squared Euclidean cost arbitrarily well (see \Cref{preliminaries on RKHS})
\end{rmk} 


\new{The preceding Riesz representation allows us to express the correction required by torsion-freeness as an element of the Riemannian tangent space. We can therefore identify the Levi-Civita covariant derivative within the family of metric-compatible connections constructed above.}

\begin{thm}[The Levi-Civita covariant derivative] \label{levi civita}
    Let $(\mu_t)_{t\in [0,1]}$ be a Sinkhorn regular curve and let
    $(v_t)_{t\in [0,1]}$ be a Sinkhorn regular tangent field along
    $(\mu_t)_{t\in [0,1]}$. For every $u\in \mathfrak{T}_\mu\mathcal P(X)$, define the functional $\Omega: \mathfrak{T}_\mu\mathcal P(X) \times \mathfrak{T}_\mu\mathcal P(X)\to \mathfrak{T}_\mu\mathcal P(X)$ given by 
    \[
        \Omega_{(u)}v
        =
        \frac12
        \Pi_\mu S_{\mu}^{-1}\left(
        (\operatorname D_{\sigma_v}S_\mu)u
        -
        C_\mu(u,v)^\sharp
        \right), \quad \sigma_v
        =
        \left(
        \operatorname D_{(\cdot)}B_\mu
        \right)^{-1}(v).
    \] 
    Then under \Cref{standing sinkhorn assumptions}, we have that
    \begin{align}\label{sinkhorn covariant deriv}
    \nabla^{\operatorname{LC}}_{
        \dot \beta_t}
        v_t
        =
        \Pi_{\mu_t}
        \bigg(
        \dot v_t
        +
        \frac12
        S_{\mu_t}^{-1}\dot S_{\mu_t}v_t
        +
        \Omega_{\dot \beta_t}
        v_t
        \bigg) \tag{LC}
    \end{align}
    is a metric-compatible and  torsion-free covariant derivative, and it therefore defines a Levi-Civita
    covariant derivative associated with $\tilde{\mathbf g}$ on $\mathfrak{T}_\mu\mathcal P(X)$. 
\end{thm}
We provide the proof of \Cref{levi civita} in \Cref{sec: proof of levi civita}.  For shorthand, we will occasionally write
\begin{equation}\label{gamma expression} \Gamma_{\mu}(u,v) = \Pi_{\mu}\left(\frac{1}{2}S_{\mu}^{-1}(\operatorname D_{\sigma_u}S_\mu)v + \Omega_u v \right)
\end{equation}
where $u,v\in\mathfrak T_\mu\mathcal P(X)$ and $\sigma_u=(\operatorname D_{(\cdot)}B_\mu)^{-1}(u)$, and we identify $\Gamma_\beta(u,v)$ with this expression when $\beta=B_\mu$. We will briefly remark on the similarities and differences between the Levi-Civita covariant derivative arising in the Sinkhorn Riemannian geometry and that of Wasserstein geometry. Recall, for a curve of measures $(\mu_t)_t$ on a boundaryless, complete and $C^\infty$ Riemannian manifold $(\man, g)$ driven by a tangent velocity field $\nabla\varphi_t$ (i.e. $\dot \mu_t = -\operatorname{div}_g(\mu_t\nabla \varphi_t)$) for some $\varphi \in C^\infty_c(\man \times [0,1])$, the Levi-Civita covariant derivative \citep[Remark 5.3]{gigli2012second} is given by

\begin{equation}\label{W2 covariant deriv}
\nabla^{W_2}_{\dot \mu_t}v_t = P_{\mu_t}\left(\dot v_t + \nabla^M_{(\nabla \varphi_t)}v_t\right)
\end{equation}
where $P_{\mu_t}$ is the $L^2(\mu_t)$-orthogonal projection onto $\overline{\{\nabla \varphi : \varphi \in C^\infty_c(\man)\}}^{L^2(\mu_t)}.$ This projection requires solving a weighted elliptic PDE \citep[Equation 2.6]{lavenant2019courbes}, and it is therefore nontrivial in both a computational and statistical sense (see \citet{saidi2026wassersteinparalleltransportpredicting} for a proposed estimator in the $\man = \mathbb R^d$ case). Unlike this orthogonal projection, the one required in the Sinkhorn covariant derivative is simply given by $v \mapsto v - \langle v, B_{\mu_t}\rangle_{\mathcal{H}_c}B_{\mu_t}$ (\Cref{Pimu g orthogonal projection}). Since both $v$ and $B_{\mu_t}$ are elements of the fixed RKHS $\mathcal H_c$, this particularly simple form suggests that the Sinkhorn covariant derivative, and the geometric objects derived from it, may be amenable to both computational and statistical treatment. As we show in \Cref{sec: linearizations}, the latter is indeed the case.

While \Cref{W2 covariant deriv} and \eqref{sinkhorn covariant deriv} are quite distinct, they share the same basic structure. In both cases, the covariant derivative is obtained by taking an ambient time derivative $\dot v_t$, adding a correction term that accounts for the variation of the tangent geometry along the base curve, and projecting the resulting ambient vector back onto the tangent space at $\mu_t$. This is directly analogous to covariant differentiation in finite-dimensional Riemannian geometry. Indeed, in local coordinates the covariant derivative associated with an affine connection satisfies
\[
    (\nabla_XY)_k
    =
    \sum_{i = 1}^dX_i \frac{\partial Y_k}{\partial x_i}
    +
    \sum_{i, j = 1}^d\Gamma_{ij}^k X_iY_j
\]
\citep[Proposition~4.6]{lee2018introduction} for vector fields $X$ and $Y$. Note that the first term differentiates the vector field in the ambient (local) coordinates and the second term corrects for the variation of the tangent spaces (where $\Gamma_{ij}^k$ are the Christoffel symbols associated to the connection $\nabla$). In \Cref{W2 covariant deriv}, this correction is given by the covariant derivative of the vector fields on the base manifold $\nabla^M_{\nabla\varphi_t}v_t$, whereas in the Sinkhorn geometry it is encoded by $\smash{\frac12 S_{\mu_t}^{-1}\dot S_{\mu_t}v_t
    +
    \Omega_{\dot\beta_t}v_t.}$


\subsection{The Geodesic Equations and Parallel Transport} \label{sec: geodesic equations and pt}

With the Levi-Civita covariant derivative in hand, we obtain intrinsic notions of geodesics and parallel transport. In particular, geodesics are characterized by vanishing covariant acceleration, while parallel tangent fields are characterized by vanishing covariant derivative along a curve \citep[Chapter 4]{lee2018introduction}. Specializing these definitions to the Sinkhorn Riemannian geometry yields the following characterization.

\begin{thm}[The geodesic equations] \label{geodesic characterization}
    Let $(\mu_t)_{t \in [0,1]}$ be a second-order Sinkhorn regular curve of measures with tangent velocity $\dot \beta_t = \operatorname{D}_{\dot \mu_t}B_{\mu_t}$. Then $(\mu_t)_{t \in [0,1]}$ is a Sinkhorn geodesic if and only if for a.e. $t \in [0,1]$ we have
    \begin{align}\label{geodesic equations}
        \Pi_{\mu_t}\left(\ddot \beta_t
        +
        \frac12
        S_{\mu_t}^{-1}\dot S_{\mu_t}\dot{\beta}_t
        +
        \Omega_{\dot \beta_t}
        \dot{\beta}_t\right) = 0. \tag{GE}
    \end{align}
\end{thm}

\begin{figure}[!htb]
    \centering
    \includegraphics[width=\textwidth]{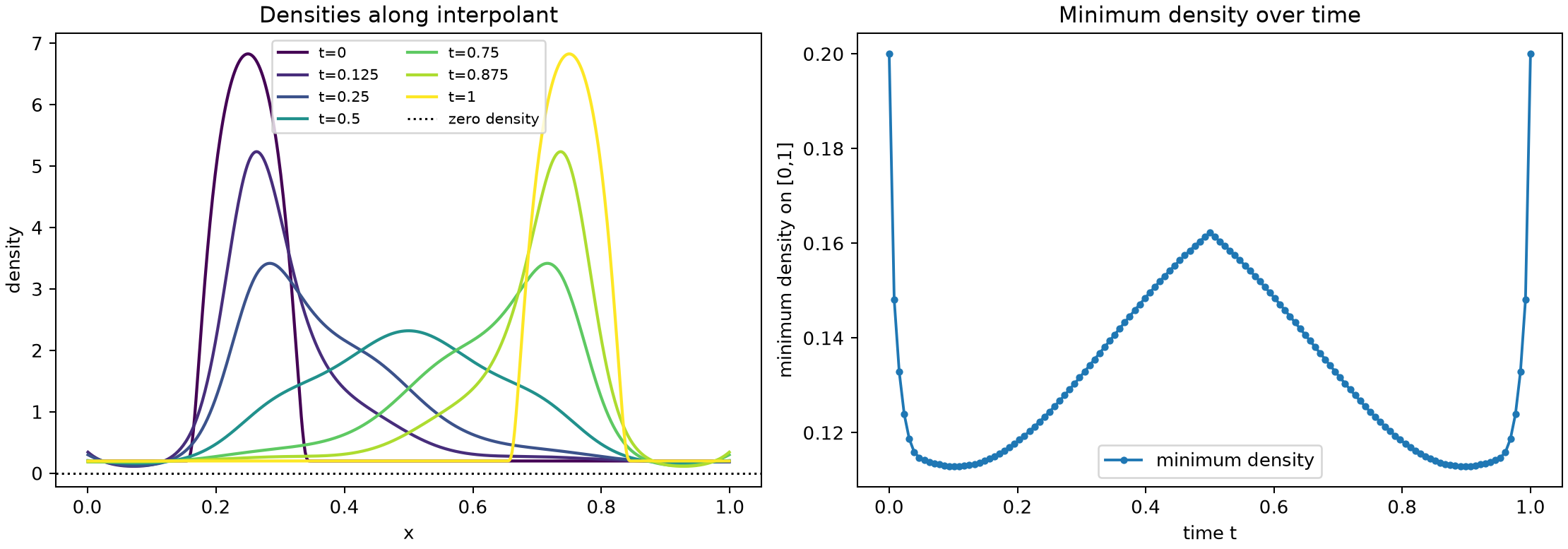}
    \caption{Eulerian chain approximation of a Sinkhorn geodesic on $X = [0,1]$, computed using the numerical scheme of \citet{lavenant2026approximation}. The endpoint densities are $\rho_0(x)=0.2+0.8\eta(x-0.25)$ and $\rho_1(x)=0.2+0.8\eta(x-0.75)$, where $\eta$ is a normalized smooth bump supported on $[-0.1,0.1]$. The approximation uses a Mat\'ern-$5/2$ kernel, $\varepsilon=0.02$, $500$ spatial cells, and $128$ temporal steps. Left: densities at selected interpolation times, with the dotted line indicating zero. Right: minimum density over the spatial grid as a function of time. We note that the intermediate densities remain strictly positive throughout the discretized interpolation.}
    
    \label{fig:eulerian-interpolation}
\end{figure}

\new{The proof follows immediately from \Cref{levi civita}. Importantly, \eqref{geodesic equations} characterizes geodesics through their covariant acceleration, but does not itself enforce the nonnegativity of the underlying measures or guarantee that its solutions minimize the Riemannian action. In contrast, \citet{lavenant2024riemannian} define geodesics variationally as action-minimizing curves of probability measures, with the associated Riemannian distance given by
\begin{equation}\label{variational geodesic}
    d^2_S(\mu_0, \mu_1) = \inf\left\{\int_0^1 \tilde{\mathbf{g}}_{\mu_t}(\dot \beta_t, \dot \beta_t)\, dt : \beta_t \in H^1((0,1); \mathcal{H}_c), \,\mu_{t=0}=\mu_0, \mu_{t=1}=\mu_1\right\}
\end{equation}
where $\beta_t = B_{\mu_t}$. They establish the existence of a minimizer through the direct method of the calculus of variations (Theorem 5.4).

The differential and variational characterizations need not coincide in general, since the nonnegativity constraint on the underlying measures may be active along a variational minimizer. Nevertheless, when a minimizer admits a uniformly positive density, sufficiently small smooth mass-preserving variations remain admissible. In this case, the weak Euler--Lagrange equations yield the unconstrained geodesic equations, as established by the following result. We emphasize that the uniform density lower bound is imposed on the minimizing interpolant itself and is not known to follow from the corresponding endpoint conditions. 
The numerical example in \Cref{fig:eulerian-interpolation} provides evidence that Sinkhorn geodesics may retain uniformly positive densities when the endpoints satisfy that property. Although the endpoint density lower bound of $0.2$ is not preserved, the minimum density along the discretized interpolant is approximately $0.113$.}

\begin{prop}[Variational minimizers satisfy the geodesic equations]\label{geodesic coincidence}
Fix $\mu_0,\mu_1$ satisfying \Cref{standing sinkhorn assumptions}.3 and let $(\mu_t^\star)_{t\in[0,1]}$ be any minimizer of \Cref{variational geodesic}. Suppose moreover that $\mu_t^\star=\rho_t^\star\,dx$ and
\[
    \inf_{t\in[0,1]}\inf_{x\in X}\rho_t^\star(x)
    \geq \rho_{\min}>0.
\]
Then \textup{(a)} $(\mu_t^\star)_{t\in[0,1]}$ is second-order Sinkhorn regular and \textup{(b)} it satisfies \eqref{geodesic equations} for a.e. $t\in[0,1]$. In particular, it is a Sinkhorn geodesic.
\end{prop}

The proof, provided in \Cref{proof of geodesic second order regularity,proof of geodesic coincidence}, uses the weak Euler--Lagrange equation \citep[Chapter 3.1]{rindler2018calculus} to establish both the regularity of the minimizer and its satisfaction of the geodesic equations.

The Levi-Civita covariant derivative also directly yields equations governing parallel transport in the Sinkhorn Riemannian geometry. Parallel transport in related geometries, including Wasserstein geometry on probability measures and Hellinger--Kantorovich geometry on nonnegative Radon measures, has recently been studied for applications in machine learning and causal inference \citep{saidi2026wassersteinparalleltransportpredicting, saidi2026differentialgeometricequivalencehellingerkantorovichconewasserstein}. These developments suggest that a computationally tractable notion of Sinkhorn parallel transport may likewise provide a useful tool for transferring tangent information between probability distributions.

\begin{thm}[The parallel transport equations]
    Let $(\mu_t)_{t \in [0,1]}$ be a Sinkhorn regular curve of measures with tangent velocity $\dot \beta_t = \operatorname{D}_{\dot \mu_t}B_{\mu_t},$
    and let $(v_t)_{t \in [0,1]}$ be a Sinkhorn regular tangent field along $(\mu_t)_{t \in [0,1]}$. Then $v_t$ is parallel along $(\mu_t)_{t \in [0,1]}$ if and only if for a.e. $t \in [0,1]$ we have
    \begin{align}\label{PT IVP}
        \Pi_{\mu_t}\left(\dot v_t
        +
        \frac12
        S_{\mu_t}^{-1}\dot S_{\mu_t}v_t
        +
        \Omega_{\dot\beta_t}v_t\right)
        =0. \tag{PT}
    \end{align}
    In particular, given $v_0\in\mathfrak{T}_{\mu_0}\mathcal{P}(X)$, the parallel transport of $v_0$ along $(\mu_t)_{t\in[0,1]}$ is characterized by the above equation together with the initial condition $v_t\big|_{t=0}=v_0$.
\end{thm}

Worth addressing are the existence and uniqueness properties of solutions to \eqref{geodesic equations} and \eqref{PT IVP}. To address this, \Cref{well posedness pt and geodesics} establishes that the coefficients of the ODEs defining the geodesic and parallel transport equations are smooth enough to guarantee the existence and uniqueness of local solutions in general (under the regularity assumptions of \Cref{standing sinkhorn assumptions}). We provide the proof of the following result in \Cref{proof of well posedness pt and geodesics}.

\begin{prop}[Local existence and uniqueness of geodesics and parallel transport]
\label{well posedness pt and geodesics}
    Suppose \Cref{standing sinkhorn assumptions} holds. Then for every
    $\mu\in\mathcal P(X)$ and $v\in\mathfrak T_\mu\mathcal P(X)$,
    there exists $\tau>0$ such that the geodesic initial-value problem
    with $\beta_0=B_\mu$ and $\dot\beta_0=v$ admits a unique solution on
    $(-\tau,\tau)$. Moreover, let $(\mu_t)_{t\in[0,T]}$ be a
    second-order Sinkhorn-regular curve. Then for every
    $t_0\in[0,T]$ and every
    $w_0\in\mathfrak T_{\mu_{t_0}}\mathcal P(X)$, there exists a
    neighborhood $I$ of $t_0$ and a unique parallel tangent field
    $(w_t)_{t\in I}$ satisfying $w_{t_0}=w_0$.
\end{prop}


\begin{rmk}[Ambient geodesics] \label{ambient geodesics}
\Cref{well posedness pt and geodesics} establishes local existence and uniqueness for the geodesic ODE, but we emphasize that it does not assert that the resulting curve $(\beta_t)$ remains in $B[\mathcal P(X)]$. Indeed, the inverse change of variables $M(\beta_t)=\beta_t H_c^{-1}(\beta_t)$ is valued in $H^s(X)^*$ (\Cref{inverse change of vars}); although $M(\beta_t)(\mathbbm{1}_X)=\|\beta_t\|_{\mathcal H_c}^2=1$, it need not be a nonnegative measure. To sidestep this, \Cref{geodesic coincidence} starts with a probability-measure-valued variational minimizer and shows, under its stated assumptions, that its image under $B$ satisfies the geodesic ODE. \new{In this sense, the equations derived here provide a local extension of the infinitesimal Sinkhorn geometry to the ambient RKHS, agreeing with the intrinsic Riemannian operations along sufficiently regular probability-measure-valued curves without requiring that their solutions remain probability-measure-valued.} 
\end{rmk}

Having remarked on this potential nonidentification of $\beta_t$ with a curve of probability measures, we can now move on to further interpretation of the result. In particular, the geodesic existence and uniqueness result restricts to measures satisfying \Cref{standing sinkhorn assumptions} (i.e. absolutely continuous measures with densities bounded away from zero), but imposes no further restrictions on the tangent elements. The parallel transport result also guarantees local existence and uniqueness, but the base curve must be second-order Sinkhorn-regular. To contextualize this discussion, we will briefly compare with analogous existence and uniqueness of geodesics and parallel transport results from the unregularized optimal transport literature. \new{We also provide a tabular description of the analogous equations in \Cref{tab: sinkhorn wasserstein comparison}. }

\begin{rmk}[Comparison with Wasserstein geometry]
    The well-posedness conditions above differ notably from those appearing in the unregularized Wasserstein geometry. In particular, \citet[Definition~2.7]{gigli2012second} calls an absolutely continuous curve $(\mu_t)_{t\in[0,1]}$ regular when its tangent velocity field $(v_t)_{t\in[0,1]}$ satisfies
    \[
        \int_0^1 \|v_t\|_{L^2(\mu_t)}^2\,dt < \infty
        \qquad\text{and}\qquad
        \int_0^1 L(v_t)\,dt < \infty,
    \]
    where $L(v_t)$ denotes the spatial Lipschitz constant of $v_t$. The latter condition is the essential additional regularity requirement used to control the variation of the spaces $L^2(\mu_t)$ and the associated Wasserstein tangent spaces; parallel transport is then constructed along curves satisfying this condition. By contrast, in the Sinkhorn geometry the change of variables $B_\mu$ realizes the tangent spaces inside the fixed RKHS $\mathcal H_c$, and the projection and tangent spaces vary smoothly with the base point. Consequently, our well-posedness argument requires no analogous spatial Lipschitz condition on the tangent velocity; second-order Sinkhorn regularity instead supplies sufficient temporal regularity of the RKHS-valued curve for the parallel transport equation to be treated as a Banach-space ODE.

    A related distinction appears for the geodesic initial-value problem. While minimizing Wasserstein geodesics between prescribed endpoints exist under very general assumptions, obtaining a geodesic from a prescribed initial tangent requires additional care. For example, \citet[Proposition~6.1]{gigli2012second} shows that, for $\mu\in\mathcal P_c(M)$ and $\varphi\in C_c^\infty(M)$, the curve
    $t\mapsto(\exp(t\nabla\varphi))_\#\mu$ is a regular Wasserstein geodesic for all sufficiently small $t$, with $\sup_t L(v_t)<\infty$. Thus the existence is established there for sufficiently spatially smooth initial directions. In contrast, \Cref{well posedness pt and geodesics} yields a unique local Sinkhorn geodesic for every $v\in\mathfrak T_\mu\mathcal P(X)$ under \Cref{standing sinkhorn assumptions}. In this sense, the RKHS representation restores more conventional geodesic initial-value properties (see Theorem 4.27 in \citet{lee2018introduction}, for example), although the regularity assumptions in the two settings should not be regarded as strictly ordered. In particular, Wasserstein geometry imposes spatial regularity on the velocity field, whereas the Sinkhorn Riemannian geometry places the corresponding regularity on the temporal evolution of RKHS-valued objects. 
\end{rmk}

\begin{table}[t]
\centering
\small
\setlength{\tabcolsep}{5pt}
\renewcommand{\arraystretch}{1.25}
\begin{tabular}{@{}m{0.17\linewidth}C{0.385\linewidth}C{0.385\linewidth}@{}}
\toprule
& \textbf{Sinkhorn} & \textbf{Wasserstein} \\
\midrule
\textbf{Geodesics}
& $\displaystyle
\begin{gathered}
\Pi_{\mu_t}\Bigl(\ddot\beta_t+
\tfrac12 S_{\mu_t}^{-1}\dot S_{\mu_t}\dot\beta_t+\Omega_{\dot\beta_t}\dot\beta_t\Bigr)=0
\end{gathered}$
\par\smallskip \emph{Second-order ODE in $\mathcal H_c$.}
& $\displaystyle
\begin{aligned}
\partial_t\mu_t+\nabla\!\cdot(\mu_t v_t)&=0,\\
\partial_t v_t+\nabla_{v_t}v_t&=0
\end{aligned}$
\par\smallskip \emph{First-order PDE system.}
\\[1em] \addlinespace[0.5em] 

\begin{tabular}{@{}l@{}} \textbf{Parallel} \\ \textbf{transport} \end{tabular}
& $\displaystyle
\begin{gathered}
\Pi_{\mu_t}\Bigl(\dot u_t+
\tfrac12 S_{\mu_t}^{-1}\dot S_{\mu_t}u_t
+\Omega_{\dot\beta_t}u_t\Bigr)=0
\end{gathered}$
\par\smallskip \emph{First-order linear ODE in $\mathcal H_c$.}
& $\displaystyle
\begin{gathered}
P_{\mu_t}\Bigl(\partial_t w_t+
\nabla_{v_t}w_t\Bigr)=0
\end{gathered}$
\par\smallskip \emph{First-order PDE.}
\\
\bottomrule
\end{tabular}
\caption{Comparison of geodesic and parallel transport equations in Sinkhorn and Wasserstein geometry. For the Wasserstein equations listed, see \citet{gigli2012second} or \citet{ambrosio2008parallel}.}
\label{tab: sinkhorn wasserstein comparison}
\end{table}

\section{Limit Theorems via Linearization} \label{sec: linearizations}

\new{While \Cref{sec: Sinkhorn riemannian geometry} was dedicated to the derivation of the Levi-Civita covariant derivative, the geodesic equations and parallel transport, the principal goal of this section is to establish $\sqrt{n}$-Gaussian limit theorems for the plugin estimators of the corresponding geometric operations. In particular, we obtain limit theorems for geodesic interpolants, the Riemannian logarithmic map, the Sinkhorn Riemannian distance (all contained in \Cref{geometric clt}), and the parallel imputation map (\Cref{parallel imputation clt}), which has found applications in causal inference and machine learning \citep{saidi2026wassersteinparalleltransportpredicting} (\Cref{sec: clts}). To establish these results, we build upon the differentiability theory of the preceding section by constructing smooth extensions of the geodesic and parallel transport equations (\Cref{sec: extensions and smoothness}), allowing us to linearize their solution maps using Banach space ODE theory (\Cref{linearizing expmap} \Cref{linearizing logmap} and \Cref{linearizing imp}). Together with a separate Hadamard differentiability argument for the variationally defined distance (\Cref{sinkhorn distance hadamard diff}), these linearizations yield the desired limit theorems through the functional delta method.}

\medskip

\emph{Empirical Process Theory.} We will briefly recall some relevant notation and definitions from \citet{vanderVaart1996weak} that will be of use in this section. Denote the empirical measure associated to $n$ i.i.d. samples $X_1, \dots, X_n$ from a measure $\mu$ as $\hat \mu_n := \frac{1}{n}\sum_{i = 1}^n \delta_{X_i}$ where $\delta_{X_i}$ is a Dirac measure at $X_i$. Given a function class $\mathcal{F}$ of measurable functions from $X \to \mathbb{R}$, a measure $\mu$ on $X$ induces a map from $\mathcal{F}$ to $\mathbb{R}$ given by $f \mapsto \int_Xf\, d\mu := \mu(f)$. Thus, for a fixed function class $\mathcal{F}$ we can define a normed space $\ell^\infty(\mathcal{F})$ as the set of functionals on $\mathcal{F}$ equipped with the supremum norm $\|\mu\|_{\ell^\infty(\mathcal{F})} = \sup_{f \in \mathcal{F}}|\mu(f)|.$

For a probability measure $\mu$, we can define the $\mathcal{F}$-indexed empirical process $(\mathbb{G}_n(f))_{f \in \mathcal F}$ where $\mathbb{G}_n$ is the  (random) functional on $\mathcal{F}$ given by $\mathbb{G}_n f := \sqrt{n}(\hat \mu_n - \mu)(f) = \frac{1}{\sqrt{n}}\sum_{i = 1}^n(f(X_i) - \mu(f))$ where $X_1, \dots, X_n$ are i.i.d. draws from $\mu$. Empirical process theory is concerned with the conditions on $\mathcal{F}$ under which the random functional $\mathbb{G}_n \in \ell^\infty(\mathcal{F})$ converges weakly to a limiting random functional $\mathbb{G}_\mu$ (also known as a $\mu$-Brownian bridge) in the normed space $\ell^\infty(\mathcal{F}).$ Recall that weak convergence in any space is tantamount to the requirement that $\mathbb{E}[\Phi(\mathbb{G}_n)] \to \mathbb{E}[\Phi(\mathbb{G}_\mu)]$ (modulo measurability qualifications) for every bounded and continuous functional $\Phi : \ell^\infty(\mathcal{F}) \to \mathbb{R}$. If we take $\mathcal{F} = B^m = \{f\in C^m(X) : \|f\|_{C^m(X)} \leq 1\}$ from \Cref{sec: hadamard diff main text}, then when $m > d/2$ it holds that $B^m$ is $\mu$-Donsker \citep[Proof of Corollary 1]{goldfeld2024limit}, and it holds that $\mathbb{G}_n = \sqrt{n}(\hat \mu_n - \mu) \rightsquigarrow \mathbb{G}_\mu$ in $\ell^\infty(B^m)$. Moreover, by \Cref{brownian bridge concentration} it also holds that $\mathbb{G}_\mu \in \mathbb T_\mu^m$ almost surely. This fact will be key for our results in this section.


\subsection{The Exponential Map} \label{sec: expmap linearization}

In finite-dimensional Riemannian geometry, the exponential map provides a canonical way to move from a point on the manifold in the direction of a tangent vector. Given a Riemannian manifold $\mathcal M$, a point $p\in\mathcal M$, and a tangent vector $v\in T_p\mathcal M$, one defines $\exp_p(v)=\gamma_v(1)$, where $\gamma_v$ is the geodesic satisfying $\gamma_v(0)=p$ and $\dot\gamma_v(0)=v$. More generally, the curve $t\mapsto\exp_p(tv)$ traces the geodesic emanating from $p$ with initial velocity $v$ \citep{lee2018introduction}. 

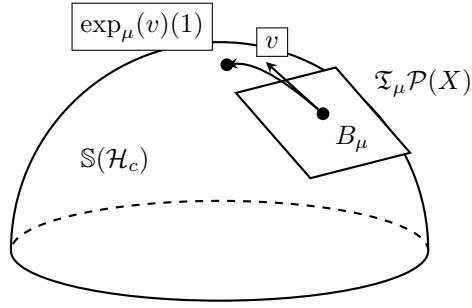
\begin{figure}
    \centering
    \begin{tikzpicture}[scale=1.2,>=stealth,every node/.style={font=\small}]

\def\R{2.3}

\draw[thick] (-\R,0) arc[start angle=180,end angle=0,radius=\R];

\draw[thick] (-\R,0) arc[start angle=180,end angle=360,x radius=\R,y radius=0.55];
\draw[thick,dashed] (\R,0) arc[start angle=0,end angle=180,x radius=\R,y radius=0.55];

\node at (-1.15,1.0) {$\mathbb{S}(\mathcal{H}_c)$};

\coordinate (mu) at (1.14,1.51);
\fill (mu) circle (2pt);

\coordinate (p1) at (1.00,0.80);
\coordinate (p2) at (0.18,1.74);
\coordinate (p3) at (1.28,2.02);
\coordinate (p4) at (2.10,1.08);

\filldraw[fill=white, draw=black, thick]
    (p1) -- (p2) -- (p3) -- (p4) -- cycle;

\node at (2.25,1.85) {$\mathfrak{T}_{\mu}\mathcal{P}(X)$};

\fill (mu) circle (2pt);
\node[below right=1pt] at (mu) {$B_\mu$};

\draw[->,thick] (mu) -- ++(-0.63,0.57);
\node at (0.58,2.3) {$\fcolorbox{black}{white}{$v$}$};

\coordinate (endpt) at (0.08,2.05);

\draw[thick,->]
    (mu) .. controls (0.60,1.99) and (0.32,2.08) .. (endpt);

\fill (endpt) circle (2pt);
\node[above left=2pt] at (endpt) {$\fcolorbox{black}{white}{$\exp_\mu(v)(1)$}$};

\end{tikzpicture}
    \caption{Visualization of the Sinkhorn exponential map in the $\mathcal{H}_c$ parameterization.}
    \label{fig:sinkhorn expoential map}
\end{figure}

This construction has a familiar interpretation in the Riemannian geometry of optimal transport: whenever $v$ induces an optimal transport map $T=\operatorname{id}+v$, the corresponding Wasserstein geodesic is the displacement interpolation $\mu_t=((1-t)\operatorname{id}+tT)_\#\mu$, so that the Wasserstein exponential map is realized by pushing the measure forward along the associated particle trajectories. We now develop the analogous exponential map for the Sinkhorn Riemannian geometry through the geodesic equations derived in \Cref{sec: geodesic equations and pt}. Let $\mu\in\mathcal P(X)$ and $v\in\mathfrak T_\mu\mathcal P(X)$, and let $\eta_{\mu,v}$ denote the solution to \eqref{geodesic equations} with initial conditions $\beta_0=B_\mu$ and $\dot\beta_0=v$. 
The exponential map
$\exp$
is then defined as the flow map of the ODE as a function of $t$,
$\exp(v)=\eta_v$. This differs from the standard
endpoint-valued definition of the exponential map, as we seek
a limit theorem for the entire traced-out curve. On the domain $\mathfrak D_{\exp}$ specified in \Cref{linearizing expmap}, define $\exp(\mu,v)=\eta_{\mu,v}\in C^1([0,1];\mathbb S(\mathcal H_c))$. For fixed $\mu$, write $\exp_\mu(v)=\eta_{\mu,v}$ for every $v$ satisfying $(\mu,v)\in\mathfrak D_{\exp}$. Unlike the usual endpoint-valued exponential map, this map represents the entire geodesic.

\begin{rmk}[Probability-valuedness of ambient geodesics]
    As discussed in \Cref{ambient geodesics}, solutions to the ambient geodesic equations are not guaranteed to remain in $B[\mathcal P(X)]$, even though they remain on the unit sphere $\mathbb S(\mathcal H_c)$. Consequently, their inverse images under $M$ are only guaranteed to be elements of $H^s(X)^*$ with unit mass, rather than probability measures. However, if $v$ is the initial velocity of a variational minimizer joining $\mu$ and $\nu$ whose associated curve of measures satisfies the uniform density lower bound of \Cref{geodesic coincidence}, then that result and uniqueness of the geodesic initial-value problem imply that $\exp_\mu(v)(t)\in B[\mathcal P(X)]$ for every $t\in[0,1]$. Thus, the ambient geodesic coincides with the probability-measure-valued variational interpolant.
\end{rmk}

To linearize the exponential map, consider perturbations $(\mu^s,v^s)$ of the initial position and velocity and let $(\beta_t^s)_{t\in[0,1]}$ denote the corresponding family of geodesics. The derivative of the exponential map will be given by the induced first variation $h_t=\frac{d}{ds}\beta_t^s|_{s=0}$. The only compatibility condition on the initial perturbation comes from tangency of the perturbed velocity. Indeed, since $v^s\in\mathfrak T_{\mu^s}\mathcal P(X)$, we have $\langle v^s,B_{\mu^s}\rangle_{\mathcal H_c}=0$, and differentiation at $s=0$ gives
\[
    \langle a,B_\mu\rangle_{\mathcal H_c}
    +
    \langle v,\operatorname D_\gamma B_\mu\rangle_{\mathcal H_c}
    =0,
\]
where $\gamma$ and $a$ denote the perturbations of $\mu$ and $v$, respectively. \new{Our strategy for linearizing the geodesic flow is to reduce the geodesic equations to a smooth ODE on an open subset of a Banach space, allowing us to invoke standard results on the differentiability of ODE solution maps. To do so, we introduce local coordinates on the unit sphere $\mathbb S(\mathcal H_c)$ and rewrite the second-order geodesic equations as a first-order system in position and velocity, following the standard construction used in the proof of Theorem 4.27 of \citet{lee2018introduction}. The smooth extensions of the geodesic coefficients then allow us to apply Banach space ODE theory to obtain the following pathwise linearization.}

Let $U_+$ be defined as in \Cref{uplus definition}, and, for an ambient geodesic $(\beta_t)_{t\in[0,1]}$ contained in $U_+\cap\mathbb S(\mathcal H_c)$, write $\mu_t=M(\beta_t)\in H^s(X)^*$. At each $t$, we use the inverse change of variables from \Cref{ambient inverse change of vars} satisfying $B_{\mu_t}=\beta_t$, and identify the operators with the corresponding compositions of this inverse and the extensions from \Cref{smoothness of geodesic coefficients}. In particular, for $h_t\in\beta_t^\perp$, writing $\sigma_{h_t}=\operatorname D_{h_t}M(\beta_t)$, we identify derivatives in distributional directions with the corresponding derivatives in $\beta$-coordinates, so that
\[
\operatorname D_{\sigma_{h_t}}\Pi_{\mu_t}
=\operatorname D_{h_t}\Pi_{\beta_t},
\qquad
\operatorname D_{\sigma_{h_t}}\Gamma_{\beta_t}
=\operatorname D_{h_t}\Gamma_{\beta_t}.
\]
These identifications agree with the original operators and derivatives whenever $\mu_t$ is a probability measure.

\begin{thm}[Linearization of the exponential map]\label{linearizing expmap}   
    Define the domain of the exponential map 
    \[
    \mathfrak D_{\exp}
    =
    \left\{
    (\mu,v)\in\mathcal P(X)\times\mathcal H_c:
    v\in\mathfrak T_\mu\mathcal P(X),\
    \eta_{\mu,v}\text{ exists on }[0,1],\
    \eta_{\mu,v}([0,1])\subset U_+
    \right\}.
    \]
    and let $(\mu, v) \in \mathfrak{D}_{\exp}$, and $(\gamma, a)$ be a tangential direction. Then the map $\exp: \mathfrak{D}_{\exp} \subset \ell^\infty(B^m) \times \mathcal{H}_c \to C^1([0,1];\mathcal{H}_c)$ given by $(\exp_\mu(v))(t)  = \eta_{\mu, v}(t)$ is Hadamard differentiable at $(\mu, v)$ tangentially to the set $\mathbb{T}_{(\mu,v)}\mathfrak{D}_{\exp}
    :=
    \left\{
    (\gamma,a)\in
    \mathbb{T}_\mu^m\times\mathcal{H}_c:
    \langle a,B_\mu\rangle_{\mathcal{H}_c}
    +
    \langle v,\operatorname{D}_\gamma B_\mu\rangle_{\mathcal{H}_c}
    =0
    \right\}.$ Moreover, its Hadamard derivative is given by $ \operatorname{D}_{(\gamma, a)}\exp_{\mu}(v) = (h_t)_{t \in [0,1]}$
    where $(h_t)_{t\in[0,1]}$ is the unique solution, subject to $\langle h_t,\beta_t\rangle_{\mathcal H_c}=0$ for every $t\in[0,1]$, of the linear initial value problem
    \begin{equation}
        (\operatorname{D}_{\sigma_{h_t}}\Pi_{\mu_t})(a_t) + \Pi_{\mu_t}\left(\ddot h_t
    +
    (\operatorname{D}_{\sigma_{h_t}}\Gamma_{\beta_t})(\dot \beta_t, \dot \beta_t) + \Gamma_{\beta_t}(\dot h_t, \dot \beta_t) + \Gamma_{\beta_t}(\dot \beta_t, \dot h_t)\right) = 0
    \end{equation}
    for every $t \in [0,1]$ where $\Gamma$ is defined in \Cref{gamma expression},  $\sigma_{h_t} = (\operatorname{D}_{(\cdot)}B_{\mu_t})^{-1}(h_t)$, $h_0 = \operatorname{D}_{\gamma}B_\mu$, $\dot h_0 = a$ and $a_t = \ddot \beta_t + \Gamma_{\beta_t}(\dot \beta_t, \dot \beta_t)$. 
\end{thm}


\new{Thus, the solution $h=(h_t)_{t\in[0,1]}$ characterizes the first-order variation of the entire geodesic trajectory with respect to perturbations of its initial position and velocity. This pathwise linearization is subsequently used to establish the Gaussian process limit for the plugin geodesic interpolant in \Cref{sec: clts}.}

\begin{proof}[Proof of \Cref{linearizing expmap}]
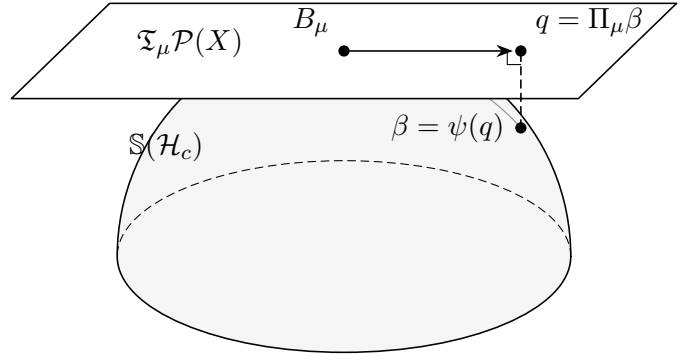
\begin{figure}
    \centering

    \begin{tikzpicture}[
        line cap=round,
        line join=round,
        >=Stealth,
        every node/.style={font=\normalsize},
        point/.style={circle,fill=black,inner sep=1.5pt}
    ]
        \def\R{3}
        \def\a{25}
        \def\b{0.78} 
    
        \pgfmathsetmacro{\H}{\R*cos(\a)}
        \pgfmathsetmacro{\E}{\R*sin(\a)}
        \pgfmathsetmacro{\Bx}{\R*\b}
        \pgfmathsetmacro{\By}{\H*sqrt(1-\b*\b)}
        \pgfmathsetmacro{\angleB}{asin(\b)}
    
        \coordinate (mu)   at (0,\H);
        \coordinate (q)    at (\Bx,\H);
        \coordinate (beta) at (\Bx,\By);
    
        \path[fill=black!4]
            (\R,0)
            arc[start angle=0,end angle=180,radius=\R]
            arc[start angle=180,end angle=360,
                x radius=\R,y radius=\E]
            -- cycle;
    
        \draw[thin,densely dashed]
            (\R,0)
            arc[start angle=0,end angle=180,
                x radius=\R,y radius=\E];
    
        \draw[semithick]
            (\R,0)
            arc[start angle=0,end angle=180,radius=\R]
            arc[start angle=180,end angle=360,
                x radius=\R,y radius=\E];
    
        \draw[thin,black!45]
            plot[domain=0:\angleB,samples=60,variable=\t]
            ({\R*sin(\t)},{\H*cos(\t)});
    
        \node at (-2.35,1.45) {$\mathbb S(\mathcal H_c)$};
    
        %
        \path[fill=white,draw=black,semithick]
            (-4.40,{\H-0.63})
            -- (3.10,{\H-0.63})
            -- (4.40,{\H+0.63})
            -- (-3.10,{\H+0.63})
            -- cycle;
    
        \node[left] at (-1.20,{\H+0.1})
            {$\mathfrak T_{\mu}\mathcal P(X)$};
    
        \draw[densely dashed,semithick]
            (beta) -- (q);
    
        \draw[-{Stealth[length=2.2mm]},semithick,
              shorten <=2pt,shorten >=3pt]
            (mu) -- (q);
    
        \draw[thin]
            ($(q)+(-0.18,0)$)
            -- ($(q)+(-0.18,-0.18)$)
            -- ($(q)+(0,-0.18)$);
    
        \node[point,label={above left:$B_\mu$}]
            at (mu) {};
    
        \node[point,label={above right:$q = \Pi_\mu \beta$}]
            at (q) {};
    
        \node[point,label={left:$\beta=\psi(q)$}]
            at (beta) {};
    
    
    \end{tikzpicture}
    \caption{Visualization of the coordinate chart $\psi: V_R \subset \mathfrak{T}_\mu \mathcal{P}(X) \to \mathbb{S}(\mathcal{H}_c)$.}
    \label{fig:sinkhorn coordinate chart}
\end{figure}

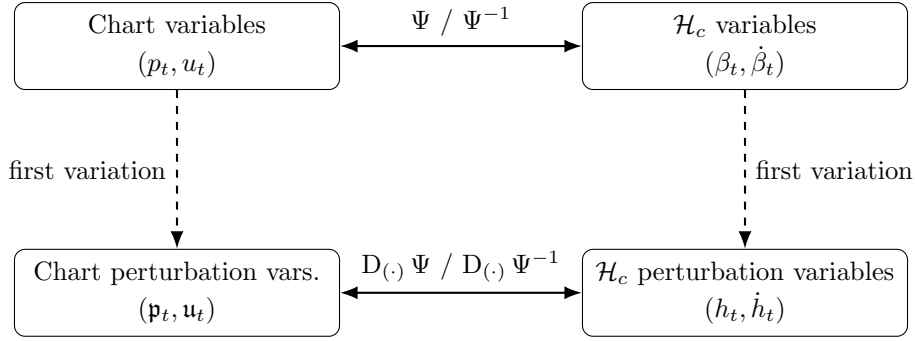
\begin{figure}[t]
\centering
\begin{tikzpicture}[
    >=Latex,
    every node/.style={font=\small},
    box/.style={
        draw,
        rounded corners=4pt,
        align=center,
        minimum width=4.3cm,
        minimum height=1.15cm
    },
    lab/.style={font=\footnotesize\itshape}
]

\node[box] (coordstate) {Chart variables\\[2pt] $(p_t,u_t)$};
\node[box, right=3.2cm of coordstate] (geomstate) {$\mathcal{H}_c$ variables\\[2pt] $(\beta_t,\dot\beta_t)$};

\node[box, below=2.1cm of coordstate] (coordvar) {Chart perturbation vars.\\[2pt] $(\mathfrak p_t,\mathfrak u_t)$};
\node[box, right=3.2cm of coordvar] (geomvar) {$\mathcal{H}_c$ perturbation variables\\[2pt] $(h_t,\dot h_t)$};

\draw[<->, thick] (coordstate) -- node[above] {$\Psi$ / $\Psi^{-1}$} (geomstate);
\draw[<->, thick] (coordvar) -- node[above] {$\operatorname D_{(\cdot)}\Psi$ / $\operatorname D_{(\cdot)}\Psi^{-1}$} (geomvar);

\draw[->, dashed, thick] (coordstate) -- node[left] {first variation} (coordvar);
\draw[->, dashed, thick] (geomstate) -- node[right] {first variation} (geomvar);

\end{tikzpicture}
\caption{Notation for the exponential-map linearization on the chart $I_{t_i}$. }
\label{fig:exp-linearization-notation}
\end{figure}

As mentioned, we will analyze the geodesic equations in a fixed local
coordinate system on the $\mathcal H_c$-unit sphere. Fix a reference
measure $\mu\in\mathcal P(X)$ and write $B_\mu\in\mathbb S(\mathcal H_c)$
for its image under the Sinkhorn change of variables. Since $\mathfrak T_\mu\mathcal P(X) = B_\mu^\perp = T_{B_\mu}\mathbb S(\mathcal H_c),$
a neighborhood of $B_\mu$ on the sphere can be parameterized by the
tangent plane $\mathfrak T_\mu\mathcal P(X)$. In particular, define
\[
    \psi(q)
    =
    B_\mu\sqrt{1-\|q\|_{\mathcal H_c}^2}+q,
    \qquad
    q\in V_R,
\]
where $V_R :=
    \left\{
        q\in\mathfrak T_\mu\mathcal P(X):
        \|q\|_{\mathcal H_c}<R<1
    \right\}$
with $R$ chosen sufficiently small so that $\psi(V_R) \subset U_+.$ Then $\psi$ parameterizes a neighborhood of $B_\mu$ contained in the
hemisphere $\left\{
        \beta\in\mathbb S(\mathcal H_c):
        \langle\beta,B_\mu\rangle_{\mathcal H_c}>0
    \right\},$
and its inverse is simply $\psi^{-1}(\beta)=\Pi_\mu\beta.$ Thus the nonlinear spherical constraint is removed locally by representing
$\beta$ through the unconstrained coordinate
$p=\Pi_\mu\beta\in V_R$.

Since the geodesic equation is second order, we must simultaneously keep
track of position and velocity. The natural state space is therefore the
tangent bundle of the sphere over $\psi(V_R)$,
\[
    T\psi(V_R)
    =
    \left\{
        (\beta,u):
        \beta\in\psi(V_R),\
        u\in T_\beta\mathbb S(\mathcal H_c)
    \right\}.
\]
The base-space chart $\psi^{-1}$ induces the standard tangent-bundle chart $\Psi:T\psi(V_R)
    \to
    V_R\times\mathfrak T_\mu\mathcal P(X)$ 
given by
\begin{equation}\label{psi chart}
    \Psi(\beta,u)
    =
    \left(
        \psi^{-1}(\beta),
        \operatorname D_u\psi^{-1}_\beta
    \right).
\end{equation}
Since $\psi^{-1}$ is the restriction of the linear projection $\Pi_\mu$,
its derivative is independent of the base point and satisfies
$\operatorname D_{(\cdot)}\psi^{-1}_\beta=\Pi_\mu(\cdot)$, and consequently, $\Psi(\beta,u) =
    (\Pi_\mu\beta,\Pi_\mu u).$ The state space in these local coordinates is therefore the open subset $U
    :=
    V_R\times\mathfrak T_\mu\mathcal P(X)$ of the Hilbert space $\mathbb E
    :=
    \mathfrak T_\mu\mathcal P(X)
    \times
    \mathfrak T_\mu\mathcal P(X).$
    
For a set of initial conditions in \eqref{geodesic equations} given by $(B_\mu,v)$ with
$v\in\mathfrak T_\mu\mathcal P(X)$, we have $\Psi(B_\mu,v)=(0,v).$ Now, for $t$ sufficiently small let $(\beta_t,\dot\beta_t)$ be a geodesic remaining in $T\psi(V_R)$ and define its coordinate representation with $ (p_t,u_t):= \Psi(\beta_t,\dot\beta_t).$ Since $p_t=\Pi_\mu\beta_t$ and $u_t=\Pi_\mu\dot\beta_t,$ we immediately obtain $\dot p_t=u_t$. The remaining task is therefore to
express $\dot u_t$ in terms of $(p_t,u_t)$ using the geodesic equations.
This yields a first-order system of the form
\[
    \frac{d}{dt}
    \begin{pmatrix}
        p_t\\
        u_t
    \end{pmatrix}
    =
    \begin{pmatrix}
        u_t\\
        \mathscr A(p_t,u_t)
    \end{pmatrix}
\]
on the fixed Hilbert space $\mathbb E$. Once the acceleration map
$\mathscr A$ is shown to be sufficiently smooth, standard Banach-space
ODE theory yields a local flow and the corresponding smooth dependence
on the initial position and velocity.

\begin{lemma}[Chart geodesic ODE] \label{chart geodesic ODE}
    The geodesic equations in the $(p, u)$-coordinates induced by the chart $\Psi$ can be expressed as 
    \begin{align*}
        \frac{d}{dt}\begin{pmatrix}
            p_t \\ u_t
        \end{pmatrix} = \begin{pmatrix}
            u_t \\
            \mathscr{A}(p_t, u_t)
        \end{pmatrix} \quad \text{s.t. $p_0 = \psi^{-1}(\beta_0)$ and $u_0=(\operatorname D\psi(p_0))^{-1}
\dot\beta_0=\Pi_\mu\dot\beta_0$}
    \end{align*}
    where $\mathscr{A}: V_R \times \mathfrak{T}_\mu \mathcal{P}(X)\to\mathfrak T_\mu\mathcal P(X)$ is given by
    \[\mathscr{A}(p, u) = -\|\operatorname{D}_{u}\psi(p)\|_{\mathcal{H}_c}^2p  - \Pi_\mu\Pi_{\psi(p)}\Gamma_{\psi(p)}(\operatorname{D}_{u}\psi(p), \operatorname{D}_{u}\psi(p)).\]
\end{lemma}
We provide the proof of \Cref{chart geodesic ODE} in \Cref{proof of chart geodesic ODE}. By \Cref{smoothness of chart acceleration}, the map $\mathscr A:V_R\times\mathfrak{T}_\mu\mathcal{P}(X)\to\mathfrak{T}_\mu\mathcal{P}(X)$ is $C^\infty$, and therefore $C^1$. Fix an $\epsilon>0$, let $J=(-\epsilon,\epsilon)$, and define the vector field $f:J\times U\to\mathbb E$ by
\[
    f(t,(p,u))
    =
    \begin{pmatrix}
        u\\
        \mathscr A(p,u)
    \end{pmatrix}.
\]
Since $\mathscr A$ is $C^1$, the vector field $f$ belongs to $C^1(J\times U,\mathbb E)$. The coordinate geodesic equations can therefore be regarded as the Banach-space initial value problem $\dot x_t=f(t,x_t)$ on the open set $U\subset\mathbb E$. We may now invoke standard Banach-space ODE theory. In particular, we recall below the \emph{Local Smoothness Theorem} of \citet{lang2012fundamentals}, which guarantees the local existence and regularity of the flow generated by a sufficiently smooth vector field on a possibly infinite-dimensional Banach space. This result will allow us to transfer the regularity of the coordinate vector field $f$ to the local geodesic flow and, subsequently, to its dependence on the initial position and velocity.

\begin{thm}[Local Smoothness Theorem, \cite{lang2012fundamentals}, Theorem 1.11] \label{local smoothness theorem}
    Let $J$ be an open interval in $\mathbb{R}$ containing 0 and let $U$ be open in the Banach space $\mathbb{E}.$ Let $f: J \times U \to \mathbb{E}$ be a $C^p$ morphism with $p \geq 1$, and let $x_0 \in U$. Then there exists a unique local flow for $f$ at $x_0$. We can select an open subinterval $J_0$ of $J$ containing $0$ and an open subset $U_0$ of $U$ containing $x_0$ such that the unique local flow
    \[\alpha: J_0 \times U_0 \to U\]
    is of class $C^p$ and such that $\operatorname{D}_2 \alpha$ satisfies
    \[\operatorname{D}_{1}\operatorname{D}_2\alpha(t,x) = \operatorname{D}_2 f(t, \alpha(t,x))\operatorname{D}_2\alpha(t,x)\]
    on $J_0 \times U_0$ with initial condition $\operatorname{D}_2\alpha(0, x) = \operatorname{id}.$
\end{thm}
Thus, we may invoke \Cref{local smoothness theorem} to conclude that there exist open sets $J_0\subset J$ and $U_0\subset U$ together with a local flow $\alpha:J_0\times U_0\to U$ of class $C^1$. Moreover, its derivative with respect to the initial condition satisfies
\[
    \frac{d}{dt}\operatorname{D}_2\alpha(t,(p,u))
    =
    \operatorname{D}_2f\bigl(t,\alpha(t,(p,u))\bigr)
    \circ
    \operatorname{D}_2\alpha(t,(p,u)),
\]
with initial condition $\operatorname{D}_2\alpha(0,(p,u)) = \operatorname{id}_{\mathbb E}.$ Write $\alpha(t,(p,u))=(p_t,u_t)$ and note that, since $f(t,(p,u)) = \begin{pmatrix}
        u & 
        \mathscr A(p,u)
    \end{pmatrix}^\top,$ its derivative with respect to the state variable is
\[
    \operatorname{D}_{2,(a,b)}f(t,(p,u))
    =
    \begin{pmatrix}
        b\\
        \operatorname{D}_{(a,b)}\mathscr A(p,u)
    \end{pmatrix}
    =
    \begin{pmatrix}
        b\\
        \operatorname{D}_{1,a}\mathscr A(p,u)
        +
        \operatorname{D}_{2,b}\mathscr A(p,u)
    \end{pmatrix}
\]
for every $(a,b)\in\mathbb E$. Now fix an initial state $(p_0,u_0)\in U_0$ and an initial perturbation $(\mathfrak p_0,\mathfrak u_0)\in\mathbb E$. Define the propagated perturbation by $(\mathfrak p_t,\mathfrak u_t) := \operatorname{D}_{2, (\mathfrak p_0,\mathfrak u_0)}\alpha(t,(p_0,u_0))
    .$
Then the equation above implies that $(\mathfrak p_t,\mathfrak u_t)$ satisfies 

\begin{equation}\label{state space ivp}
    \frac{d}{dt}
    \begin{pmatrix}
        \mathfrak p_t\\
        \mathfrak u_t
    \end{pmatrix}
    =
    \begin{pmatrix}
        \mathfrak u_t\\
        \operatorname{D}_{1,\mathfrak p_t}\mathscr A(p_t,u_t)
        +
        \operatorname{D}_{2,\mathfrak u_t}\mathscr A(p_t,u_t)
    \end{pmatrix},
\end{equation}
with prescribed initial condition $(\mathfrak p_0,\mathfrak u_0)$. Thus, the local geodesic flow $\alpha_t:(p_0,u_0)\mapsto(p_t,u_t)$ is $C^1$ in its initial condition, and its derivative is characterized by the linear equation \Cref{state space ivp}. In particular, because $\mathscr A$ is $C^\infty$, its Fr\'echet derivative $\operatorname{D}\mathscr A(p_t,u_t)$ exists, is bounded and linear in $(\mathfrak p_t,\mathfrak u_t)$, and depends smoothly on the base trajectory $(p_t,u_t)$.

\begin{rmk}[Chart locality]\label{tagged partition}
    Note that our argument implicitly assumes that $\Psi(\beta_t,\dot\beta_t)\in V_R\times\mathfrak T_\mu\mathcal P(X)$ for every $t\in J$, which we have not established. To circumvent this, we employ the following argument. For every $t\in[0,1]$, choose a sphere chart $\psi_t$ centered at $\beta_t$ with image contained in $U_+$, and let $\Psi_t$ be its tangent-bundle chart. By continuity of the state and \Cref{local smoothness theorem}, we may choose a relative open interval $I_t$ containing $t$ such that $\Psi_t(\beta_s,\dot\beta_s)\in V_{R_t}\times\beta_t^\perp$ for every $s\in I_t$. Shrinking $I_t$ further, we may ensure that these coordinate states lie in the initial-condition neighborhood of the local flow and that the length of $I_t$ is smaller than its common positive existence time. Thus, $\{I_t\}_{t \in [0,1]}$ is an open cover of $[0,1]$; by Cousin's lemma \citep[Theorem 1.4]{bartle2001modern}, there exists a tagged partition $0=s_0<s_1<\dots<s_N=1$, with tags $t_i \in [s_{i-1},s_i]$, such that $[s_{i-1},s_i]\subset I_{t_i}$. In particular, $I_{t_i}\cap I_{t_{i+1}}\neq \varnothing$, so composing the associated flow maps and smooth chart transitions gives $C^1$ dependence on the initial condition along the entire trajectory. We make this rigorous in the stitching lemma (\Cref{stitching lemma}).
\end{rmk}

Now we need to stitch together the local flows $\alpha_i$ obtained by each partition $t_i$. Since this is technical and doesn't offer much by way of insight, we defer its proof to the Appendix, \Cref{proof of stitching lemma}.


\begin{lemma}[Stitching lemma]\label{stitching lemma}
Let $(\beta_t)_{t\in[0,1]}$ be a solution of the ambient geodesic equations satisfying $\beta_t\in U_+\cap\mathbb S(\mathcal H_c)$ for every $t\in[0,1]$ where $U_+$ is defined in \Cref{uplus definition}. Then there exists an open neighborhood $\mathcal N_0$ of $(\beta_0,\dot\beta_0)$ in $T\mathbb S(\mathcal H_c)$ such that the ambient geodesic flow is defined on $[0,1]$ for every initial state in $\mathcal N_0$, and the resulting trajectories remain in $U_+$. Moreover, the trajectory map
\[
G:\mathcal N_0\to C^0([0,1];\mathcal H_c\times\mathcal H_c)
\]
is $C^1$, and its first component is $C^1$ as a map into $C^1([0,1];\mathcal H_c)$. Its derivative is obtained by solving the chartwise variational equations \Cref{state space ivp} and transporting the perturbations between adjacent charts by the differentials of the transition maps.
\end{lemma}



Finally, we identify the equation satisfied by $h_t$. Every nearby geodesic satisfies $\Pi_{\mu_t^n}(a_t^n)=0$, with the operators interpreted through the local extensions above. Set $\sigma_{h_t}=\operatorname D_{h_t}M(\beta_t)$, and note that differentiating the first-order coordinate system and reconstructing the ambient variation shows that the full first variation of $a_t=\ddot\beta_t+\Gamma_{\beta_t}(\dot\beta_t,\dot\beta_t)$ is
\[
\ddot h_t
+(\operatorname D_{\sigma_{h_t}}\Gamma_{\beta_t})(\dot\beta_t,\dot\beta_t)
+\Gamma_{\beta_t}(\dot h_t,\dot\beta_t)
+\Gamma_{\beta_t}(\dot\beta_t,\dot h_t).
\]
Differentiating $\Pi_{\mu_t}(a_t)=0$ therefore gives exactly the projected linearized equation in \Cref{linearizing expmap}. The initial conditions are $h_0=\operatorname D_\gamma B_\mu$ and $\dot h_0=a$, while differentiation of the unit-sphere constraint gives $\langle h_t,\beta_t\rangle_{\mathcal H_c}=0$. Uniqueness follows from \Cref{uniqueness linearized geodesic}, and bounded linearity follows from \Cref{bounded linearity}, completing the proof. \end{proof}

\subsection{The Logarithmic Map}\label{sec: logmap linearization}

In finite-dimensional Riemannian geometry, the logarithmic map solves the inverse problem to that of the exponential map. Given two sufficiently close points $p,q\in\mathcal M$, the logarithmic map $\log_{p}(q)\in T_{p}\mathcal M$ is the unique tangent vector satisfying $\exp_p(\log_{p}(q))=q.$ Equivalently, the curve $t\mapsto\exp_{p}(t\log_{p}(q))$ is the geodesic emanating from $p$ that reaches $q$ at time one. The analogous object in Wasserstein geometry is the displacement induced by the Brenier map. In particular, if $T$ is the optimal transport map pushing $\mu$ to $\nu$, then $\log_{\mu}(\nu)=T-\operatorname{id}$, and the curve $\mu_t=((1-t)\operatorname{id}+tT)_{\#}\mu$ is the corresponding Wasserstein geodesic.

\begin{defn}[The Sinkhorn logarithmic map]
    For $\mu,\nu\in\mathcal P(X)$, we define $\log_\mu(\nu) = v$ where $\exp_\mu(v)|_{t = 1} = B_\nu$, whenever such a $v \in \mathfrak{T}_\mu \mathcal{P}(X)$ exists and is unique.
\end{defn}

We define the Sinkhorn logarithmic map in direct analogy to that of Wasserstein geometry. The only distinction is that our exponential map is valued in the $\beta$-parameterization of probability measures. So for $\mu,\nu\in\mathcal P(X)$, the logarithmic map seeks an initial velocity $v\in\mathfrak T_\mu\mathcal P(X)$ whose geodesic satisfies $\exp_\mu(v)(1)=B_\nu$. As in finite-dimensional Riemannian geometry, the logarithmic map need not be globally well defined or regular.
We will therefore restrict attention to pairs of measures $(\mu,\nu)$ for which the minimizing tangent vector $v=\log_\mu(\nu)$ exists uniquely and the exponential map is locally invertible with respect to its initial velocity at $v$. More precisely, suppose
\[
    \operatorname D_2 E(\mu,v):
    \mathfrak T_\mu\mathcal P(X)
    \to
    \mathfrak T_\nu\mathcal P(X)
\]
is a bounded linear isomorphism, where $ E(\mu,v):=\exp_\mu(v)(1)$ denotes the endpoint map. In analogy with the finite-dimensional characterization of conjugate points \citep[Proposition~10.20]{lee2018introduction}, we say that $\mu$ and $\nu$ are \emph{nonconjugate along the geodesic} $t\mapsto\exp_\mu(v)(t)$ whenever this condition holds. We then define the regular domain of the logarithmic map by
\[
    \mathfrak D_{\log} := \left\{ (\mu,\nu)\in\mathcal P(X)\times \mathcal{P}(X): \begin{array}{c} \log_\mu(\nu)\text{ exists uniquely, and}\\ \operatorname D_2E(\mu,\log_\mu(\nu)) \text{ is a bounded linear isomorphism} \end{array} \right\}.
\]

The defining relation between the exponential and logarithmic maps is therefore $ E(\mu,\log_\mu(\nu))=B_\nu.$ Note that the appearance of $B_\nu$ rather than $\nu$ reflects the fact that our exponential map is valued in the $\beta$-parameterization of probability measures. This identity provides a convenient route to the linearization of the logarithmic map: rather than differentiating the minimizing representation directly, we may differentiate the inverse relation between the exponential and logarithmic maps.

The implicit function theorem used below establishes local rather than global uniqueness of the logarithmic map. In particular, near a nonconjugate pair $(\mu,\nu)$, it yields a unique initial velocity in a neighborhood of $v=\log_\mu(\nu)$ satisfying the endpoint equation for each sufficiently close pair $(\mu',\nu')$. We continue to denote this locally defined map by $\log_{\mu'}(\nu')$, which coincides with the original logarithmic map wherever the latter is uniquely defined.


\begin{thm}[Linearization of the logarithmic map]
\label{linearizing logmap}

Let $(\mu,\nu)\in\mathfrak D_{\log}$ and let $v=\log_\mu(\nu)$, where $v=\dot\beta_0^\star$ for a variational minimizing geodesic $(\mu_t^\star)_{t\in[0,1]}$ satisfying the assumptions of \Cref{geodesic coincidence}, where $\beta_t^\star=B_{\mu_t^\star}$. Then the logarithmic map $(\mu',\nu')\mapsto\log_{\mu'}(\nu')$, regarded as an $\mathcal H_c$-valued map, is Hadamard differentiable at $(\mu,\nu)$ tangentially to $\mathbb T_\mu^m\times\mathbb T_\nu^m$. For $(\gamma,\eta)\in
\mathbb T_\mu^m\times\mathbb T_\nu^m$,  define
$a_{\gamma,\eta}^{\mathfrak{T}}\in\mathfrak T_\mu\mathcal P(X)$ by
\begin{equation}
\label{coordinate log map derivative}
a_{\gamma,\eta}^{\mathfrak{T}}
=
\left[
    \operatorname D_{2,(\cdot)}E(\mu,v)
\right]^{-1}
\left[
    \operatorname D_\eta B_\nu
    -
    \operatorname D_{
        \left(
            \gamma,
            -\langle v, \operatorname{D}_\gamma B_\mu\rangle_{\mathcal H_c}B_\mu
        \right)
    }E(\mu,v)
\right].
\end{equation}
Then
\begin{equation}
\label{log map derivative}
\operatorname D_{(\gamma,\eta)}\log_\mu(\nu)
=
a_{\gamma,\eta}^{\mathfrak{T}}
-
\langle v,\operatorname D_\gamma B_\mu\rangle_{\mathcal H_c}B_\mu.
\end{equation}

\end{thm}
We regard $a_{\gamma, \eta}^{\mathfrak{T}}$ as the tangent variation of the logarithmic map as induced by the variation in $\mu, \nu$, while the final derivative is obtained by adding the normal component. The distinction between the ambient variation of the logarithmic map
and its variation in fixed tangent coordinates is important when the
base point varies. Indeed, if
$v=\log_\mu(\nu)$ and
$a=\operatorname D_{(\gamma,\eta)}\log_\mu(\nu)$, then
$\log_{\mu'}(\nu')\in\mathfrak T_{\mu'}\mathcal P(X)$ implies
\[
    \langle a,B_\mu\rangle_{\mathcal H_c}
    +
    \langle v,\operatorname D_\gamma B_\mu\rangle_{\mathcal H_c}
    =0.
\]
Thus, writing
$a=a^{\mathfrak{T}}-\langle v,\operatorname D_\gamma B_\mu\rangle_{\mathcal H_c}B_\mu$
with $a^{\mathfrak{T}}\in\mathfrak T_\mu\mathcal P(X)$ separates the variation
of the velocity coordinate from the variation of the tangent space.
The pair
\[
    \left(
        \gamma,
        -\langle v,\operatorname D_\gamma B_\mu\rangle_{\mathcal H_c}
        B_\mu
    \right)
\]
is a tangent direction to $\mathfrak D_{\exp}$, as is $(0,a^{\mathfrak{T}})$.
Differentiating the identity
$E(\mu,\log_\mu(\nu))=B_\nu$ therefore yields
\[
    \operatorname D_{
        \left(
            \gamma,
            -\langle v,\operatorname D_\gamma B_\mu\rangle_{\mathcal H_c}
            B_\mu
        \right)
    }E(\mu,v)
    +
    \operatorname D_{2,a^{\mathfrak{T}}}E(\mu,v)
    =
    \operatorname D_\eta B_\nu.
\]
The nonconjugacy assumption allows us to solve uniquely for
$a^{\mathfrak{T}}$, yielding \Cref{coordinate log map derivative}, and the
ambient derivative then follows from the decomposition above. The proof below makes this formal calculation rigorous by applying the implicit function theorem to the defining relation between the exponential and logarithmic maps.

\begin{proof}[Proof of \Cref{linearizing logmap}]

As in the proof of \Cref{linearizing expmap}, we work in the
spherical coordinate charts centered at $B_\mu$ and $B_\nu$.
Throughout the proof, the subscripts on $\psi$ and $\Psi$
indicate the measures around which the corresponding charts
are centered. Let $v=\log_\mu(\nu)$, and write
$E_{\mathrm{amb}}(\beta,w)$ such that for
$(\mu',w)\in\mathfrak D_{\exp}$, we have $E(\mu',w)=E_{\mathrm{amb}}(B_{\mu'},w).$

By \Cref{geodesic coincidence,ambient coefficient domain},
the reference geodesic solves the ambient geodesic equations
and remains in $U_+$ throughout $[0,1]$. Thus,
\Cref{stitching lemma} gives an open neighborhood
$\mathscr D_{\exp}$ of $(0,v)$ in
$V_\mu\times\mathfrak T_\mu\mathcal P(X)$ on which the
ambient geodesic flow with initial state
$\Psi_\mu^{-1}(p,u)$ exists on $[0,1]$ and depends
$C^1$-smoothly on $(p,u)$. Since $E_{\mathrm{amb}}(B_\mu,v)=B_\nu$,
continuity of the endpoint map allows us to restrict
$\mathscr D_{\exp}$ so that all corresponding endpoints
belong to $\psi_\nu(V_\nu)$.

\medskip

\emph{Applying the Implicit Function Theorem.} To apply the implicit function theorem, define $\mathscr E_{\mu,\nu}
=\psi_\nu^{-1}\circ E_{\mathrm{amb}}\circ\Psi_\mu^{-1}$ and note that $\mathscr E_{\mu,\nu}(0,v)=0$. Moreover, define the map $\mathscr M_{\mu,\nu}(p,y,u)=\mathscr E_{\mu,\nu}(p,u)-y$. Then $\mathscr M_{\mu,\nu}(0,0,v)=0$. Recall that
\[
    \Psi_\mu^{-1}(p,u)
    =\bigl(\psi_\mu(p),\operatorname D_u\psi_\mu(p)\bigr).
\]
Since $\operatorname D_a\psi_\mu(0)=a$, for every
$a\in\mathfrak T_\mu\mathcal P(X)$ we have $\operatorname D_{2,a}\Psi_\mu^{-1}(0,v)=(0,a)$ and $\operatorname D_{(0,a)}E_{\mathrm{amb}}(B_\mu,v) =\operatorname D_{2,a}E(\mu,v).$ Thus, by the chain rule, the derivative of $\mathscr M_{\mu,\nu}$
with respect to the unknown $u$ at $(0,0,v)$ is
\[
    \operatorname D_{3,(\cdot)}\mathscr M_{\mu,\nu}(0,0,v)
    =
    \operatorname D_{2,(\cdot)}\mathscr E_{\mu,\nu}(0,v)
    =
    \operatorname D\psi_\nu^{-1}\big|_{B_\nu}
    \circ
    \operatorname D_{2,(\cdot)}E(\mu,v).
\]
By \Cref{coordinate chart regularity}, the differentials of $\psi_\nu^{-1}$ and $\Psi_\mu^{-1}$ appearing above are bounded linear isomorphisms. Moreover, since $(\mu,\nu)$ is nonconjugate, $\operatorname D_{2,(\cdot)}E(\mu,v)$ is a bounded linear isomorphism as well. Hence $\operatorname D_{3,(\cdot)}\mathscr M_{\mu,\nu}(0,0,v)$ is a bounded linear isomorphism. The Banach-space implicit function theorem \citep[Theorem 2.5.7]{abraham2012manifolds} therefore yields neighborhoods $U_\mu\subset V_\mu$ of $0$, $W_\nu\subset V_\nu$ of $0$, and a neighborhood $O_v\subset\mathfrak T_\mu\mathcal P(X)$ of $v$, together with a unique $C^1$ map $g:U_\mu\times W_\nu\to O_v$ satisfying $\mathscr E_{\mu,\nu}(p,g(p,y))=y$ for every $(p,y)\in U_\mu\times W_\nu$.

\medskip

\emph{Regularity of the logarithmic map.} Now let $\mu',\nu'$ be sufficiently close to $\mu,\nu$ that $p'=\psi_\mu^{-1}(B_{\mu'})\in U_\mu$ and $y'=\psi_\nu^{-1}(B_{\nu'})\in W_\nu$. By construction, $E_{\mathrm{amb}}(\Psi_\mu^{-1}\bigl(p',g(p',y')\bigr))=B_{\nu'}.$
Consequently, the second component of $\Psi_\mu^{-1}(p',g(p',y'))$ is the locally unique initial velocity near $v$ satisfying the endpoint equation, which we continue to denote by $\log_{\mu'}(\nu')$. We note, however, that the implicit function theorem does \textit{not} exclude other initial velocities outside $O_v$ reaching the same endpoint.
 Since $g$ and $\Psi_\mu^{-1}$ are $C^1$, the logarithmic map is $C^1$ as a function of the local coordinates $(p',y')$. By \Cref{hadamard deriv of change of var}, the maps $\mu'\mapsto B_{\mu'}$ and $\nu'\mapsto B_{\nu'}$ are Hadamard differentiable tangentially to $\mathbb T_\mu^m$ and $\mathbb T_\nu^m$, respectively. Composing these maps with $\psi_\mu^{-1}$, $\psi_\nu^{-1}$, $g$, and $\Psi_\mu^{-1}$, and applying the Hadamard chain rule, shows that $(\mu',\nu')\mapsto\log_{\mu'}(\nu')$ is Hadamard differentiable at $(\mu,\nu)$ tangentially to $\mathbb T_\mu^m\times\mathbb T_\nu^m$.

\medskip

\medskip

\emph{Identifying the Hadamard derivative.}
Let $a =
    \operatorname D_{(\gamma,\eta)}\log_\mu(\nu)
    \in\mathcal H_c.$ Since $\log_{\mu'}(\nu')\in\mathfrak T_{\mu'}\mathcal P(X)$ for every
nearby pair $(\mu',\nu')$, we have
$\langle\log_{\mu'}(\nu'),B_{\mu'}\rangle_{\mathcal H_c}=0$.
Differentiating this identity at $(\mu,\nu)$ gives $\langle a,B_\mu\rangle_{\mathcal H_c}
    =
    -
    \langle v,\operatorname D_\gamma B_\mu\rangle_{\mathcal H_c}.$ Therefore, writing $a^{\mathfrak{T}}=\Pi_\mu a$, we have
\[
    a
    =
    a^{\mathfrak{T}}
    -
    \langle v,\operatorname D_\gamma B_\mu\rangle_{\mathcal H_c}
    B_\mu,
    \qquad
    a^{\mathfrak{ T}} \in\mathfrak T_\mu\mathcal P(X).
\]
Now differentiate the defining identity
$E(\mu,\log_\mu(\nu))=B_\nu$. Since
$(\gamma,a)$ is a tangent direction to $\mathfrak D_{\exp}$,
\Cref{linearizing expmap} gives $\operatorname D_{(\gamma,a)}E(\mu,v)=
\operatorname D_\eta B_\nu.$ By linearity of the derivative and the preceding decomposition of $a$,
\[
    \operatorname D_{
        \left(
            \gamma,
            -\langle v,\operatorname D_\gamma B_\mu\rangle_{\mathcal H_c}
            B_\mu
        \right)
    }E(\mu,v)
    +
    \operatorname D_{2,a^{\mathfrak{T}}}E(\mu,v)
    =
    \operatorname D_\eta B_\nu.
\]
Both directions appearing on the left-hand side belong to
$\mathbb T_{(\mu,v)}\mathfrak D_{\exp}$: the first satisfies the
linearized tangency condition by construction, while the second does
so because $a^{\mathfrak{T}}\perp B_\mu$. Since $(\mu,\nu)$ is nonconjugate,
$\operatorname D_{2,(\cdot)}E(\mu,v)$ is a bounded linear isomorphism.
Hence
\[
    a^{\mathfrak{T}}
    =
    \left[
        \operatorname D_{2,(\cdot)}E(\mu,v)
    \right]^{-1}
    \left[
        \operatorname D_\eta B_\nu
        -
        \operatorname D_{
            \left(
                \gamma,
                -\langle v,\operatorname D_\gamma B_\mu\rangle_{\mathcal H_c}
                B_\mu
            \right)
        }E(\mu,v)
    \right],
\]
which is \Cref{coordinate log map derivative}. Substituting this into
the decomposition of $a$ gives \Cref{log map derivative}. Finally, the right-hand side of \Cref{log map derivative} is bounded
and linear in $(\gamma,\eta)$ by
\Cref{linearizing expmap}, \Cref{hadamard deriv of change of var}, and
the bounded invertibility of
$\operatorname D_{2,(\cdot)}E(\mu,v)$. This completes the proof. 

\end{proof}

\subsection{Parallel Transport} \label{sec: pt linearization}

In finite-dimensional Riemannian geometry, parallel transport provides a canonical way to compare tangent vectors attached to different points of a manifold. Given a smooth curve $\gamma_t$ from $p$ to $q$ and a tangent vector $v\in T_p\mathcal M$, parallel transport evolves $v$ along $\gamma_t$ through the equation $\nabla_{\dot\gamma_t}v_t=0$ with $v_0=v$, producing a tangent vector $\operatorname{PT}_{p\to q}(v) =v_1\in T_q\mathcal M$. In the Sinkhorn geometry, we apply the same construction along the geodesic $t\mapsto\exp_\mu(\log_\mu(\nu))(t)$ joining $\mu$ to $\nu$, thereby obtaining a map that transports tangent directions in $\mathfrak T_\mu\mathcal P(X)$ to tangent directions in $\mathfrak T_\nu\mathcal P(X)$.

Combining the exponential map, logarithmic map, and parallel transport yields a natural operation that we refer to as \emph{parallel imputation}. Given three measures $(\mu,\nu,\rho)$, the tangent $\log_\mu(\rho)$ represents the initial direction of the geodesic from $\mu$ to $\rho$. We transport this direction along the geodesic from $\mu$ to $\nu$, and then use the resulting tangent at $\nu$ as the initial velocity of a new geodesic. In this sense, parallel imputation transfers the displacement from $\mu$ to $\rho$ onto the new base point $\nu$. Formally, writing $\operatorname{PT}_{\mu\to\nu}$ for parallel transport along the geodesic from $\mu$ to $\nu$, this construction takes the form $\operatorname{Imp}(\mu,\nu,\rho)=\exp_\nu\!\left(\operatorname{PT}_{\mu\to\nu}(\log_\mu(\rho))\right)$. Closely related constructions have been used to transfer distributional effects across populations in causal inference and machine learning \citep{saidi2026wassersteinparalleltransportpredicting,saidi2026differentialgeometricequivalencehellingerkantorovichconewasserstein}.

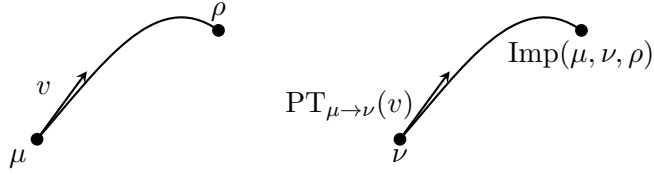
\begin{figure}
    \centering
    \begin{tikzpicture}[scale=1.2, >=stealth]

\coordinate (mu) at (0,0);
\coordinate (rho) at (2,1.2);
\coordinate (nu) at (4,0);
\coordinate (imp) at (6,1.2);

\fill (mu) circle (2pt) node[below left] {$\mu$};
\fill (rho) circle (2pt) node[above] {$\rho$};
\fill (nu) circle (2pt) node[below] {$\nu$};
\fill (imp) circle (2pt) node[below] {$\operatorname{Imp}(\mu,\nu,\rho)$};

\draw[thick]
    (mu) .. controls (0.7,0.8) and (1.3,1.7) .. (rho);

\draw[thick]
    (nu) .. controls (4.7,0.8) and (5.3,1.7) .. (imp);

\draw[->, thick]
    (0,0) -- (0.55,0.75)
    node[midway, above left] {$v$};

\draw[->, thick]
    (4.0,0.0) -- (4.55,0.75)
    node[midway, left] {$\PT_{\mu\to\nu}(v)$};

\end{tikzpicture}
    \caption{Visualization of the parallel imputation map, $\operatorname{Imp}(\mu, \nu, \rho).$}
    \label{fig:imputation map}
\end{figure}

For the linearization below, we introduce notation for the constituent maps. As in the preceding section, write $L(\mu,\rho)=\log_\mu(\rho)$ and $E(\mu,v)=\exp_\mu(v)|_{t=1}$. For initial geodesic data $(\mu,v)$, let $\beta_t=\exp_\mu(v)(t)$, and for $u\in\mathfrak T_\mu\mathcal P(X)$ let $(w_t)_{t\in[0,1]}$ solve the parallel transport equation
\[
    \Pi_{\mu_t}\left(\dot w_t+\Gamma_{\beta_t}(\dot\beta_t,w_t)\right)=0,
    \qquad w_0=u.
\]
We denote the resulting parallel transport map by $\mathscr T(\mu,v,u)=(w_t)_{t\in[0,1]}$. In particular, if $(\mu,\nu)$ belongs to the regular domain of the logarithmic map, then parallel transport from $\mu$ to $\nu$ is obtained along the geodesic $\beta_t^{\mu,\nu}=\exp_\mu(\log_\mu(\nu))(t)$. With this notation, the parallel imputation map introduced above can be written as
\begin{equation}\label{imputation map}
    \operatorname{Imp}(\mu,\nu,\rho)
    =
    \exp_\nu\left(
        \mathscr T\left(
            \mu,\log_\mu(\nu),\log_\mu(\rho)
        \right)\Big|_{t=1}
    \right).
\end{equation}
Thus, $\operatorname{Imp}$ is obtained by composing the logarithmic map, parallel transport, and the exponential map. We will study the linearization of this composition about a fixed triplet $(\mu,\nu,\rho)$; this will subsequently yield a Gaussian limiting process for its plug-in estimator.

\medskip 
\noindent{\emph{A heuristic calculation.}} To further understand what is required to linearize $\operatorname{Imp}$, we will proceed with an informative, albeit informal, calculation. For a formal linearization, see the proof of \Cref{linearizing imp} below. For shorthand, we'll write $\mathscr T_1=\mathscr T(\mu,\log_\mu(\nu),\log_\mu(\rho))|_{t=1}$. Differentiating \Cref{imputation map} in the direction $(\gamma,\eta,\zeta)$ suggests $\operatorname D_{(\gamma,\eta,\zeta)}\operatorname{Imp}(\mu,\nu,\rho)=\operatorname D_{(\eta,k_1)}\exp_\nu(\mathscr T_1)$, where $k_1$ is the first variation of $\mathscr T_1$. Applying the chain rule gives
\begin{equation*}
    k_1
    =
    \operatorname D_{(\gamma,\dot h_0,k_0)}
        \mathscr T(\mu,\log_\mu(\nu),\log_\mu(\rho))
        \big|_{t=1},
\end{equation*}
where $h_0=\operatorname D_\gamma B_\mu$, $\dot h_0=\operatorname D_{(\gamma,\eta)}\log_\mu(\nu)$, and $k_0=\operatorname D_{(\gamma,\zeta)}\log_\mu(\rho)$. The joint derivative of $\mathscr T$ will be obtained by linearizing the coupled geodesic and parallel transport equations, using a modified version of the argument in \Cref{linearizing expmap}.

\begin{thm}[Linearization of $\operatorname{Imp}$] \label{linearizing imp}
Let $(\mu,\nu,\rho)\in\mathcal P(X)^3$ and suppose that both $(\mu,\rho)$ and $(\mu,\nu)$ are nonconjugate elements of $\mathfrak D_{\log}$. Write $w=\mathscr T(\mu,\log_\mu(\nu),\log_\mu(\rho))|_{t=1}$ and suppose that $(\nu,w)\in\mathfrak D_{\exp}$. Suppose moreover that the geodesics associated with $\log_\mu(\nu)$ and $\log_\mu(\rho)$ are variational minimizing geodesics satisfying the assumptions of \Cref{geodesic coincidence}. Then the map $\operatorname{Imp}$ described in \Cref{imputation map}, regarded as a $C^1([0,1];\mathcal H_c)$-valued map, is Hadamard differentiable at $(\mu,\nu,\rho)$ tangentially to $\mathbb T_\mu^m\times\mathbb T_\nu^m\times\mathbb T_\rho^m$, with derivative
\begin{equation*}
    \operatorname D_{(\gamma,\eta,\zeta)}
        \operatorname{Imp}(\mu,\nu,\rho)
    =
    \operatorname D_{(\eta,k_1)}\exp_\nu(w).
\end{equation*}
In the charts of \Cref{three coordinate ode}, let $(\mathfrak p_t^i,\mathfrak u_t^i)$ be the chart variations of the geodesic from $\mu$ to $\nu$ induced by $(\gamma,\eta)$. The chart variation $\mathfrak r_t^i$ of the parallel transported field is the unique solution of
\begin{align*}
    \dot{\mathfrak r}_t^i
    &=
    \operatorname D_{1,\mathfrak p_t^i}
        \mathscr R_i(p_t^i,u_t^i,r_t^i)
    +
    \operatorname D_{2,\mathfrak u_t^i}
        \mathscr R_i(p_t^i,u_t^i,r_t^i) +
    \operatorname D_{3,\mathfrak r_t^i}
        \mathscr R_i(p_t^i,u_t^i,r_t^i),
    \qquad t\in I_{t_i},
\end{align*}
with the initial and chart-transition conditions of \Cref{chart perturbation boundary condition}. The variation $k_t$ is reconstructed from the chart variations by \eqref{ambient parallel variation}; in particular, $k_0=\operatorname D_{(\gamma,\zeta)}\log_\mu(\rho)$.
\end{thm}


\begin{rmk}[On the assumptions of \Cref{linearizing imp}]
We emphasize that the preceding result does \emph{not} require the population imputation trajectory to be probability-measure-valued. Rather, the assumption $(\nu,w)\in\mathfrak D_{\exp}$ requires only that the associated ambient geodesic exist throughout $[0,1]$ and remain in $U_+$, where its coefficients are smooth by \Cref{smoothness of geodesic coefficients}. In particular, \Cref{small tangent full time existence} guarantees this condition whenever $\|w\|_{\mathcal H_c}<\delta_\nu$ for some $\delta_\nu>0$. Moreover, since parallel transport preserves the metric and the source geodesic joining $\mu$ to $\rho$ is constant-speed minimizing, we have
\[
\frac{\varepsilon}{2}\|w\|_{\mathcal H_c}^2
\leq\tilde{\mathbf g}_\nu(w,w)
=d_S^2(\mu,\rho).
\]
Consequently, $d_S(\mu,\rho)<\sqrt{\varepsilon/2}\,\delta_\nu$ is a sufficient condition for $(\nu,w)\in\mathfrak D_{\exp}$. Thus, our limit theorem applies even when the population imputation trajectory leaves $B[\mathcal P(X)]$.
\end{rmk}

Our proof strategy will closely mirror that used for the exponential map in \Cref{linearizing expmap}. There, the coordinate charts allowed us to represent the position and velocity of a geodesic by coordinates $(p_t,u_t)$ taking values in a fixed Hilbert space, so that the geodesic equations became a first-order Banach-space ODE. Here, we augment this coordinate system with a third variable $r_t$ that encodes the parallel transported tangent field $w_t$ in the same chart. More precisely, if $\psi$ is a local chart along the base geodesic, we write $r_t=(\operatorname D_{(\cdot)}\psi(p_t))^{-1}(w_t)$, so that $(p_t,u_t,r_t)$ jointly encode the position, velocity, and transported tangent vector. The coupled geodesic and parallel transport equations can then be written locally as a first-order system of the form $\dot p_t=u_t$, $\dot u_t=\mathscr A(p_t,u_t)$, and $\dot r_t=\mathscr R(p_t,u_t,r_t)$. Once the smoothness of $\mathscr A$ and $\mathscr R$ is established, the same Banach-space ODE argument used for the exponential map yields a smooth local flow in these three coordinates. Differentiating this flow with respect to its initial data then gives the linearized parallel transport equation, which can be combined with the linearizations of the logarithmic and exponential maps to obtain the derivative of the parallel imputation map.
\begin{proof}[Proof of \Cref{linearizing imp}]

Let $\{I_{t_i}\}_{i \in \{1, \dots, N\}}$ be the finite covering of $[0,1]$ from \Cref{linearizing expmap} and let $\psi_{t_i}$ be the associated coordinate charts. Let $i$ be any element of the set $\{1, \dots, N\}$ and define $(p_t^i, u_t^i)$ such that $p_t^i = \Pi_{\mu_{t_i}}\beta_t$, $u_t^i = \Pi_{\mu_{t_i}}\dot \beta_{t}$  for $t \in I_{t_i}$. Observe that $\dot p_t^i = \Pi_{\mu_{t_i}} \dot \beta_t = u_t^i$, and recall the identities \[\beta_t = \psi_{t_i}(p_t^i) \quad \text{and }\quad \dot \beta_t = \operatorname{D}_{u_t^i}\psi_{t_i}(p_t^i).\] 
Now we'll add a new coordinate (denoted with the variable $r$) to encode the parallel transported field induced by the chart $\psi_{t_i}$. In particular, define $r_t^i = (\operatorname{D}_{(\cdot)}\psi_{t_i}(p_t^i))^{-1}(w_t)$ where $w_t$ is the solution to \eqref{PT IVP}, which exists due to our assumptions and \Cref{well posedness pt and geodesics}. In addition, note that the formula from \Cref{chart differential} implies that $\Pi_{\mu_{t_i}} w_t = \Pi_{\mu_{t_i}} \operatorname{D}_{r_t^i}\psi_{t_i}(p_t^i) = r_t^i$ for $t \in I_{t_i}$. Note that in this section we will abuse notation and redefine $\Psi$ (\Cref{psi chart}) to be $\Psi_{t_i}(\beta, u_1, u_2) = (\Pi_{\mu_{t_i}}\beta, \Pi_{\mu_{t_i}}u_1, \Pi_{\mu_{t_i}}u_2)$. This simply extends the former definition to include the additional variable $r$ in the chart.

\begin{lemma}\label{three coordinate ode}
    The coordinates $(p_t^i, u_t^i, r_t^i)$ have a time derivative that can be described as
    \begin{align} \label{three coordinate ode expression}
        \frac{d}{dt}\begin{pmatrix}
            p_t^i \\
            u_t^i \\
            r_t^i
        \end{pmatrix} = \begin{pmatrix}
            u_{t}^i \\
            \mathscr{A}_i(p_t^i, u_t^i) \\
            \mathscr{R}_i(p_t^i, u_t^i, r_t^i)
        \end{pmatrix}
    \end{align}
    where $\mathscr{R}_i:V_{R_{t_i}}\times(\mathfrak{T}_{\mu_{t_i}}\mathcal{P}(X))^2 \to \mathcal{H}_c$ and $\mathscr{A}_i: V_{R_{t_i}} \times \mathfrak{T}_{\mu_{t_i}} \mathcal{P}(X)\to\mathfrak T_{\mu_{t_i}}\mathcal P(X)$ are given by
    \begin{gather*}
        \mathscr{R}_i(p, u, r) =  - \langle \operatorname{D}_{r}\psi_{t_i}(p), \operatorname{D}_{u}\psi_{t_i}(p)\rangle_{\mathcal{H}_c} p - \Pi_{\mu_{t_i}} \Pi_{\psi_{t_i}(p)} \Gamma_{\psi_{t_i}(p)}(\operatorname{D}_{u}\psi_{t_i}(p), \operatorname{D}_{r}\psi_{t_i}(p)) \\
        \mathscr{A}_i(p, u) = -\|\operatorname{D}_{u}\psi_{t_i}(p)\|_{\mathcal{H}_c}^2p  - \Pi_{\mu_{t_i}}\Pi_{\psi_{t_i}(p)}\Gamma_{\psi_{t_i}(p)}(\operatorname{D}_{u}\psi_{t_i}(p), \operatorname{D}_{u}\psi_{t_i}(p)).
    \end{gather*}
\end{lemma}

\begin{figure}[t]
\centering
\begin{tikzpicture}[
    >=Latex,
    every node/.style={font=\small},
    box/.style={
        draw,
        rounded corners=4pt,
        align=center,
        minimum width=4.8cm,
        minimum height=1.2cm
    },
    lab/.style={font=\footnotesize\itshape}
]

\node[box] (coordstate) {Chart variables\\[2pt] $(p_t^i,u_t^i,r_t^i)$};
\node[box, right=3.4cm of coordstate] (geomstate) {$\mathcal{H}_c$ variables\\[2pt] $(\beta_t,\dot\beta_t,w_t)$};

\node[box, below=2.2cm of coordstate] (coordvar) {Chart perturbations vars.\\[2pt] $(\mathfrak p_t^i,\mathfrak u_t^i,\mathfrak r_t^i)$};
\node[box, right=3.4cm of coordvar] (geomvar) {$\mathcal{H}_c$ perturbation variables\\[2pt] $(h_t,\dot h_t,k_t)$};

\draw[<->, thick] (coordstate) -- node[above] {$\Psi_i$ / $\Psi_i^{-1}$} (geomstate);
\draw[<->, thick] (coordvar) -- node[above] {$\operatorname D_{(\cdot)}\Psi_i$ / $\operatorname D_{(\cdot)}\Psi_i^{-1}$} (geomvar);

\draw[->, dashed, thick] (coordstate) -- node[left] {first variation} (coordvar);
\draw[->, dashed, thick] (geomstate) -- node[right] {first variation} (geomvar);

\end{tikzpicture}
\caption{Notation for the parallel transport / $\operatorname{Imp}$ linearization on the chart $I_{t_i}$, in direct analogy with that of \Cref{fig:exp-linearization-notation}.}
\label{fig:imp-linearization-notation}
\end{figure}
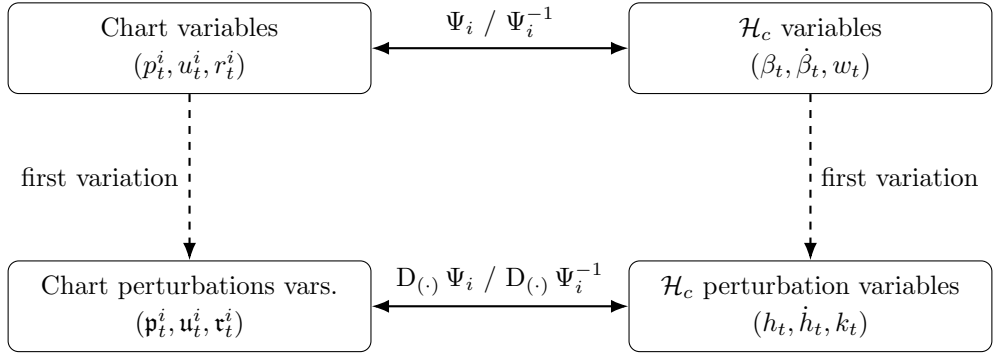

We provide the proof of \Cref{three coordinate ode} in \Cref{proof of three coordinate ode}. By \Cref{smoothness of chart acceleration} we know that $\mathscr{A}_i: V_{R_{t_i}}\times \mathfrak{T}_{\mu_{t_i}}\mathcal P(X) \to \mathfrak{T}_{\mu_{t_i}} \mathcal{P}(X)$ is $C^1$ and by \Cref{smoothness of chart parallel transport} we know that $\mathscr{R}_i: V_{R_{t_i}}
\times\mathfrak T_{\mu_{t_i}}\mathcal P(X)
\times\mathfrak T_{\mu_{t_i}}\mathcal P(X)
\to
\mathfrak T_{\mu_{t_i}}\mathcal P(X)$ is $C^1$ as well. Thus, if we define $\Xi_i(p_t^i, u_t^i, r_t^i)$ as the right-hand side of \Cref{three coordinate ode expression}, then $\Xi_i$ is $C^1$ as well. Therefore, one can invoke the local smoothness theorem \Cref{local smoothness theorem} as was done in \Cref{linearizing expmap} to conclude that the local flow $\alpha_i$ generated by $\Xi_i$ is $C^1$-dependent on its initial condition. As in the proof of \Cref{linearizing expmap}, the finitely many local
flows may be composed with the $C^1$ transition maps
$\Psi_{i+1}\circ\Psi_i^{-1}$ and pulled back through the inverse charts to define the augmented trajectory map \begin{equation}
    G_{\operatorname{PT}}: U_{\PT} \to C^1([0,1];\mathcal H_c\times \mathcal{H}_c \times \mathcal{H}_c) \quad \text{given by $(\beta_0,\dot\beta_0,w_0)
    \to
    \bigl((\beta_t,\dot\beta_t,w_t)\bigr)_{t\in[0,1]}$}.
\end{equation}
Here $U_{\PT}$ is a relatively open neighborhood of $(\beta_0,\dot\beta_0,w_0)$ in \[\{(\beta,u,w)\in\mathcal H_c^3:\beta\in U_+\cap\mathbb S(\mathcal H_c),\ u,w\in\beta^\perp\}.\] Its image under $\Psi_1$ is open in the corresponding coordinate space $V_{R_{t_1}}\times(\beta_{t_1}^\perp)^2$. More precisely, the initial
coordinate state is
$\Psi_1(\beta_0,\dot\beta_0,w_0)$, each coordinate trajectory is
generated by the corresponding local flow $\alpha_i$, and adjacent
coordinate trajectories are joined through
$\Psi_{i+1}\circ\Psi_i^{-1}$. Since the local flows, transition maps,
and inverse charts are $C^1$, the map $G_{\operatorname{PT}}$ is $C^1$.

Now let $(\mathfrak p_t^i,\mathfrak u_t^i,\mathfrak r_t^i)$ denote the first variation of $(p_t^i,u_t^i,r_t^i)$ for $t \in I_{t_i}$. Taking the first variation of \Cref{three coordinate ode expression} and swapping the order of the spatial and temporal differentiation implies
\begin{align}
    \frac{d}{dt}
    \begin{pmatrix}
        \mathfrak p_t^i\\
        \mathfrak u_t^i\\
        \mathfrak r_t^i
    \end{pmatrix}
    =
    \begin{pmatrix}
        \mathfrak u_t^i\\
        \operatorname{D}_{1,\mathfrak p_t^i}
            \mathscr A_i(p_t^i,u_t^i)
        +
        \operatorname{D}_{2,\mathfrak u_t^i}
            \mathscr A_i(p_t^i,u_t^i)
        \\
        \operatorname{D}_{1,\mathfrak p_t^i}
            \mathscr R_i(p_t^i,u_t^i,r_t^i)
        +
        \operatorname{D}_{2,\mathfrak u_t^i}
            \mathscr R_i(p_t^i,u_t^i,r_t^i)
        +
        \operatorname{D}_{3,\mathfrak r_t^i}
            \mathscr R_i(p_t^i,u_t^i,r_t^i)
    \end{pmatrix}.
    \label{three coordinate variational ode}
\end{align}
The first two equations are precisely the chartwise perturbation
equations appearing in the proof of \Cref{linearizing expmap}. We defer the derivation of the temporal boundary conditions to the Appendix, in \Cref{proof of chart perturbation boundary condition}.

\begin{lemma}[Chart perturbation boundary condition] \label{chart perturbation boundary condition}
    Let $(\mathfrak{p}_t^i, \mathfrak{u}_t^i, \mathfrak{r}_t^i)_{t \in I_{t_i}}$ be the perturbation of the $(p^i, u^i, r^i)$ coordinates of the geodesic and tangent induced by the perturbation $(\gamma, \eta, \zeta)$ of $(\mu, \nu, \rho)$. Then the chart perturbations $(\mathfrak{p}_t^i, \mathfrak{u}_t^i, \mathfrak{r}_t^i)_{t \in I_{t_i}}$ satisfy the boundary conditions
    \begin{equation}
    \begin{pmatrix}
        \mathfrak{p}_{s_{i-1}}^i \\
        \mathfrak u_{s_{i - 1}}^i \\
        \mathfrak r_{s_{i - 1}}^i
    \end{pmatrix} = \begin{cases}\operatorname{D}_{(\mathfrak{p}_{s_{i-1}}^{i-1},\mathfrak u_{s_{i - 1}}^{i - 1}, \mathfrak{r}_{s_{i - 1}}^{i - 1})}(\Psi_i \circ \Psi^{-1}_{i - 1})(p^{i-1}_{s_{i-1}}, u^{i-1}_{s_{i-1}}, r^{i-1}_{s_{i-1}}) & \text{if $i > 1$} \\
    \operatorname{D}_{(h_0, \dot h_0, k_0)}\Psi_1(\beta_0, \dot \beta_0, w_0) & \text{otherwise}
    \end{cases}
\end{equation}
    where 
    \[(\beta_0, \dot \beta_0, w_0) = (B_\mu, \log_\mu(\nu), \log_\mu(\rho)), \quad \text{and} \quad (h_0, \dot h_0, k_0) = (\operatorname{D}_\gamma B_\mu, \operatorname{D}_{(\gamma, \eta)}\log_\mu(\nu), \operatorname{D}_{(\gamma, \zeta)}\log_\mu(\rho)), \]
    respectively.
\end{lemma}

Now we will convert the chart perturbations $(\mathfrak{p}_t^i, \mathfrak{u}_t^i, \mathfrak{r}_t^i)$ back to the $\mathcal{H}_c$ perturbations $(h_t, \dot h_t, k_t).$ For $k_t$, recall that $w_t=\operatorname{D}_{r_t^i}\psi_{t_i}(p_t^i) := F_1(r_t^i, p_t^i).$ If we regard $k_t$ as the variation of $w_t$ induced by the perturbation triplet, then we can differentiate $F_1( r_t^i, p_t^i)$ in the perturbation direction $( \mathfrak{r}_t^i, \mathfrak{p}_t^i)$ to write $k_t = \operatorname{D}_{( \mathfrak{r}_t^i, \mathfrak{p}_t^i)}\operatorname{D}_{r_t^i}\psi_{t_i}(p_t^i)$. For $\dot \beta_t$, recall that $\dot \beta_t = \operatorname{D}_{u_t^i}\psi_{t_i}(p_t^i) := F_2(u_t^i, p_t^i)$. Then, differentiating in the direction $(\mathfrak{u}_t^i, \mathfrak{p}_t^i)$ implies $\dot h_t = \operatorname{D}_{(\mathfrak{u}_t^i, \mathfrak{p}_t^i)}\operatorname{D}_{u_t^i}\psi_{t_i}(p_t^i)$. Finally, because $\beta_t = \psi_{t_i}(p_t^i)$, we differentiate to get $h_t = \operatorname{D}_{\mathfrak{p}_t^i}\psi_{t_i}(p_t^i)$. Summarizing, we have that the $\mathcal{H}_c$ perturbation variables are characterized by
\begin{equation} \label{ambient parallel variation}
    \begin{pmatrix}
        h_t \\ \dot h_t \\ k_t
    \end{pmatrix} = \begin{pmatrix}
        \operatorname{D}_{\mathfrak{p}_t^i}\psi_{t_i}(p_t^i) \\\operatorname{D}_{(\mathfrak{u}_t^i, \mathfrak{p}_t^i)}\operatorname{D}_{u_t^i}\psi_{t_i}(p_t^i) \\
        \operatorname{D}_{( \mathfrak{r}_t^i, \mathfrak{p}_t^i)}\operatorname{D}_{r_t^i}\psi_{t_i}(p_t^i) 
    \end{pmatrix} 
    \quad \text{for $t \in I_{t_i}$.}
\end{equation}

Now consider the initial-condition map $ I(\mu,\nu,\rho) =
    \bigl(
        B_\mu,
        \log_\mu(\nu),
        \log_\mu(\rho)
    \bigr).$ By \Cref{hadamard deriv of change of var} and
\Cref{linearizing logmap}, $I$ is Hadamard differentiable at
$(\mu,\nu,\rho)$ tangentially to
$\mathbb T_\mu^m\times\mathbb T_\nu^m\times\mathbb T_\rho^m$, with
derivative $\operatorname{D}_{(\gamma,\eta,\zeta)}I(\mu,\nu,\rho) = (h_0,\dot h_0,k_0).$
Also, recall that the trajectory map $G_{\PT}$ constructed above is $C^1$ in these
initial data and is therefore Hadamard differentiable. The Hadamard chain rule consequently yields Hadamard differentiability of $(\mu,\nu,\rho)\to\mathscr T
    \left(
        \mu,
        \log_\mu(\nu),
        \log_\mu(\rho)
    \right)$, with derivative $(k_t)_{t\in[0,1]}$
obtained from \Cref{three coordinate variational ode} and
\Cref{ambient parallel variation}.

In particular, if $w=\mathscr T(\mu,\log_\mu(\nu),\log_\mu(\rho))|_{t=1}$, then its first variation induced by $(\gamma,\eta,\zeta)$ is $k_1$. Since $w\in\mathfrak T_\nu\mathcal P(X)$, differentiating $\langle w,B_\nu\rangle_{\mathcal H_c}=0$ gives $\langle k_1,B_\nu\rangle_{\mathcal H_c}+\langle w,\operatorname D_\eta B_\nu\rangle_{\mathcal H_c}=0$. Thus $(\eta,k_1)$ is an admissible joint direction for the derivative of the exponential map at $(\nu,w)$. Recalling that $\operatorname{Imp}(\mu,\nu,\rho)=\exp_\nu(w)$, the Hadamard chain rule yields $\operatorname D_{(\gamma,\eta,\zeta)}\operatorname{Imp}(\mu,\nu,\rho)=\operatorname D_{(\eta,k_1)}\exp_\nu(w)$. Linearity of the Hadamard derivative follows from \Cref{linearity of parallel imputation variation}, completing the proof.
\end{proof}

\subsection{Sinkhorn Distance}

In this subsection we will present results regarding our final linearization. In particular, we will obtain the first variation of the squared Sinkhorn distance $d_S^2$ defined in \Cref{variational geodesic}. To do so, consider a one parameter family of perturbations of $\mu$ and $\nu$ respectively given by
\begin{equation} \label{exponential reweighting perturbations}
\mu^r = \frac{e^{ru_0}\mu}{\mu ( e^{ru_0})}, \quad \nu^r = \frac{e^{ru_1}\nu}{\nu(e^{ru_1})}
\end{equation}
where $u_0,u_1\in C^\infty(X)$, and let $\boldsymbol{\mu}^r$ be the associated minimizing curve connecting $\mu^r$ and $\nu^r$. Unlike earlier linearizations, we now need to perturb a curve $\boldsymbol{\mu}^r$ connecting $\mu^r$ to $\nu^r$ whose regularity is \emph{unknown}. Additive perturbations therefore may violate probability-valuedness, but the perturbations above are guaranteed not to. These perturbations instead exponentially ``tilt" the mass according to a scalar-valued function $u_i: X \to \mathbb R.$

\begin{lemma}[First variation of the Sinkhorn distance]
\label{first variation sinkhorn distance}
Suppose $m > d/2$, the pair $(\mu, \nu)$ is nonconjugate with $\mu \neq \nu$, and the population variational geodesic $(\mu_t)_{t\in[0,1]}$ is the unique constant-speed minimizing geodesic joining $\mu$ and $\nu$ with a uniformly positive Lebesgue density. Then $r\mapsto d_S^2(\mu^r,\nu^r)$ is differentiable at $r = 0$, with
\[
\left.\frac{d}{dr}d_S^2(\mu^r,\nu^r)\right|_{r=0}
=
-2\tilde{\mathbf g}_\mu(v,\operatorname D_hB_\mu)
+2\tilde{\mathbf g}_\nu(\dot\beta_1,\operatorname D_kB_\nu),
\]
where $h=(u_0-\mu(u_0))\mu$, $k=(u_1-\nu(u_1))\nu$, and $v=\dot\beta_0$.
\end{lemma}

We provide the proof of \Cref{first variation sinkhorn distance} in \Cref{proof of first variation sinkhorn distance}, where we employ the following argument. We construct admissible interpolant competitors for both the original $(\mu, \nu)$ and perturbed $(\mu^r, \nu^r)$ variational problems. The strong stability of minimizing geodesics (\Cref{strong stability variational geodesics}) then allows us to identify the first variation of the \emph{optimal} value without differentiating the minimizers themselves. Invoking the Euler--Lagrange equation derived in \Cref{geodesic coincidence} and integration by parts ensures that the interior variation vanishes, leaving only contributions from the endpoint perturbations. Since this argument initially establishes differentiability only along directions described in \Cref{exponential reweighting perturbations}, we next exploit their density in $H^s(X)^*_0$, together with the local Lipschitz continuity of $d_S^2$, to establish full Hadamard differentiability. The proof of \Cref{sinkhorn distance hadamard diff} is provided in \Cref{proof of sinkhorn distance hadamard diff}.

\begin{lemma}[Hadamard differentiability of the Sinkhorn distance]
\label{sinkhorn distance hadamard diff}
Under the assumptions of \Cref{first variation sinkhorn distance,linearizing logmap}, the map $(\mu,\nu)\mapsto d_S^2(\mu,\nu)$ is fully Hadamard differentiable at $(\mu,\nu)$ tangentially to $H^s(X)^*_0\times H^s(X)^*_0$, with bounded linear derivative
\begin{align*}
\operatorname D_{(\gamma,\eta)}d_S^2(\mu,\nu)
=
\frac{\varepsilon}{2}
\left\langle
(\operatorname D_\gamma S_\mu)v,v
\right\rangle_{\mathcal H_c}
+
\varepsilon
\left\langle
S_\mu v,
\operatorname D_{(\gamma,\eta)}\log_\mu(\nu)
\right\rangle_{\mathcal H_c},
\end{align*}
where $v=\log_\mu(\nu)=\dot\beta_0$. In particular, the result holds tangentially to $\mathbb T_\mu^m\times\mathbb T_\nu^m$.
\end{lemma}

\subsection{Central Limit Theorems}\label{sec: clts}

The linearization results of the preceding sections allow us to transfer the empirical process limit to statistical inference for the corresponding plug-in estimators. Let $X_1,\ldots,X_n\sim\mu$, $Y_1,\ldots,Y_n\sim\nu$, and $Z_1,\ldots,Z_n\sim\rho$ be three mutually independent i.i.d.\ samples, with associated empirical measures $\widehat\mu_n$, $\widehat\nu_n$, and $\widehat\rho_n$. Since $m>d/2$, the corresponding empirical processes converge jointly as
\[
    \sqrt n
    \begin{pmatrix}
        \widehat\mu_n-\mu\\
        \widehat\nu_n-\nu\\
        \widehat\rho_n-\rho
    \end{pmatrix}
    \rightsquigarrow
    \begin{pmatrix}
        \mathbb G_\mu\\
        \mathbb G_\nu\\
        \mathbb G_\rho
    \end{pmatrix}
    \qquad\text{in }\ell^\infty(B^m) \times \ell^\infty(B^m)  \times \ell^\infty(B^m) ,
\]
where $\mathbb G_\mu$, $\mathbb G_\nu$, and $\mathbb G_\rho$ are independent tight Brownian bridges. Moreover, by \Cref{brownian bridge concentration}, it holds that $\mathbb{G}_\mu, \mathbb G_\nu$ and $\mathbb G_\rho$ are elements of $\mathbb T_\mu^m$, $\mathbb T_\nu^m$ and $\mathbb T_\rho^m$ almost surely. The preceding linearization results show that the operators considered below are fully Hadamard differentiable with bounded linear derivatives. Their plug-in limit distributions therefore follow immediately from the functional delta method. 

\subsubsection{Geodesics, logarithmic map and Sinkhorn distance}

For the following section, we suppose that $(\beta_t^{\mu, \nu})_{t \in [0,1]}$ is the variational geodesic obtained from \Cref{variational geodesic} with endpoints $\mu, \nu$ such that $(\mu_t)_{t \in [0,1]} = (B^{-1}(\beta_t^{\mu, \nu}))_{t \in [0,1]}$ admits a Lebesgue density uniformly bounded away from zero. Recall that under these assumptions, $(\beta_t^{\mu, \nu})_{t \in [0,1]}$ is second-order Sinkhorn regular and satisfies \eqref{geodesic equations}. Write
$v=\log_\mu(\nu) = \dot \beta_0^{\mu, \nu}$ and define the geodesic
$\beta^{\mu,\nu}=(\beta_t^{\mu,\nu})_{t\in[0,1]}$ by
$\beta_t^{\mu,\nu}=\exp_\mu(v)(t)$. Its natural plug-in estimator is
therefore
$\widehat\beta_t=\exp_{\widehat\mu_n}\!
\bigl(\log_{\widehat\mu_n}(\widehat\nu_n)\bigr)(t)$. We further assume that $(\mu_t)_{t\in[0,1]}$ is the unique constant-speed minimizer of \Cref{variational geodesic} among all admissible probability-measure-valued curves joining $\mu$ and $\nu$.

\begin{thm}[Limit theory for the geodesic, logarithmic map and Sinkhorn distance]
\label{geometric clt}
Suppose $m>d/2$, the assumptions of \Cref{linearizing logmap} hold at the
nonconjugate pair $(\mu,\nu)$ with $\mu \neq \nu$, and the assumptions of
\Cref{linearizing expmap} hold at
$(\mu,v)$, where $v=\log_\mu(\nu)$. Define 
\begin{equation}
\label{log map clt limit}
    \mathbb Z_{\mu,\nu}^{\log}
    =
    \operatorname D_{(\mathbb G_\mu,\mathbb G_\nu)}
        \log_\mu(\nu)
    =
    a_{\mathbb G_\mu,\mathbb G_\nu}^{\mathfrak{T}}
    -
    \langle v,\operatorname D_{\mathbb G_\mu}B_\mu\rangle_{\mathcal H_c}B_\mu,
\end{equation}
where $a_{\mathbb G_\mu,\mathbb G_\nu}^{\mathfrak{T}}$ is defined by \Cref{coordinate log map derivative}. Let $\mathbb Z_{\mu,\nu}^{\exp}=\operatorname D_{(\mathbb G_\mu,\mathbb Z_{\mu,\nu}^{\log})}\exp_\mu(v)$ and define
\begin{equation}
\label{sinkhorn distance clt limit}
    \mathbb Z_{\mu,\nu}^{d}
    =
    \frac{\varepsilon}{2}
    \left\langle
        (\operatorname D_{\mathbb G_\mu}S_\mu)v,
        v
    \right\rangle_{\mathcal H_c}
    +
    \varepsilon
    \left\langle
        S_\mu v,
        \mathbb Z_{\mu,\nu}^{\log}
    \right\rangle_{\mathcal H_c}.
\end{equation}
Then
\begin{equation} 
\sqrt n \begin{pmatrix} 
\widehat\beta-\beta^{\mu,\nu} \\[0.3ex]
\log_{\widehat\mu_n}(\widehat\nu_n)-\log_\mu(\nu) \\[0.3ex]
\tilde{\mathbf g}_{\widehat\mu_n} (\log_{\widehat\mu_n}(\widehat\nu_n), \log_{\widehat\mu_n}(\widehat\nu_n)) - d_S^2(\mu,\nu) \\[0.3ex]
d_S^2(\widehat\mu_n,\widehat\nu_n)-d_S^2(\mu,\nu) 
\end{pmatrix} 
\rightsquigarrow 
\begin{pmatrix} 
\mathbb Z_{\mu,\nu}^{\exp} \\[0.3ex]
\mathbb Z_{\mu,\nu}^{\log} \\[0.3ex]
\mathbb Z_{\mu,\nu}^{d}  \\[0.3ex]
\mathbb Z_{\mu,\nu}^{d} 
\end{pmatrix} 
\end{equation}
in
$C^1([0,1];\mathcal H_c)\times\mathcal H_c\times\mathbb R \times \mathbb R$ where the limiting element is a centered Gaussian with dependent entries. In particular,
$\mathbb Z_{\mu,\nu}^{d}\sim N(0,\sigma_{\mu,\nu}^2)$, where
$\sigma_{\mu,\nu}^2
=\operatorname{Var}(\mathbb Z_{\mu,\nu}^{d})$.
\end{thm}

\begin{proof}
Define
\[
\Theta(\mu,\nu)
=
\left(
    \beta^{\mu,\nu},
    \log_\mu(\nu),
    \tilde{\mathbf g}_\mu(\log_\mu(\nu),\log_\mu(\nu)),
    d_S^2(\mu,\nu)
\right)
\]
as a map into $\mathbb Y:=C^1([0,1];\mathcal H_c)\times\mathcal H_c\times\mathbb R^2$. By \Cref{linearizing logmap}, the map $(\mu,\nu)\mapsto\log_\mu(\nu)$ is fully Hadamard differentiable at $(\mu,\nu)$ tangentially to $\mathbb T_\mu^m\times\mathbb T_\nu^m$. Since $\beta^{\mu,\nu}=\exp_\mu(\log_\mu(\nu))$, \Cref{linearizing expmap} together with the Hadamard chain rule implies that $(\mu,\nu)\mapsto\beta^{\mu,\nu}$ is fully Hadamard differentiable, with derivative in the direction $(\gamma,\eta)$ given by
\[
    \operatorname D_{(\gamma,\operatorname D_{(\gamma,\eta)}\log_\mu(\nu))}
        \exp_\mu(v),
    \qquad v=\log_\mu(\nu).
\]
Moreover, \Cref{metric hadamard diff}, \Cref{linearizing logmap} and the Hadamard chain rule imply that the map $(\mu,\nu)\mapsto\tilde{\mathbf g}_\mu(\log_\mu(\nu),\log_\mu(\nu))$ is fully Hadamard differentiable, with derivative
\[
    \frac{\varepsilon}{2}
    \left\langle
        (\operatorname D_\gamma S_\mu)v,v
    \right\rangle_{\mathcal H_c}
    +
    \varepsilon
    \left\langle
        S_\mu v,
        \operatorname D_{(\gamma,\eta)}\log_\mu(\nu)
    \right\rangle_{\mathcal H_c}.
\]
By \Cref{sinkhorn distance hadamard diff}, the squared Sinkhorn distance is fully Hadamard differentiable with the same derivative. Furthermore, \Cref{norm and distance coincide} gives $\tilde{\mathbf g}_\mu(v,v)=d_S^2(\mu,\nu)$. Consequently, $\Theta$ is fully Hadamard differentiable at $(\mu,\nu)$, tangentially to $\mathbb T_\mu^m\times\mathbb T_\nu^m$, with bounded linear derivative given by the four components above. Since $m>d/2$, the joint empirical process satisfies
\[
    \sqrt n
    \left(
        \widehat\mu_n-\mu,
        \widehat\nu_n-\nu
    \right)
    \rightsquigarrow
    (\mathbb G_\mu,\mathbb G_\nu)
    \qquad\text{in }\ell^\infty(B^m)\times\ell^\infty(B^m).
\]
\Cref{brownian bridge concentration} and the functional delta method (\Cref{functional delta method} or \citet[Theorem~3.9.4]{vanderVaart1996weak}) therefore yield 
\[
    \sqrt n
    \left(
        \Theta(\widehat\mu_n,\widehat\nu_n)
        -
        \Theta(\mu,\nu)
    \right)
    \rightsquigarrow
    \operatorname D_{(\mathbb G_\mu,\mathbb G_\nu)}
        \Theta(\mu,\nu).
\]
The four components of the derivative are precisely $\mathbb Z_{\mu,\nu}^{\exp}$, $\mathbb Z_{\mu,\nu}^{\log}$, $\mathbb Z_{\mu,\nu}^{d}$, and $\mathbb Z_{\mu,\nu}^{d}$. Since $(\mathbb G_\mu,\mathbb G_\nu)$ is centered Gaussian and $\operatorname D\Theta$ is bounded and linear, the limiting random element is centered Gaussian. In particular, $\mathbb Z_{\mu,\nu}^{d}$ is a centered Gaussian random variable with variance $\sigma_{\mu,\nu}^2$. Moreover, since the last two components have the same limit, their difference satisfies
\[
    \tilde{\mathbf g}_{\widehat\mu_n}
    (\log_{\widehat\mu_n}(\widehat\nu_n),
    \log_{\widehat\mu_n}(\widehat\nu_n))
    -d_S^2(\widehat\mu_n,\widehat\nu_n)
    =o_{\mathbb P}(n^{-1/2}).
\] \end{proof}

\begin{rmk}[Two plug-in estimators of the Sinkhorn distance]
\label{two sinkhorn distance estimators}
\Cref{geometric clt} provides two asymptotically equivalent estimators of the squared Sinkhorn distance. The first evaluates the empirical metric tensor at the empirical logarithmic map, which can be obtained by solving the geodesic boundary-value problem in the ambient RKHS. The second directly evaluates the intrinsic Sinkhorn distance between the empirical measures, which admits a variational characterization through \eqref{variational geodesic}. Although the two estimators need not coincide at finite sample sizes, their difference is $o_{\mathbb P}(n^{-1/2})$, and both admit the same limiting Gaussian distribution. 

\end{rmk}

\begin{rmk}[Convergence of the induced distribution-valued curve] \label{distribution valued convergence}
The preceding limit theorem is stated in the $\beta$-parameterization, but it also yields a limit theorem in the original distributional coordinates. Write $\mu_t=M(\beta_t)$ for the population geodesic and define the empirical interpolant by $\widehat\mu_{n,t}=M(\widehat\beta_{n,t})$, where $M(\beta)=\beta H_c^{-1}(\beta)$. \Cref{smoothness of geodesic coefficients}, the functional delta method and \Cref{geometric clt} give
\[
(\sqrt n(\widehat\mu_{n,t}-\mu_t))_{t \in [0,1]}
\rightsquigarrow
\left(\operatorname D_{h_t}M(\beta_t)\right)_{t\in[0,1]}
\quad\text{in }C^1([0,1];H^s(X)^*),
\]
where $h_t=\mathbb Z_{\mu,\nu}^{\exp}(t)$ and
$\operatorname D_{h_t}M(\beta_t)=h_tH_c^{-1}(\beta_t)+\beta_tH_c^{-1}(h_t)$. In particular,
\[
\sup_{t\in[0,1]}
\|\widehat\mu_{n,t}-\mu_t\|_{H^s(X)^*}
=O_{\mathbb P}(n^{-1/2}).
\]
Although $\widehat\mu_{n,t}(\mathbbm{1}_X)=1$ for every $t\in[0,1]$, the empirical geodesic $\widehat\mu_{n,t}$ is \emph{not} guaranteed to be a finite signed measure. Thus the convergence above takes place in the Sobolev dual $H^s(X)^*$ rather than in $(\mathcal P(X),d_S)$, while the population curve $(\mu_t)_{t\in[0,1]}$ indeed remains probability-measure-valued. 
In particular, let $\Phi:C^1([0,1];H^s(X)^*)\to C^1([0,1];\mathbb R)$ be any fully Hadamard differentiable functional at $\boldsymbol\mu=(\mu_t)_{t\in[0,1]}$. Writing $\widehat{\boldsymbol\mu}_n=(\widehat\mu_{n,t})_{t\in[0,1]}$, the functional delta method gives
\[
\sqrt n\left(\Phi(\widehat{\boldsymbol\mu}_n)-\Phi(\boldsymbol\mu)\right)
\rightsquigarrow
\operatorname D_{(\operatorname D_{h_t}M(\beta_t))_{t\in[0,1]}}\Phi(\boldsymbol\mu)
\quad\text{in }C^1([0,1];\mathbb R),
\]
where $h_t=\mathbb Z_{\mu,\nu}^{\exp}(t)$ and the limiting random element is a centered Gaussian process.

\end{rmk}

For the corresponding bootstrap result, let
$\widehat\mu_n^B$ and $\widehat\nu_n^B$ denote empirical measures obtained
from independent size-$n$ nonparametric bootstrap samples drawn from
$\widehat\mu_n$ and $\widehat\nu_n$, conditional on the original data.
Define $\mathbb Y
    =
    C^1([0,1];\mathcal H_c)
    \times\mathcal H_c
    \times\mathbb R \times \mathbb R$, and
\[
\Theta(\mu,\nu)
=
\left(
    \beta^{\mu,\nu},
    \log_\mu(\nu),
    \tilde{\mathbf g}_\mu(\log_\mu(\nu),\log_\mu(\nu)),
    d_S^2(\mu,\nu)
\right).
\]
Write $\widehat\Theta_n=\Theta(\widehat\mu_n,\widehat\nu_n)$ and
$\widehat\Theta_n^B
=\Theta(\widehat\mu_n^B,\widehat\nu_n^B)$.

\begin{thm}[Bootstrap consistency for the geodesic, logarithmic map and Sinkhorn distance]
\label{bootstrap geometric maps}
Under the assumptions of \Cref{geometric clt}, write
$\mathbb Z_{\mu,\nu}
=
(\mathbb Z_{\mu,\nu}^{\exp},
\mathbb Z_{\mu,\nu}^{\log},
\mathbb Z_{\mu,\nu}^{d},
\mathbb Z_{\mu,\nu}^{d})$.
Then $\sqrt n(\widehat\Theta_n^B-\widehat\Theta_n)
\rightsquigarrow_B \mathbb Z_{\mu,\nu}$ in outer probability. Equivalently,
\begin{equation}
\label{bootstrap geometric convergence}
    \sup_{h\in\operatorname{BL}_1(\mathbb Y)}
    \left|
        \mathbb E_B
        \left[
            h\left(
                \sqrt n
                (\widehat\Theta_n^B-\widehat\Theta_n)
            \right)
        \right]
        -
        \mathbb E
        \left[
            h(\mathbb Z_{\mu,\nu})
        \right]
    \right|
    \to 0
\end{equation}
in outer probability as $n\to\infty$, where $\mathbb E_B$ denotes conditional expectation given the original samples and $\operatorname{BL}_1(\mathbb Y)$ denotes the class of functions $h:\mathbb Y\to[-1,1]$ with Lipschitz constant at most one.
\end{thm}

\begin{proof}
Since $B^m$ is Donsker and the two bootstrap samples are drawn independently conditional on the original data, the nonparametric bootstrap empirical processes satisfy
\[
    \sqrt n
    \left(
        \widehat\mu_n^B-\widehat\mu_n,
        \widehat\nu_n^B-\widehat\nu_n
    \right)
    \rightsquigarrow_B
    (\mathbb G_\mu,\mathbb G_\nu)
    \qquad\text{in }\ell^\infty(B^m)\times\ell^\infty(B^m),
\]
in outer probability by Theorem 3.6.1 of \citet{vanderVaart1996weak}. By the proof of \Cref{geometric clt}, the map $\Theta$ is fully Hadamard differentiable at $(\mu,\nu)$ tangentially to $\mathbb T_\mu^m\times\mathbb T_\nu^m$, with bounded linear derivative $\operatorname D\Theta$. In particular, \Cref{sinkhorn distance hadamard diff} guarantees that the last two components have the same derivative.

The bootstrap functional delta method \citep[Theorem~3.9.11]{vanderVaart1996weak} therefore gives conditional weak convergence of $\sqrt n(\widehat\Theta_n^B-\widehat\Theta_n)$ to
$\operatorname D_{(\mathbb G_\mu,\mathbb G_\nu)}
\Theta(\mu,\nu)=\mathbb Z_{\mu,\nu}$. This can be equivalently written as
\[
    \sup_{h\in\operatorname{BL}_1(\mathbb Y)}
    \left|
        \mathbb E_B
        \left[
            h\left(
                \sqrt n(\widehat\Theta_n^B-\widehat\Theta_n)
            \right)
        \right]
        -
        \mathbb E
        \left[
            h(\mathbb Z_{\mu,\nu})
        \right]
    \right|
    \to 0
\]
in outer probability, which is \Cref{bootstrap geometric convergence}.
\end{proof}

\begin{corollary}[Bootstrap consistency for functionals of the geodesic]
\label{bootstrap functionals geodesic}
Under the same assumptions, define the bootstrap geodesic and its image by
\[
    \widehat\beta_n^B
    :=\exp_{\widehat\mu_n^B}\!\bigl(\log_{\widehat\mu_n^B}(\widehat\nu_n^B)\bigr),
    \qquad
    \widehat{\boldsymbol\mu}_n^B
    :=\bigl(M(\widehat\beta_n^B(t))\bigr)_{t\in[0,1]}.
\]
Then the bootstrap functional delta method gives
\[
\sqrt n\left(\Phi(\widehat{\boldsymbol\mu}_n^B)-\Phi(\widehat{\boldsymbol\mu}_n)\right)
\rightsquigarrow_B
\operatorname D_{(\operatorname D_{h_t}M(\beta_t))_{t\in[0,1]}}\Phi(\boldsymbol\mu)
\quad\text{in }C^1([0,1];\mathbb R),
\]
in outer probability, where $h_t=\mathbb Z_{\mu,\nu}^{\exp}(t)$.
\end{corollary}
\begin{proof}
By \Cref{smoothness of geodesic coefficients}, the map $\boldsymbol\beta\mapsto(M(\beta_t))_{t\in[0,1]}$ is continuously Fr\'echet differentiable from $C^1([0,1];\mathcal H_c)$ into $C^1([0,1];H^s(X)^*)$, with derivative $(h_t)_{t\in[0,1]}\mapsto(\operatorname D_{h_t}M(\beta_t))_{t\in[0,1]}$. The result therefore follows from \Cref{bootstrap geometric maps} and the bootstrap functional delta method.
\end{proof}

\subsubsection{Parallel imputation via parallel transport}

For parallel imputation, let
$X_1,\ldots,X_n\sim\mu$,
$Y_1,\ldots,Y_n\sim\nu$, and
$Z_1,\ldots,Z_n\sim\rho$ be three mutually independent i.i.d.\ samples
with empirical measures
$\widehat\mu_n$, $\widehat\nu_n$, and $\widehat\rho_n$. Recall that the
parallel imputation map returns the entire imputed geodesic,
\[
    \operatorname{Imp}(\mu,\nu,\rho)
    =
    \exp_\nu\left(
        \operatorname{PT}_{\mu\to\nu}
        (\log_\mu(\rho))
    \right)
    \in C^1([0,1];\mathcal H_c),
\]
where, as throughout this section, $\exp_\nu(q)$ denotes the entire curve,
whose value at time $t$ is $\exp_\nu(q)(t)$.

\begin{thm}[Parallel imputation CLT]
\label{parallel imputation clt}
Suppose $m>d/2$ and that the assumptions of
\Cref{linearizing imp} hold at $(\mu,\nu,\rho)$. Then
\begin{equation}
    \sqrt n
    \left(
        \operatorname{Imp}
        (\widehat\mu_n,\widehat\nu_n,\widehat\rho_n)
        -
        \operatorname{Imp}(\mu,\nu,\rho)
    \right)
    \rightsquigarrow
    \mathbb Z_{\mu,\nu,\rho}^{\operatorname{Imp}}
    \qquad\text{in }C^1([0,1];\mathcal H_c),
\end{equation}
where $\mathbb Z_{\mu,\nu,\rho}^{\operatorname{Imp}}
    =
    \operatorname D_{(
        \mathbb G_\mu,
        \mathbb G_\nu,
        \mathbb G_\rho)}
        \operatorname{Imp}(\mu,\nu,\rho).$ More explicitly, let
$q=\operatorname{PT}_{\mu\to\nu}(\log_\mu(\rho))$, and let
$(k_t)_{t\in[0,1]}$ denote the variation of the parallel transported
field characterized in \Cref{linearizing imp} under the perturbation
$(\mathbb G_\mu,\mathbb G_\nu,\mathbb G_\rho)$. Then
\begin{equation}
\label{parallel imputation clt limit}
    \mathbb Z_{\mu,\nu,\rho}^{\operatorname{Imp}}
    =
    \operatorname D_{(\mathbb G_\nu,k_1)}
        \exp_\nu(q)
    \qquad\text{in }C^1([0,1];\mathcal H_c).
\end{equation}
In particular,
$\mathbb Z_{\mu,\nu,\rho}^{\operatorname{Imp}}$ is a centered Gaussian
random element of $C^1([0,1];\mathcal H_c)$.
\end{thm}

\begin{proof}
By \Cref{linearizing imp}, the map
$(\mu,\nu,\rho)\mapsto\operatorname{Imp}(\mu,\nu,\rho)$ is fully
Hadamard differentiable at $(\mu,\nu,\rho)$ tangentially to
$\mathbb T_\mu^m\times\mathbb T_\nu^m\times\mathbb T_\rho^m$, with
bounded linear derivative taking values in
$C^1([0,1];\mathcal H_c)$. Since the three samples are mutually
independent,
\[
    \sqrt n
    \left(
        \widehat\mu_n-\mu,
        \widehat\nu_n-\nu,
        \widehat\rho_n-\rho
    \right)
    \rightsquigarrow
    (\mathbb G_\mu,\mathbb G_\nu,\mathbb G_\rho)
    \qquad\text{in }\ell^\infty(B^m) \times \ell^\infty(B^m) \times \ell^\infty(B^m).
\]
\Cref{brownian bridge concentration} and the functional delta method
\citep[Theorem~3.9.4]{vanderVaart1996weak} therefore yield
\[
    \sqrt n
    \left(
        \operatorname{Imp}
        (\widehat\mu_n,\widehat\nu_n,\widehat\rho_n)
        -
        \operatorname{Imp}(\mu,\nu,\rho)
    \right)
    \rightsquigarrow
    \operatorname D_{(
        \mathbb G_\mu,
        \mathbb G_\nu,
        \mathbb G_\rho)}
        \operatorname{Imp}(\mu,\nu,\rho)
\]
in $C^1([0,1];\mathcal H_c)$. By \Cref{linearizing imp}, the
perturbation of the transported tangent
$q=\operatorname{PT}_{\mu\to\nu}(\log_\mu(\rho))$ is $k_1$. Applying
the linearization of the exponential map to the perturbed initial data
$(\nu,q)$ therefore gives
\[
    \operatorname D_{(
        \mathbb G_\mu,
        \mathbb G_\nu,
        \mathbb G_\rho)}
        \operatorname{Imp}(\mu,\nu,\rho)
    =
    \operatorname D_{(\mathbb G_\nu,k_1)}
        \exp_\nu(q),
\]
which proves \Cref{parallel imputation clt limit}. Since this derivative
is bounded and linear and
$(\mathbb G_\mu,\mathbb G_\nu,\mathbb G_\rho)$ is jointly centered
Gaussian, the limiting random element is centered Gaussian in
$C^1([0,1];\mathcal H_c)$.
\end{proof}

\begin{rmk}[Distribution-valued parallel imputation]
\label{distribution valued parallel imputation}
The preceding limit theorem also admits an interpretation in distributional coordinates, in direct analogy with that discussed in \Cref{distribution valued convergence}. Write $I_t=\operatorname{Imp}(\mu,\nu,\rho)(t)$, $\lambda_t=M(I_t)$, and $\widehat\lambda_{n,t}=M(\operatorname{Imp}(\widehat\mu_n,\widehat\nu_n,\widehat\rho_n)(t))$. Since $M$ is continuously Fr\'echet differentiable by \Cref{smoothness of geodesic coefficients}, the functional delta method and \Cref{parallel imputation clt} give
\[
\sqrt n(\widehat{\boldsymbol\lambda}_n-\boldsymbol\lambda)
\rightsquigarrow
\mathbb Z_{\mu,\nu,\rho}^{\lambda}
\quad\text{in }C^1([0,1];H^s(X)^*),
\]
where
$\mathbb Z_{\mu,\nu,\rho}^{\lambda}
=(\operatorname D_{\mathbb Z_{\mu,\nu,\rho}^{\operatorname{Imp}}(t)}M(I_t))_{t\in[0,1]}$.
Although $\lambda_t(\mathbbm{1}_X)=\widehat\lambda_{n,t}(\mathbbm{1}_X)=1$, neither curve is guaranteed to be probability-measure-valued, or even finite signed measure-valued. Unlike the geodesic interpolant, this qualification applies even to the \emph{population} imputation curve. 
Nevertheless, for any fully Hadamard differentiable functional $\Phi:C^1([0,1];H^s(X)^*)\to C^1([0,1];\mathbb R)$ at $\boldsymbol\lambda$, the functional delta method gives
\[
\sqrt n\left(\Phi(\widehat{\boldsymbol\lambda}_n)-\Phi(\boldsymbol\lambda)\right)
\rightsquigarrow
\operatorname D_{\mathbb Z_{\mu,\nu,\rho}^{\lambda}}\Phi(\boldsymbol\lambda)
\quad\text{in }C^1([0,1];\mathbb R),
\]
where the limiting random element is a centered Gaussian process.
\end{rmk}

Let $\widehat\mu_n^B$, $\widehat\nu_n^B$, and
$\widehat\rho_n^B$ denote the empirical measures obtained from
independent size-$n$ nonparametric bootstrap samples from
$\widehat\mu_n$, $\widehat\nu_n$, and $\widehat\rho_n$, respectively,
conditional on the original samples. Write
$\widehat I_n=\operatorname{Imp}
(\widehat\mu_n,\widehat\nu_n,\widehat\rho_n)$ for the imputed curve
obtained from the original samples and let
$\widehat I_n^B=\operatorname{Imp}
(\widehat\mu_n^B,\widehat\nu_n^B,\widehat\rho_n^B)$ denote its bootstrap
analogue.

\begin{thm}[Bootstrap consistency for parallel imputation]
\label{bootstrap parallel imputation}
Under the assumptions of \Cref{parallel imputation clt},
\begin{equation}
\label{bootstrap parallel imputation convergence}
    \sup_{h\in\operatorname{BL}_1(C^1([0,1];\mathcal H_c))}
    \left|
        \mathbb E_B
        \left[
            h\left(
                \sqrt n
                (\widehat I_n^B-\widehat I_n)
            \right)
        \right]
        -
        \mathbb E
        \left[
            h\left(
                \mathbb Z_{\mu,\nu,\rho}^{\operatorname{Imp}}
            \right)
        \right]
    \right|
    \to 0
\end{equation}
in outer probability.
\end{thm}

\begin{proof}
Since $B^m$ is Donsker and the three bootstrap samples are drawn
independently conditional on the original data, Theorem~3.6.1 of
\citet{vanderVaart1996weak} yields
\[
    \sqrt n
    \left(
        \widehat\mu_n^B-\widehat\mu_n,
        \widehat\nu_n^B-\widehat\nu_n,
        \widehat\rho_n^B-\widehat\rho_n
    \right)
    \rightsquigarrow_B
    (\mathbb G_\mu,\mathbb G_\nu,\mathbb G_\rho)
    \qquad\text{in }\ell^\infty(B^m)\times \ell^\infty(B^m) \times \ell^\infty(B^m)
\]
in outer probability. By \Cref{linearizing imp}, the parallel
imputation map is fully Hadamard differentiable at $(\mu,\nu,\rho)$,
with bounded linear derivative taking values in
$C^1([0,1];\mathcal H_c)$. Hence the bootstrap functional delta method
\citep[Theorem~3.9.11]{vanderVaart1996weak} gives
\[
    \sup_{h\in\operatorname{BL}_1(C^1([0,1];\mathcal H_c))}
    \left|
        \mathbb E_B
        \left[
            h\left(
                \sqrt n(\widehat I_n^B-\widehat I_n)
            \right)
        \right]
        -
        \mathbb E
        \left[
            h\left(
                \mathbb Z_{\mu,\nu,\rho}^{\operatorname{Imp}}
            \right)
        \right]
    \right|
    \to 0
\]
in outer probability, proving
\Cref{bootstrap parallel imputation convergence}.
\end{proof}

\begin{rmk}[Bootstrap inference for functionals of parallel imputation]
\label{bootstrap functionals parallel imputation}
The bootstrap consistency in \Cref{bootstrap parallel imputation} extends to the induced distribution-valued curve and smooth functionals of it. Define $\widehat\lambda_{n,t}^B=M(\widehat I_n^B(t))$, and note that by the smoothness of $M$ established in \Cref{smoothness of geodesic coefficients} and the bootstrap functional delta method, we have
\[
\sqrt n(\widehat{\boldsymbol\lambda}_n^B-\widehat{\boldsymbol\lambda}_n)
\rightsquigarrow_B
\mathbb Z_{\mu,\nu,\rho}^{\lambda}
\quad\text{in }C^1([0,1];H^s(X)^*)
\]
in outer probability. Thus, for any fully Hadamard differentiable functional $\Phi:C^1([0,1];H^s(X)^*)\to C^1([0,1];\mathbb R)$ at $\boldsymbol\lambda$,
\[
\sqrt n\left(
\Phi(\widehat{\boldsymbol\lambda}_n^B)
-\Phi(\widehat{\boldsymbol\lambda}_n)
\right)
\rightsquigarrow_B
\operatorname D_{\mathbb Z_{\mu,\nu,\rho}^{\lambda}}\Phi(\boldsymbol\lambda)
\quad\text{in }C^1([0,1];\mathbb R),
\]
in outer probability. Note that bootstrap inference for these functionals does not require the imputed curves to be probability-measure-valued. We leave a more detailed investigation of the behavior of functionals of parallel-imputation curves, such as their means and higher-order moments, and the development of corresponding application-specific inferential procedures to future work. \end{rmk}

\section{Discussion}
In this work, we extended the Riemannian geometry of the Sinkhorn divergence by deriving the Levi-Civita covariant derivative and the associated geodesic and parallel transport equations. Our strategy centered around proving a sufficiently strong notion of differentiability to both enable the construction of covariant differentiation (\Cref{sec: Sinkhorn riemannian geometry}) and obtain limit theorems for relevant Sinkhorn geometric objects (\Cref{sec: clts}). These results leave several directions for future work. From a statistical and computational perspective, the plug-in estimators studied here remain infinite-dimensional objects, motivating the development of computationally tractable finite-dimensional approximations to the geodesic, logarithmic map, and parallel transport equations. Of particular interest would be approximation schemes whose error is $o_{\mathbb P}(n^{-1/2})$, so that the limiting distributions and bootstrap guarantees established here are inherited by their computational implementations. From an analytical perspective, it remains to understand the extent to which some of our regularity assumptions are intrinsic to the geometry. In particular, our lower bound on the density (\Cref{standing sinkhorn assumptions}.3) removes many singularities and boundary issues in the geometry; determining whether this assumption can be weakened, or whether genuinely new phenomena arise near measures whose densities vanish, is a relevant analytical open problem. Finally, these developments suggest several applied directions. Combining computational approximations of the geometric objects with the root-$n$ inference developed here could provide uncertainty quantification for procedures based on distributional interpolation and trajectory estimation. Single-cell biology, where optimal transport is frequently used to interpolate between observed cell populations and infer developmental trajectories, provides one particularly natural setting in which such inferential procedures could be useful.

\section{Acknowledgments}
The authors would like to thank Dejan Slep\v{c}ev and Hugo Lavenant for their suggestions regarding analytical results, and Larry Wasserman for his comments on the overall direction of this manuscript. The authors would also like to thank Nikola Pavlovic for fruitful conversations regarding RKHS regularity. Finally, the authors acknowledge that ChatGPT 5.6 Sol was used extensively to assist with proof auditing and writing, though the authors take full responsibility for the results presented in this manuscript.

\break

\begin{appendices}

\section{Technical tools}

\subsection{Normed-space Differentiability}\label{normed space differentiability}

In this section, suppose that $\mathbb{D}, \mathbb{B}$ are two (possibly infinite-dimensional) normed spaces (unless otherwise specified)  with $\phi: \mathbb{D} \to \mathbb{B}$ a map between them. Following the notation and definitions of \citet{van2000asymptotic}, we will examine different notions of differentiability of $\phi$.

\paragraph{G\^ateaux Differentiability.} We will start with the weakest notion of differentiability. The map $\phi: \mathbb{D} \to \mathbb{B}$ is said to be \emph{G\^ateaux differentiable} at some $\theta \in \mathbb{D}$ if for every fixed $\gamma \in \mathbb{D}$ there exists an element $\text{D}_{\gamma}\phi_{\theta} \in \mathbb{B}$ such that
\[\left\|\frac{\phi(\theta + t\gamma) - \phi(\theta)}{t} - \text{D}_{\gamma}\phi_{\theta}\right\|_{\mathbb{B}} \to 0\]
as $t \downarrow 0.$ G\^ateaux differentiability is often considered \say{directional differentiability} for normed spaces.

\paragraph{Hadamard differentiability.} In some cases, one may require a stronger notion of differentiability than that of G\^ateaux. Again, let $\mathbb{D}, \mathbb{B}$ be normed spaces, and let $\phi: \Theta \subset \mathbb{D} \to \mathbb{B}$ be a map. Define $\mathfrak{T}_\Theta(\theta)$ to be the tangent cone to $\Theta$ at $\theta$,
\[\mathfrak{T}_{\Theta}(\theta) = \left\{ \gamma \in \mathbb{D} : \gamma = \lim_{t \downarrow 0}\frac{\theta_t - \theta}{t}\, \text{ for some $\theta_t \to \theta$ in $\Theta$ with $t \downarrow 0$}\right\}.\]
We say that $\phi$ is \emph{Hadamard directionally differentiable} at $\theta \in \Theta$ if there exists a map $\text{D}_{(\cdot)}\phi_{\theta}: \mathfrak{T}_\Theta(\theta) \to \mathbb{B}$ such that 
\[\left\| \frac{\phi(\theta + t\gamma_t) - \phi(\theta)}{t} - \text{D}_{\gamma}\phi_{\theta}\right\|_{\mathbb{B}} \to 0 \]
for any sequence $(\theta + t\gamma_t)_{t > 0} \subset \Theta$ with $\gamma_t \to \gamma$ as $t \downarrow 0$. If the limit above holds only for a subset $\mathfrak{D}_0 \subset \mathfrak{T}_\Theta(\theta)$ then we say that $\phi$ is Hadamard directionally differentiable at $\theta$ tangentially to $\mathfrak{D}_0$, in which case the derivative $\text{D}_{(\cdot)}\phi_{\theta}$ is defined only on $\mathfrak{D}_0.$ Moreover, if the map $\operatorname{D}_{\gamma}\phi_\theta$ is also linear then we say $\phi$ is \textit{Hadamard differentiable.}

\begin{lemma}[Functional delta method, \cite{romisch2014delta}] \label{functional delta method}
    Let $\mathbb{D}, \mathbb{B}$ be normed spaces and let $\phi: \Theta \subset \mathbb{D} \to \mathbb{B}$ be a map that is Hadamard directionally differentiable at a $\theta \in \Theta$ tangentially to a set $\mathfrak{D}_0 \subset \mathfrak{T}_\Theta (\theta)$. Let $T_n: \Omega \to \Theta$ be maps such that $r_n(T_n - \theta) \overset{d}{\rightarrow} T$ for some $r_n \to \infty$ and some Borel measurable map $T: \Omega \to \mathbb{D}$ with values in a separable subset of $\mathfrak{D}_0$. Then it holds that $r_n(\phi(T_n) - \phi(\theta)) \overset{d}{\rightarrow } \operatorname{D}_{T}\phi_{\theta}.$
\end{lemma}

\paragraph{Fr\'echet differentiability.} The final notion of differentiability we will discuss is the strongest. Let $\mathbb{D}$ and $\mathbb{B}$ be two normed spaces and let $\Theta$ be a nonempty subset of $\mathbb{D}$. A map $\phi: \Theta \to \mathbb{B}$ is said to be \emph{Fr\'echet differentiable} at a $\theta \in \Theta$ if 
\[\lim_{\substack{\|\gamma\|_{\mathbb{D}} \to 0 \\ \theta + \gamma \in \Theta}}\frac{\|\phi(\theta + \gamma) - \phi(\theta) - \text{D}_{\gamma}\phi_{\theta}\|_{\mathbb{B}}}{\|\gamma\|_{\mathbb{D}}} = 0\]
for some continuous linear map $\gamma \mapsto \text{D}_{\gamma}\phi_{\theta}$ \citep{van2000asymptotic}.

\begin{prop}[\citet{hajek2014smooth}, Proposition 79] \label{multilinear hadamard diff}
    Let $X_1,\ldots,X_n$ and $Y$ be normed linear spaces, and let
    \[
    M:X_1\times\cdots\times X_n\to Y
    \]
    be a continuous $n$-linear map. Then $M$ is infinitely Fr\'echet
    differentiable as a map from
    \[
    X:=X_1\oplus\cdots\oplus X_n
    \]
    into $Y$. In particular, for
    $x=(x_1,\ldots,x_n)\in X$ and
    $\gamma=(\gamma_1,\ldots,\gamma_n)\in X$,
    \[
    \operatorname{D}_{\gamma}M(x)
    =
    \sum_{j=1}^n
    M(x_1,\ldots,x_{j-1},\gamma_j,x_{j+1},\ldots,x_n).
    \]
    Moreover, $\operatorname{D}^kM\equiv 0$ for every $k>n$. Consequently, every
    continuous multilinear map between normed linear spaces is Hadamard
    differentiable.
\end{prop}

\section{Proofs for \Cref{sec: Sinkhorn riemannian geometry}} \label{sec: proofs for Sinkhorn riemannian geometry}

\subsection{Proof of \Cref{uniqueness lemma}}

Note that the symmetry of $c$ straightforwardly guarantees that $(g_{\mu, \nu}, f_{\mu, \nu})$ are solutions to the Schr\"odinger system for $(\nu, \mu)$. Let $(f_{\nu,\mu},g_{\nu,\mu})$ be the independently normalized potentials for $(\nu,\mu)$, satisfying $f_{\nu,\mu}(x_0)=g_{\nu,\mu}(x_0)$. By the uniqueness of the potential pairs modulo constants, we know that there exist $\lambda_1, \lambda_2 \in \mathbb{R}$ such that
    \[g_{\nu, \mu} = f_{\mu, \nu} + \lambda_1 \quad \text{and} \quad f_{\nu, \mu} = g_{\mu, \nu} + \lambda_2. \]
    In fact, satisfaction of the Schr\"odinger system guarantees that $\lambda_1 = -\lambda_2$, allowing us to redefine $\lambda := \lambda_1 = -\lambda_2.$ Evaluating this system at $x_0 \in X$, combining the equations, and using the representative pairs characterized by $f_{\mu, \nu}(x_0) = g_{\mu, \nu}(x_0)$, $f_{\nu, \mu}(x_0) = g_{\nu, \mu}(x_0)$ implies
    \[f_{\mu, \nu}(x_0)  - g_{\mu, \nu}(x_0) = -2\lambda.\]
    By our selection principle the left side is zero, and thus $\lambda = 0$. Thus, we can conclude that $f_{\mu, \nu} = g_{\nu, \mu}$ as desired. \qed{}

\subsection{Proofs for \Cref{sec: parameterizing action}}

\begin{prop} \label{well defined r j}
    The operators $R_{\mu}$ and $J_{\mu}$ are independent of the chosen representative from the equivalence class $\mathcal{Q}_{\mu}$ and are therefore well-defined maps on $\mathcal{Q}_{\mu}$.
\end{prop}

\begin{proof}
Due to \Cref{Qt existence}, $(\operatorname{id}_{\mathcal{Q}_{\mu}} - K_{\mu})^{-1}$ is also a well defined operator on $\mathcal{Q}_{\mu}$, which completes the proof for $R_{\mu}$. Now we'll prove that $J_{\mu}$ is well defined. Let $[z] \in \mathcal{Q}_{\mu}$ and let $v, w$ be two distinct representatives for $[z].$ Thus, there exists a $\lambda \in \mathbb{R}$ such that $v \equiv w + \lambda$. It follows that 
\begin{align*}
    \Pi_{\mu}(a_{\mu}^{-1}v) &= \Pi_{\mu}(a_{\mu}^{-1} w + \lambda a_{\mu}^{-1} \mathbbm{1}_X) = \Pi_{\mu}(a_{\mu}^{-1} w) + \lambda \Pi_{\mu}(a_{\mu}^{-1} ).
\end{align*}
Since $a_{\mu}^{-1} \in E_{\mu}$ by construction, the second term on the right hand side is $0.$ Hence $\Pi_{\mu}(a_{\mu}^{-1}v) = \Pi_{\mu}(a_{\mu}^{-1}w)$ and $J_{\mu}[z]$ is independent of the choice of representative for $[z].$ 
\end{proof}

\begin{lemma} \label{theta inverse}
    The map $\Theta_{\mu}: \mathcal{H}_{\mu}/\mathbb{R} \to \mathfrak{T}_\mu \mathcal{P}(X)$ is well defined, and it admits an inverse $\Theta_{\mu}^{-1}: \mathfrak{T}_\mu \mathcal{P}(X)  \to \mathcal{H}_{\mu}/\mathbb{R}$ given by $\Theta_{\mu}^{-1} u  = [\iota_{\mu} u] = [a_{\mu}u]$. Consequently, the pulled back operator $\mathbf{J}_{\mu} = J_{\mu}\Theta_{\mu}^{-1}\Pi_{\mu}$ satisfies $\mathbf{J}_{\mu}u = H_c(a_{\mu}^2 \Pi_{\mu}u \, \mu)$ for any $u \in \mathcal{H}_c.$
\end{lemma}

\begin{proof}
    Recall that $\Theta_{\mu}[f] = \Pi_{\mu} \iota_{\mu}^{-1}f$, and fix a $u \in \mathfrak{T}_\mu \mathcal{P}(X)$. We will first show that the value of $\Theta_{\mu}$ is independent of the equivalence class representative chosen. To see this, fix an $f \in \mathcal{H}_\mu$, $\lambda \in \mathbb{R}$ and note that
\begin{align*}
    \Pi_{\mu}\iota_{\mu}^{-1}(f + \lambda \mathbbm{1}_X) &= \Pi_{\mu}\iota_{\mu}^{-1} f + \lambda\Pi_{\mu}( \iota_{\mu}^{-1} \mathbbm{1}_X) = \Pi_{\mu}(\iota_{\mu}^{-1}f)
\end{align*}
since constant functions pull back to $E_{\mu}$ under $\iota_{\mu}^{-1}$ and $\Pi_{\mu}$ is the orthogonal projection onto $\mathfrak{T}_\mu \mathcal{P}(X).$ Thus $\Theta_{\mu}$ indeed can be regarded as a function on the quotient $\mathcal{H}_{\mu}/\mathbb{R}$. Now we consider the function $\Theta_{\mu}^{-1}$ defined above. Fix a $u \in \mathfrak{T}_\mu \mathcal{P}(X)$ and observe that we have
\begin{align*}
    \Theta_{\mu}\Theta_{\mu}^{-1}u = \Theta_{\mu}([a_{\mu}u]) = \Pi_{\mu}\iota_{\mu}^{-1}a_{\mu}u = \Pi_{\mu}u = u.
\end{align*}
Conversely, fix an $[f] \in \mathcal{H}_{\mu}/\mathbb{R}.$ Then we have
\begin{align}
    \Theta_{\mu}^{-1}\Theta_{\mu}([f]) = \Theta_{\mu}^{-1}(\Pi_{\mu}\iota_{\mu}^{-1}f) = [a_{\mu}\Pi_{\mu}\iota_{\mu}^{-1}f]. \label{theta inverse expression}
\end{align}
We will return to this expression shortly. For now, observe that $\iota_{\mu}^{-1}f - \Pi_{\mu}\iota_{\mu}^{-1}f \in E_{\mu}$, which implies $\iota_{\mu}^{-1}f - \Pi_{\mu}\iota_{\mu}^{-1}f = \lambda a_{\mu}^{-1}$ for some $\lambda \in \mathbb{R}.$ Multiplying both sides by $a_{\mu}$ yields the implication
\begin{align*}
    f - a_{\mu}\Pi_{\mu}\iota_{\mu}^{-1} f = \lambda \mathbbm{1}_X,
\end{align*}
which further implies $[f] = [a_{\mu}\Pi_{\mu}\iota_{\mu}^{-1}f].$ Returning to \Cref{theta inverse expression} yields the conclusion that $\Theta_{\mu}^{-1}\Theta_{\mu}([f]) = [f]$, which proves that $\Theta_{\mu}^{-1}$ is indeed the inverse of the map $\Theta_{\mu}.$ Finally, for the expression of $\mathbf{J}_{\mu}$ it simply follows from the definition and the fact that $\Pi_{\mu}^2 = \Pi_{\mu}$,
\begin{align*}
    \mathbf{J}_{\mu} u = J_{\mu}[a_{\mu}\Pi_{\mu} u] = H_c(a_{\mu}^2\Pi_{\mu}(a_{\mu}^{-1}a_{\mu}\Pi_{\mu}u)\mu)= H_c(a_{\mu}^2\Pi_{\mu}u \, \mu).
\end{align*}
\end{proof}

\begin{lemma}[Invertibility of $S_{\mu}$ on $\mathcal{H}_c$]\label{invertibility of st}
    It holds that $S_{\mu}$ is invertible with an inverse $S_{\mu}^{-1}: \mathcal{H}_c \to \mathcal{H}_c$ given by \[S_{\mu}^{-1} = \operatorname{id}_{\mathcal{H}_c} - 2J_{\mu}(\operatorname{id}_{\mathcal{Q}_{\mu}} + K_{\mu})^{-1}A_{\mu}.\] 
    Moreover, $\|S_{\mu}^{-1}\|_{\mathcal{L}(\mathcal{H}_c, \mathcal{H}_c)} \leq 1$ for every $\mu \in \mathcal{P}(X).$
\end{lemma}
\begin{proof}
    
Write $Q_{\mu}:= (\operatorname{id}_{\mathcal{Q}_{\mu}} + K_{\mu})^{-1}$, which exists as an operator from $\mathcal{H}_{\mu}/\mathbb{R} \to \mathcal{H}_{\mu}/\mathbb{R}$ due to \Cref{Qt existence}. Briefly, we will recall the domains and codomains of the operators in the expression above to ensure compatibility. We have
\[A_{\mu}: \mathcal{H}_c \to \mathcal{H}_{\mu}/\mathbb{R}, \quad Q_{\mu}: \mathcal{H}_{\mu}/\mathbb{R} \to \mathcal{H}_{\mu}/\mathbb{R}, \quad J_{\mu}: \mathcal{H}_{\mu}/\mathbb{R} \to \mathcal{H}_c. \]
We will prove the statement by showing that $S_{\mu}(\operatorname{id}_{\mathcal{H}_c} - 2J_{\mu}Q_{\mu}A_{\mu}) = \operatorname{id}_{\mathcal{H}_c}$. To see this, we will expand the definition of $S_{\mu}$,
\begin{align*}
    (\operatorname{id}_{\mathcal{H}_c} + 2J_{\mu}R_{\mu}A_{\mu})(\operatorname{id}_{\mathcal{H}_c} -2J_{\mu}Q_{\mu}A_{\mu}) &= \operatorname{id}_{\mathcal{H}_c} -2J_{\mu}Q_{\mu}A_{\mu} + 2J_{\mu}R_{\mu}A_{\mu} - 4J_{\mu}R_{\mu}A_{\mu}J_{\mu}Q_{\mu}A_{\mu} \\
    &= \operatorname{id}_{\mathcal{H}_c} - 2J_{\mu}\left(Q_{\mu} - R_{\mu} + 2R_{\mu}A_{\mu}J_{\mu}Q_{\mu}\right)A_{\mu}.
\end{align*}
Since $A_{\mu}J_{\mu} = K_{\mu}$, it remains to show that $Q_{\mu} - R_{\mu} + 2R_{\mu}K_{\mu}Q_{\mu} = 0$. Using the standard inverse identity $A^{-1} - B^{-1} = B^{-1}(B-A)A^{-1}$ with $A = \operatorname{id}_{\mathcal{Q}_{\mu}} - K_{\mu}$ and $B = \operatorname{id}_{\mathcal{Q}_{\mu}} + K_{\mu}$ we conclude that
\begin{align*}
    Q_{\mu} - R_{\mu} = -R_{\mu}(\operatorname{id}_{\mathcal{Q}_{\mu}} + K_{\mu} - \operatorname{id}_{\mathcal{Q}_{\mu}} + K_{\mu})Q_{\mu} = -2R_{\mu}K_{\mu}Q_{\mu}
\end{align*}
as desired, proving that $S_{\mu}$ is the left inverse of $\operatorname{id}_{\mathcal{H}_c} - 2J_{\mu}Q_{\mu}A_{\mu}$. To prove the right inverse property, note that we have
\begin{align*}
    (\operatorname{id}_{\mathcal{H}_c} -2J_{\mu}Q_{\mu}A_{\mu})(\operatorname{id}_{\mathcal{H}_c} + 2J_{\mu}R_{\mu}A_{\mu}) &= \operatorname{id}_{\mathcal{H}_c} + 2J_{\mu}R_{\mu}A_{\mu} - 2J_{\mu} Q_{\mu}A_{\mu} - 4J_{\mu}Q_{\mu}A_{\mu}J_{\mu}R_{\mu}A_{\mu} \\
    &= \operatorname{id}_{\mathcal{H}_c} + 2J_{\mu}R_{\mu}A_{\mu} - 2J_{\mu} Q_{\mu}A_{\mu} - 4J_{\mu}Q_{\mu}K_{\mu}R_{\mu}A_{\mu} \tag{$A_{\mu}J_{\mu} = K_{\mu}$} \\
    &= \operatorname{id}_{\mathcal{H}_c} + 2J_{\mu}(R_{\mu} - Q_{\mu} - 2Q_{\mu}K_{\mu}R_{\mu})A_{\mu}.
\end{align*}
Using another standard inverse identity $A^{-1} - B^{-1} = A^{-1}(B-A)B^{-1}$ we conclude that
\begin{align*}
    Q_{\mu} - R_{\mu} = -Q_{\mu}(\operatorname{id}_{\mathcal{Q}_{\mu}} + K_{\mu} - \operatorname{id}_{\mathcal{Q}_{\mu}} + K_{\mu})R_{\mu} = -2Q_{\mu}K_{\mu}R_{\mu}.
\end{align*}
Combining this with the panel above proves that $S_{\mu}$ is also the right inverse of $\operatorname{id}_{\mathcal{H}_c} - 2J_{\mu}Q_{\mu}A_{\mu}$, completing the proof of the first statement. The second statement regarding the operator norm of the inverse is proved in \Cref{St inverse bound}.
\end{proof}

\subsubsection{Proof of \Cref{operator metric equivalence}}\label{sec: operator metric equivalence}

By the definition $S_{\mu}
=
\operatorname{id}_{\mathcal H_c}
+
2J_{\mu}R_{\mu}A_{\mu}$ we have
\begin{align*}
\frac{\varepsilon}{2}
\left\langle S_{\mu}v,w\right\rangle_{\mathcal H_c}
&=
\frac{\varepsilon}{2}
\left(
\langle v,w\rangle_{\mathcal H_c}
+
2
\left\langle
J_{\mu}R_{\mu}A_{\mu}v,w
\right\rangle_{\mathcal H_c}
\right).
\end{align*}
By \Cref{inner prod equiv}, $
\left\langle
J_{\mu}R_{\mu}A_{\mu}v,w
\right\rangle_{\mathcal H_c}
=
\left\langle
[a_{\mu}v],
R_{\mu}[a_{\mu}w]
\right\rangle_{L^2(\mu)/\mathbb R}.$ Since $a_{\mu}v$ has zero $\mu$ mean by
\Cref{quotient inner prod}, let $r\in L^2(\mu)$ be any
representative of the quotient element $R_{\mu}[a_{\mu}w]
\in L^2(\mu)/\mathbb R.$ The inner product on $L^2(\mu)/\mathbb R$ is obtained from \citet{fricain2016theoryvol2}, Chapter 16.1. In particular,
\begin{multline*}
\left\langle
[a_{\mu}v],
R_{\mu}[a_{\mu}w]
\right\rangle_{L^2(\mu)/\mathbb R} =
\left\langle
a_{\mu}v,
r-\int_X r\,d\mu
\right\rangle_{L^2(\mu)}\\
=
\left\langle
a_{\mu}v,r
\right\rangle_{L^2(\mu)}
-
\left(\int_X a_{\mu}v\,d\mu\right)
\left(\int_X r\,d\mu\right) =
\left\langle
a_{\mu}v,r
\right\rangle_{L^2(\mu)}.
\end{multline*}
Moreover, this expression is independent of the chosen representative.
Indeed, any other representative is of the form $r+\lambda$ for some
$\lambda\in\mathbb R$, and
\[
\left\langle
a_{\mu}v,r+\lambda
\right\rangle_{L^2(\mu)} = 
\left\langle
a_{\mu}v,r
\right\rangle_{L^2(\mu)}
+
\lambda\int_Xa_{\mu}v,d\mu
\left\langle
a_{\mu}v,r
\right\rangle_{L^2(\mu)}.
\]
Thus, since any representative of
$R_{\mu}[a_{\mu}w]$ may be used, we may write
\[
\left\langle
[a_{\mu}v],
R_{\mu}[a_{\mu}w]
\right\rangle_{L^2(\mu)/\mathbb R} = 
\left\langle
a_{\mu}v,
R_{\mu}[a_{\mu}w]
\right\rangle_{L^2(\mu)}.
\]
where the expression is independent of the representative chosen
for $R_{\mu}[a_{\mu}w]$ since $a_{\mu}v$ is orthogonal to constants in $L^2(\mu)$. Consequently,
\begin{align*}
\frac{\varepsilon}{2}
\left\langle S_{\mu}v,w\right\rangle_{\mathcal H_c}
&=
\frac{\varepsilon}{2}
\left(
\langle v,w\rangle_{\mathcal H_c}
+
2
\left\langle
a_{\mu}v,
R_{\mu}[a_{\mu}w]
\right\rangle_{L^2(\mu)}
\right)\\
&=
\widetilde{\mathbf g}_{\mu}(v,w),
\end{align*}
where the last equality follows from
\Cref{metric under change of vars thm}. This completes the proof of the expression for the metric tensor. Now we will show the self-adjointness of $S_{\mu}$ on the appropriate subspace. By \Cref{S tangent invariance}, $S_{\mu}$ maps $\mathfrak{T}_\mu \mathcal{P}(X)$\
onto itself. Moreover, the symmetry of the metric tensor gives
\begin{align*}
\left\langle
S_{\mu}v,w
\right\rangle_{\mathcal H_c}
&=
\frac{2}{\varepsilon}
\widetilde{\mathbf g}_{\mu}(v,w) =
\frac{2}{\varepsilon}
\widetilde{\mathbf g}_{\mu}(w,v) =
\left\langle
S_{\mu}w,v
\right\rangle_{\mathcal H_c} =
\left\langle
v,S_{\mu}w
\right\rangle_{\mathcal H_c}.
\end{align*}
Thus the restriction of \(S_{\mu}\) to \(\mathfrak{T}_\mu \mathcal{P}(X)\) is self-adjoint. Full self-adjointness on $\mathcal{H}_c$ is given by \Cref{ambient self adjointness}.
\qed{}

\subsubsection{Supplemental Results}

\begin{lemma} \label{inner prod equiv}
    Let $v, w \in \mathfrak{T}_{\mu}\mathcal{P}(X)$ and recall the operator $R_{\mu}: L^2(\mu)/\mathbb{R} \to L^2(\mu)/\mathbb{R}$ given by $R_{\mu} := (\operatorname{id}_{L^2(\mu)/\mathbb{R}} - K_{\mu})^{-1}$. It holds that
    \[\langle J_{\mu} R_{\mu} A_{\mu} v, w \rangle_{\mathcal{H}_c} = \langle R_{\mu}[a_{\mu}v], [a_{\mu}w]\rangle_{L^2(\mu)/\mathbb{R}}.\]
    Equivalently, using the self-adjointness of $R_{\mu}$,
    \[
    \left\langle
    J_{\mu}R_{\mu}A_{\mu}v,w
    \right\rangle_{\mathcal H_c}
    =
    \left\langle
    [a_{\mu}v],
    R_{\mu}[a_{\mu}w]
    \right\rangle_{L^2(\mu)/\mathbb R}.
    \]
\end{lemma}
\begin{proof} 
    Define $u := \Theta_{\mu} R_{\mu}A_{\mu}v \in \mathfrak{T}_\mu \mathcal{P}(X)$.  Plugging the expression for $J_{\mu}$ into the inner product in the statement yields
    \begin{multline*}
        \langle J_{\mu} R_{\mu} A_{\mu} v, w \rangle_{\mathcal{H}_c} = \langle H_c(a_{\mu}^2 \Pi_{\mu}(a_{\mu}^{-1}[R_{\mu}A_{\mu}v])\mu), w\rangle_{\mathcal{H}_c} \\
        = \langle H_c(a_{\mu}^2 \Theta_{\mu}([R_{\mu}A_{\mu}v])\mu), w\rangle_{\mathcal{H}_c} 
        = \langle H_c(a_{\mu}^2 u\mu), w\rangle_{\mathcal{H}_c}.
    \end{multline*}
    Applying the definition for $H_c$ yields,
    \begin{multline*}
        \langle J_{\mu} R_{\mu} A_{\mu} v, w \rangle_{\mathcal{H}_c} = \int_X a_{\mu}^2(x) u(x) \langle k_c(x,\cdot), w\rangle_{\mathcal{H}_c} \, d\mu(x) 
        \\ = \int_X a_{\mu}^2 u w \, d\mu 
        = \int_X (a_{\mu} u)(a_{\mu} w) \, d\mu 
        = \langle a_{\mu}u, a_{\mu}w\rangle_{L^2(\mu)}.
    \end{multline*}
    We will now show that the two arguments of the inner product can be replaced with their equivalence class elements $[a_{\mu}u], [a_{\mu}w] \in \mathcal{Q}_{\mu}$ respectively. 
    \begin{rmk} \label{quotient inner prod}
        We can write $\langle a_{\mu}u, a_{\mu}w\rangle_{L^2(\mu)} = \langle [a_{\mu}u], [a_{\mu}w]\rangle_{L^2(\mu)/\mathbb R}.$
    \end{rmk}
    \begin{proof}
        Since $u, w \in \mathfrak{T}_\mu \mathcal{P}(X)$, it follows that $[a_{\mu}w] = \Theta_{\mu}^{-1}w \in \mathcal{Q}_{\mu}$ and $[a_{\mu}u] = \Theta_{\mu}^{-1}u \in \mathcal{Q}_{\mu}$. Thus, since constant functions pull back to $E_{\mu}$ under the isometric isomorphism $\iota_{\mu}^{-1} = a_{\mu}^{-1}$, it follows that $\langle a_{\mu}w, \mathbbm{1}_X\rangle_{\mathcal{H}_{\mu}} = 0$ and $\langle a_{\mu}u, \mathbbm{1}_X\rangle_{\mathcal{H}_{\mu}} = 0.$ Applying the Riesz map $H_{\mu}$ implies the condition
        \begin{align*}
            0 = \langle a_{\mu}w, \mathbbm{1}_X\rangle_{\mathcal{H}_{\mu}} = \int_X \langle k_{\mu}(x,\cdot) ,a_{\mu}w\rangle_{\mathcal{H}_{\mu}} \, d\mu = \int_Xa_{\mu}(x)w(x)\, d\mu(x).
        \end{align*}
        Applying the same logic yields the analogous conclusion $\int_X a_{\mu}u\, d\mu = 0.$
        Hence $a_{\mu}w, a_{\mu}u$ are orthogonal to constants. Write $f:= a_{\mu}u$, $g:= a_{\mu}w$ and observe that $f + g, f-g$ are also orthogonal to constants. Applying the polarization identities \citep{thomas2014polarization} allows us to say
        \begin{align*}
            \langle [f], [g]\rangle_{L^2(\mu)/\mathbb{R}} &= \frac{1}{4}\left(\|[f+g]\|_{L^2(\mu)/\mathbb{R}}^2 - \|[f - g]\|_{L^2(\mu)/\mathbb{R}}^2\right).
        \end{align*}
        For any $h$ that is orthogonal to constants, $\|[h]\|_{L^2(\mu)/\mathbb{R}} = \inf_{\lambda \in \mathbb R}\|h + \lambda \|_{L^2(\mu)} = \|h\|_{L^2(\mu)}$. Since $f + g, f-g$ are orthogonal to constants, it holds that
        \begin{multline*}
            \langle [f], [g]\rangle_{L^2(\mu)/\mathbb{R}} = \frac{1}{4}\left(\|[f+g]\|_{L^2(\mu)/\mathbb{R}}^2 - \|[f - g]\|_{L^2(\mu)/\mathbb{R}}^2\right) \\
            = \frac{1}{4}\left(\|f+g\|_{L^2(\mu)}^2 - \|f - g\|_{L^2(\mu)}^2\right) 
            = \langle f, g\rangle_{L^2(\mu)}
        \end{multline*}
        via the polarization identities.
    \end{proof}
    Thus, by \Cref{quotient inner prod}, 
    \begin{align*}
        \langle J_{\mu} R_{\mu} A_{\mu} v, w \rangle_{\mathcal{H}_c} &= \langle a_{\mu}u, a_{\mu}w\rangle_{L^2(\mu)} = \langle [a_{\mu}u], [a_{\mu}w]\rangle_{L^2(\mu)/\mathbb{R}}.
    \end{align*}
    By \Cref{theta inverse}, it holds that
    \[[a_{\mu} u] = \Theta_{\mu}^{-1}u = R_{\mu}A_{\mu} v = R_{\mu}[a_{\mu}v].\]
    Thus, $[a_{\mu} u] = R_{\mu}[a_{\mu}v]$ as elements of $\mathcal{Q}_{\mu}$. Now, applying $\overline{j}_{\mu}$ from \Cref{rkhs and lebesgue quotient} implies
    \begin{align*}
        [a_{\mu}u]_{L^2(\mu)/\mathbb{R}} = \overline{j}_{\mu}[a_{\mu} u]_{\mathcal{Q}_{\mu}} 
        = \overline{j}_{\mu} R_{\mu}^{\mathcal{Q}}[a_{\mu}v]_{\mathcal{Q}_{\mu}} 
        = R_{\mu}^{L^2}\overline{j}_{\mu}[a_{\mu}v]_{\mathcal{Q}_{\mu}} 
        = R^{L^2}_{\mu}[a_{\mu}v]_{L^2(\mu)/\mathbb R}.
    \end{align*}
    Thus $\langle J_{\mu} R_{\mu} A_{\mu} v, w \rangle_{\mathcal{H}_c} = \langle R_{\mu}[a_{\mu}v], [a_{\mu}w]\rangle_{L^2(\mu)/\mathbb{R}}.$ Finally, $R_{\mu}$ is self-adjoint on
$L^2(\mu)/\mathbb R$, and hence $\left\langle
R_{\mu}[a_{\mu}v],
[a_{\mu}w]
\right\rangle_{L^2(\mu)/\mathbb R}
=
\left\langle
[a_{\mu}v],
R_{\mu}[a_{\mu}w]
\right\rangle_{L^2(\mu)/\mathbb R}.$
\end{proof}

\begin{lemma}[Invariance of the tangent space]
\label{S tangent invariance}
The operator $S_{\mu}
=
\operatorname{id}_{\mathcal H_c}
+
2J_{\mu}R_{\mu}A_{\mu}$ maps $\mathfrak{T}_\mu \mathcal{P}(X)$ onto itself.
\end{lemma}

\begin{proof}
By the definition of $S_{\mu}$, it suffices to show that $J_{\mu}(\mathcal Q_{\mu})\subseteq \mathfrak{T}_\mu \mathcal{P}(X).$ To see this, fix \(q\in\mathcal Q_{\mu}\) and recall $J_{\mu}q
=
H_c\left(
a_{\mu}^2\Theta_{\mu}[q]\mu
\right).$ Therefore,
\begin{multline*}
\left\langle
J_{\mu}q,
a_{\mu}^{-1}
\right\rangle_{\mathcal H_c}
=
\left\langle
H_c\left(
a_{\mu}^2\Theta_{\mu}[q]\mu
\right),
a_{\mu}^{-1}
\right\rangle_{\mathcal H_c} =
\int_X
a_{\mu}^2
\Theta_{\mu}[q]
a_{\mu}^{-1}
\,d\mu \\ =
\int_X
a_{\mu}\Theta_{\mu}[q]
\,d\mu =
\left\langle
H_c(a_{\mu}\mu),
\Theta_{\mu}[q]
\right\rangle_{\mathcal H_c}
=
\left\langle
a_{\mu}^{-1},
\Theta_{\mu}[q]
\right\rangle_{\mathcal H_c}=
0,
\end{multline*}
because $\Theta_{\mu}[q]\in \mathfrak{T}_\mu \mathcal{P}(X)$ and $E_{\mu} = \operatorname{span}\{a_{\mu}^{-1}\}$. Hence $J_{\mu}q\in \mathfrak{T}_\mu \mathcal{P}(X).$ Now let $v\in \mathfrak{T}_\mu \mathcal{P}(X)$ and note that, since $J_{\mu}R_{\mu}A_{\mu}v\in \mathfrak{T}_\mu \mathcal{P}(X),$ we have $S_{\mu}v
=
v+2J_{\mu}R_{\mu}A_{\mu}v
\in \mathfrak{T}_\mu \mathcal{P}(X).$ Thus $S_{\mu}(\mathfrak{T}_\mu \mathcal{P}(X))$ indeed is a subset of $\mathfrak{T}_\mu \mathcal{P}(X)$.
\end{proof}

\begin{prop} \label{rkhs and lebesgue quotient}
    Define the operator $\overline{j}_{\mu}: \mathcal{H}_{\mu}/\mathbb R \to L^2(\mu)/\mathbb R$ given by 
    \[\overline{j}_{\mu}([f]_{\mathcal{Q}_{\mu}}) := [f]_{L^2(\mu)/\mathbb{R}}.\]
    Then $\overline{j}_{\mu}$ is well defined. Moreover, if $R^{\mathcal{Q}}:= (\operatorname{id}_{\mathcal{Q}_{\mu}} - K_{\mu})^{-1}$ and $R^{L^2}:= (\operatorname{id}_{L^2(\mu)/\mathbb{R}} - K_{\mu})^{-1}$, then $\overline{j}_{\mu} R_{\mu}^{\mathcal{Q}_{\mu}} = R^{L^2} \overline{j}_{\mu}$.
\end{prop}
\begin{proof}
    Fix $f \in \mathcal{H}_{\mu}$ and observe that
    \[|f(x)| \leq \|f\|_{\mathcal{H}_{\mu}}\|k_{\mu}(x, \cdot)\|_{\mathcal{H}_{\mu}} = \|f\|_{\mathcal{H}_{\mu}}\sqrt{k_{\mu}(x,x)}\]
    where we applied the reproducing property and Cauchy--Schwarz. Thus,
    \begin{align*}
        \|j_{\mu}f\|_{L^2(\mu)}^2 = \int_X|f(x)|^2 \, d\mu(x) \leq \left(\int_X k_{\mu}(x,x)\, d\mu(x)\right)\|f\|_{\mathcal{H}_{\mu}}^2.
    \end{align*}
    Thus, the assumption that $k_{\mu}$ is bounded and $X$ is compact guarantees that every element $f \in \mathcal{H}_{\mu}$ also belongs to $L^2(\mu)$. Now, we will prove that $\overline{j}_{\mu}$ is a well defined operator from $\mathcal{Q}_{\mu} \to L^2(\mu)/\mathbb{R}.$ Indeed, define
    \[\overline{j}_{\mu}: \mathcal{Q}_{\mu} \to L^2(\mu)/\mathbb{R}, \qquad \overline{j}_{\mu}\left([f]_{\mathcal{Q}_{\mu}}\right):= [f]_{L^2(\mu)/\mathbb{R}}.\]
    We will show that this is well defined. Indeed, if $[f]_{\mathcal{Q}_{\mu}} = [g]_{\mathcal{Q}_{\mu}}$ then $f - g = \lambda$ for some $\lambda \in \mathbb{R}$, and thus $[f]_{L^2(\mu)/\mathbb R} = [g]_{L^2(\mu)/\mathbb R}$. Now note that the two realizations of $K_{\mu}$ in $R^{L^2}$ and $R^{\mathcal{Q}}$ are induced by the same kernel integral operator. Hence, for any $[f] \in \mathcal{Q}_{\mu}$, 
    \begin{align*}
        \overline{j}_{\mu} K^{\mathcal{Q}}_{\mu}[f] = [K_{\mu}f]_{L^2(\mu)/\mathbb{R}} 
        = K_{\mu}^{L^2}[f]_{L^2(\mu)/\mathbb{R}}
        = K^{L^2}_{\mu}\overline{j}_{\mu}[f]
    \end{align*}
    where the superscript on the operator $K_{\mu}$ is used to distinguish the realizations. Thus $\overline{j}_{\mu}K_{\mu}^{\mathcal{Q}} = K_{\mu}^{L^2}\overline{j}_{\mu}$ and consequently
    \[\overline{j}_{\mu} (\operatorname{id}_{\mathcal{Q}_{\mu}} - K_{\mu}^{\mathcal{Q}}) = (\operatorname{id}_{L^2(\mu)/\mathbb{R}} - K_{\mu}^{L^2})\overline{j}_{\mu}.\]
    Left-composition with $R^{L^2}_{\mu}$ and right-composition with $R_{\mu}^\mathcal{Q}$ yields $\overline{j}_{\mu} R_{\mu}^{\mathcal{Q}} = R^{L^2}\overline{j}_{\mu}$ as desired.
\end{proof}

\begin{lemma}[Ambient self-adjointness of the metric operators]
\label{ambient self adjointness}
Under \Cref{standing sinkhorn assumptions}, $S_\mu$ and $S_\mu^{-1}$ are self-adjoint operators on the full space $\mathcal H_c$. Moreover, if $(\mu_t)_{t\in[0,1]}$ satisfies the assumptions of \Cref{metric time deriv}, then $\dot S_{\mu_t}$ is self-adjoint on $\mathcal H_c$ for almost every $t\in[0,1]$.
\end{lemma}

\begin{proof}
By \Cref{operator metric equivalence}, $S_\mu$ preserves $\mathfrak T_\mu\mathcal P(X)$ and its restriction to $\mathfrak T_\mu\mathcal P(X)$ is self-adjoint. On the other hand, if $z\in\mathfrak T_\mu\mathcal P(X)^\perp$, then $\Pi_\mu z=0$, and hence the definition of $\mathbf R_\mu$ in \eqref{reparameterization} gives $\mathbf R_\mu z=0$. Therefore $S_\mu z=z$. Thus, with respect to the orthogonal decomposition $\mathcal H_c=\mathfrak T_\mu\mathcal P(X)\oplus\mathfrak T_\mu\mathcal P(X)^\perp$, the operator $S_\mu$ is block diagonal, with a self-adjoint restriction on the first factor and the identity on the second. Hence $S_\mu$ is self-adjoint on $\mathcal H_c$. Since $S_\mu$ is invertible by \Cref{invertibility of st}, we have $(S_\mu^{-1})^*=(S_\mu^*)^{-1}=S_\mu^{-1}$, so $S_\mu^{-1}$ is also self-adjoint.

Finally, by \Cref{metric time deriv}, $t\mapsto S_{\mu_t}$ is differentiable in $\mathcal L(\mathcal H_c,\mathcal H_c)$ for almost every $t$. At every such $t$, the difference quotients
$\big(S_{\mu_{t+h}}-S_{\mu_t}\big)/h$ are self-adjoint, and the space of bounded self-adjoint operators is closed in the operator norm. Their operator-norm limit $\dot S_{\mu_t}$ is therefore self-adjoint.
\end{proof}

\begin{thm}[Differentiability of $f_{\mu, \nu}$ for smooth cost, \citep{goldfeld2024limit}, Theorem 3] \label{differentiability of EOT potentials}
    Let $X$ be a closed Euclidean ball of positive radius, and let $c \in C^\infty(X \times X)$. Then for every integer $m>d/2$ and $(\mu, \nu) \in \mathcal{P}(X) \times \mathcal{P}(X)$ each satisfying \Cref{standing sinkhorn assumptions}.3, the map $\mathcal{P}(X) \times \mathcal{P}(X) \subset \ell^\infty(B^m) \times \ell^\infty(B^m) \to C^m(X) \times C^m(X)$ given by $(\mu, \nu) \mapsto (f_{\mu, \nu}, g_{\mu, \nu})$ is Hadamard differentiable at $(\mu, \nu)$ tangentially to $\mathbb{T}_{\mu}^m \times \mathbb{T}_{\nu}^m$.
\end{thm}
\begin{proof}
    Since our definition of $\mathbb{T}_\mu^m$ differs slightly from that of \citet{goldfeld2024limit}, the above statement requires a qualification. Because $C^m(X) \hookrightarrow H^m(X)$, duality implies $H^m(X)^* \subset C^m(X)^*$ and $\|\cdot \|_{C^m(X)^* }\lesssim \|\cdot\|_{H^m(X)^*}$ (\Cref{duality of continuous embedding}). Consequently, $\mathbb{T}^m_\mu$ is contained in the closure of $\mathfrak{M}_\mu$ with respect to the $C^m(X)^*$ norm. Note that the result of this closure is the tangent space of perturbations chosen by \citet{goldfeld2024limit}. Since Hadamard differentiability tangentially to a set $\mathbb{T}$ passes to subsets $\mathbb{T}' \subset \mathbb{T}$ with the same topology, the statement follows. 
    
\end{proof}

\begin{prop}[Finite-smoothness extension of \citet{goldfeld2024limit}]
    \label{finite smoothness goldfeld}
    The assumptions on the cost $c$ in \Cref{differentiability of EOT potentials} can be weakened to the requirement that $c \in C^{(m, m)}(X \times X).$
\end{prop}
\begin{proof}
    The proof follows that of \citet[Theorem 3]{goldfeld2024limit}, with one modification. On page~1026, \citet{goldfeld2024limit} invoke the Arzel\`a--Ascoli theorem to obtain uniform convergence, along a subsequence, of the derivatives $(\operatorname{D}^{\alpha}f_{\mu_n,\nu_n},\operatorname{D}^{\alpha}g_{\mu_n,\nu_n})$. Their $C^\infty$ assumption on $c$ provides uniform bounds on derivatives of the Schr\"odinger potentials of arbitrary finite order. In particular, when $|\alpha|\leq m$, uniform control at order $|\alpha|+1$ yields equicontinuity of $(\operatorname{D}^{\alpha}f_{\mu_n,\nu_n},\operatorname{D}^{\alpha}g_{\mu_n,\nu_n})_n$, so that Arzel\`a--Ascoli may be applied to obtain uniform convergence. At the highest order $|\alpha|=m$, this argument implicitly uses regularity beyond $C^{(m,m)}(X\times X)$.

We will instead show that $c\in C^{(m,m)}(X\times X)$ is sufficient for the uniform convergence required at order $m$. Recall that the Schr\"odinger potentials satisfy
\begin{align*}
f_{\mu_n,\nu_n}(x)
&=
-\varepsilon
\log\left(
\int_X
\exp\left(
\frac{g_{\mu_n,\nu_n}(y)-c(x,y)}{\varepsilon}
\right)
d\nu_n(y)
\right), \\
g_{\mu_n,\nu_n}(y)
&=
-\varepsilon
\log\left(
\int_X
\exp\left(
\frac{f_{\mu_n,\nu_n}(x)-c(x,y)}{\varepsilon}
\right)
d\mu_n(x)
\right).
\end{align*}

\emph{Convergence at order $0$.} We first establish convergence at order zero. The uniform $C^0$ bounds in \citet[Lemma 1(ii)]{goldfeld2024limit} depend only on boundedness of $c$ and therefore remain valid under the present assumptions. Moreover, the Schr\"odinger system and the $L^\infty$ Lipschitz property of the log-sum-exp functional
\[ \left| \log\int_X e^{u}\,d\nu - \log\int_X e^{v}\,d\nu \right| \leq \|u-v\|_{L^\infty(X)} \]
directly give
\begin{align*}
    |f_{\mu_n,\nu_n}(x)-f_{\mu_n,\nu_n}(x')|
    &\leq
    \sup_{y\in X}|c(x,y)-c(x',y)|\\
    |g_{\mu_n,\nu_n}(y)-g_{\mu_n,\nu_n}(y')|
    &\leq
    \sup_{x\in X}|c(x,y)-c(x,y')|.
\end{align*}
Since $c$ is uniformly continuous on the compact set $X\times X$, the families $(f_{\mu_n,\nu_n})_n$ and $(g_{\mu_n,\nu_n})_n$ are uniformly bounded and equicontinuous. Thus, by Arzel\`a--Ascoli, every subsequence $(f_{\mu_{n'},\nu_{n'}},g_{\mu_{n'},\nu_{n'}})$ admits a further subsequence $(f_{\mu_{n''},\nu_{n''}},g_{\mu_{n''},\nu_{n''}})$ converging uniformly to some $(f^*,g^*)\in C(X)\times C(X)$. Supposing that $\mu_n\rightharpoonup\mu$ and $\nu_n\rightharpoonup\nu$, then uniform convergence of the potentials implies
\[
    \exp\left(
        \frac{g_{\mu_{n''},\nu_{n''}}(y)-c(x,y)}{\varepsilon}
    \right)
    \to
    \exp\left(
        \frac{g^*(y)-c(x,y)}{\varepsilon}
    \right)
\]
uniformly on $X\times X$. The same conclusion holds for the second Schr\"odinger equation. Moreover, if $q\in C(X\times X)$ and $\nu_n\rightharpoonup\nu$, then
\[
    \sup_{x\in X}
    \left|
        \int_X q(x,y)\,d(\nu_n-\nu)(y)
    \right|
    \to 0.
\]
Indeed, the map $x\mapsto q(x,\cdot)$ is continuous from compact $X$ into $C(X)$ and therefore has compact image. Since the linear functionals $\nu_n-\nu\in C(X)^*$ converge pointwise to zero and satisfy $|\nu_n-\nu|_{C(X)^*}\leq2$, their convergence is uniform on compact subsets of $C(X)$. We may therefore pass to the limit in both Schr\"odinger equations and conclude that $(f^*,g^*)$ solves the Schr\"odinger system associated with $(\mu,\nu)$. The normalization convention is preserved under uniform convergence, so uniqueness of the normalized Schr\"odinger potentials implies $(f^*,g^*)=(f_{\mu,\nu},g_{\mu,\nu}).$ Since every subsequence admits a further subsequence converging uniformly to the same limit, it follows that $f_{\mu_n,\nu_n}\to f_{\mu,\nu}$ and $g_{\mu_n,\nu_n}\to g_{\mu,\nu}$ uniformly on $X$.

\emph{Convergence at order $|\alpha| > 0$.} We now consider positive derivative orders. Fix a multi-index $\alpha$ with $1\leq|\alpha|\leq m$ and define
\[
    Z_n(x)
    =
    \int_X
    \exp\left(
        \frac{g_{\mu_n,\nu_n}(y)-c(x,y)}{\varepsilon}
    \right)
    d\nu_n(y),
\]
so that $f_{\mu_n,\nu_n}=-\varepsilon\log Z_n$. Repeated differentiation and the chain rule show that $\operatorname{D}^{\alpha}f_{\mu_n,\nu_n}$ is a finite combination of products and quotients of $Z_n$ and its derivatives $\operatorname{D}^{\alpha'}Z_n$, $1\leq|\alpha'|\leq|\alpha|$. Each such derivative can be written as
\[    \operatorname{D}^{\alpha'}Z_n(x)
    =
    \int_X q_{\alpha',n}(x,y)\,d\nu_n(y),
\]
where $q_{\alpha',n}$ is the product of $\exp((
g_{\mu_n,\nu_n}(y)-c(x,y))/\varepsilon)$ with a polynomial in the derivatives \[\left\{
\operatorname{D}_1^\gamma c(x,y):
1\leq|\gamma|\leq|\alpha'|
\right\}.\] 
Since $c\in C^{(m,m)}(X\times X)$ and $g_{\mu_n,\nu_n}\to g_{\mu,\nu}$ uniformly, we have $q_{\alpha', n} \to q_{\alpha'}$ uniformly on $X \times X$ for every $|\alpha'|\leq m$. Consequently,
\begin{multline*}
\sup_{x\in X}
\left|
\int_X q_{\alpha',n}(x,y)\,d\nu_n(y)
-
\int_X q_{\alpha'}(x,y)\,d\nu(y)
\right|\
\leq
\|q_{\alpha',n}-q_{\alpha'}\|_{L^\infty(X\times X)}
\\+
\sup_{x\in X}
\left|
\int_X q_{\alpha'}(x,y)\,d(\nu_n-\nu)(y)
\right|
\to 0.
\end{multline*}
The first term converges to zero by uniform convergence of the integrands, while the second converges to zero by the compactness of $\{q_{\alpha'}(x, \cdot): x \in X\}$. Hence $\operatorname{D}^{\alpha'}Z_n \to \operatorname{D}^{\alpha'}Z$ uniformly on $X$ for every $|\alpha'|\leq m$. 

Note that the denominators introduced by differentiating the logarithm are uniformly bounded away from zero. Indeed, the Schr\"odinger equation gives $Z_n(x) = \exp\left(-f_{\mu_n,\nu_n}(x)/\varepsilon
\right),$ while the uniform $C^0$ bound established above therefore implies $\inf_{n\in\mathbb{N}}\inf_{x\in X} Z_n(x)>0.$ It follows by continuity of the finite algebraic combinations appearing in the derivatives of $\log Z_n$ that $\|\operatorname{D}^{\alpha}f_{\mu_n,\nu_n}
  -
  \operatorname{D}^{\alpha}f_{\mu,\nu}\|_{L^\infty(X)}
\to 0$ for every $|\alpha|\leq m$. Applying the same argument to the second Schr\"odinger equation, with the roles of $\mu_n$ and $\nu_n$ and of the two arguments of $c$ reversed, gives $\|\operatorname{D}^{\alpha}g_{\mu_n,\nu_n} -\operatorname{D}^{\alpha}g_{\mu,\nu}\|_{L^\infty(X)}
\to 0$ for every $|\alpha|\leq m$. Therefore, $(f_{\mu_n,\nu_n},g_{\mu_n,\nu_n}) \to (f_{\mu,\nu},g_{\mu,\nu})$ in $C^m(X)\times C^m(X)$ as desired.
    
It remains to verify that the remainder of the proof of \citet[Theorem~3]{goldfeld2024limit} does not require additional regularity. For a multi-index $\alpha$ with $|\alpha|\leq m$, the differentiated Schr\"odinger equations used in their Step 1 involve functions of the form
\[
    q_t(x_1,x_2)
    =
    \zeta_\alpha(x_1,x_2)
    \exp\left(
        \frac{
            f_{\mu_t,\nu_t}(x_1)
            +
            g_{\mu_t,\nu_t}(x_2)
            -
            c(x_1,x_2)
        }{\varepsilon}
    \right),
\]
where $\zeta_\alpha$ is a polynomial in derivatives of $c$ of order at most $|\alpha|$ in the variable being differentiated. Their argument requires the resulting function, with the differentiated variable fixed, to belong to $C^m(X)$ in the opposite variable. Differentiating up to order $m$ in that opposite variable produces mixed derivatives $\operatorname{D}_1^\gamma\operatorname{D}_2^{\alpha'} c$ for $|\gamma| \leq m$ and $|\alpha'|\leq m$. These derivatives exist and are continuous under $c\in C^{(m,m)}(X\times X)$. 

This, taken together with (1) the $C^m$-continuity of the Schr\"odinger potentials established above and (2) the fact that $C^m(X)$ is a Banach algebra under pointwise multiplication \citep[Equation 1.11]{moore2017global}, yields the $C^m$ convergence of the test functions $q_t$ required in Step 1 of their proof (they use the notation $g_t$, which we avoid due to clashes with existing notation in our manuscript). Step 2 then assembles the derivatives of orders $|\alpha|\leq m$ using completeness of $C^m(X)$ and requires no additional derivatives of $c$. Thus the remainder of the proof of \citet[Theorem 3]{goldfeld2024limit} applies at the fixed order $m$, completing the proof. 
\end{proof}

\begin{corollary}[Hadamard differentiability of the change of variable map]\label{hadamard deriv of change of var}
   The map $B: \mathcal{P}(X) \subset \ell^\infty(B^m) \to \mathcal{H}_c$ is Hadamard differentiable at $\mu$ tangentially to $\mathbb{T}_\mu^m$ with derivative
   \[\operatorname{D}_{\gamma}B_{\mu} = H_c\left(\frac{a_\mu \mu}{\varepsilon}\operatorname{D}_\gamma f_{\mu, \mu} + a_\mu \gamma\right).\]
\end{corollary}
\begin{proof}
    This result follows from \Cref{hadamard diff of emu} because $B_{\mu} = e_\mu$.
\end{proof}

\subsection{Proof of \Cref{statistical riemannian tangent embedding}}

We first characterize the statistical tangent space. Since every element of $\mathfrak M_\mu$ has zero total mass and $H^m(X)^*_0$ is closed in $H^m(X)^*$, we immediately have $\mathbb T_\mu^m\subseteq H^m(X)^*_0$ \citep[Proposition 1.5]{brezis2011functional}. For the reverse inclusion, let $\sigma\in H^m(X)^*_0$. By the density of functionals induced by smooth, compactly supported functions in $H^m(X)^*$ \citep[Theorem 3.29(ii) and 3.30(i)]{mclean2000strongly}, there exists a sequence $g_n\in C^\infty_c(\operatorname{int}(X))$ such that $g_n\,dx\to\sigma$ in $H^m(X)^*$. Fix $\psi\in C^\infty_c(\operatorname{int}(X))$ satisfying $\int_X\psi\,dx=1$ and define $h_n=g_n-(\int_Xg_n\,dx)\psi$. Since $\sigma(\mathbbm{1}_X)=0$, we have $\int_Xg_n\,dx\to0$, and consequently $h_n\,dx\to\sigma$ in $H^m(X)^*$. Moreover, $\int_Xh_n\,dx=0$. Recall that $\mu=\rho\,dx$ with $\rho\geq\rho_{\min}>0$. For each $n$, choose $t_n>0$ sufficiently large that $\rho(x)+h_n(x)/t_n\geq0$ for every $x\in X$, and set $\nu_n=(\rho+h_n/t_n)\,dx$. Then $\nu_n\in\mathcal P(X)$ and, since $\operatorname{supp}(\mu)=X$, we have $\nu_n\in\mathcal P_\mu$. Hence $h_n\,dx=t_n(\nu_n-\mu)\in\mathfrak M_\mu$. It follows that $\sigma\in\mathbb T_\mu^m$, and therefore $\mathbb T_\mu^m=H^m(X)^*_0$.

We next characterize the Riemannian tangent space. By \citet[Theorem 4.6]{lavenant2024riemannian}, $T_\mu\mathcal P(X)=\mathcal H_{\mu,0}^*$ with equivalent norms. Moreover, $\mathcal H_\mu$ is isomorphic to $\mathcal H_c$ through multiplication by the Schr\"odinger scaling $a_\mu$ \citep[Proposition 4.7]{lavenant2024riemannian}. Since $\mathcal H_c\simeq H^s(X)$ and multiplication by $a_\mu$ and $a_\mu^{-1}=B_\mu$ are bounded on $H^s(X)$, it follows that $\mathcal H_\mu$ and $H^s(X)$ coincide as sets with equivalent norms. Therefore $T_\mu\mathcal P(X)\cong H^s(X)^*_0$. Finally, since $s>m$ and $X$ is bounded, the Sobolev embedding $H^s(X)\hookrightarrow H^m(X)$ is continuous and dense \citep[Theorem 3.27(ii) and Theorem 3.29 (ii)]{mclean2000strongly}. By duality, the canonical inclusion $H^m(X)^*\hookrightarrow H^s(X)^*$ is therefore continuous (one can employ the same argument as that of \Cref{duality of continuous embedding}). To see that the corresponding inclusion $H^m(X)^*_0\hookrightarrow H^s(X)^*_0$ has dense range, let $\sigma\in H^s(X)^*_0$. By \citet[Theorems 3.29(ii) and 3.30(i)]{mclean2000strongly}, there exists a sequence $g_n\in C^\infty_c(\operatorname{int}(X))$ such that $g_n\,dx\to\sigma$ in $H^s(X)^*$. Fix $\psi\in C^\infty_c(\operatorname{int}(X))$ satisfying $\int_X\psi\,dx=1$ and set $h_n=g_n-(\int_Xg_n\,dx)\psi$. Since $\sigma(\mathbbm{1}_X)=0$, we have $\int_Xg_n\,dx\to0$, so $h_n\,dx\to\sigma$ in $H^s(X)^*$. Moreover, $\int_Xh_n\,dx=0$ and $h_n\,dx\in H^m(X)^*_0$ for every $n$. Thus the inclusion $H^m(X)^*_0\hookrightarrow H^s(X)^*_0$ is continuous with dense range. Combining the preceding identities gives the claimed continuous dense embedding $\mathbb T_\mu^m\hookrightarrow T_\mu\mathcal P(X)$.
\qed{}

\subsection{Proof of \Cref{self schrodinger hadamard diff}} \label{proof of self schrodinger hadamard diff}
In order to obtain this result, we need to compose with a map $\Delta: \mathcal{P}(X) \to \mathcal{P}(X) \times \mathcal{P}(X)$ given by $\mu \mapsto (\mu, \mu)$. Then, writing the Schr\"odinger potential map as $f :=  ((\mu, \nu) \mapsto f_{\mu, \nu})$, the self-Schr\"odinger map $\mathfrak{f}: \mathcal{P}(X) \to C^m(X)$ given by $\mathfrak{f}_\mu = f_{\mu, \mu}$ can be written as $\mathfrak{f} = f \circ \Delta$. Note that $\Delta$ is Hadamard differentiable at $\mu$ tangentially to $\mathbb{T}_{\mu}^m$ due to linearity (\Cref{multilinear hadamard diff}), with Hadamard derivative $\operatorname{D}_{\gamma}\Delta_\mu = (\gamma, \gamma)$. By the Hadamard chain rule \citep[Lemma 3.9.3]{van2000asymptotic} and \Cref{finite smoothness goldfeld} it therefore follows that $\mathfrak{f}_\mu$ is Hadamard differentiable at $\mu$ tangentially to $\mathbb{T}_\mu^m$. \qed{}


\subsection{Proof of \Cref{metric hadamard diff}} \label{proof of metric hadamard deriv}
We will begin with Hadamard differentiability of $S_\mu$. By \Cref{tilde J hadamard diff} and \Cref{tilde R hadamard diff}, $\mu \mapsto \mathbf{J}_\mu$ and $\mu \mapsto \mathbf{R}_\mu$ are Hadamard differentiable as maps from $\mathcal{P}(X) \subset \ell^\infty(B^m) \to \mathcal{L}(\mathcal{H}_c, \mathcal{H}_c).$ In particular, they are Hadamard differentiable at $\mu \in \mathcal{P}(X)$ tangentially to $\mathbb{T}^m_\mu$. Now, define the map $\mathsf{S} := \mathfrak{S} \circ (\mathbf{J}_{(\cdot )}, \mathbf{R}_{(\cdot)})$ where $\mathfrak{S}: \mathcal{L}(\mathcal{H}_c, \mathcal{H}_c) \times \mathcal{L}(\mathcal{H}_c, \mathcal{H}_c) \to \mathcal{L}(\mathcal{H}_c, \mathcal{H}_c)$ is given by  $\mathfrak{S}(A,B):= \operatorname{id}_{\mathcal{H}_c} + 2AB$. Observe that $S_{(\cdot)} \equiv \mathfrak{S} \circ  (\mathbf{J}_{(\cdot )}, \mathbf{R}_{(\cdot)})$, and note that by \Cref{multilinear hadamard diff}, $\mathfrak{S}$ is Hadamard differentiable at $(A, B)$ tangentially to $\mathcal{L}(\mathcal{H}_c, \mathcal{H}_c) \times \mathcal{L}(\mathcal{H}_c, \mathcal{H}_c)$ with Hadamard derivative
    \[\operatorname{D}_{(\dot A, \dot B)}\mathfrak{S}_{(A,B)} = 2\dot A B + 2A \dot B.\]
    Thus, we can apply the Hadamard chain rule \citep[Lemma 3.9.3]{van2000asymptotic} to complete the derivation of the derivative. The observation that the derivative is a linear function of $\gamma$ completes the proof.
    
    Now we will prove the Hadamard differentiability of the metric tensor. Define the map $\mathsf{G}: \mathcal{L}(\mathcal{H}_c, \mathcal{H}_c) \times \mathcal{H}_c \times \mathcal{H}_c \to \mathbb{R}$ given by $\mathsf{G}(\mathsf{S}, \mathsf{v}, \mathsf{w}) = \frac{\varepsilon}{2}\langle \mathsf{S}\mathsf{v}, \mathsf{w}\rangle_{\mathcal{H}_c},$ which is clearly trilinear and bounded. As such, it is also Hadamard differentiable (\Cref{multilinear hadamard diff}) with Hadamard derivative 
    \begin{align*}
        \operatorname{D}_{(\dot{\mathsf{S}}, \dot{\mathsf{v}}, \dot{\mathsf{w}})}\mathsf{G}(\mathsf{S}, \mathsf{v}, \mathsf{w}) = \frac{\varepsilon}{2}\left[\langle \dot{\mathsf{S}}\mathsf{v}, \mathsf{w}\rangle_{\mathcal{H}_c} + \langle \mathsf{S}\dot{\mathsf{v}}, \mathsf{w}\rangle_{\mathcal{H}_c} + \langle \mathsf{S}\mathsf{v}, \dot{\mathsf{w}}\rangle_{\mathcal{H}_c}\right].
    \end{align*}
    Furthermore, define $\mathsf{F}: \mathcal{P}(X) \times \mathcal{H}_c \times \mathcal{H}_c \to \mathcal{L}(\mathcal{H}_c, \mathcal{H}_c) \times \mathcal{H}_c \times \mathcal{H}_c$ given by $\mathsf{F}(\mu, v, w) = (S_\mu, v, w)$, which (by the proof above) is Hadamard differentiable tangentially to $\mathbb{T}_\mu^m \times \mathcal{H}_c \times \mathcal{H}_c$ with Hadamard derivative 
    \[\operatorname{D}_{(\gamma, \dot v, \dot w)}\mathsf{F}(\mu, v, w) = (\operatorname{D}_\gamma S_\mu, \dot v, \dot w).\]
    Since $\tilde{\mathbf{g}} = \mathsf{G}\circ \mathsf{F}$, we can apply the Hadamard chain rule \citep[Lemma 3.9.3]{van2000asymptotic} to complete the proof. \qed{}

\subsection{Proof of \Cref{metric time deriv}} \label{proof of metric time deriv}

We begin with proving time differentiability of $S_{\mu_t}$. Since $\mu_t(\mathbbm{1}_X)=1$ for every $t$, we have
$\dot\mu_t(\mathbbm{1}_X)=0$ for almost every
$t\in[0,1]$. Consequently, by
\Cref{statistical riemannian tangent embedding},
$\dot\mu_t\in H^m(X)^*_0=\mathbb T_{\mu_t}^m$
for almost every $t$. Moreover, the
positive-density assumption implies
$\operatorname{supp}(\mu_t)=X$, so that
$\mathcal P_{\mu_t}=\mathcal P(X)$. Since $C^m(X)\hookrightarrow H^m(X)$
continuously, duality yields
$H^m(X)^*\hookrightarrow C^m(X)^*$
(\Cref{duality of continuous embedding}).
Thus, $\mu_t$ is differentiable in
$C^m(X)^*$ for almost every $t$, with
derivative belonging to $\mathbb T_{\mu_t}^m$.
The conclusion follows from the Hadamard
chain rule and \Cref{metric hadamard diff}.

Now we will prove the statement concerning the metric tensor. Since $v_t, w_t \in H^{1}([0,1]; \mathcal{H}_c)$, they admit absolutely continuous representatives that are $\mathcal{H}_c$-norm differentiable \citep[Theorem 5.9.2]{evans2022partial}. As such, $\operatorname{D}_\lambda v_t = \lambda \dot v_t$ and $\operatorname{D}_\lambda w_t = \lambda \dot w_t$, $\lambda \in \mathbb{R}$ are the Hadamard derivatives of $t \mapsto v_t$ and $t \mapsto w_t$ respectively, where $\dot v_t$ and $\dot w_t$ are the weak time derivatives of $v_t, w_t$ respectively. Now define the map $\mathsf{H}: [0,1] \to \mathcal{P}(X) \times \mathcal{H}_c \times \mathcal{H}_c$ given by $\mathsf{H}(t) = (\mu_t, v_t, w_t),$ and note that
     its Hadamard derivative is given by
     \[\operatorname{D}_{\lambda}\mathsf{H}(t) = (\lambda\dot \mu_t , \lambda \dot v_t, \lambda \dot w_t) \quad \text{for a.e. $t \in [0,1]$}\]
     since $\mu_t$ is $C^m(X)^*$-norm differentiable for a.e. $t \in [0,1]$. Observing that $\tilde{\mathbf{g}}_{\mu_t}(v_t, w_t) = (\tilde{\mathbf{g}} \circ \mathsf{H})(t)$, we can apply the Hadamard chain rule and \Cref{metric hadamard diff} to obtain the Hadamard derivative of $t\mapsto \tilde{\mathbf{g}}_{\mu_t}(v_t, w_t)$ which is given by 
     \[\operatorname{D}_{\lambda}\tilde{\mathbf{g}}_{\mu_t}(v_t, w_t) = \frac{\lambda \varepsilon}{2}\bigg[\big\langle (\operatorname{D}_{\dot \mu_t} S_{\mu_t})v_t, w_t\big\rangle_{\mathcal{H}_c} + \big\langle S_{\mu_t} \dot v_t, w_t\big\rangle_{\mathcal{H}_c} + \big\langle S_{\mu_t} v_t, \dot w_t\big\rangle_{\mathcal{H}_c}\bigg].\]
     Setting $\lambda = 1$ gives $\frac{d}{dt}\tilde{\mathbf{g}}_{\mu_t}(v_t,w_t).$ \qed{}

\subsection{Proof of \Cref{metric compatible family}}\label{sec: metric compatible family}

\begin{lemma}
    For every $\Omega \in \mathcal{L}(\mathcal{H}_c, \mathcal{H}_c)$ such that $S_{\mu_t}\Omega_{\mu_t} + \Omega_{\mu_t}^* S_{\mu_t} = 0$, the operator $\nabla^{\Omega}$ is a valid covariant derivative along the curve $(\mu_t)_t$.
\end{lemma}
\begin{proof}
    As alluded to in the text preceding \Cref{metric compatible family}, we will verify the first two (but not the third) properties described in \citet[Theorem 4.24]{lee2018introduction}. In particular, a covariant derivative along a curve needs to satisfy (i) linearity over $\mathbb{R}$, (ii) the Leibniz rule. We will verify each in turn.

    \begin{enumerate}
        \item[(i)] Let $a, b \in \mathbb{R}$ and let $v_t, w_t \in H^1([0,1]; \mathcal{H}_c)$ be two Sinkhorn regular tangent vector fields along the curve of measures $(\mu_t)_{t \in [0,1]}.$ Then 
        \begin{align*}
            \nabla^{\Omega}_{\dot\beta_t}(av_t + bw_t) &= \Pi_{\mu_t}\left(
            \partial_t (av_t + bw_t)
            +
            \left(\frac{1}{2}S_{\mu_t}^{-1}\dot S_{\mu_t}
            +
            \Omega_{\mu_t} 
            \right)(av_t + bw_t)\right) \\
            &= \Pi_{\mu_t}\bigg(
            \partial_t (av_t) + \partial_t(bw_t)
            +
            \left(\frac{1}{2}S_{\mu_t}^{-1}\dot S_{\mu_t}
            +
            \Omega_{\mu_t} 
            \right)(av_t) \\
            &\hspace{60mm}+ \left(\frac{1}{2}S_{\mu_t}^{-1}\dot S_{\mu_t}
            +
            \Omega_{\mu_t} 
            \right)(bw_t)\bigg) \\
            &= a \nabla^{\Omega}_{\dot\beta_t} v_t + b \nabla^{\Omega}_{\dot \mu_t} w_t
        \end{align*}
        by the linearity of $\Pi_{\mu_t}, S_{\mu_t}$ and $\dot S_{\mu_t}$.
        \item[(ii)] For any $f_t \in C^\infty([0,1])$ we have 
        \begin{align*}
            \nabla^{\Omega}_{\dot\beta_t}(f_tv_t) &= \Pi_{\mu_t}\left(\partial_t(f_tv_t)+ \left(\frac{1}{2}S_{\mu_t}^{-1}\dot S_{\mu_t}
            +
            \Omega_{\mu_t} 
            \right)(f_tv_t)\right) \\
            &= \Pi_{\mu_t}\left(\dot f_t v_t + f_t \dot v_t + f_t\left(\frac{1}{2}S_{\mu_t}^{-1}\dot S_{\mu_t}
            +
            \Omega_{\mu_t} 
            \right)(v_t)\right) \\
            &= \dot f_t v_t + f_t \nabla _{\dot {\mu_t}}^\Omega v_t.
        \end{align*}
        
    \end{enumerate}
\end{proof}

Now suppose $v_t, w_t \in H^{1}([0,1]; \mathcal{H}_c)$ are two tangent vector fields along $(\mu_t)_{t \in [0,1]}$. Metric compatibility of a connection $\nabla$ with respect to $\tilde{\mathbf{g}}_{\mu_t}$ is then characterized by the identity
\[\frac{d}{dt}\tilde{\mathbf{g}}_{\mu_t}(v_t, w_t) = \tilde{\mathbf{g}}_{\mu_t}\left(\nabla_{\dot \mu_t}v_t, w_t\right) + \tilde{\mathbf{g}}_{\mu_t}\left(v_t, \nabla_{\dot \mu_t}w_t\right).\]
Note that by \Cref{metric time deriv} we have for a.e. $t \in [0,1]$,
\begin{align*}
\frac{d}{dt}\tilde{\mathbf{g}}_{\mu_t}(v_t,w_t) &= \frac{\varepsilon}{2}\langle \dot S_{\mu_t}v_t, w_t\rangle_{\mathcal{H}_c} + \frac{\varepsilon}{2}\langle S_{\mu_t}\partial_tv_t, w_t\rangle_{\mathcal{H}_c} + \frac{\varepsilon}{2}\langle S_{\mu_t}v_t, \partial_tw_t\rangle_{\mathcal{H}_c} \\
&= \frac{\varepsilon}{2}\langle \dot S_{\mu_t}v_t, w_t\rangle_{\mathcal{H}_c} + \tilde{\mathbf{g}}_{\mu_t}(\partial_tv_t, w_t) + \tilde{\mathbf{g}}_{\mu_t}(v_t, \partial_tw_t).
\end{align*}
By \Cref{Pimu g orthogonal projection}, $\Pi_{\mu_t}$ is the $\tilde{\mathbf g}_{\mu_t}$-orthogonal projection onto the tangent space. Since $v_t,w_t$ are tangent,
\begin{align*}
\frac{d}{dt}\tilde{\mathbf{g}}_{\mu_t}(v_t,w_t) &= \frac{\varepsilon}{2}\left\langle  \Pi_{\mu_t}\left( \dot S_{\mu_t}v_t\right), w_t \right\rangle_{\mathcal{H}_c} + \tilde{\mathbf{g}}_{\mu_t}\left(\Pi_{\mu_t}(\partial_tv_t), w_t\right) + \tilde{\mathbf{g}}_{\mu_t}(v_t, \Pi_{\mu_t}(\partial_tw_t)) \\
&= \frac{\varepsilon}{2}\left\langle  \dot S_{\mu_t}v_t, w_t \right\rangle_{\mathcal{H}_c} + \tilde{\mathbf{g}}_{\mu_t}\left(\Pi_{\mu_t}(\partial_tv_t), w_t\right) + \tilde{\mathbf{g}}_{\mu_t}(v_t, \Pi_{\mu_t}(\partial_tw_t)).
\end{align*}
It therefore suffices to show that
\begin{multline*}
\frac{\varepsilon}{2}\left\langle  \Pi_{\mu_t}\left( \dot S_{\mu_t}v_t\right), w_t \right\rangle_{\mathcal{H}_c}
=
\tilde{\mathbf{g}}_{\mu_t}\left(\Pi_{\mu_t}\left(\left(\frac{1}{2}S_{\mu_t}^{-1} \dot{S}_{\mu_t} + \Omega_{\mu_t}\right)v_t\right),w_t\right)
\\ + 
\tilde{\mathbf{g}}_{\mu_t}\left(v_t,\Pi_{\mu_t}\left(\left(\frac{1}{2}S_{\mu_t}^{-1} \dot{S}_{\mu_t} + \Omega_{\mu_t}\right)w_t\right)\right)
\end{multline*}
for any $\Omega_{\mu_t}$ satisfying $S_{\mu_t}\Omega_{\mu_t} + \Omega_{\mu_t}^*S_{\mu_t} = 0.$ To simplify notation, write $\mathcal A_{\mu_t}:=\frac12S_{\mu_t}^{-1}\dot S_{\mu_t}+\Omega_{\mu_t}$. Since $v_t,w_t$ are tangent and $S_{\mu_t}\Pi_{\mu_t}=\Pi_{\mu_t}S_{\mu_t}$, the right-hand side equals $\frac{\varepsilon}{2}\langle(S_{\mu_t}\mathcal A_{\mu_t}+\mathcal A_{\mu_t}^*S_{\mu_t})v_t,w_t\rangle_{\mathcal H_c}$. Thus metric compatibility is equivalent to $\Pi_{\mu_t}(S_{\mu_t}\mathcal A_{\mu_t}+\mathcal A_{\mu_t}^*S_{\mu_t}-\dot S_{\mu_t})\Pi_{\mu_t}=0$. For the stated choice of $\mathcal A_{\mu_t}$, we will verify the stronger identity $S_{\mu_t}\mathcal A_{\mu_t}+\mathcal A_{\mu_t}^*S_{\mu_t}=\dot S_{\mu_t}$. Indeed,
\begin{align*}
    S_{\mu_t}\mathcal{A}_{\mu_t} + \mathcal{A}_{\mu_t}^*S_{\mu_t} &= S_{\mu_t}\left(\frac{1}{2}S_{\mu_t}^{-1}\dot{S}_{\mu_t} + \Omega_{\mu_t}\right) + \left(\frac{1}{2}S_{\mu_t}^{-1}\dot{S}_{\mu_t} + \Omega_{\mu_t}\right)^* S_{\mu_t} \\
    &= \frac{1}{2}\dot S_{\mu_t} + S_{\mu_t}\Omega_{\mu_t} + \frac{1}{2}(S_{\mu_t}^{-1}\dot S_{\mu_t})^*S_{\mu_t} + \Omega_{\mu_t}^*S_{\mu_t} \\
    &= \frac{1}{2}\dot S_{\mu_t} + S_{\mu_t}\Omega_{\mu_t} + \frac{1}{2}(\dot S_{\mu_t})^*(S_{\mu_t}^{-1})^*S_{\mu_t} + \Omega_{\mu_t}^*S_{\mu_t}.
\end{align*}
By \Cref{ambient self adjointness}, $S_{\mu_t}$, $S_{\mu_t}^{-1}$ and $\dot S_{\mu_t}$ are self-adjoint on $\mathcal H_c$. Thus 
\begin{align*}
    S_{\mu_t}\mathcal{A}_{\mu_t} + \mathcal{A}_{\mu_t}^*S_{\mu_t} &= \frac{1}{2}\dot S_{\mu_t} + S_{\mu_t}\Omega_{\mu_t} + \frac{1}{2}\dot S_{\mu_t}S_{\mu_t}^{-1}S_{\mu_t} + \Omega_{\mu_t}^*S_{\mu_t} = \dot S_{\mu_t} + S_{\mu_t}\Omega_{\mu_t} + \Omega_{\mu_t}^*S_{\mu_t}.
\end{align*}
By the definition of $\Omega_{\mu_t}$, it follows that $S_{\mu_t}\mathcal{A}_{\mu_t}+\mathcal{A}_{\mu_t}^*S_{\mu_t}=\dot S_{\mu_t}$ as desired. Hence $\frac{d}{dt}\tilde{\mathbf{g}}_{\mu_t}(v_t,w_t)
=
\tilde{\mathbf{g}}_{\mu_t}
(
\nabla^{\Omega}_{\dot\beta_t}v_t,w_t
)
+
\tilde{\mathbf{g}}_{\mu_t}
(
v_t,\nabla^{\Omega}_{\dot\beta_t}w_t
)$ for any $\Omega_{\mu_t} \in \mathcal{L}(\mathcal{H}_c, \mathcal{H}_c)$ such that $S_{\mu_t}\Omega_{\mu_t} + \Omega_{\mu_t}^*S_{\mu_t} = 0$. \qed{} 


\subsubsection{Supplemental Results}

\begin{lemma}\label{Pimu g orthogonal projection}
    The map $\Pi_{\mu}: \mathcal{H}_c \to \mathfrak{T}_{\mu}\mathcal{P}(X)$ is the $\tilde{\mathbf{g}}_{\mu}$-orthogonal projection onto $\mathfrak{T}_{\mu}\mathcal{P}(X)$.
\end{lemma}

\begin{proof}
    To verify this we need to show that (i) $\Pi_{\mu}$ is self adjoint, (ii) $\operatorname{Ran}(\Pi_\mu) = \mathfrak{T}_\mu \mathcal{P}(X)$, and (iii) $\Pi_{\mu}^2 = \Pi_\mu$. Since $\Pi_\mu$ was constructed as the $\mathcal{H}_c$-orthogonal projection onto $\mathfrak{T}_\mu \mathcal{P}(X)$, the latter two properties hold. It remains to prove that $\Pi_\mu$ is self-adjoint with respect to $\tilde{\mathbf{g}}_\mu$. 
    
    We know that $S_\mu(E_\mu^\perp)\subseteq E_\mu^\perp$, and for $z\in E_\mu$, $A_\mu z=0$, so $S_\mu z=z$. Hence $S_\mu\Pi_\mu=\Pi_\mu S_\mu.$ Since $\Pi_\mu$ is the usual $\mathcal H_c$-orthogonal projection, $\Pi_\mu^*=\Pi_\mu$; thus, for $u,v\in\mathcal H_c$,
    \begin{equation*} \tilde{\mathbf g}_\mu(\Pi_\mu u,v) = \frac{\varepsilon}{2} \langle S_\mu\Pi_\mu u,v\rangle_{\mathcal H_c} = \frac{\varepsilon}{2} \langle \Pi_\mu S_\mu u,v\rangle_{\mathcal H_c} = \frac{\varepsilon}{2} \langle S_\mu u,\Pi_\mu v\rangle_{\mathcal H_c} = \tilde{\mathbf g}_\mu(u,\Pi_\mu v). \end{equation*}
    Thus $\Pi_\mu$ is $\tilde{\mathbf g}_\mu$-self-adjoint.     
\end{proof}


\subsection{Proof of \Cref{levi civita}}\label{sec: proof of levi civita}
We will first verify metric compatibility. Fix $t\in[0,1]$ and let $v,w\in\mathfrak T_{\mu_t}\mathcal P(X)$. Writing $\Omega_{\dot\beta_t}=S_{\mu_t}^{-1}L_{\dot\beta_t}$ as in \Cref{Omega skew}, we have
\[
    \langle S_{\mu_t}\Omega_{\dot\beta_t}v,w\rangle_{\mathcal H_c}
    +
    \langle S_{\mu_t}\Omega_{\dot\beta_t}w,v\rangle_{\mathcal H_c}
    =
    \langle L_{\dot\beta_t}v,w\rangle_{\mathcal H_c}
    +
    \langle L_{\dot\beta_t}w,v\rangle_{\mathcal H_c}
    =
    0,
\]
where the final equality follows from \Cref{Omega skew}. Thus $\Omega_{\dot\beta_t}$ satisfies the condition of \Cref{metric compatible family}, and hence the stated covariant derivative is metric compatible.

It remains to verify torsion-freeness. We first
consider Sinkhorn regular curves $(\mu_t)$ and
$(\nu_t)$ that are differentiable at $t=0$,
with $\mu_0=\nu_0=\mu$, and cross tangent
fields $u^\mu$ and $u^\nu$; note that the superscript on the tangent field denotes which curve's initial velocity is encoded by that tangent field. As such, suppose the fields satisfy the initial conditions $u_0^\mu = \operatorname D_{\dot\mu_0}B_\mu,$ and $u_0^\nu = \operatorname D_{\dot\nu_0}B_\mu.$

Now fix $\varphi\in\mathcal H_c$ and consider the functional
$F_\varphi:\mathcal P(X)\to\mathbb R$ defined by
$F_\varphi(\eta):=\langle B_\eta,\varphi\rangle_{\mathcal H_c}$.
For any $u\in\mathfrak T_\eta\mathcal P(X)$, let
$\sigma_u=(\operatorname D_{(\cdot)}B_\eta)^{-1}(u)$. Then the action of
$u$ on $F_\varphi$ is naturally defined by
\[u(F_\varphi)
    :=
    \operatorname D_{\sigma_u}F_\varphi(\eta)
    =
    \left\langle
    \operatorname D_{\sigma_u}B_\eta,\varphi
    \right\rangle_{\mathcal H_c}
    =
    \langle u,\varphi\rangle_{\mathcal H_c}.\]
Applying this identity to the cross tangent fields gives
\[u_0^\nu\big(u^\mu(F_\varphi)\big)
    :=
    \frac{d}{dt}\bigg|_{t=0}
    \langle u_t^\mu,\varphi\rangle_{\mathcal H_c}
    =
    \langle\dot u_0^\mu,\varphi\rangle_{\mathcal H_c}\]
and similarly,
\[u_0^\mu\big(u^\nu(F_\varphi)\big)
    =
    \langle\dot u_0^\nu,\varphi\rangle_{\mathcal H_c}.\]
Following the functional characterization of the commutator used by
\citet{gigli2012second}, we therefore define its action on $F_\varphi$ by
\begin{align*}
[u^\nu,u^\mu](F_\varphi)(\mu)
:=
u_0^\nu\big(u^\mu(F_\varphi)\big)
-
u_0^\mu\big(u^\nu(F_\varphi)\big)=
\left\langle
\dot u_0^\mu-\dot u_0^\nu,\varphi
\right\rangle_{\mathcal H_c}.
\end{align*}
By \Cref{tangency of commutator},
$\dot u_0^\mu-\dot u_0^\nu\in\mathfrak T_\mu\mathcal P(X)$.
By the Riesz representation theorem, the continuous linear functional $\varphi\mapsto
    \left\langle
    \dot u_0^\mu-\dot u_0^\nu,\varphi
    \right\rangle_{\mathcal H_c}$ is uniquely represented by the element
$\dot u_0^\mu-\dot u_0^\nu\in\mathcal H_c$. Hence the tangent vector
representing the commutator is uniquely given by
\[
    [u^\nu,u^\mu](\mu)
    =
    \dot u_0^\mu-\dot u_0^\nu.
\]
Consequently, torsion-freeness amounts to verifying
\[\nabla_{u_0^\nu}^{\Omega}u_0^\mu
    -
    \nabla_{u_0^\mu}^{\Omega}u_0^\nu
    =
    [u^\nu,u^\mu](\mu).\]
We now compute the difference of the two covariant
derivatives to check this torsion free identity. Fix
$w\in\mathfrak T_\mu\mathcal{P}(X)$. Since $u_0^\nu = \operatorname D_{\dot\nu_0}B_\mu,$ and $u_0^\mu = \operatorname D_{\dot\mu_0}B_\mu,$ the bijectivity of
$\operatorname D_{(\cdot)}B_\mu$ (\Cref{change of vars bijectivity})
implies $\sigma_{u_0^\nu}=\dot\nu_0,$ and $\sigma_{u_0^\mu}=\dot\mu_0.$ Hence,
\begin{multline} \label{torsion difference}
    \nabla^{\operatorname{LC}}_{
    \operatorname D_{\dot\nu_0}B_\mu}
    u_0^\mu
    -
    \nabla^{\operatorname{LC}}_{
    \operatorname D_{\dot\mu_0}B_\mu}
    u_0^\nu
    =
    \Pi_\mu\bigg[
    \dot u_0^\mu-\dot u_0^\nu
    +
    \frac12 S_\mu^{-1}
    \left(
    (\operatorname D_{\dot\nu_0}S_\mu)u_0^\mu
    -
    (\operatorname D_{\dot\mu_0}S_\mu)u_0^\nu
    \right)
    \\ +
    \Omega_{(u_0^\nu)}u_0^\mu
    -
    \Omega_{(u_0^\mu)}u_0^\nu
    \bigg].
\end{multline}
By the definition of $\Omega$ and the identities
$\sigma_{u_0^\mu}=\dot\mu_0$ and
$\sigma_{u_0^\nu}=\dot\nu_0$, we have
\begin{equation*}
    \Omega_{(u_0^\nu)}u_0^\mu
    -
    \Omega_{(u_0^\mu)}u_0^\nu
    =
    \frac12 \Pi_\mu S_\mu^{-1}
    \left(
    (\operatorname D_{\dot\mu_0}S_\mu)u_0^\nu
    -
    C_\mu(u_0^\nu,u_0^\mu)^\sharp
    \right)
     -
    \frac12 \Pi_\mu  S_\mu^{-1}
    \left(
    (\operatorname D_{\dot\nu_0}S_\mu)u_0^\mu
    -
    C_\mu(u_0^\mu,u_0^\nu)^\sharp
    \right).
\end{equation*}
Since $C_\mu$ is symmetric by \Cref{C symmetry}, its Riesz representatives satisfy
$C_\mu(u_0^\nu,u_0^\mu)^\sharp
=
C_\mu(u_0^\mu,u_0^\nu)^\sharp$. Therefore,
\begin{align*}
    \Omega_{(u_0^\nu)}u_0^\mu
    -\Omega_{(u_0^\mu)}u_0^\nu
    =
    -\frac12\Pi_\mu S_\mu^{-1}
    \left(
        (\operatorname D_{\dot\nu_0}S_\mu)u_0^\mu
        -
        (\operatorname D_{\dot\mu_0}S_\mu)u_0^\nu
    \right).
\end{align*}
Combining this identity with \Cref{torsion difference}, the derivative terms involving $S_\mu$ cancel, yielding
\begin{align*}
    \nabla^{\operatorname{LC}}_{
    \operatorname D_{\dot\nu_0}B_\mu}
    u_0^\mu
    -
    \nabla^{\operatorname{LC}}_{
    \operatorname D_{\dot\mu_0}B_\mu}
    u_0^\nu
    =
    \Pi_\mu
    \left(
    \dot u_0^\mu-\dot u_0^\nu
    \right).
\end{align*}
By \Cref{tangency of commutator},
$\dot u_0^\mu-\dot u_0^\nu\in\mathfrak T_\mu\mathcal P(X)$, so the projection can be dropped. By the preceding characterization of the commutator,
$[u^\nu,u^\mu](\mu)=\dot u_0^\mu-\dot u_0^\nu$. Consequently,
\begin{align*}
    \nabla^{\operatorname{LC}}_{
    \operatorname D_{\dot\nu_0}B_\mu}
    u_0^\mu
    -
    \nabla^{\operatorname{LC}}_{
    \operatorname D_{\dot\mu_0}B_\mu}
    u_0^\nu
    =
    [u^\nu,u^\mu](\mu).
\end{align*}
Therefore the torsion-free identity holds. Since metric compatibility follows from \Cref{metric compatible family}, $\nabla^{\operatorname{LC}}$ is the Levi-Civita covariant derivative associated with $\tilde{\mathbf g}$ on $\mathfrak T_\mu\mathcal P(X)$. The preceding argument establishes torsion-freeness
for tangent directions induced by smooth bounded
zero-mass density perturbations. Under
\Cref{standing sinkhorn assumptions}, such
perturbations generate probability-valued curves
for sufficiently small times and are dense in
the completed tangent space by
\Cref{statistical riemannian tangent embedding}.
Since the left hand side of \Cref{torsion difference} is a continuous
bilinear function of the initial tangent
directions (as follows from the continuity
of the constituent operators) the identity
extends to all tangent directions.

\qed{}

\subsubsection{Supplemental Results}

\begin{lemma} \label{C symmetry}
        For every $u,v,w \in \mathfrak{T}_\mu \mathcal{P}(X)$ we have $C_\mu(u,v)(w) = C_\mu(v,u)(w).$
    \end{lemma}
    \begin{proof}
Fix $w\in\mathfrak T_\mu\mathcal P(X)$ and write
$\sigma_w=(\operatorname D_{(\cdot)}B_\mu)^{-1}(w)\in
T_\mu\mathcal P(X)\cong H^s(X)^*_0$. By
\Cref{statistical riemannian tangent embedding}, there exists a sequence
$(\gamma_n)_n\subset\mathbb T_\mu^m$ such that
$\gamma_n\to\sigma_w$ in $H^s(X)^*$.

For each $\gamma_n\in\mathbb T_\mu^m$, the Hadamard derivative
$\operatorname D_{\gamma_n}S_\mu$ is self-adjoint on $\mathcal H_c$.
Indeed, it is obtained as an operator-norm limit of difference quotients
of the self-adjoint operators $S_\eta$. By
\Cref{agreement of base derivatives}, the map $\sigma\mapsto\operatorname D_\sigma S_\mu$ is bounded from $H^s(X)^*$ into
$\mathcal L(\mathcal H_c,\mathcal H_c)$. Hence
$\operatorname D_{\gamma_n}S_\mu\to
\operatorname D_{\sigma_w}S_\mu$ in operator norm. Since the space of
bounded self-adjoint operators on $\mathcal H_c$ is closed in the
operator norm, $\operatorname D_{\sigma_w}S_\mu$ is self-adjoint.
Therefore,
\[
    C_\mu(u,v)(w)
    =
    \left\langle
    (\operatorname D_{\sigma_w}S_\mu)u,v
    \right\rangle_{\mathcal H_c}
    =
    \left\langle
    u,(\operatorname D_{\sigma_w}S_\mu)v
    \right\rangle_{\mathcal H_c}
    =
    C_\mu(v,u)(w),
\]
which proves the claim.
\end{proof}
    \begin{lemma} \label{Omega skew}
        Define $L_{(u)}$ such that $\Omega_{(u)} = S_{\mu}^{-1}L_{(u)}$. Then for every
    $u,v,w\in\mathfrak T_\mu\mathcal{P}(X)$, we have $\left\langle L_{(u)}v, w\right\rangle_{\mathcal{H}_c}
        +
        \left\langle L_{(u)}w, v\right\rangle_{\mathcal{H}_c} = 0.$
    \end{lemma}
    \begin{proof}
        Plugging in the definition of $L$ directly, and using the fact that $C_\mu(u,v)^\sharp$ is the Riesz representative of $C_\mu(u,v)$ yields
    \begin{multline*}
        \left\langle L_{(u)}v, w\right\rangle_{\mathcal{H}_c}
        +
        \left\langle L_{(u)}w,v\right\rangle_{\mathcal{H}_c}
        =
        \frac12
        \left(
        \left\langle
        (\operatorname D_{\sigma_v}S_\mu)u,w
        \right\rangle_{\mathcal H_c}
        -
        \left\langle
        (\operatorname D_{\sigma_w}S_\mu)u,v
        \right\rangle_{\mathcal H_c}
        \right)\\
        +
        \frac12
        \left(
        \left\langle
        (\operatorname D_{\sigma_w}S_\mu)u,v
        \right\rangle_{\mathcal H_c}
        -
        \left\langle
        (\operatorname D_{\sigma_v}S_\mu)u,w
        \right\rangle_{\mathcal H_c}
        \right)
        =0.
    \end{multline*}
    Thus $\left\langle L_{(u)}v, w\right\rangle_{\mathcal{H}_c}
        +
        \left\langle L_{(u)}w, v\right\rangle_{\mathcal{H}_c} = 0$. 
    \end{proof}

\begin{lemma}
\label{tangency of commutator}
Let $\epsilon > 0$ and let $(\mu_t)_{t\in(-\epsilon,\epsilon)}$ and
$(\nu_t)_{t\in(-\epsilon,\epsilon)}$ be Sinkhorn regular curves satisfying
$\mu_0=\nu_0=\mu$ and differentiable at $t = 0$. Let $u^\mu,u^\nu$ be Sinkhorn regular tangent fields along $\nu_t$ and $\mu_t$ respectively, such that they satisfy the initial
conditions
\[
    u_0^\mu=\operatorname D_{\dot\mu_0}B_\mu,
    \qquad
    u_0^\nu=\operatorname D_{\dot\nu_0}B_\mu
\]
and such that they are also differentiable at $t = 0$. Then $\dot u_0^\mu-\dot u_0^\nu\in \mathfrak{T}_\mu\mathcal P(X).$
\end{lemma}

\begin{proof}
Since $u_t^\mu\in \mathfrak{T}_{\nu_t}\mathcal P(X) = B_{\nu_t}^{\perp},$ we have $\left\langle u_t^\mu,B_{\nu_t} \right\rangle_{\mathcal H_c} =0.$ Differentiating at $t=0$ and using
\[
    \frac{d}{dt}\bigg|_{t=0}B_{\nu_t}
    =
    \operatorname D_{\dot\nu_0}B_\mu
\]
gives $\left\langle \dot u_0^\mu,B_\mu \right\rangle_{\mathcal H_c} + \left\langle
    u_0^\mu,
    \operatorname D_{\dot\nu_0}B_\mu
    \right\rangle_{\mathcal H_c}
    =0.$ 
By the initial condition stipulation, we know that $\operatorname D_{\dot\nu_0}B_\mu=u_0^\nu,$ and therefore
\[
    \left\langle
    \dot u_0^\mu,B_\mu
    \right\rangle_{\mathcal H_c}
    +
    \left\langle
    u_0^\mu,u_0^\nu
    \right\rangle_{\mathcal H_c}
    =0.
\]
Similarly, since $u_t^\nu\in \mathfrak T_{\mu_t}\mathcal P(X) = B_{\mu_t}^{\perp},$ we obtain
\[
    \left\langle
    \dot u_0^\nu,B_\mu
    \right\rangle_{\mathcal H_c}
    +
    \left\langle
    u_0^\nu,u_0^\mu
    \right\rangle_{\mathcal H_c}
    =0.
\]
Subtracting the two identities and using symmetry of the
$\mathcal H_c$ inner product yields $\left\langle \dot u_0^\mu-\dot u_0^\nu, B_\mu \right\rangle_{\mathcal H_c} =0.$ Hence $\dot u_0^\mu-\dot u_0^\nu \in B_\mu^\perp = \mathfrak T_\mu\mathcal P(X),$ as claimed.
\end{proof}

\subsection{Proof of \Cref{change of vars bijectivity}} \label{sec: proof of change of vars bijectivity}

Fix $\mu\in\mathcal P(X)$. By
\Cref{ambient inverse change of vars}, the map $B$ admits a smooth
local extension to an open neighborhood of $\mu$ in $H^s(X)^*$,
and its derivative is the inverse of $\operatorname DM(B_{\mu})$. Moreover, \Cref{ambient inverse change of vars} establishes that the restriction of $\operatorname DM(B_\mu)$ to
$B_\mu^\perp$ is a bounded linear isomorphism onto
$H^s(X)^*_0$. Its inverse is precisely the restriction
of $\operatorname DB_\mu$ to $H^s(X)^*_0$. Since
$T_\mu\mathcal P(X)\cong H^s(X)^*_0$ by
\Cref{statistical riemannian tangent embedding} and
$\mathfrak T_\mu\mathcal P(X)=B_\mu^\perp$ by definition,
we conclude that $\operatorname D_{(\cdot)}B_\mu$
is a bounded linear bijection from $T_\mu\mathcal P(X)$
onto $\mathfrak T_\mu\mathcal P(X)$.
 It remains to identify its inverse. Fix
$w\in\mathfrak T_\mu\mathcal P(X)=B_\mu^\perp$
and set $\gamma=\operatorname D_wM(B_\mu)$. By
\Cref{ambient inverse change of vars},
$\gamma=wH_c^{-1}(B_\mu)+B_\mu H_c^{-1}(w)$.
Since $H_c^{-1}(B_\mu)=a_\mu\mu$ and
$B_\mu=a_\mu^{-1}$, \Cref{Hmu Hc identity} gives
\begin{align*}
H_\mu(\gamma)
&=a_\mu H_c(a_\mu\gamma)\\
&=a_\mu H_c\left(a_\mu^2w\mu+H_c^{-1}(w)\right)\\
&=K_\mu(a_\mu w)+a_\mu w
=(\operatorname{id}_{\mathcal H_\mu}+K_\mu)(a_\mu w).
\end{align*}
Consequently,
$\gamma=H_\mu^{-1}\left((\operatorname{id}_{\mathcal H_\mu}
+K_\mu)(a_\mu w)\right)$, which is the claimed inverse formula.
Finally, $\gamma$ has zero total mass, since
\[
\gamma(\mathbbm{1}_X)
=2\langle w,B_\mu\rangle_{\mathcal H_c}=0.
\]
This completes the proof.
\qed{}

\subsection{Proof of \Cref{well posedness pt and geodesics}} \label{proof of well posedness pt and geodesics}
The geodesic statement follows from \Cref{chart geodesic ODE},
    \Cref{smoothness of chart acceleration}, and
    \Cref{local smoothness theorem}. Indeed, in the tangent-bundle
    coordinates \Cref{psi chart}, the geodesic equation is equivalent to
    a first-order ODE on $\mathfrak T_\mu\mathcal P(X)^2$ with a $C^1$
    vector field, and \Cref{local smoothness theorem} therefore yields a
    unique local solution for every coordinate initial condition $(0,v)$,
    corresponding to the ambient initial state $(B_\mu,v)$.

    For parallel transport, fix $t_0\in[0,T]$ and choose a local chart
    containing the base curve on a neighborhood $I$ of $t_0$. By
    \Cref{proof of three coordinate ode}, writing
    $w_t=\operatorname D_{r_t}\psi(p_t)$ transforms the parallel transport
    equation into
    \[
        \dot r_t=\mathscr R(p_t,u_t,r_t).
    \]
    Since the base curve is second-order Sinkhorn regular,
    $\beta\in H^2([0,T];\mathcal H_c)$, and therefore
    $\beta\in C^1([0,T];\mathcal H_c)$ and
    $\dot\beta\in C^0([0,T];\mathcal H_c)$. Consequently, the corresponding
    chart coordinates $p_t$ and $u_t$ are continuous, with $p_t$ continuously
    differentiable. By \Cref{smoothness of chart parallel transport},
    $\mathscr R$ is smooth in $(p,u,r)$ and bounded linear in $r$ for fixed
    $(p,u)$. Thus the parallel transport equation is a locally Lipschitz
    Banach-space ODE with continuous time-dependent coefficients; by the Cauchy-Lipschitz theorem for Banach-space ODEs \citep[Proposition 1.1]{lang2012fundamentals}, it
    admits a unique local solution through the prescribed value $r_{t_0}$.
    Pulling this solution back through the chart yields the unique local
    parallel tangent field satisfying $w_{t_0}=w_0$. \qed{}

\subsection{Additional Results for \Cref{sec: deriving the connection}}

\begin{lemma}[Sinkhorn regular curves are admissible]
\label{sinkhorn regular implies admissible}
Under \Cref{standing sinkhorn assumptions}, every first-order Sinkhorn regular curve is admissible in the sense of \citet[Definition 5.1(1)]{lavenant2024riemannian}. In particular, if
$\mu\in H^1([0,1];H^m(X)^*)$, then its associated curve
$\beta_t=B_{\mu_t}$ satisfies
$\beta\in H^1([0,1];\mathcal H_c)$.
\end{lemma}
\begin{proof}
Let $\mu\in H^1([0,1];H^m(X)^*)$. By
\citet[Theorem 5.2]{lavenant2024riemannian}, for any
$\mu,\nu\in\mathcal P(X)$,
\[
    \sqrt{\frac{\varepsilon}{2}}
    \|B_\mu-B_\nu\|_{\mathcal H_c}
    \leq
    d_S(\mu,\nu).
\]
Moreover, \citet[Proposition 5.9]{lavenant2024riemannian} gives
$d_S(\mu,\nu)\lesssim_{c,\varepsilon,m}
\|\mu-\nu\|_{C^m(X)^*}$. Since
$C^m(X)\hookrightarrow H^m(X)$ continuously, duality yields
$H^m(X)^*\hookrightarrow C^m(X)^*$ (\Cref{duality of continuous embedding}), and therefore
\[
    \|B_\mu-B_\nu\|_{\mathcal H_c}
    \lesssim_{c,\varepsilon,m}
    \|\mu-\nu\|_{H^m(X)^*}.
\]
Thus $B$ is Lipschitz from $\mathcal P(X)\subset H^m(X)^*$ into
$\mathcal H_c$. Since
$\mu\in H^1([0,1];H^m(X)^*)$, the equivalence between Hilbert-valued
$H^1$ and $AC^2$ curves implies
$\mu\in AC^2([0,1];H^m(X)^*)$
\citep[Theorem C.2]{lavenant2024riemannian}. The Lipschitz composition
property for absolutely continuous curves then gives
$B_{\mu_\cdot}\in AC^2([0,1];\mathcal H_c)$, and another application
of the same equivalence yields
$B_{\mu_\cdot}\in H^1([0,1];\mathcal H_c)$. Hence
$(\mu_t)_{t\in[0,1]}$ is admissible in the sense of
\citet[Definition 5.1(1)]{lavenant2024riemannian}.
\end{proof}

\begin{lemma}\label{Hmu Hc identity}
Let $\mu\in\mathcal P(X)$ and let $\sigma\in\mathcal{H}_\mu^*$. Define
$a_{\mu}\sigma\in\mathcal{H}_c^*$ by $(a_\mu\sigma)(\varphi):=\sigma(a_\mu\varphi)$ for any $\varphi\in\mathcal H_c.$ Then
\[
H_\mu(\sigma)
=
a_\mu H_c(a_\mu\sigma).
\]

\end{lemma}

\begin{proof}
Recalling that $k_\mu(x,y)
=
a_\mu(x)a_\mu(y)k_c(x,y)$ yields for every $y\in X$,
\begin{multline*}
H_\mu(\sigma)(y) =
\sigma\bigl(k_\mu(\cdot,y)\bigr)=
\sigma\bigl(a_\mu(\cdot)a_\mu(y)k_c(\cdot,y)\bigr) =
a_\mu(y)
\sigma\bigl(a_\mu(\cdot)k_c(\cdot,y)\bigr)\\
=
a_\mu(y)
(a_\mu\sigma)\bigl(k_c(\cdot,y)\bigr)=
a_\mu(y)H_c(a_\mu\sigma)(y).
\end{multline*}
Since this holds for every $y\in X$, we conclude that $H_\mu(\sigma)=a_\mu H_c(a_\mu\sigma).$
\end{proof}

\begin{lemma} \label{Cmu riesz rep}
    For every $\mu\in\mathcal P(X)$ and for every $u,v\in \mathfrak{T}_\mu\mathcal P(X)$, the functional $C_\mu(u,v)\in \mathfrak{T}_\mu\mathcal{P}(X)^*$ admits the Riesz representative
    \[C_\mu(u,v)^\sharp = 2\left( a_\mu (\mathbf{R}_\mu u)(\mathbf{R}_\mu v) - H_c\!\left[ a_\mu^3 (\mathbf{R}_\mu u)(\mathbf{R}_\mu v)\mu \right] \right). \]
\end{lemma}
For directions $\sigma_w\in T_\mu\mathcal P(X)$ that need not belong
to $\mathbb T_\mu^m$, all base-point derivatives below are understood
as derivatives of the $H^s(X)^*$ smooth extensions from
\Cref{agreement of base derivatives}. On $\mathbb T_\mu^m$, these
coincide with the Hadamard derivatives derived above.

\begin{proof}
    Recall that $\mathbf{K}_\mu=\Pi_\mu\mathbf{J}_\mu$. Since $\mathbf{J}_\mu$ maps into $\mathfrak{T}_\mu\mathcal{P}(X)=E_\mu^\perp$, we have $\Pi_\mu\mathbf{J}_\mu=\mathbf{J}_\mu$, and hence $\mathbf{K}_\mu=\mathbf{J}_\mu$. Moreover, $\mathbf{J}_\mu$ is self-adjoint, since for any $h,g\in\mathcal{H}_c$,
    \[
        \langle \mathbf{J}_\mu h,g\rangle_{\mathcal{H}_c}
        =
        \int_X a_\mu^2(\Pi_\mu h)(\Pi_\mu g)\,d\mu
        =
        \langle h,\mathbf{J}_\mu g\rangle_{\mathcal{H}_c}.
    \]
    In particular, $\mathbf{K}_\mu$ is self-adjoint. By the definition of $\mathbf{R}_\mu$, we have
    $(\operatorname{id}_{\mathcal{H}_c}-\mathbf{K}_\mu)\mathbf{R}_\mu=\Pi_\mu$.
    Thus $\mathbf{K}_\mu\mathbf{R}_\mu=\mathbf{R}_\mu-\Pi_\mu$, and consequently
    $S_\mu=\operatorname{id}_{\mathcal{H}_c}+2(\mathbf{R}_\mu-\Pi_\mu)$.
    Differentiating in the direction $\sigma_w$ gives
    \[
        \operatorname{D}_{\sigma_w}S_\mu
        =
        2\left(
        \operatorname{D}_{\sigma_w}\mathbf{R}_\mu
        -
        \operatorname{D}_{\sigma_w}\Pi_\mu
        \right).
    \]
    Furthermore, differentiating
    $(\operatorname{id}_{\mathcal{H}_c}-\mathbf{K}_\mu)\mathbf{R}_\mu=\Pi_\mu$
    yields
    \[
        (\operatorname{id}_{\mathcal{H}_c}-\mathbf{K}_\mu)
        \operatorname{D}_{\sigma_w}\mathbf{R}_\mu
        =
        (\operatorname{D}_{\sigma_w}\mathbf{K}_\mu)\mathbf{R}_\mu
        +
        \operatorname{D}_{\sigma_w}\Pi_\mu.
    \]
    Since $u,v\in\mathfrak{T}_\mu\mathcal{P}(X)$, we have $\Pi_\mu u=u$ and
    $\Pi_\mu v=v$, and therefore
    $(\operatorname{id}_{\mathcal{H}_c}-\mathbf{K}_\mu)\mathbf{R}_\mu v=v$.
    Using the self-adjointness of
    $\operatorname{id}_{\mathcal{H}_c}-\mathbf{K}_\mu$, it follows that
    \begin{align*}
        \frac12 C_\mu(u,v)(w)
        &=
        \left\langle
        \left(
        \operatorname{D}_{\sigma_w}\mathbf{R}_\mu
        -
        \operatorname{D}_{\sigma_w}\Pi_\mu
        \right)u,v
        \right\rangle_{\mathcal{H}_c} \\
        &=
        \left\langle
        (\operatorname{D}_{\sigma_w}\mathbf{K}_\mu)\mathbf{R}_\mu u,
        \mathbf{R}_\mu v
        \right\rangle_{\mathcal{H}_c}
        +
        \left\langle
        (\operatorname{D}_{\sigma_w}\Pi_\mu)u,
        \mathbf{R}_\mu v-v
        \right\rangle_{\mathcal{H}_c}.
    \end{align*}
    By definition of $\sigma_w$, we have
    $\operatorname D_{\sigma_w}B_\mu=w$. Since
    $\Pi_\mu=\operatorname{id}_{\mathcal H_c}
    -B_\mu\otimes B_\mu$ and
    $w\in B_\mu^\perp$, differentiation of the smooth extension of
    $\Pi_\mu$ gives
    \[
        (\operatorname{D}_{\sigma_w}\Pi_\mu)u
        =
        -\langle u,w\rangle_{\mathcal{H}_c}e_\mu
        -
        \langle u,e_\mu\rangle_{\mathcal{H}_c}w
        =
        -\langle u,w\rangle_{\mathcal{H}_c}e_\mu.
    \]
    On the other hand,
    $\mathbf{R}_\mu v-v=\mathbf{K}_\mu\mathbf{R}_\mu v\in E_\mu^\perp$.
    Hence
    \[
        \left\langle
        (\operatorname{D}_{\sigma_w}\Pi_\mu)u,
        \mathbf{R}_\mu v-v
        \right\rangle_{\mathcal{H}_c}
        =0.
    \]
    The smooth extensions satisfy
    $\mathbf K_\lambda=\mathbf J_\lambda$ in a neighborhood of $\mu$,
    since $\mathbf J_\lambda$ takes values in
    $B_\lambda^\perp$ and hence
    $\Pi_\lambda\mathbf J_\lambda=\mathbf J_\lambda$. Therefore,
    \[
        \operatorname D_{\sigma_w}\mathbf K_\mu
        =
        \operatorname D_{\sigma_w}\mathbf J_\mu.
    \]
    By \Cref{agreement of base derivatives}, the map
    $\lambda\mapsto\mathbf J_\lambda$ is Fr\'echet differentiable in the
    $H^s(X)^*$ topology. Differentiating $\mathbf J_\lambda z
        =
        H_c\left(a_\lambda^2\Pi_\lambda z\,\lambda\right)$ in the direction $\sigma_w$ gives
    \[
        (\operatorname D_{\sigma_w}\mathbf J_\mu)z
        =
        H_c\left((a_\mu^2\Pi_\mu z)\sigma_w\right)
        +
        H_c\left(
            \left[
                2a_\mu(\operatorname D_{\sigma_w}a_\mu)\Pi_\mu z
                +
                a_\mu^2(\operatorname D_{\sigma_w}\Pi_\mu)z
            \right]\mu
        \right).
    \] Since
    $\mathbf{R}_\mu u,\mathbf{R}_\mu v\in E_\mu^\perp$, the projection
    terms simplify to give
    \begin{equation*}
        \frac12 C_\mu(u,v)(w)
        =
        \sigma_w\left(
        a_\mu^2(\mathbf{R}_\mu u)(\mathbf{R}_\mu v)
        \right) 
        +
        \int_X
        \left[
        2a_\mu(\operatorname{D}_{\sigma_w}a_\mu)(\mathbf{R}_\mu u)
        +
        a_\mu^2
        (\operatorname{D}_{\sigma_w}\Pi_\mu)(\mathbf{R}_\mu u)
        \right]
        (\mathbf{R}_\mu v)\,d\mu.
    \end{equation*}
    Since $a_\mu=e_\mu^{-1}$ and $\operatorname{D}_{\sigma_w}e_\mu=w$,
    the chain rule gives
    $\operatorname{D}_{\sigma_w}a_\mu=-a_\mu^2w$. Moreover,
    \[
        (\operatorname{D}_{\sigma_w}\Pi_\mu)(\mathbf{R}_\mu u)
        =
        -\langle \mathbf{R}_\mu u,w\rangle_{\mathcal{H}_c}e_\mu,
    \]
    because $\mathbf{R}_\mu u\in E_\mu^\perp$. The contribution of this
    projection term vanishes, since
    \[
        \int_X a_\mu^2e_\mu(\mathbf{R}_\mu v)\,d\mu
        =
        \int_X a_\mu(\mathbf{R}_\mu v)\,d\mu
        =
        \langle e_\mu,\mathbf{R}_\mu v\rangle_{\mathcal{H}_c}
        =
        0.
    \]
    Hence
    \[
        \frac12 C_\mu(u,v)(w)
        =
        \sigma_w\left(
        a_\mu^2(\mathbf{R}_\mu u)(\mathbf{R}_\mu v)
        \right)
        -
        2\int_X
        a_\mu^3(\mathbf{R}_\mu u)(\mathbf{R}_\mu v)w\,d\mu.
    \]
    By \Cref{change of vars bijectivity},
    $H_\mu(\sigma_w)
    =(\operatorname{id}_{\mathcal{H}_\mu}+K_\mu)(a_\mu w)$.
    Since multiplication by $a_\mu$ is the isometric isomorphism from
    $\mathcal{H}_c$ to $\mathcal{H}_\mu$, we obtain
    \begin{align*}
        \sigma_w\left(
        a_\mu^2(\mathbf{R}_\mu u)(\mathbf{R}_\mu v)
        \right)
        &=
        \left\langle
        H_\mu(\sigma_w),
        a_\mu^2(\mathbf{R}_\mu u)(\mathbf{R}_\mu v)
        \right\rangle_{\mathcal{H}_\mu} \\
        &=
        \left\langle
        (\operatorname{id}_{\mathcal{H}_\mu}+K_\mu)(a_\mu w),
        a_\mu^2(\mathbf{R}_\mu u)(\mathbf{R}_\mu v)
        \right\rangle_{\mathcal{H}_\mu} \\
        &=
        \left\langle
        w,
        a_\mu(\mathbf{R}_\mu u)(\mathbf{R}_\mu v)
        \right\rangle_{\mathcal{H}_c}
        +
        \int_X
        a_\mu^3(\mathbf{R}_\mu u)(\mathbf{R}_\mu v)w\,d\mu.
    \end{align*}
    Combining the previous two displays gives
    \begin{align*}
        \frac12 C_\mu(u,v)(w)
        &=
        \left\langle
        w,
        a_\mu(\mathbf{R}_\mu u)(\mathbf{R}_\mu v)
        \right\rangle_{\mathcal{H}_c}
        -
        \int_X
        a_\mu^3(\mathbf{R}_\mu u)(\mathbf{R}_\mu v)w\,d\mu \\
        &=
        \left\langle
        w,\,
        a_\mu(\mathbf{R}_\mu u)(\mathbf{R}_\mu v)
        -
        H_c\!\left[
        a_\mu^3(\mathbf{R}_\mu u)(\mathbf{R}_\mu v)\mu
        \right]
        \right\rangle_{\mathcal{H}_c}.
    \end{align*}
    Therefore,
    \[
        C_\mu(u,v)(w)
        =
        \left\langle
        2\left(
        a_\mu(\mathbf{R}_\mu u)(\mathbf{R}_\mu v)
        -
        H_c\!\left[
        a_\mu^3(\mathbf{R}_\mu u)(\mathbf{R}_\mu v)\mu
        \right]
        \right),
        w
        \right\rangle_{\mathcal{H}_c}.
    \]
    Finally, it remains to show that the element in the first argument is tangent. Since
    \begin{multline*}
        \left\langle
        a_\mu(\mathbf{R}_\mu u)(\mathbf{R}_\mu v)
        -
        H_c\!\left[
        a_\mu^3(\mathbf{R}_\mu u)(\mathbf{R}_\mu v)\mu
        \right],
        e_\mu
        \right\rangle_{\mathcal{H}_c} 
        \\ =
        \int_X a_\mu^2(\mathbf{R}_\mu u)(\mathbf{R}_\mu v)\,d\mu
        -
        \int_X a_\mu^2(\mathbf{R}_\mu u)(\mathbf{R}_\mu v)\,d\mu
        =0.
    \end{multline*}
    Thus it belongs to $\mathfrak{T}_\mu\mathcal{P}(X)$, and the claimed
    expression is the Riesz representative of $C_\mu(u,v)$.
\end{proof}

\subsection{Additional Results for \Cref{sec: geodesic equations and pt}}

\begin{lemma}\label{temporal spatial interpolation}
Let $d/2<r<m<s$, and let $p\in\mathbb N$ satisfy
$p(m-r)>s-m$. Suppose $(\mu_t)_{t\in[0,1]}\subset\mathcal P(X)$ is
$C^{p+1}$ as an $H^s(X)^*$-valued curve. Then
$\mu\in C^1([0,1];H^m(X)^*)$. In particular,
$\mu\in H^1([0,1];H^m(X)^*)$.
\end{lemma}

\begin{proof}
Set $\theta=(m-r)/(s-r)\in(0,1)$. By the Sobolev interpolation theorem
\citep[Theorem 6.4.5]{bergh1976interpolation}, there exists a constant
$C>0$ such that
\[
    \|z\|_{H^m(X)^*}
    \leq
    C
    \|z\|_{H^r(X)^*}^{1-\theta}
    \|z\|_{H^s(X)^*}^{\theta}
\]
for every $z\in H^r(X)^*$. Since $r>d/2$, the Sobolev embedding
$H^r(X)\hookrightarrow C(X)$ implies
\[
    \sup_{t\in[0,1]}
    \|\mu_t\|_{H^r(X)^*}
    \lesssim
    \sup_{t\in[0,1]}\|\mu_t\|_{\operatorname{TV}}
    =1.
\]
In particular, applying the preceding interpolation inequality to
$\mu_t-\mu_q$ shows that continuity of $t\mapsto\mu_t$ in
$H^s(X)^*$ implies continuity in $H^m(X)^*$.

Now choose constants $c_0,\ldots,c_p$ satisfying
$\sum_{j=0}^p c_j=0$, $\sum_{j=0}^p jc_j=1$, and
$\sum_{j=0}^p j^\ell c_j=0$ for $2\leq\ell\leq p$ (noting that such constants
necessarily exist by invertibility of the corresponding Vandermonde matrix). For
$t\in[0,1/2]$ and $0<h\leq 1/(2p)$, define
\[
    Q_h(t)
    :=
    \frac1h\sum_{j=0}^p c_j\mu_{t+jh}.
\]
Taylor expansion in the Banach space $H^s(X)^*$ gives, uniformly for
$t\in[0,1/2]$, $\|Q_h(t)-\dot\mu_t\|_{H^s(X)^*}
    \lesssim
    h^p.$ On the other hand, the uniform $H^r(X)^*$ bound on the probability
measures gives
$\|Q_h(t)\|_{H^r(X)^*}\lesssim h^{-1}$. Now let $h_n=2^{-n}h_0$ for some fixed $h_0\leq1/(2p)$. The preceding
bounds imply
\[
    \|Q_{h_n}(t)-Q_{h_{n+1}}(t)\|_{H^r(X)^*}
    \lesssim h_n^{-1} \quad \text{and} \quad 
    \|Q_{h_n}(t)-Q_{h_{n+1}}(t)\|_{H^s(X)^*}
    \lesssim h_n^p.
\]
Hence, by the interpolation inequality,
\[
    \|Q_{h_n}(t)-Q_{h_{n+1}}(t)\|_{H^m(X)^*}
    \lesssim
    h_n^\delta,
    \qquad
    \delta
    :=
    p\theta-(1-\theta)
    =
    \frac{p(m-r)-(s-m)}{s-r}
    >0.
\]
Thus $(Q_{h_n})_n$ is uniformly Cauchy in
$C([0,1/2];H^m(X)^*)$. Its limit agrees with $\dot\mu$ in
$H^s(X)^*$, and therefore $\dot\mu_t\in H^m(X)^*$ for
$t\in[0,1/2]$. The same argument using backward differences applies on
$[1/2,1]$, and the two limits agree at $t=1/2$ because they both agree
with the $H^s(X)^*$ derivative there. Consequently,
$\dot\mu\in C([0,1];H^m(X)^*)$. Finally, the identity
$\mu_t-\mu_0=\int_0^t\dot\mu_q\,dq$, initially valid in
$H^s(X)^*$, is also an identity in $H^m(X)^*$ because the integrand is
continuous in $H^m(X)^*$. Hence
$\mu\in C^1([0,1];H^m(X)^*)$.
\end{proof}

\subsubsection{Proof of \Cref{geodesic coincidence} (a)} \label{proof of geodesic second order regularity}

Since $(\mu_t^\star)_{t\in[0,1]}$ is a minimizer of
\Cref{variational geodesic}, its associated curve
$\beta_t^\star=B_{\mu_t^\star}$ belongs to
$H^1([0,1];\mathcal H_c)$. Recall that
$M(\beta)=\beta H_c^{-1}(\beta)$ satisfies
$M(B_\mu)=\mu$. Since multiplication
$\mathcal H_c\times\mathcal H_c^*\to\mathcal H_c^*$ is bounded
bilinear, $M:\mathcal H_c\to\mathcal H_c^*$ is $C^\infty$. Using
$\mathcal H_c^*\simeq H^s(X)^*$ and the Banach-valued chain rule
therefore gives
\[
    \boldsymbol{\mu}^\star
    \in
    H^1([0,1];H^s(X)^*).
\]

\medskip

\emph{The variational criterion.} For each $\mu$ along the curve, define $A_\mu:H^s(X)^*_0\to H^s(X)^*_0$ by Riesz representation through $\left\langle A_\mu u,v\right\rangle_{H^s(X)^*}
=
\left\langle S_\mu\operatorname{D}_{u}B_\mu,\operatorname{D}_{v}B_\mu\right\rangle_{\mathcal H_c}.$ This is a bounded self-adjoint operator since $\operatorname{D}_{u}B_\mu,\operatorname{D}_{v}B_\mu\in\mathfrak T_\mu\mathcal P(X)$ and $S_\mu$ is self-adjoint on $\mathfrak T_\mu\mathcal P(X)$ by \Cref{operator metric equivalence}. The proof of \Cref{ambient inverse change of vars} establishes that $B$ admits a local $C^\infty$ extension and is a local $C^\infty$ diffeomorphism, while \Cref{agreement of base derivatives} establishes that $S_\mu$ depends $C^\infty$-smoothly on the base point in operator norm. Consequently, $\mu\mapsto A_\mu$ is locally $C^\infty$ as a map into $\mathcal L(H^s(X)^*_0)$. Since $\boldsymbol{\mu}^\star\in H^1([0,1];H^s(X)^*)$ has a continuous representative (and therefore compact image), the operator norms of $A_{\mu_t^\star}$ and $\operatorname{D}A_{\mu_t^\star}$ are uniformly bounded along the path. With this notation, the energy can be written as
\begin{equation}\label{energy}
    \mathscr E(\boldsymbol{\mu})
    =
    \frac{\varepsilon}{2}\int_0^1
    \left\langle
    A_{\mu_t}\dot\mu_t,\dot\mu_t
    \right\rangle_{H^s(X)^*}\,dt.
\end{equation}
Indeed, $\dot\beta_t=\operatorname{D}_{\dot\mu_t}B_{\mu_t}$, so this is precisely the objective in \Cref{variational geodesic}. Now consider perturbations of the form
\begin{equation*}
    \sigma_t=\zeta(t)h\,dx,
    \qquad
    \zeta\in C_c^\infty((0,1)),
    \qquad
    h\in C^\infty(X),
    \qquad
    \int_Xh\,dx=0.
\end{equation*}
Since $\rho_t^\star\geq\rho_{\min}>0$ uniformly in $(t,x)$ and
$\zeta h$ is bounded, $\mu_t^\star+\epsilon\sigma_t$ remains a
probability measure for all sufficiently small $|\epsilon|$. Moreover,
$\boldsymbol{\sigma}\in H^1([0,1];H^s(X)^*)$. By the local
$H^s(X)^*$-smoothness of the change of variables established in
\Cref{ambient inverse change of vars}, the corresponding curve
$B_{\mu^\star+\epsilon\sigma}$ belongs to
$H^1([0,1];\mathcal H_c)$ for sufficiently small $|\epsilon|$.
Therefore $\boldsymbol{\mu}^\star+\epsilon\boldsymbol{\sigma}$ is an
admissible competitor in \Cref{variational geodesic}, and minimality
implies
$\delta\mathscr E(\boldsymbol{\mu}^\star)[\boldsymbol{\sigma}]=0$ \citep[Proposition 5.5]{troutman2012variational}. Differentiating the pulled-back energy gives
\begin{equation*}
    0
    =
    \int_0^1
    \zeta(t)
    \left\langle
    (\operatorname{D}_{h\,dx}A_{\mu_t^\star})\dot\mu_t^\star,
    \dot\mu_t^\star
    \right\rangle_{H^s(X)^*}\,dt
    +
    2\int_0^1
    \dot\zeta(t)
    \left\langle
    A_{\mu_t^\star}\dot\mu_t^\star,
    h\,dx
    \right\rangle_{H^s(X)^*}\,dt.
\end{equation*}
Since $\zeta\in C_c^\infty((0,1))$ is arbitrary, the definition of the distributional derivative yields the variational criterion
\begin{equation}\label{pulled back distributional identity}
    \frac{d}{dt}
    \left\langle
    A_{\mu_t^\star}\dot\mu_t^\star,
    h\,dx
    \right\rangle_{H^s(X)^*}
    =
    \frac12
    \left\langle
    (\operatorname{D}_{h\,dx}A_{\mu_t^\star})\dot\mu_t^\star,
    \dot\mu_t^\star
    \right\rangle_{H^s(X)^*} \tag{\textbf{VC}}
\end{equation}
in the distributional sense. Uniform boundedness of $\operatorname{D}A_{\mu_t^\star}$ along the path gives a constant $C>0$ such that
\begin{equation}\label{quadratic bound A}
    \left|
    \left\langle
    (\operatorname{D}_{\psi}A_{\mu_t^\star})\dot\mu_t^\star,
    \dot\mu_t^\star
    \right\rangle_{H^s(X)^*}
    \right|
    \leq
    C\|\psi\|_{H^s(X)^*}\|\dot\mu_t^\star\|_{H^s(X)^*}^2.
\end{equation}
Since $\boldsymbol{\mu}^\star\in H^1([0,1];H^s(X)^*)$, we have $\dot\mu^\star\in L^2([0,1];H^s(X)^*_0)$, so the right-hand side of \Cref{quadratic bound A} is integrable in time.

\medskip

\emph{Regularity of $\dot \mu_t^\star.$} Now, by the density argument in the proof of \Cref{statistical riemannian tangent embedding}, smooth zero-mass densities $h\,dx$ are dense in $H^s(X)^*_0$. Since both sides of \eqref{pulled back distributional identity} depend continuously and linearly on the spatial test direction, the identity extends uniquely to every $\psi\in H^s(X)^*_0$ by the bounded linear extension theorem \citep[Theorem 2.7-11]{kreyszig1991introductory}. Thus
\begin{equation} \label{distributional identities}
    \frac{d}{dt}
    \left\langle
    A_{\mu_t^\star}\dot\mu_t^\star,\psi
    \right\rangle_{H^s(X)^*}
    =
    \frac12
    \left\langle
    (\operatorname{D}_{\psi}A_{\mu_t^\star})\dot\mu_t^\star,
    \dot\mu_t^\star
    \right\rangle_{H^s(X)^*}
    \qquad
    \text{for every }\psi\in H^s(X)^*_0.
\end{equation}
For almost every $t$, the right-hand side defines a bounded linear functional of $\psi$. Let $\mathfrak F_t\in H^s(X)^*_0$ denote its Riesz representative, so that
\begin{equation}
    \left\langle
    \mathfrak F_t,\psi
    \right\rangle_{H^s(X)^*}
    =
    \frac12
    \left\langle
    (\operatorname{D}_{\psi}A_{\mu_t^\star})\dot\mu_t^\star,
    \dot\mu_t^\star
    \right\rangle_{H^s(X)^*}.
\end{equation}
By \Cref{quadratic bound A}, $ \|\mathfrak F_t\|_{H^s(X)^*}
    \leq
    C\|\dot\mu_t^\star\|_{H^s(X)^*}^2,$
and therefore $\mathfrak F\in L^1([0,1];H^s(X)^*_0)$. Moreover, uniform boundedness of $A_{\mu_t^\star}$ and $\dot\mu^\star\in L^2$ imply $A_{\mu^\star}\dot\mu^\star\in L^2([0,1];H^s(X)^*_0)\subset L^1([0,1];H^s(X)^*_0)$. Hence the distributional identity in \Cref{distributional identities} is precisely the statement that $\mathfrak F$ is the $H^s(X)^*_0$-valued weak derivative of $A_{\mu^\star}\dot\mu^\star$. Consequently,
\begin{equation}\label{momentum W11}
    A_{\mu^\star}\dot\mu^\star
    \in
    W^{1,1}([0,1];H^s(X)^*_0).
\end{equation}
We next verify that $A_{\mu_t^\star}$ is uniformly coercive, which will allow us to transfer the regularity on $A_{\mu^\star}\dot\mu^\star$ to $\dot\mu^\star$. By \Cref{operator metric equivalence} and \Cref{invertibility of st}, $S_{\mu_t^\star}$ is positive on $\mathfrak T_{\mu_t^\star}\mathcal P(X)$ and satisfies $\langle
S_{\mu_t^\star}v,v
\rangle_{\mathcal H_c}
\geq
\|v\|_{\mathcal H_c}^2.$ Furthermore, the proof of \Cref{ambient inverse change of vars} shows that $B$ is a local $C^\infty$ diffeomorphism, so $(\operatorname{D}_{(\cdot)}B_{\mu})^{-1}$ depends continuously on the base point. Since $\boldsymbol{\mu}^\star\in H^1([0,1];H^s(X)^*)$, it admits a continuous representative as an $H^s(X)^*$-valued curve by the characterization $H^1=AC^2$ for Hilbert-valued curves \citep[Theorem~2.11]{gigli2021korevaar}. Hence the image $\{\mu_t^\star:t\in[0,1]\}$ is compact in $H^s(X)^*$ as the continuous image of the compact interval $[0,1]$. Compactness of the image of $(\mu_t^\star)_{t\in[0,1]}$ therefore gives a constant $C_B>0$ such that $\|u\|_{H^s(X)^*}
    \leq
    C_B
    \|\operatorname{D}_{u}B_{\mu_t^\star}\|_{\mathcal H_c}$
uniformly in $t$. Consequently,
\begin{equation}
    \left\langle
    A_{\mu_t^\star}u,u
    \right\rangle_{H^s(X)^*}
    \geq
    C_B^{-2}\|u\|_{H^s(X)^*}^2.
\end{equation}
The Lax--Milgram theorem \citep[Section 6.2.1, Theorem 1]{evans2022partial} therefore implies that $A_{\mu_t^\star}$ is invertible and that $A_{\mu_t^\star}^{-1}$ is uniformly bounded along the path. By \Cref{momentum W11}, $A_{\mu^\star}\dot\mu^\star$ admits an absolutely continuous representative and is therefore bounded on $[0,1]$. Hence $\dot\mu_t^\star = A_{\mu_t^\star}^{-1} \left(A_{\mu_t^\star}\dot\mu_t^\star\right)$ implies $\dot\mu^\star\in L^\infty([0,1];H^s(X)^*_0)$, and therefore
\begin{equation}\label{mu W1 infinity}
    \mu^\star\in W^{1,\infty}([0,1];H^s(X)^*).
\end{equation}

\emph{Regularity of $ \mu_t^\star.$} Returning to \Cref{quadratic bound A}, the improved regularity $\dot\mu^\star\in L^\infty$ now gives $\mathfrak F\in L^\infty([0,1];H^s(X)^*_0)$. Thus
\begin{equation}\label{momentum W1 infinity}
    A_{\mu^\star}\dot\mu^\star
    \in
    W^{1,\infty}([0,1];H^s(X)^*_0).
\end{equation}
Since $\mu^\star\in W^{1,\infty}([0,1];H^s(X)^*)$ and $\mu\mapsto A_\mu$ is $C^1$ in operator norm, Banach-valued chain rule gives $A_{\mu^\star} \in W^{1, \infty}([0,1]; \mathcal{L}(H^s(X)^*_0, H^s(X)^*_0))$. Using the argument of \Cref{tilde R hadamard diff}, inversion is
Fr\'echet differentiable on the open subset of invertible operators, so
applying the chain rule again gives
$A_{\mu^\star}^{-1}\in
W^{1,\infty}([0,1];
\mathcal L(H^s(X)^*_0,H^s(X)^*_0))$.

Combining \Cref{momentum W1 infinity} with the
$W^{1,\infty}$ regularity of $A_{\mu^\star}^{-1}$ gives
\[
    \dot\mu^\star
    =
    A_{\mu^\star}^{-1}
    \left(A_{\mu^\star}\dot\mu^\star\right)
    \in
    W^{1,\infty}([0,1];H^s(X)^*_0).
\]
In particular,
$\mu^\star\in W^{2,\infty}([0,1];H^s(X)^*)$. Now, for $\mu$ in a neighborhood of the path and
$v\in H^s(X)^*_0$, let $\mathfrak F(\mu,v)\in H^s(X)^*_0$ denote the
Riesz representative characterized by
\[
    \left\langle
        \mathfrak F(\mu,v),\psi
    \right\rangle_{H^s(X)^*}
    =
    \frac12
    \left\langle
        (\operatorname D_\psi A_\mu)v,v
    \right\rangle_{H^s(X)^*}.
\]
Since $\mu\mapsto A_\mu$ is $C^\infty$ in operator norm,
$(\mu,v)\mapsto\mathfrak F(\mu,v)$ is $C^\infty$. The variational
identity may therefore be written in strong form as
\[
    \ddot\mu_t^\star
    =
    A_{\mu_t^\star}^{-1}
    \left[
        \mathfrak F(\mu_t^\star,\dot\mu_t^\star)
        -
        (\operatorname D_{\dot\mu_t^\star}A_{\mu_t^\star})
        \dot\mu_t^\star
    \right]
    \qquad\text{for a.e. }t.
\]
The right-hand side is a $C^\infty$ function of
$(\mu_t^\star,\dot\mu_t^\star)$ as an
$H^s(X)^*\times H^s(X)^*_0$-valued variable. Since
$(\mu^\star,\dot\mu^\star)\in W^{1,\infty}$, the right-hand side is
continuous, and hence $(\mu^\star,\dot\mu^\star)$ is $C^1$.
Iterating the chain rule then gives
\[
    \mu^\star
    \in
    C^\infty([0,1];H^s(X)^*).
\]
Now choose any $r\in(d/2,m)$ and then choose $p\in\mathbb N$ sufficiently
large that $p(m-r)>s-m$. Since
$\mu_t^\star\in\mathcal P(X)$ for every $t$ and
$\mu^\star\in C^\infty([0,1];H^s(X)^*)$,
\Cref{temporal spatial interpolation} yields
\[
    \mu^\star
    \in
    C^1([0,1];H^m(X)^*)
    \subset
    H^1([0,1];H^m(X)^*).
\]
Thus $(\mu_t^\star)_{t\in[0,1]}$ is first-order Sinkhorn regular.

\medskip

\medskip

\emph{Putting it together.} Finally, the map $B$ is locally
$C^\infty$ from $H^s(X)^*$ into $\mathcal H_c$. Since
$\mu^\star\in C^\infty([0,1];H^s(X)^*)$, it follows that
$\beta^\star=B_{\mu^\star}$ is $C^\infty$ as an
$\mathcal H_c$-valued curve. In particular,
$\dot\beta^\star\in H^1([0,1];\mathcal H_c)$, so
$\dot\beta^\star$ is a Sinkhorn regular tangent field. Together with
the first-order Sinkhorn regularity established above, this shows that
$(\mu_t^\star)_{t\in[0,1]}$ is second-order Sinkhorn regular.
\qed{}

\subsubsection{Proof of \Cref{geodesic coincidence} (b)} \label{proof of geodesic coincidence}

Consider the variational criterion \eqref{pulled back distributional identity} for any minimizer of the objective in \Cref{variational geodesic}:
\[
    \frac{d}{dt}
    \left\langle
        A_{\mu_t^\star}\dot\mu_t^\star,\psi
    \right\rangle_{H^s(X)^*}
    =
    \frac12
    \left\langle
        (\operatorname D_\psi A_{\mu_t^\star})\dot\mu_t^\star,
        \dot\mu_t^\star
    \right\rangle_{H^s(X)^*}
    \qquad
    \text{for every }\psi\in H^s(X)^*_0.
\]
In \Cref{proof of geodesic second order regularity} we showed that
$t\mapsto A_{\mu_t^\star}\in W^{1,\infty}([0,1];\mathcal L(H^s(X)^*_0,H^s(X)^*_0))$ and
$t\mapsto\dot\mu_t^\star\in W^{1,\infty}([0,1];H^s(X)^*)$. Thus, we can differentiate the variational criterion to say
    \begin{align*}
        \left\langle
        (\operatorname{D}_{\dot \mu_t^\star}A_{\mu_t^\star})\dot\mu_t^\star, \psi\right\rangle_{H^s(X)^*} + \left\langle A_{\mu_t^\star}\ddot \mu_t^\star,\psi
        \right\rangle_{H^s(X)^*} =
        \frac12
        \left\langle
        (\operatorname{D}_{\psi}A_{\mu_t^\star})\dot\mu_t^\star,
        \dot\mu_t^\star
        \right\rangle_{H^s(X)^*}
    \end{align*}
    for all $\psi \in H^s(X)^*_0$ and a.e. $t \in [0,1].$ Using the definition of $A_\mu$ and differentiating it in the base point, the first term on the left-hand side satisfies
\begin{multline*}
    \left\langle
    (\operatorname{D}_{\dot\mu_t^\star}A_{\mu_t^\star})\dot\mu_t^\star,\psi
    \right\rangle_{H^s(X)^*}
    =
    \left\langle
    (\operatorname{D}_{\dot\mu_t^\star}S_{\mu_t^\star})\dot\beta_t^\star,
    \operatorname{D}_{\psi}B_{\mu_t^\star}
    \right\rangle_{\mathcal H_c}
    +
    \left\langle
    S_{\mu_t^\star}
    \operatorname{D}_{\dot\mu_t^\star}\operatorname{D}_{\dot\mu_t^\star}B_{\mu_t^\star},
    \operatorname{D}_{\psi}B_{\mu_t^\star}
    \right\rangle_{\mathcal H_c}\\
    +
    \left\langle
    S_{\mu_t^\star}\dot\beta_t^\star,
    \operatorname{D}_{\dot\mu_t^\star}\operatorname{D}_{\psi}B_{\mu_t^\star}
    \right\rangle_{\mathcal H_c}.
\end{multline*}
Moreover, $\langle A_{\mu_t^\star}\ddot\mu_t^\star,\psi\rangle_{H^s(X)^*}
=
\langle S_{\mu_t^\star}\operatorname{D}_{\ddot\mu_t^\star}B_{\mu_t^\star},
\operatorname{D}_{\psi}B_{\mu_t^\star}\rangle_{\mathcal H_c}$. Since $\operatorname{D}_{\dot\mu_t^\star}S_{\mu_t^\star}=\dot S_{\mu_t^\star}$ and
$\ddot\beta_t^\star
=
\operatorname{D}_{\ddot\mu_t^\star}B_{\mu_t^\star}
+
\operatorname{D}_{\dot\mu_t^\star}\operatorname{D}_{\dot\mu_t^\star}B_{\mu_t^\star}$, the left-hand side of the variational identity therefore becomes
\begin{equation*}
    \left\langle
    S_{\mu_t^\star}\ddot\beta_t^\star
    +
    \dot S_{\mu_t^\star}\dot\beta_t^\star,
    \operatorname{D}_{\psi}B_{\mu_t^\star}
    \right\rangle_{\mathcal H_c}
    +
    \left\langle
    S_{\mu_t^\star}\dot\beta_t^\star,
    \operatorname{D}_{\dot\mu_t^\star}\operatorname{D}_{\psi}B_{\mu_t^\star}
    \right\rangle_{\mathcal H_c}.
\end{equation*}
Similarly, expanding the right-hand side using the definition of $A_\mu$ gives
\begin{multline*}
    \frac12
    \left\langle
    (\operatorname{D}_{\psi}A_{\mu_t^\star})\dot\mu_t^\star,
    \dot\mu_t^\star
    \right\rangle_{H^s(X)^*}
    =
    \frac12
    \left\langle
    (\operatorname{D}_{\psi}S_{\mu_t^\star})\dot\beta_t^\star,
    \dot\beta_t^\star
    \right\rangle_{\mathcal H_c}\\
    +
    \frac12
    \left\langle
    S_{\mu_t^\star}
    \operatorname{D}_{\psi}\operatorname{D}_{\dot\mu_t^\star}B_{\mu_t^\star},
    \dot\beta_t^\star
    \right\rangle_{\mathcal H_c}
    +
    \frac12
    \left\langle
    S_{\mu_t^\star}\dot\beta_t^\star,
    \operatorname{D}_{\psi}\operatorname{D}_{\dot\mu_t^\star}B_{\mu_t^\star}
    \right\rangle_{\mathcal H_c}.
\end{multline*}
By the smoothness of $B$ established in the proof of \Cref{ambient inverse change of vars}, the mixed derivatives commute. Moreover, $S_{\mu_t^\star}$ is self-adjoint on $\mathcal H_c$, which follows from \Cref{operator metric equivalence}, \Cref{S tangent invariance}, and the fact that $S_\mu B_\mu=B_\mu$. Hence the latter two terms above coincide, and their sum cancels with the final term on the left-hand side. We therefore obtain
\begin{equation*}
    \left\langle
    S_{\mu_t^\star}\ddot\beta_t^\star
    +
    \dot S_{\mu_t^\star}\dot\beta_t^\star,
    \operatorname{D}_{\psi}B_{\mu_t^\star}
    \right\rangle_{\mathcal H_c}
    =
    \frac12
    \left\langle
    (\operatorname{D}_{\psi}S_{\mu_t^\star})\dot\beta_t^\star,
    \dot\beta_t^\star
    \right\rangle_{\mathcal H_c}.
\end{equation*}
Recalling the definition of $C_\mu$, the right-hand side can be written as $\frac12\langle C_{\mu_t^\star}(\dot\beta_t^\star,\dot\beta_t^\star)^\sharp,\operatorname{D}_{\psi}B_{\mu_t^\star}\rangle_{\mathcal H_c}$. Thus
\begin{equation*}
    \left\langle
    S_{\mu_t^\star}\ddot\beta_t^\star
    +
    \dot S_{\mu_t^\star}\dot\beta_t^\star
    -
    \frac12
    C_{\mu_t^\star}(\dot\beta_t^\star,\dot\beta_t^\star)^\sharp,
    \operatorname{D}_{\psi}B_{\mu_t^\star}
    \right\rangle_{\mathcal H_c}
    =0.
\end{equation*}
Since $\operatorname{D}_{(\cdot)}B_{\mu_t^\star}$ is a bijection from $T_{\mu_t^\star}\mathcal P(X)$ onto $\mathfrak T_{\mu_t^\star}\mathcal P(X)$ by \Cref{change of vars bijectivity}, this holds against every element of $\mathfrak T_{\mu_t^\star}\mathcal P(X)$. Equivalently,
\begin{equation*}
    \Pi_{\mu_t^\star}S_{\mu_t^\star}^{-1}
    \left(
    S_{\mu_t^\star}\ddot\beta_t^\star
    +
    \dot S_{\mu_t^\star}\dot\beta_t^\star
    -
    \frac12
    C_{\mu_t^\star}(\dot\beta_t^\star,\dot\beta_t^\star)^\sharp
    \right)
    =0,
\end{equation*}
where we used that $\Pi_{\mu_t^\star}$ is also the $\tilde{\mathbf g}_{\mu_t^\star}$-orthogonal projection onto $\mathfrak T_{\mu_t^\star}\mathcal P(X)$ by \Cref{Pimu g orthogonal projection}. Finally, since
$\operatorname{D}_{\dot\mu_t^\star}S_{\mu_t^\star}=\dot S_{\mu_t^\star}$ and
\[
    \Omega_{\dot\beta_t^\star}\dot\beta_t^\star
    =
    \frac12
    \Pi_{\mu_t^\star}S_{\mu_t^\star}^{-1}
    \left(
    \dot S_{\mu_t^\star}\dot\beta_t^\star
    -
    C_{\mu_t^\star}(\dot\beta_t^\star,\dot\beta_t^\star)^\sharp
    \right),
\]
the preceding identity is exactly
\begin{equation*}
    \Pi_{\mu_t^\star}
    \left(
    \ddot\beta_t^\star
    +
    \frac12S_{\mu_t^\star}^{-1}\dot S_{\mu_t^\star}\dot\beta_t^\star
    +
    \Omega_{\dot\beta_t^\star}\dot\beta_t^\star
    \right)
    =0.
\end{equation*}
Thus $(\mu_t^\star)_{t\in[0,1]}$ satisfies the geodesic equations in \eqref{geodesic equations}. \qed{}

\subsection{Operator Regularity} \label{sec: operator regularity}

\subsubsection{Main results}

\begin{lemma}[Uniform inverse bound for $S_{\mu}$] \label{St inverse bound}
    Under the hypotheses of \Cref{metric time deriv}, it holds that $\|S_{\mu}^{-1}\|_{\mathcal{L}(\mathcal{H}_c, \mathcal{H}_c)} \leq 1.$
\end{lemma}
\begin{proof}
    By \citet[Proposition 4.15]{lavenant2024riemannian} we know that for any $z \in \mathcal{H}_c$ we have
    \[\frac{\varepsilon}{2}\|z\|_{\mathcal{H}_c}^2 \leq \tilde{\mathbf{g}}_{\mu}(z, z) = \frac{\varepsilon}{2}\langle S_{\mu}z, z\rangle_{\mathcal{H}_c}.\]
    Canceling the $\varepsilon/2$ yields the bound $\|z\|_{\mathcal{H}_c}^2 \leq \langle S_{\mu} z, z\rangle_{\mathcal{H}_c}$, which implies $S_{\mu}$ is strongly coercive with coercivity constant $1$. By the Lax-Milgram theorem \citet[Section 6.2.1, Theorem 1]{evans2022partial}, for every $y \in \mathcal{H}_c$ there exists a unique $u \in \mathcal{H}_c$ such that 
    \[\langle S_{\mu}u, v\rangle_{\mathcal{H}_c} = \langle y, v\rangle_{\mathcal{H}_c}\]
    for every $v \in \mathcal{H}_c$. Thus, $S_{\mu}^{-1} y = u$. Taking $v = u$, applying uniform coercivity and Cauchy--Schwarz allows us to say
    \[\|u\|_{\mathcal{H}_c}^2 \leq \langle y, u\rangle_{\mathcal{H}_c} \leq \|y\|_{\mathcal{H}_c}\|u\|_{\mathcal{H}_c}\]
    and thus
    \[\|S_{\mu}^{-1}y\|_{\mathcal{H}_c} =\|u\|_{\mathcal{H}_c}\leq \|y\|_{\mathcal{H}_c}\]
    for every $y \in \mathcal{H}_c$.
\end{proof}

\begin{lemma} \label{hadamard diff of amu in Cm}
    The map $\mu \mapsto a_{\mu}$ from $\mathcal{P}(X) \subset \ell^\infty(B^m)$ to $C^m(X)$ is Hadamard  differentiable at any $\mu \in \mathcal{P}(X)$ tangentially to the set $\mathbb{T}_\mu^m.$ Moreover, the Hadamard derivative is given by 
    \[\operatorname{D}_{\gamma}a_{\mu} = \frac{a_{\mu}}{\varepsilon} \operatorname{D}_{\gamma}f_{\mu, \mu}.\]
    Writing $e_{\mu} = a_{\mu}^{-1}$ as the pointwise reciprocal of $a_{\mu}$, it moreover holds that the map $\mu \mapsto e_{\mu}$ from $\mathcal{P}(X) \subset \ell^\infty(B^m)$ to $C^m(X)$ is also Hadamard differentiable tangentially to $\mathbb{T}_{\mu}^m$ with Hadamard derivative
    \[\operatorname{D}_{\gamma} e_{\mu} = -\frac{e_{\mu}}{\varepsilon} \operatorname{D}_{\gamma}f_{\mu, \mu}.\]
\end{lemma}
\begin{proof}
    Define $\mathsf{a}:C^m(X)\to C^m(X)$ by $\mathsf{a}(f)=e^{f/\varepsilon}$; note that the codomain is correct because composition with the $C^\infty$ function $z \mapsto e^{z/\varepsilon}$ preserves $C^m(X)$ \citep[Theorem 1.3.7]{abraham1978foundations}. We first show that $\mathsf{a}$ is continuously Fr\'echet differentiable. Fix $f\in C^m(X)$ and choose $R>\|f\|_\infty$. Let $V:=(-R,R)$ and define $g:\operatorname{int}(X)\times V\to\mathbb{R}$ by $g(x,z)=e^{z/\varepsilon}$. By assumption, $X=\overline{\operatorname{int}(X)}$, and $\operatorname{int}(X)\subset\mathbb{R}^d$ is open and bounded. Moreover, $V$ is open, bounded, and convex, and $\mathbb{R}$ is a Banach space. Since $g$ is the restriction of the smooth function $(x,z)\mapsto e^{z/\varepsilon}$ on $\mathbb{R}^d\times\mathbb{R}$, we have $g\in C^m(\overline{\operatorname{int}(X)\times V})$. Furthermore,
    \[
    \operatorname{D}_{2,a}g(x,z)=\frac{1}{\varepsilon}e^{z/\varepsilon}a,
    \]
    so $\operatorname{D}_2g\in C^m(\overline{\operatorname{int}(X)\times V},\mathcal{L}(\mathbb{R},\mathbb{R}))$. Thus the hypotheses of the local $\Omega$-lemma \cite[Lemma~2.3]{Wittmann2019} are satisfied with $r=m$ and $l=1$. Consequently, $\mathsf{a}$ is continuously Fr\'echet differentiable at $f$, with
    \[
    \operatorname{D}_\gamma\mathsf{a}_f=\frac{1}{\varepsilon}e^{f/\varepsilon}\gamma=\frac{1}{\varepsilon}\mathsf{a}(f)\gamma.
    \]
    Since $f\in C^m(X)$ was arbitrary, $\mathsf{a}$ is continuously Fr\'echet differentiable on $C^m(X)$ and hence Hadamard differentiable with the same derivative.
    
    Now, recall that $a_\mu=\mathsf{a}(f_{\mu,\mu})$. By the Hadamard differentiability of the map $\mu\mapsto f_{\mu,\mu}$ at $\mu$ tangentially to $\mathbb{T}_\mu^m$, together with the Hadamard chain rule, the map $\mu\mapsto a_\mu$ is Hadamard differentiable at $\mu$ tangentially to $\mathbb{T}_\mu^m$. For every $\gamma\in\mathbb{T}_\mu^m$,
    \[
    \operatorname{D}_\gamma a_\mu
    =
    \operatorname{D}_{\operatorname{D}_\gamma f_{\mu,\mu}}\mathsf{a}_{f_{\mu,\mu}}
    =
    \frac{1}{\varepsilon}e^{f_{\mu,\mu}/\varepsilon}\operatorname{D}_\gamma f_{\mu,\mu}
    =
    \frac{1}{\varepsilon}a_\mu\operatorname{D}_\gamma f_{\mu,\mu}.
    \]
    Hence $\operatorname{D}_{\gamma}a_\mu
    =
    \frac{1}{\varepsilon}a_\mu\operatorname{D}_{\gamma}f_{\mu,\mu},$ as claimed. For the second claim, choose $L>\|f_{\mu,\mu}\|_\infty$, which is finite since $X$ is compact and $f_{\mu,\mu}\in C^m(X)$ \citep[Proposition 3.7]{lavenant2024riemannian}. Set $V:=(-L,L)$, which contains the range of $f_{\mu,\mu}$, and define $g:\operatorname{int}(X)\times V\to\mathbb R$ by $g(x,z)=e^{-z/\varepsilon}$. Since $g$ is the restriction of an infinitely differentiable function on $\mathbb{R}^d \times \mathbb{R}$, it holds that $g \in C^m (\overline{\operatorname{int}(X) \times V})$ for any $m \geq 0.$ Moreover,
    \[\operatorname{D}_{2,a}g(x,z) = -\frac{1}{\varepsilon}e^{-z/\varepsilon}a\]
    and thus $\operatorname{D}_2g\in C^m(\overline{\operatorname{int}(X)\times V},\mathcal{L}(\mathbb{R},\mathbb{R}))$. Thus the hypotheses of the local $\Omega$-lemma \cite[Lemma 2.3]{Wittmann2019} are again satisfied with $r=m$ and $l=1$. Applying the same logic as above yields the second part of the claim. The observation that the derivatives are linear in $\gamma$ yields full Hadamard differentiability and completes the proof.
\end{proof}

\begin{lemma} \label{hadamard diff of emu}
    Write $e_{\mu} = a_{\mu}^{-1}$ as the pointwise reciprocal of $a_{\mu}$. Then the map $\mu \mapsto e_{\mu}$ from $\mathcal{P}(X) \subset \ell^\infty(B^m)$ to $ \mathcal{H}_c$ is Hadamard differentiable at any $\mu \in \mathcal{P}(X)$ tangentially to the set $\mathbb{T}_\mu^m.$ Moreover, the Hadamard derivative is given by 
    \[\operatorname{D}_{\gamma}e_{\mu} =  H_c\left(\frac{a_\mu \mu}{\varepsilon}\operatorname{D}_{\gamma}f_{\mu, \mu} + a_\mu \gamma\right).\]
\end{lemma}
\begin{proof}
    By \Cref{a_t reciprocal} we know that $e_\mu = H_c(a_\mu \mu)$, where $H_c(\nu) = \int_Xk_c(x, \cdot) \, d\nu(x)$. To prove the Hadamard differentiability of $\mu \mapsto e_\mu$ regarded as a map into $\mathcal{H}_c$ we will employ a direct analytical argument. 
    Fix $\gamma\in\mathbb T_\mu^m$, let $t\downarrow0$
    and let $\gamma_t\in\ell^\infty(B^m)$ be arbitrary
    admissible perturbations satisfying
    $\mu+t\gamma_t\in\mathcal P(X)$ and
    $\|\gamma_t-\gamma\|_{C^m(X)^*}\to0$.
     Then we have 
    \begin{align} \label{Hc decomp}
        \frac{e_{\mu + t\gamma_t} - e_\mu}{t} &= H_c\left(\frac{a_{\mu + t\gamma_t}(\mu + t\gamma_t) - a_\mu \mu}{t}\right) = H_c\left(\mu\frac{ (a_{\mu + t\gamma_t} - a_\mu)}{t}\right) +  H_c \left(a_{\mu + t\gamma_t}\gamma_t\right).
    \end{align}
    Now we will analyze the above. Regarding the first term, by \Cref{hadamard diff of amu in Cm} and the definition of Hadamard differentiability, we know that 
    \[\frac{a_{\mu + t\gamma_t} - a_\mu}{t} \to \frac{a_\mu}{\varepsilon}\operatorname{D}_{\gamma}f_{\mu, \mu} \quad \text{in $C^m(X)$ as $t \downarrow 0$.}\]
    Moreover, we have 
    \[\left\|\mu \left(\frac{a_{\mu + t\gamma_t} - a_\mu}{t} - \frac{a_{\mu}}{\varepsilon}\operatorname{D}_{\gamma}f_{\mu, \mu}\right)\right\|_{C^m(X)^*} \lesssim_{m, d} \|\mu\|_{C^m(X)^*}\left\|\frac{a_{\mu + t\gamma_t} - a_\mu}{t} - \frac{a_{\mu}}{\varepsilon}\operatorname{D}_{\gamma}f_{\mu, \mu}\right\|_{C^m(X)}\]
    due to \Cref{dual product estimate}. Note that the object in the parentheses is necessarily $C^m(X)$ valued as $a_\mu^{-1} \in C^m(X)$ due to \citet[Proof of Proposition 3.7]{lavenant2024riemannian} and inversion, multiplication and summation preserve $C^m(X)$ \citep[Section 6.5]{makarov2021smooth} (for inversion the function must be nonvanishing, which is trivially true for $a_\mu$ for all $\mu$). Furthermore, by \citet[Lemma B.1]{lavenant2024riemannian}, we know that
    \[\left\|\mu \left(\frac{a_{\mu + t\gamma_t} - a_\mu}{t} - \frac{a_{\mu}}{\varepsilon}\operatorname{D}_{\gamma}f_{\mu, \mu}\right)\right\|_{\mathcal{H}_c^*} \lesssim_{c, m} \left\|\mu \left(\frac{a_{\mu + t\gamma_t} - a_\mu}{t} - \frac{a_{\mu}}{\varepsilon}\operatorname{D}_{\gamma}f_{\mu, \mu}\right)\right\|_{C^m(X)^*}.\]
    Combining the previous three panels therefore implies 
    \begin{align*}
        \mu \frac{a_{\mu + t\gamma_t} - a_{\mu}}{t} \rightarrow \frac{a_\mu \mu}{\varepsilon}\operatorname{D}_{\gamma}f_{\mu, \mu} \quad \text{in $\mathcal{H}_c^*$ as $t \downarrow 0$.}
    \end{align*}
    Since $H_c$ is the Riesz map for $\mathcal{H}_c$, it holds that for any $\nu  \in \mathcal{H}_c^*$ we have $\|H_c(\nu)\|_{\mathcal{H}_c} = \|\nu\|_{\mathcal{H}_c^*}$ and thus,
    \begin{align*}
        \left\|H_c\left(\mu\frac{(a_{\mu + t\gamma_t} - a_\mu)}{t}\right) - H_c\left(\frac{a_\mu \mu}{\varepsilon}\operatorname{D}_\gamma f_{\mu, \mu}\right)\right\|_{\mathcal{H}_c}  = \left\|\mu \frac{a_{\mu + t\gamma_t} - a_{\mu}}{t} - \frac{a_\mu \mu}{\varepsilon}\operatorname{D}_{\gamma}f_{\mu, \mu}\right\|_{\mathcal{H}_c^*}
    \end{align*}
    which we have already established goes to $0$ as $t \downarrow0.$ For the second term in \Cref{Hc decomp}, the Hadamard differentiability of $\mu \mapsto a_{\mu}$ as a map from $\mathcal P(X)\subset\ell^\infty(B^m) \to C^m(X)$ directly guarantees that $a_{\mu + t\gamma_t} \to a_\mu$. To see this, simply observe that $\sup_t \|t^{-1}(a_{\mu + t\gamma_t} - a_\mu)\|_{C^m(X)} < \infty$ since it has a finite $C^m(X)$ limit -- this directly implies that $t\|t^{-1} (a_{\mu + t\gamma_t} - a_\mu)\|_{C^m(X)}  = \|a_{\mu + t\gamma_t} - a_\mu\|_{C^m(X)}\to 0$ as $t \downarrow0$. This,  the stipulation that $\gamma_t \to \gamma$ in $\mathbb{T}_\mu^m$ and \Cref{dual product estimate} implies
    \begin{align*}
    \|a_{\mu + t\gamma_t} \gamma_t -a_{\mu}\gamma\|_{C^m(X)^*} &\leq \|a_{\mu + t\gamma_t}(\gamma_t - \gamma)\|_{C^m(X)^*} + \|(a_{\mu + t\gamma_t} - a_{\mu})\gamma\|_{C^m(X)^*} \\
    &\lesssim \|a_{\mu + t\gamma_t}\|_{C^m(X)}\|\gamma_t - \gamma\|_{C^m(X)^*} + \|\gamma\|_{C^m(X)^*}\|a_{\mu + t\gamma_t} - a_{\mu}\|_{C^m(X)} \\&\to 0.
    \end{align*}
    Thus, $a_{\mu + t\gamma_t} \gamma_t  \to a_{\mu}\gamma$ in $C^m(X)^*$. By the same arguments employed above, this further implies convergence in $\mathcal{H}_c^*$ and convergence of the second term in \Cref{Hc decomp} to $H_c(a_\mu \gamma)$ in $\mathcal{H}_c$. The observation that the obtained derivative is linear completes the proof. 
\end{proof}

\begin{lemma} \label{Pi hadamard differentiability}
    The map $\mu \mapsto \Pi_{\mu}$ from $\mathcal P(X)\subset\ell^\infty(B^m)$ to $\mathcal{L}(\mathcal{H}_c, \mathcal{H}_c)$ is Hadamard differentiable at any $\mu \in \mathcal{P}(X)$ tangentially to the set $\mathbb{T}_\mu^m$ with a Hadamard  derivative given by
    \[(\operatorname{D}_{\gamma}\Pi_{\mu})u =  - \langle u, \operatorname{D}_{\gamma}e_{\mu} \rangle_{\mathcal{H}_c}e_{\mu} - \langle u, e_\mu\rangle_{\mathcal{H}_c}\operatorname{D}_{\gamma}e_\mu.\]
\end{lemma}

\begin{proof}
    Define the map $\mathsf{P}: \mathcal{H}_c \to \mathcal{L}(\mathcal{H}_c, \mathcal{H}_c)$ defined by $\mathsf{P}(e) u = u - \langle u, e\rangle_{\mathcal{H}_c}e$. It holds that 
    \begin{align*}
        \left[\frac{\mathsf{P}(e + th_t) - \mathsf{P}(e)}{t}\right]u = -\langle u, h_t\rangle_{\mathcal{H}_c}e - \langle u, e\rangle_{\mathcal{H}_c}h_t  - t\langle u, h_t\rangle_{\mathcal{H}_c}h_t
    \end{align*}
    for any $h_t \to h$ in $\mathcal{H}_c$ as $t \downarrow 0$. Taking the supremum over $\|u\|_{\mathcal H_c}\leq1$
in the preceding expression gives
\[
\left\|
\frac{\mathsf P(e+th_t)-\mathsf P(e)}{t}
-\operatorname D_h\mathsf P(e)
\right\|_{\operatorname{op}}
\leq
2\|e\|_{\mathcal H_c}\|h_t-h\|_{\mathcal H_c}
+t\|h_t\|_{\mathcal H_c}^2
\to 0.
\]
Thus, $\mathsf P$ is Hadamard differentiable in
operator norm, with
\[
(\operatorname D_h\mathsf P(e))u
=
-\langle u,h\rangle_{\mathcal H_c}e
-\langle u,e\rangle_{\mathcal H_c}h.
\]
Applying the Hadamard chain rule to this fact and \Cref{hadamard diff of emu} completes the proof, since $(\mu \mapsto \Pi_\mu) = \mathsf{P} \circ(\mu \mapsto e_\mu)$.
\end{proof}

\begin{lemma}\label{tilde J hadamard diff}
    The map $\mu \mapsto \mathbf{J}_{\mu}$ is Hadamard differentiable as a map from $\mathcal{P}(X) \subset \ell^\infty(B^m) \to \mathcal{L}(\mathcal{H}_c, \mathcal{H}_c)$ tangentially to the set $\mathbb{T}^m_{\mu}$ with a Hadamard derivative given by
    \[(\operatorname{D}_{\gamma}\mathbf{J}_{\mu})u = H_c((a_{\mu}^2\Pi_{\mu}u)\gamma) + H_c\left(\left[2a_{\mu}\operatorname{D}_{\gamma}a_{\mu} \Pi_{\mu}u + a_{\mu}^2(\operatorname{D}_{\gamma}\Pi_{\mu})u\right]\mu\right).\]
\end{lemma}
\begin{proof}
    Recall the characterization $\mathbf{J}_{\mu} u = H_c(a_{\mu}^2\Pi_\mu u\, \mu)$ from \Cref{theta inverse}, and define the operator $\mathsf{J}: C^m(X) \times \mathcal{L}(\mathcal{H}_c , \mathcal{H}_c) \times C^m(X)^* \to \mathcal{L}(\mathcal{H}_c, \mathcal{H}_c)$ characterized by $\mathsf{J}(a, \mathsf{P}, \sigma)u = H_c(a(\mathsf{P}u)\sigma)$ for all $u \in \mathcal{H}_c.$ It is clear that $\mathsf{J}$ is tri-linear, and its operator norm is given by
    \begin{align*}
        \|\mathsf{J}\|_{\operatorname{op}} &= \sup\left\{\|\mathsf{J}(a, \mathsf{P}, \sigma)\|_{\mathcal{L}(\mathcal{H}_c, \mathcal{H}_c)} \, :\, \|a\|_{C^m(X)} \leq 1, \|\mathsf{P}\|_{\mathcal{L}(\mathcal{H}_c, \mathcal{H}_c)} \leq 1, \|\sigma\|_{\ell^\infty(B^m)} \leq 1 \right\} .
    \end{align*}
    For $a\in C^m(X)$,
    $\mathsf P\in\mathcal L(\mathcal H_c,\mathcal H_c)$,
    and $\sigma\in C^m(X)^*$, the reproducing
    property and the definition of dual multiplication
    give
    \begin{align*}
    \|\mathsf J(a,\mathsf P,\sigma)\|_{\operatorname{op}}
    &=
    \sup_{\|u\|_{\mathcal H_c},\|v\|_{\mathcal H_c}\leq1}
    \left|
    \left\langle
    H_c(a(\mathsf Pu)\sigma),v
    \right\rangle_{\mathcal H_c}
    \right|\\
    &=
    \sup_{\|u\|_{\mathcal H_c},\|v\|_{\mathcal H_c}\leq1}
    |\sigma(a(\mathsf Pu)v)|\\
    &\lesssim
    \|a\|_{C^m(X)}
    \|\mathsf P\|_{\operatorname{op}}
    \|\sigma\|_{C^m(X)^*},
    \end{align*}
    where the last inequality follows from the
    Banach algebra property of $C^m(X)$ and
    $\mathcal H_c\hookrightarrow C^m(X)$ (\Cref{continuous embedding}).
    Thus, $\mathsf J$ is bounded trilinear in
    operator norm.
     Thus, it is also Hadamard differentiable tangentially to the normed space $C^m(X) \times \mathcal{L}(\mathcal{H}_c, \mathcal{H}_c) \times C^m(X)^*$ equipped with the norm $\|\cdot\|_{C^m(X)} \lor \|\cdot\|_{\mathcal{L}(\mathcal{H}_c, \mathcal{H}_c)} \lor \|\cdot\|_{C^m(X)^*}$ by \Cref{multilinear hadamard diff}. In particular, its Hadamard derivative is given by
    \begin{equation*}
        \operatorname{D}_{(b,\mathsf{G}, \gamma)}\mathsf{J}(a,\mathsf{P}, \sigma) = \mathsf{J}(b, \mathsf{P}, \sigma) + \mathsf{J}(a, \mathsf{G}, \sigma) + \mathsf{J}(a, \mathsf{P}, \gamma).
    \end{equation*}
    Now, note that $\mathbf{J}_\mu = \mathsf{J}(a_{\mu}^2, \Pi_\mu, \mu)$ and the map $\mu \mapsto (a_{\mu}^2, \Pi_\mu, \mu)$ is Hadamard differentiable with Hadamard derivative
    \[\gamma \mapsto (2a_\mu \operatorname{D}_\gamma a_\mu, \operatorname{D}_\gamma \Pi_\mu, \gamma).\]
    To see this, note that $a \mapsto a^2$ can be regarded as the map $(a,b) \mapsto ab$ from $C^m(X) \times C^m(X) \to C^m(X)$, which is clearly bilinear and bounded, and thus Hadamard differentiable (\Cref{multilinear hadamard diff}). The Hadamard differentiability of $\mu \mapsto a_\mu$
    due to \Cref{hadamard diff of amu in Cm} and the Hadamard differentiability of $\mu \mapsto \Pi_\mu$ \Cref{Pi hadamard differentiability} complete the justification of $\mu \mapsto (a_\mu^2, \Pi_\mu, \mu)$. Finally, we can apply the Hadamard chain rule to conclude that
    \begin{align*}
        (\operatorname{D}_{\gamma}\mathbf{J}_{\mu})u &= \left(\mathsf{J}(2a_\mu \operatorname{D}_\gamma a_\mu,  \Pi_\mu, \mu) + \mathsf{J}(a_\mu^2, \operatorname{D}_\gamma\Pi_\mu, \mu) + \mathsf{J}(a_\mu^2, \Pi_\mu, \gamma)\right)u
    \end{align*}
    for any $u \in \mathcal{H}_c$. Substituting in the definition of $\mathsf{J}$ and simplifying completes the derivation of the expression in the statement. The observation that the derivative is indeed a linear function of $\gamma$ completes the proof of full Hadamard differentiability.
\end{proof}

\begin{corollary} \label{tilde K hadamard diff}
    The map $\mu \mapsto \mathbf{K}_\mu$ is Hadamard  differentiable as a map from $\mathcal P(X)\subset\ell^\infty(B^m) \to \mathcal{L}(\mathcal{H}_c, \mathcal{H}_c)$ tangentially to the set $\mathbb{T}_\mu^m$ with a Hadamard  derivative given by
    \[\operatorname{D}_{\gamma}\mathbf{K}_{\mu} = (\operatorname{D}_{\gamma}\Pi_{\mu})\mathbf{J}_\mu + \Pi_\mu (\operatorname{D}_\gamma \mathbf{J}_\mu).\]
\end{corollary}
\begin{proof}
    By \Cref{tilde K expression}, we can write $\mathbf{K}_\mu = \Pi_\mu \mathbf{J}_\mu = \mathsf{K}(\Pi_\mu, \mathbf{J}_\mu)$ where $\mathsf{K}: \mathcal{L}(\mathcal{H}_c, \mathcal{H}_c) \times \mathcal{L}(\mathcal{H}_c, \mathcal{H}_c) \to \mathcal{L}(\mathcal{H}_c, \mathcal{H}_c)$ is given by $\mathsf{K}(A,B) = AB.$ Since operator multiplication is bounded and bilinear, it is Hadamard differentiable (\Cref{multilinear hadamard diff}) with Hadamard derivative \[\operatorname{D}_{(\dot A, \dot B)}\mathsf{K}(A,B) = \dot AB + A\dot B.\] Now note that \Cref{tilde J hadamard diff} and \Cref{Pi hadamard differentiability} guarantee that the map $\mu \mapsto (\Pi_\mu, \mathbf{J}_\mu)$ is Hadamard differentiable with Hadamard derivative $\gamma \mapsto (\operatorname{D}_{\gamma}\Pi_\mu, \operatorname{D}_\gamma \mathbf{J}_\mu)$. Finally, since $\mathbf{K}_\mu = \mathsf{K} \circ (\mu \mapsto (\Pi_\mu, \mathbf{J}_\mu))$ we can apply the Hadamard product rule to conclude that $\mathbf{K}_\mu$ is indeed Hadamard differentiable with a Hadamard derivative 
    \[\operatorname{D}_\gamma \mathbf{K}_\mu = (\operatorname{D}_\gamma \Pi_\mu) \mathbf{J}_\mu + \Pi_\mu (\operatorname{D}_\gamma \mathbf{J}_\mu)\]
    as desired.
\end{proof}

\begin{lemma} \label{tilde R hadamard diff}
    The map $\mu \mapsto \mathbf{R}_\mu$ is Hadamard differentiable as a map from $\mathcal P(X)\subset\ell^\infty(B^m) \to \mathcal{L}(\mathcal{H}_c, \mathcal{H}_c)$ tangentially to the set $\mathbb{T}_\mu^m$ with a Hadamard derivative given by
    \[\operatorname{D}_{\gamma}\mathbf{R}_{\mu} = \overline{R}_\mu (\operatorname{D}_\gamma \mathbf{K}_\mu )\overline{R}_\mu \Pi_\mu + \overline{R}_\mu (\operatorname{D}_\gamma \Pi_\mu).\]
\end{lemma}

\begin{proof}
    By \Cref{Rbar expression}, we can write $\mathbf{R}_\mu = \overline{R}_\mu \Pi_\mu$ where $\overline{R}_\mu = (\operatorname{id}_{\mathcal{H}_c} - \mathbf{K}_\mu)^{-1}$. Since $\operatorname{id}_{\mathcal{H}_c} - \mathbf{K}_\mu$ is invertible (\Cref{Rbar expression}) and since $\mathbf{K}_\mu \in \mathcal{L}(\mathcal{H}_c, \mathcal{H}_c)$, by the inverse mapping theorem \citep[Theorem 3.1.6.]{vershynin2010functional} we know that $\overline{R}_\mu \in \mathcal{L}(\mathcal{H}_c, \mathcal{H}_c).$ By \citet[Example 2.3]{driver2020functional}, since $\mathcal{H}_c$ is a Banach space then $\mathcal{L}(\mathcal{H}_c, \mathcal{H}_c)$ is a Banach algebra in operator norm with identity, and since the operator norm is submultiplicative, the algebra is also unital. Thus, by \citet[Theorem 1.1.23.]{garcia2018non}, the map $\operatorname{inv}: \mathrm{GL}(\mathcal H_c) \to \mathcal{L}(\mathcal{H}_c, \mathcal{H}_c)$ given by $\operatorname{inv}(A) = A^{-1}$ is Fr\'echet differentiable at any invertible $A$ with Fr\'echet derivative
    \[\operatorname{D}_{\dot A}\operatorname{inv}(A) = -A^{-1}\dot A A^{-1}.\]
    Since Fr\'echet is a stronger notion of differentiation than Hadamard, this too is the Hadamard derivative of $A \mapsto A^{-1}.$ Because $\mu \mapsto \operatorname{id}_{\mathcal{H}_c} - \mathbf{K}_\mu$ is Hadamard differentiable tangentially to $\mathbb{T}_\mu^m$ with derivative $\gamma \mapsto -\operatorname{D}_\gamma \mathbf{K}_\mu$, and since $\overline{R}_\mu = \operatorname{inv} \circ (\mu \mapsto (\operatorname{id}_{\mathcal{H}_c} - \mathbf{K}_\mu))$, we can apply the Hadamard chain rule to conclude that $\mu \mapsto \overline{R}_\mu$ is Hadamard differentiable tangentially to $\mathbb{T}_\mu^m$ with derivative 
    \[\operatorname{D}_{\gamma}\overline{R}_\mu = \overline{R}_\mu (\operatorname{D}_\gamma \mathbf{K}_\mu)\overline{R}_\mu.\]
    Finally, because $\mathbf{R}_\mu = \overline{R}_\mu \Pi_\mu$, we can apply the argument involving the multiplication operator $\mathsf{K}(A,B) = AB$ employed in \Cref{tilde K hadamard diff} to obtain the expression for the Hadamard derivative. Since the derivative is linear in $\gamma$, the map $\mu \mapsto \mathbf{R}_\mu$ is indeed fully Hadamard differentiable.
\end{proof}

\subsubsection{Supplemental results}
\begin{lemma} \label{a_t reciprocal}
    The pointwise reciprocal of the map $a_{\mu} \in \mathcal{H}_c$ is simply the kernel mean embedding $a_{\mu}^{-1} = H_c(a_{\mu}\mu).$ Furthermore, it satisfies $\|a_{\mu}^{-1}\|_{\mathcal{H}_c} = 1.$
\end{lemma}
\begin{proof}
    Recall that the optimal coupling of $\mu$ with itself, denoted $\pi$, satisfies
    \[\frac{d\pi}{d(\mu \otimes \mu)}(x,y) = \exp\left(\frac{f_{\mu,\mu}(x) + f_{\mu,\mu}(y) - c(x,y)}{\varepsilon}\right) = k_c(x,y)a_{\mu}(x) a_{\mu}(y).\]
    Since the coupling has marginals $\mu$, for each fixed $y$ we know that
    \[\int_X k_c(x,y) a_{\mu}(x) a_{\mu}(y)\, d\mu(x) = 1.\]
    Factoring out $a_{\mu}(y)$ and bringing it to the other side finishes the proof that $a_{\mu}^{-1}$ is the kernel mean embedding. For the proof of the second fact, note that the integral operator representation of $a_{\mu}^{-1}$ as $H_c(a_{\mu}\mu)$ straightforwardly implies
    \begin{align*}
        \langle a_{\mu}^{-1}, a_{\mu}^{-1}\rangle_{\mathcal{H}_c} &= \int_{X \times X}k_c(x,y)a_{\mu}(x)a_{\mu}(y) \,d\mu(x) d\mu(y) \\
        &= \int_{X \times X} \exp\left(\frac{f_{\mu,\mu}(x) + f_{\mu,\mu}(y) - c(x,y)}{\varepsilon}\right) \, d(\mu \otimes \mu)(x,y) \\
        &= \int_{X \times X} d\pi(x,y) = 1.
    \end{align*}
\end{proof}

\begin{lemma}[Existence of $Q_{\mu}$ on $\mathcal{Q}_{\mu}$] \label{Qt existence}
    The operators $\operatorname{id}_{\mathcal{Q}_{\mu}} + K_{\mu}, \operatorname{id}_{\mathcal{Q}_{\mu}} - K_{\mu}$ are invertible on $\mathcal{H}_{\mu}/\mathbb{R}$, and hence $Q_{\mu} = (\operatorname{id} + K_{\mu})^{-1}$ is well defined on $\mathcal{H}_{\mu}/\mathbb{R}$. Moreover, $\|Q_{\mu}\|_{\mathcal{L}(\mathcal{H}_{\mu}/\mathbb{R})} \leq 1$.
\end{lemma}
\begin{proof}
    The spectral bounds of \citet[Proposition 4.2]{lavenant2024riemannian} indicate that $\sigma(K_{\mu}|_{\mathcal{H}_{\mu}/\mathbb{R}}) \subset [0, q_{\mu}]$ for some $q_{\mu} < 1$. Thus, 
    \[\sigma\left(\left(\operatorname{id} + K_{\mu}\right)\Big|_{\mathcal{H}_{\mu} /\mathbb{R}}\right) \subset [1, 1+q_{\mu}]\]
    and hence $Q_{\mu}^{-1} = \operatorname{id} + K_{\mu}$ and the operator $\operatorname{id}_{\mathcal{Q}_{\mu}} - K_{\mu}$ are invertible on $\mathcal{H}_{\mu}/\mathbb{R}$. Furthermore, the spectral bound above guarantees \[\|Q_{\mu}\|_{\mathcal{L}(\mathcal{H}_{\mu}/\mathbb{R})} \leq 1.\]
\end{proof}

\begin{lemma} \label{Rbar expression}
    The operator $\overline{R}_{\mu} := \mathbf{R}_{\mu} + \operatorname{id}_{\mathcal{H}_c} - \Pi_{\mu}$ can be expressed as \[\overline{R}_{\mu} = (\operatorname{id}_{\mathcal{H}_c} - \mathbf{K}_{\mu})^{-1} \quad \text{where} \quad \mathbf{K}_{\mu} := \Theta_{\mu} K_{\mu}\Theta_{\mu}^{-1}\Pi_{\mu} :\mathcal{H}_c \to \mathcal{H}_c\]
    and satisfies $\mathbf{R}_{\mu} = \overline{R}_{\mu} \Pi_{\mu}.$
\end{lemma}
\begin{proof}
    Since $\overline R_{\mu} = \mathbf{R}_{\mu} + \operatorname{id}_{\mathcal{H}_c} - \Pi_{\mu}$,
    \begin{align*}
        \overline{R}_{\mu} \Pi_{\mu} = (\mathbf{R}_{\mu} + \operatorname{id}_{\mathcal{H}_c} - \Pi_{\mu})\Pi_{\mu} = \mathbf{R}_{\mu}
    \end{align*}
    which completes the proof of the second property. For the first, recall the definition $R_{\mu} = (\operatorname{id}_{\mathcal{Q}_{\mu}} - K_{\mu})^{-1}$, and note that $\Theta_{\mu}K_{\mu}\Theta_{\mu}^{-1}$ sends functions in $\mathfrak{T}_\mu \mathcal{P}(X) \subset \mathcal{H}_c$ to $\mathcal{H}_c$. Through standard manipulations, we have
    \begin{align} \label{t eqn 1}
        (\operatorname{id}_{\mathfrak{T}_\mu \mathcal{P}(X)} - \Theta_{\mu}K_{\mu}\Theta_{\mu}^{-1})\Theta_{\mu} R_{\mu}\Theta_{\mu}^{-1} =  \Theta_{\mu}(\operatorname{id}_{\mathcal{Q}_{\mu}} - K_{\mu})\Theta_{\mu}^{-1} \Theta_{\mu} R_{\mu} \Theta_{\mu}^{-1} = \Theta_{\mu}R_{\mu}^{-1}R_{\mu}\Theta_{\mu}^{-1} = \operatorname{id}_{\mathfrak{T}_\mu \mathcal{P}(X)}.
    \end{align}
    Moreover,
    \begin{align} \label{t eqn 2}
        \Theta_{\mu}R_{\mu} \Theta_{\mu}^{-1}(\operatorname{id}_{E^\perp_{\mu}} - \Theta_{\mu} K_{\mu}\Theta_{\mu}^{-1}) = \Theta_{\mu}R_{\mu}(\operatorname{id}_{\mathcal{Q}_{\mu}} -K_{\mu})\Theta_{\mu}^{-1} = \operatorname{id}_{\mathfrak{T}_\mu \mathcal{P}(X)}.
    \end{align}
    Thus, if we write $T_{\mu} := \Theta_{\mu}R_{\mu}\Theta_{\mu}^{-1}$, then \Cref{t eqn 1} and \Cref{t eqn 2} imply 
    \begin{equation}\label{t inverse} 
    T_{\mu} = (\operatorname{id}_{E^\perp_{\mu}} - \Theta_{\mu}K_{\mu}\Theta_{\mu}^{-1})^{-1}.
    \end{equation}
    We will return to this identity shortly. For now, through standard manipulations
    \begin{align*}(\operatorname{id}_{\mathcal{H}_c} - \mathbf{K}_{\mu})\overline{R}_{\mu} = (\operatorname{id}_{\mathcal{H}_c} - \mathbf{K}_{\mu})(\mathbf{R}_{\mu} +\operatorname{id}_{\mathcal{H}_c} - \Pi_{\mu}) = \mathbf{R}_{\mu} + \operatorname{id}_{\mathcal{H}_c} - \Pi_{\mu} - \mathbf{K}_{\mu}\mathbf{R}_{\mu}  - \mathbf{K}_{\mu}(\operatorname{id}_{\mathcal{H}_c} -   \Pi_{\mu}).
    \end{align*}
    But $\mathbf{K}_{\mu} \mathbf{R}_{\mu} = \Theta_{\mu}K_{\mu}\Theta_{\mu}^{-1}\Pi_{\mu} \mathbf{R}_{\mu} =  \Theta_{\mu}K_{\mu}\Theta_{\mu}^{-1} \mathbf{R}_{\mu}$ since $\operatorname{Im}(\mathbf{R}_{\mu}) \subset \mathfrak{T}_\mu \mathcal{P}(X)$, and \[\mathbf{K}_{\mu}(\operatorname{id}_{\mathcal{H}_c} -   \Pi_{\mu}) = \Theta_{\mu}K_{\mu}\Theta_{\mu}^{-1}\Pi_{\mu} (\operatorname{id}_{\mathcal{H}_c} -   \Pi_{\mu}) = 0.\]
    Using these identities to further manipulate the right-hand side above yields
    \begin{align*}
        (\operatorname{id}_{\mathcal{H}_c} - \mathbf{K}_{\mu})\overline{R}_{\mu} &= \operatorname{id}_{\mathcal{H}_c} - \Pi_{\mu}  + \mathbf{R}_{\mu}- \Theta_{\mu}K_{\mu}\Theta_{\mu}^{-1} \mathbf{R}_{\mu} \\
        &= \operatorname{id}_{\mathcal{H}_c} - \Pi_{\mu}  + (\operatorname{id}_{\mathfrak{T}_\mu \mathcal{P}(X)}- \Theta_{\mu}K_{\mu}\Theta_{\mu}^{-1} )\Theta_{\mu}R_{\mu}\Theta_{\mu}^{-1}\Pi_{\mu} \\
        &= \operatorname{id}_{\mathcal{H}_c} - \Pi_{\mu} + \Pi_{\mu} = \operatorname{id}_{\mathcal{H}_c}. \tag{\Cref{t inverse}}
    \end{align*}
    Thus, $\overline{R}_{\mu}$ is the right inverse of $\operatorname{id}_{\mathcal{H}_c} - \mathbf{K}_{\mu}.$ Now we will prove that it is also the left inverse. Write $F_{\mu} := \Theta_{\mu}K_{\mu}\Theta_{\mu}^{-1}$, and observe that we have
    \begin{align*}
        \overline{R}_{\mu}(\operatorname{id}_{\mathcal{H}_c} - \mathbf{K}_{\mu}) &= (\mathbf{R}_{\mu} + \operatorname{id}_{\mathcal{H}_c} - \Pi_{\mu})(\operatorname{id}_{\mathcal{H}_c} - \mathbf{K}_{\mu}) \\
        &= (T_{\mu}\Pi_{\mu} + (\operatorname{id}_{\mathcal{H}_c} - \Pi_{\mu}))(\operatorname{id}_{\mathcal{H}_c} - F_{\mu} \Pi_{\mu})\\
        &= T_{\mu}\Pi_{\mu} -T_{\mu}\Pi_{\mu}F_{\mu}\Pi_{\mu} + \operatorname{id}_{\mathcal{H}_c} - \Pi_{\mu} - (\operatorname{id}_{\mathcal{H}_c} - \Pi_{\mu})F_{\mu}\Pi_{\mu}.
    \end{align*}
    Since $\operatorname{Im}(\Theta_{\mu}) \subset \mathfrak{T}_\mu \mathcal{P}(X)$ we have $(\operatorname{id}_{\mathcal{H}_c}-\Pi_{\mu})F_{\mu} = 0$, which kills the last term. Moreover, this implies that $\Pi_{\mu}F_{\mu} = F_{\mu}$. Continuing yields
    \begin{align*}
        \overline{R}_{\mu}(\operatorname{id}_{\mathcal{H}_c} - \mathbf{K}_{\mu}) &= T_{\mu}\Pi_{\mu} -T_{\mu}F_{\mu}\Pi_{\mu} + \operatorname{id}_{\mathcal{H}_c} - \Pi_{\mu} \\
        &= T_{\mu}(\operatorname{id}_{\mathfrak{T}_\mu \mathcal{P}(X)} - F_{\mu})\Pi_{\mu} +\operatorname{id}_{\mathcal{H}_c} - \Pi_{\mu} \\
        &= \Pi_{\mu} +\operatorname{id}_{\mathcal{H}_c} - \Pi_{\mu} \\
        &= \operatorname{id}_{\mathcal{H}_c}
    \end{align*}
    which completes the proof that it is also a left inverse. Thus, $\overline{R}_{\mu} = (\operatorname{id}_{\mathcal{H}_c} - \mathbf{K}_{\mu})^{-1}$, completing the proof.
\end{proof}

\begin{lemma} \label{tilde K expression}
    The operator $\mathbf{K}_\mu: \mathcal{H}_c \to \mathcal{H}_c$ can be expressed as $\mathbf{K}_\mu = \Pi_\mu \mathbf{J}_\mu$.
\end{lemma}
\begin{proof}
    By \Cref{Rbar expression}, we know that $\mathbf{R}_{\mu} =\overline{R}_{\mu} \Pi_{\mu}$ and $\overline{R}_{\mu} = (\operatorname{id} - \mathbf{K}_{\mu})^{-1}$ where $\mathbf{K}_{\mu} = \Theta_{\mu}K_{\mu}\Theta_{\mu}^{-1}\Pi_{\mu}.$ We will manipulate $\mathbf{K}_{\mu}$ like so,
    \begin{align*}
        \mathbf{K}_{\mu} = \Theta_{\mu}K_{\mu}\Theta_{\mu}^{-1}\Pi_{\mu} = \Theta_{\mu}L_{\mu}J_{\mu}\Theta_{\mu}^{-1}\Pi_{\mu}
    \end{align*}
    where $L_{\mu}: \mathcal{H}_c \to \mathcal{Q}_{\mu}$ is given by $L_{\mu}z = [a_{\mu}z]$. Now note that $\Theta_{\mu}L_{\mu}z = \Pi_{\mu}\iota_{\mu}^{-1}(a_{\mu}z) = \Pi_{\mu}z$, and thus $\mathbf{K}_{\mu} = \Pi_{\mu}J_{\mu}\Theta_{\mu}^{-1}\Pi_{\mu} = \Pi_{\mu}\mathbf{J}_{\mu}.$
\end{proof}

\begin{lemma}
    Define $L_{\mu}: \mathcal{H}_c \to \mathcal{Q}_{\mu}$ given by $L_{\mu}z = [a_{\mu}z]$. Then $K_{\mu} = L_{\mu}J_{\mu}$ as operators on $\mathcal{Q}_{\mu}$. 
\end{lemma}
\begin{proof}
    Fix $[\ell]\in\mathcal Q_\mu$, and choose its representative
    $\ell\in\mathcal H_\mu$ satisfying $\mu(\ell)=0$.
    Since $\langle a_\mu^{-1}\ell,B_\mu\rangle_{\mathcal H_c}=\mu(\ell)=0$,
    we have $\Pi_\mu(a_\mu^{-1}\ell)=a_\mu^{-1}\ell$.
    Thus $L_{\mu}J_{\mu}[\ell] = [a_{\mu}J_{\mu}[\ell]]$. Note that
    \begin{multline*}
        (a_{\mu}J_{\mu}[\ell])(y) = a_{\mu}(y) \int_X k_c(x,y)a_{\mu}(x)\ell(x)\, d\mu(x)
        =  \int_X a_{\mu}(y)k_c(x,y)a_{\mu}(x)\ell(x)\, d\mu(x) \\
        = \int_X k_{\mu}(x,y) \ell(x) \, d\mu(x) 
        = K_{\mu}\ell(y).
    \end{multline*}
    Thus, as quotient elements $L_{\mu}J_{\mu}[\ell] = [K_{\mu}\ell]$. To complete the proof, we will quickly show that $K_{\mu}$ descends as an operator on $\mathcal{Q}_{\mu}$. Let $[z] \in \mathcal{Q}_{\mu}$ and suppose $v, w \in [z]$. In particular, this means there exists a $\lambda \in \mathbb{R}$ such that $v + \lambda \equiv w$. By linearity, we also know that 
    \[K_{\mu}w = K_{\mu}(v + \lambda) = K_{\mu}v +\lambda K_{\mu}\mathbbm{1}_X.\]
    Analyzing the second term,
    \begin{align*}
        (K_{\mu}\mathbbm{1}_X) (y) = \int_X k_{\mu}(x,y) \, d\mu(x) = a_{\mu}(y) \int_X k_c(x,y)a_{\mu}(x)\, d\mu(x) = a_{\mu}(y)(H_c(a_{\mu} \mu))(y).
    \end{align*}
    By \Cref{a_t reciprocal} we know that $a_{\mu}^{-1} = H_c(a_{\mu}\mu)$, and thus $K_{\mu}\mathbbm{1}_X \equiv 1$. Moreover, by the panel above, $K_{\mu}w = K_{\mu}v + \lambda$ and thus $K_{\mu}w \sim K_{\mu}v$. Furthermore, the assignment $K_{\mu}[z] := [K_{\mu}z]$ is independent of the chosen representative and defines an operator $K_{\mu}: \mathcal{Q}_{\mu} \to \mathcal{Q}_{\mu}.$
\end{proof}

\begin{lemma}\label{dual product estimate}
    Let $f\in C^m(X)$ and $\sigma\in C^m(X)^*$.
    Define $f\sigma\in C^m(X)^*$ by
    $(f\sigma)(\phi)=\sigma(f\phi)$ for
    $\phi\in C^m(X)$. Then
    \[
        \|f\sigma\|_{C^m(X)^*}
        \lesssim_{m,d}
        \|f\|_{C^m(X)}\|\sigma\|_{C^m(X)^*}.
    \]
\end{lemma}
\begin{proof}
    By the definition of the dual norm and the
    Banach algebra property of $C^m(X)$
    \citep[Equation 1.11]{moore2017global},
    \begin{align*}
        \|f\sigma\|_{C^m(X)^*}
        &=\sup_{\|\phi\|_{C^m(X)}\leq1}
        |\sigma(f\phi)|\\
        &\leq\|\sigma\|_{C^m(X)^*}
        \sup_{\|\phi\|_{C^m(X)}\leq1}
        \|f\phi\|_{C^m(X)}\\
        &\lesssim_{m,d}
        \|f\|_{C^m(X)}\|\sigma\|_{C^m(X)^*}.
    \end{align*}
\end{proof}

\section{Proofs for \Cref{sec: linearizations}} \label{regularity theory for linearizations}

\subsection{Proof of \Cref{chart geodesic ODE}} \label{proof of chart geodesic ODE}

We will derive the vector field induced by the geodesic equations and the chart $\Psi$ in the variables $(p, u)$. In order to do this, we need to obtain an expression for $\ddot \beta_t$ in terms of the two state space variables. Observe that the geodesic equation states
\begin{equation*}
    \Pi_{\mu_t}\left(\ddot \beta_t
    +
    \Gamma_{\beta_t}(\dot \beta_t, \dot \beta_t)\right) = 0
\end{equation*}
combined with the unit-sphere constraint $\|\beta_t\|_{\mathcal{H}_c}^2 = 1$. Differentiating the unit sphere condition twice implies $\langle \beta_t, \ddot \beta_t\rangle_{\mathcal{H}_c} = -\|\dot \beta_t\|_{\mathcal{H}_c}^2.$ Decomposing $\ddot{\beta}_t$ into its $\mathfrak{T}_{\mu_t} \mathcal{P}(X)$ and $\mathfrak{T}_{\mu_t} \mathcal{P}(X)$--orthogonal parts gives
\[\ddot{\beta_t} = \langle \ddot{\beta}_t, \beta_t\rangle_{\mathcal{H}_c}\beta_t + (\operatorname{id} - \beta_t \otimes \beta_t)\ddot{\beta}_t = -\|\dot \beta_t\|_{\mathcal{H}_c}^2\beta_t + \Pi_{\mu_t}\ddot{\beta}_t.\]
The second term -- the tangential acceleration -- is determined by the geodesic equations. In particular, plugging in the expression above into \eqref{geodesic equations} gives
\begin{align*}
    0 = \Pi_{\mu_t}\left(-\|\dot \beta_t\|_{\mathcal{H}_c}^2 \beta_t + \Pi_{\mu_t}\ddot{\beta}_t + \Gamma_{\beta_t}(\dot \beta_t, \dot \beta_t)\right)
\end{align*}
which implies 
\begin{align*}
    \Pi_{\mu_t}\ddot{\beta_t} = \|\dot \beta_t\|^2_{\mathcal{H}_c} \Pi_{\mu_t}(\beta_t) - \Pi_{\mu_t}\left(\Gamma_{\beta_t}(\dot \beta_t, \dot \beta_t)\right).  
\end{align*}
Since $\beta_t \perp \mathfrak{T}_{\mu_t}\mathcal{P}(X)$, the first term on the right is zero. Thus, the tangential acceleration satisfies
\begin{align*}
    \Pi_{\mu_t}\ddot{\beta}_t = - \Pi_{\mu_t}\Gamma_{\beta_t}(\dot \beta_t, \dot \beta_t)
\end{align*}
and the aggregate acceleration satisfies $\ddot \beta_t = -\|\dot{\beta}_t\|_{\mathcal{H}_c}^2 \beta_t - \Pi_{\mu_t}\Gamma_{\beta_t}(\dot \beta_t, \dot \beta_t).$ Thus, we get the ambient differential equation
\begin{align*}
    \frac{d}{dt}\begin{pmatrix}
        \beta_t \\
        \dot{\beta}_t
    \end{pmatrix} &= \begin{pmatrix}
        \dot \beta_t \\
        -\|\dot{\beta}_t\|_{\mathcal{H}_c}^2 \beta_t- \Pi_{\mu_t}\Gamma_{\beta_t}(\dot \beta_t, \dot \beta_t).
    \end{pmatrix}
\end{align*}
Now we will convert this into state-space variables $(p_t, u_t) = (\Pi_\mu \beta_t, \Pi_\mu \dot \beta_t).$ In particular, we have $\dot p_t = \Pi_\mu \dot \beta_t$ and $\dot u_t = \Pi_\mu \ddot \beta_t$. Note that $\dot u_t$ can be equivalently written as 
\begin{align*}
    \dot u_t =  -\|\operatorname{D}_{u_t}\psi(p_t)\|_{\mathcal{H}_c}^2p_t  - \Pi_\mu \Pi_{\psi(p_t)}\Gamma_{\psi(p_t)}(\operatorname{D}_{u_t}\psi(p_t), \operatorname{D}_{u_t}\psi(p_t)).
\end{align*}
Now if we define 
\begin{equation} \label{first order term chart geodesic equations}
    \mathscr{A}(p, u) = -\|\operatorname{D}_{u}\psi(p)\|_{\mathcal{H}_c}^2p  - \Pi_\mu\Pi_{\psi(p)}\Gamma_{\psi(p)}(\operatorname{D}_{u}\psi(p), \operatorname{D}_{u}\psi(p))
\end{equation}
then the geodesic equations are clearly characterized by the state-space ODE
\begin{align*}
    \frac{d}{dt}\begin{pmatrix}
        p_t \\
        u_t
    \end{pmatrix} = \begin{pmatrix}
        u_t  \\ \mathscr{A}(p_t, u_t).
    \end{pmatrix}
\end{align*}
as desired. \qed{}

\subsection{Proof of the Stitching Lemma} \label{proof of stitching lemma}

Since $\beta_t\in U_+$ for every $t\in[0,1]$ and $U_+$ is open, for every $t$ there exists a sphere chart centered at $\beta_t$ whose image is contained in $U_+$. By \Cref{smoothness of geodesic coefficients}, the geodesic coefficients are $C^\infty$ on these chart images. Consequently, \Cref{smoothness of chart acceleration,local smoothness theorem} provide smooth local flows in the corresponding tangent-bundle coordinates. Since the population trajectory is continuous and $[0,1]$ is compact, the tagged-partition argument of \Cref{tagged partition} gives a finite collection of charts covering the trajectory. We may therefore compose the associated local flows using their smooth transition maps.

Now suppose the tagged partition gives $[s_{i-1},s_i]\subset I_{t_i}$. Write $\Psi_i:=\Psi_{t_i}$ for the corresponding tangent-bundle chart and let $\alpha_i:[0,s_i-s_{i-1}]\times U_i\to(\beta_{t_i}^\perp)^2$ denote its local flow, where $U_i$ is an open neighborhood of $x_i:=\Psi_i(\beta_{s_{i-1}},\dot\beta_{s_{i-1}})$. Since the partition is finite, we may successively restrict to an open neighborhood $\mathcal U$ of $(\beta_0,\dot\beta_0)$ in $T\mathbb S(\mathcal H_c)$ so that all local flows and chart transitions remain well-defined and their position components remain in the prescribed chart images contained in $U_+$. Then the geodesic flow satisfies
\begin{align*}
    ((\beta_t,\dot\beta_t))_{t\in[0,1]}
    =\bigoplus_{i=1}^N
    \left(\Psi_i^{-1}
    (\alpha_i(t-s_{i-1},x_i))\right)_{t\in[s_{i-1},s_i]}.
\end{align*}
Since each local flow $\alpha_i$, each chart and its inverse $\Psi_i,\Psi_i^{-1}$, and each transition map $\Psi_{i+1}\circ\Psi_i^{-1}$ is $C^1$, and their derivatives are continuous uniformly over the corresponding compact time intervals, their finite composition implies that the trajectory map $G:\mathcal U\to C^0([0,1];\mathcal H_c\times\mathcal H_c)$ given by $(\beta_0,\dot\beta_0)\mapsto((\beta_t,\dot\beta_t))_{t\in[0,1]}$ is $C^1$, and hence Hadamard differentiable. Moreover, since the second component is the time derivative of the first, the first-component map is $C^1$ into $C^1([0,1];\mathcal H_c)$. Its Hadamard derivative is given by
\[
\operatorname D_{(h_0,\dot h_0)}G(\beta_0,\dot\beta_0)
=((h_t,\dot h_t))_{t\in[0,1]},
\]
where, writing $(\mathfrak p_t^i,\mathfrak u_t^i)$ for the coordinate variation on $[s_{i-1},s_i]$, we have $(\mathfrak p_0^1,\mathfrak u_0^1)=\operatorname D_{(h_0,\dot h_0)}\Psi_1(\beta_0,\dot\beta_0)$ and
\[
(\mathfrak p_t^i,\mathfrak u_t^i)
=\operatorname D_{2,(\mathfrak p_{s_{i-1}}^i,\mathfrak u_{s_{i-1}}^i)}
\alpha_i(t-s_{i-1},x_i)
\]
for $t\in[s_{i-1},s_i]$. At each chart transition, we have
\[
(\mathfrak p_{s_i}^{i+1},\mathfrak u_{s_i}^{i+1})
=\operatorname D_{(\mathfrak p_{s_i}^i,\mathfrak u_{s_i}^i)}
(\Psi_{i+1}\circ\Psi_i^{-1})
\big(\alpha_i(s_i-s_{i-1},x_i)\big),
\]
and the corresponding variation in the $\beta$ variable is recovered by
\[
(h_t,\dot h_t)
=\operatorname D_{(\mathfrak p_t^i,\mathfrak u_t^i)}
\Psi_i^{-1}
\big(\alpha_i(t-s_{i-1},x_i)\big)
\quad\text{for }t\in[s_{i-1},s_i].
\]
Finally, since $\exp_\mu(v)=(\beta_t)_{t\in[0,1]}$ is obtained by composing $(\mu,v)\mapsto(B_\mu,v)$, the global trajectory map $G$, and projection onto the first component, the Hadamard chain rule implies $\operatorname D_{(\gamma,a)}\exp_\mu(v)=(h_t)_{t\in[0,1]}$, where $h_0=\operatorname D_\gamma B_\mu$ and $\dot h_0=a$. \qed{}

\subsection{Extension and Smoothness of ODE Coefficients} \label{sec: extensions and smoothness}

In this section we establish the regularity of the coefficients appearing in the ODEs of \Cref{sec: linearizations}. We first construct the relevant operators directly in $\beta$-coordinates on an open subset of $\mathcal H_c$ containing $B[\mathcal P(X)]$ denoted $U_+$. The inverse of the map $M(\beta)=\beta H_c^{-1}(\beta)$, which we call $B$, then transfers the extensions to distribution (i.e. $H^s(X)^*$) indexed operators. Note that we intentionally denote the local inverse of $M$ by $\lambda\mapsto B_\lambda$, with $B_{M(\beta)}=\beta$; we choose this notation because it agrees with the original change of variables at probability measures in its domain. All constructions use the domain and kernel conditions in \Cref{standing sinkhorn assumptions}.

The basic calculus for $a_\beta$, $M$, and $\Pi_\beta$ is established in \Cref{ambient basic smoothness}, while the detailed operator-norm smoothness proofs for $\mathbf J_\beta$, $\mathbf K_\beta$, $\mathbf R_\beta$, and $C_\beta(u,v)^\sharp$ are collected in \Cref{smoothness of J beta,smoothness of K beta,smoothness of C beta} in the Supplemental Results. The resulting ambient operator properties are summarized in \Cref{ambient coefficient domain,ambient resolvent regularity}. Smoothness of the extended inverse change of variables (\Cref{ambient inverse change of vars}) then transfers these constructions to $H^s(X)^*$; this is proved in \Cref{agreement of base derivatives} and used in \Cref{sinkhorn distance hadamard diff}. Finally, smoothness of the geodesic and parallel transport ODE coefficients is established in \Cref{smoothness of geodesic coefficients}.

\subsubsection{The Ambient Domain $U_+$}

Define $V_+=\{\beta\in\mathcal H_c:\min_{x\in X}\beta(x)>0\}$. For $\beta\in V_+$, set $a_\beta=1/\beta$, $M(\beta)=\beta H_c^{-1}(\beta)$, and $\Pi_\beta=\operatorname{id}_{\mathcal H_c}-\beta\otimes\beta/\|\beta\|_{\mathcal H_c}^2$ where $(\beta \otimes \beta)(u) = \langle \beta, u\rangle_{\mathcal{H}_c}\beta$. The reciprocal belongs to $\mathcal H_c$ by \Cref{sobolev reciprocal and logarithm} and \Cref{RKHS norm equivalence}. We extend the operators in \Cref{reparameterization,tilde K expression} by defining $\mathbf J_\beta u=H_c\bigl(a_\beta^2(\Pi_\beta u)M(\beta)\bigr)$ and $\mathbf K_\beta=\Pi_\beta\mathbf J_\beta$. Finally, define
\begin{equation}\label{uplus definition}
U_+=\left\{\beta\in V_+:\|\mathbf J_\beta\|_{\operatorname{op}}<1\right\}.
\end{equation}
It should be clear from the definitions that $\mathbf{J}_\mu = \mathbf{J}_{B(\mu)}$ and $\mathbf{K}_\mu = \mathbf{K}_{B(\mu)}$. This construction extends the indexing variable from $B[\mathcal P(X)]$ to $U_+$ (where the operators continue to act on $\mathcal H_c$, of course).

\begin{lemma}[Elementary ambient smoothness]
\label{ambient basic smoothness}
The set $V_+$ is open in $\mathcal H_c$, and the maps $a\in C^\infty(V_+;\mathcal H_c)$, $M\in C^\infty(\mathcal H_c;H^s(X)^*)$, and $\Pi\in C^\infty(V_+;\mathcal L(\mathcal H_c,\mathcal H_c))$ are smooth in their respective norm topologies.
\end{lemma}

\begin{proof}
We start by showing that $V_+$ is indeed open. Fix a $\beta \in V_+$; we claim that there exists an $r > 0$ such that for all $h \in \mathcal{H}_c$ such that $\|h\|_{\mathcal{H}_c} < r$, we have $\beta + h \in V_+.$ To see why this holds, note that by \Cref{continuous embedding}, $\mathcal H_c\hookrightarrow C^m(X)\hookrightarrow C(X)$ continuously, so there exists $C>0$ such that $\|f\|_{C(X)}\leq C\|f\|_{\mathcal H_c}$ for all $f \in \mathcal{H}_c$. Moreover, by continuity of $\beta$ we know that $\min_{x \in X}\beta(x) = \delta > 0$. Thus, set $r = \delta/2C$ and note that whenever $\|h\|_{\mathcal H_c}<r$, we have $\min_X(\beta+h)\geq\delta-\|h\|_{C(X)}>\delta/2$ and thus $\beta + h \in V_+$. This completes the proof of openness.

To establish smoothness of the reciprocal map, consider $F:V_+\times\mathcal H_c\to\mathcal H_c$ given by $F(\beta,g)=\beta g-\mathbbm{1}_X$. By \Cref{rkhs algebra calculus}, $F$ is $C^\infty$, and $F(\beta,1/\beta)=0$. Its Fr\'echet derivative in the second argument is 
\[h \mapsto \operatorname{D}_{2,h}F(\beta, g) = \beta h\]
and thus, at $\beta, 1/\beta$ we have  
$\operatorname D_{2,h}F(\beta,1/\beta)=\beta h$, which is a bounded linear isomorphism with bounded inverse $h\mapsto h/\beta$. 
The Banach-space Implicit Function Theorem \citep[Theorem 2.5.7]{abraham2012manifolds} therefore gives open neighborhoods $U_\beta\subset V_+$ of $\beta$ and $W_0\subset\mathcal H_c$ of $0$, together with an implicit solution $g\in C^\infty(U_\beta\times W_0;\mathcal H_c)$ satisfying $F(\alpha,g(\alpha,z))=z$ for every $(\alpha,z)\in U_\beta\times W_0$. Taking $z=0$ gives $g(\alpha,0)=1/\alpha$ for every $\alpha\in U_\beta$, so the reciprocal map is $C^\infty$ on $U_\beta$. Since $\beta\in V_+$ was arbitrary, the reciprocal map belongs to $C^\infty(V_+;\mathcal H_c)$.
 This establishes smoothness of the reciprocal map. 
Now, by \Cref{rkhs algebra calculus}, the map
\[
\begin{aligned}
M:\mathcal H_c &\rightarrow H^s(X)^*\\
\beta &\mapsto \beta H_c^{-1}(\beta)
\end{aligned}
\]
is a continuous quadratic map induced by bounded bilinear multiplication, and hence belongs to $C^\infty(\mathcal H_c;H^s(X)^*)$ \citep[Lemma 5.5]{kriegl1997convenient}. By the same reasoning, the map
\[
\begin{aligned}
\Pi:V_+ &\rightarrow \mathcal L(\mathcal H_c,\mathcal H_c)\\
\beta &\mapsto \operatorname{id}_{\mathcal H_c}
-\frac{\langle\cdot,\beta\rangle_{\mathcal H_c}}{\|\beta\|_{\mathcal H_c}^2}\beta
\end{aligned}
\]
belongs to $C^\infty(V_+;\mathcal L(\mathcal H_c,\mathcal H_c))$, since its numerator depends quadratically on $\beta$ and $\beta\mapsto\|\beta\|_{\mathcal H_c}^{-2}$ is $C^\infty$ on $V_+$.
\end{proof}

\begin{lemma}[Ambient domain]
\label{ambient coefficient domain}
The sets $V_+$ and $U_+$ are open in $\mathcal H_c$, and $B[\mathcal P(X)]\subset U_+$. More precisely, writing $k_{\min}=\min_{x,y\in X}k_c(x,y)>0$ and $k_{\max}=\max_{x\in X}k_c(x,x)$, we have $\min_X B_\mu\geq k_{\min}/\sqrt{k_{\max}}$ and $\|\mathbf J_{B_\mu}\|_{\operatorname{op}}<1$ for every $\mu\in\mathcal P(X)$.
Moreover, the maps
\[
\begin{aligned}
a: V_+ &\rightarrow \mathcal H_c\\
\beta &\mapsto a_\beta
\end{aligned}
\qquad
\begin{aligned}
\Pi: V_+ &\rightarrow \mathcal L(\mathcal H_c,\mathcal H_c)\\
\beta &\mapsto \Pi_\beta
\end{aligned}
\qquad
\begin{aligned}
\mathbf J: V_+ &\rightarrow \mathcal L(\mathcal H_c,\mathcal H_c)\\
\beta &\mapsto \mathbf J_\beta
\end{aligned}
\qquad
\begin{aligned}
\mathbf K: V_+ &\rightarrow \mathcal L(\mathcal H_c,\mathcal H_c)\\
\beta &\mapsto \mathbf K_\beta
\end{aligned}
\]
are $C^\infty$ in their respective norm topologies.
Finally, for every $\beta\in V_+$, $\mathbf J_\beta$ is self-adjoint and satisfies $\Pi_\beta\mathbf J_\beta=\mathbf J_\beta\Pi_\beta=\mathbf J_\beta=\mathbf K_\beta$.
\end{lemma}

\begin{proof}
By \Cref{ambient basic smoothness,smoothness of J beta,smoothness of K beta}, $V_+$ is open, the maps $a$, $\Pi$, $\mathbf J$, and $\mathbf K$ are $C^\infty$ in their respective norm topologies, and $\mathbf K_\beta=\mathbf J_\beta$ for every $\beta\in V_+$. In particular, $\beta\mapsto\|\mathbf J_\beta\|_{\operatorname{op}}$ is continuous on $V_+$, so $U_+$ is open. Moreover, $\Pi_\beta\mathbf J_\beta=\mathbf J_\beta$ by \Cref{smoothness of K beta}, while $\mathbf J_\beta\Pi_\beta=\mathbf J_\beta$ follows from the definition of $\mathbf J_\beta$. Thus, for every $u,v\in\mathcal H_c$, the Riesz identity and dual multiplication yield
\[
\langle\mathbf J_\beta u,v\rangle_{\mathcal H_c}
=M(\beta)\bigl(a_\beta^2(\Pi_\beta u)(\Pi_\beta v)\bigr),
\]
which is symmetric in $u,v$. Hence $\mathbf J_\beta=\mathbf K_\beta$ is self-adjoint.

Finally, we aim to show that for all $\mu\in \mathcal{P}(X)$ we have $B_\mu \in U_+$. Fix $\mu\in\mathcal P(X)$, set $\beta=B_\mu$, and observe that by \Cref{a_t reciprocal}, $\|B_\mu\|_{\mathcal H_c}=1$ and $B_\mu=H_c((1/B_\mu)\mu)$, so $M(B_\mu)=\mu$ and $a_{(B_\mu)}=a_\mu$. The reproducing property gives $\beta(x)\leq\sqrt{k_{\max}}$, and hence $\beta(x)=\int_X k_c(x,y)/\beta(y)\,d\mu(y)\geq k_{\min}/\sqrt{k_{\max}}$. Thus $\beta\in V_+$, and \Cref{theta inverse} identifies $\mathbf J_\beta$ with $\mathbf J_\mu$. The preceding quadratic-form identity now shows that $\mathbf J_\mu$ is a positive semidefinite operator, because
\[\langle\mathbf J_\mu u,u\rangle_{\mathcal H_c}=\mu\left(a_\mu^2(\Pi_\mu u)^2\right) = \int_X a_\mu^2(\Pi_\mu u)^2\,d\mu\geq 0\]
for every $u\in\mathcal H_c.$ For $u\in\mathcal H_c$, set $f=a_\mu\Pi_\mu u$. Since $\iota_\mu$ is an isometry and $\iota_\mu B_\mu=\mathbbm{1}_X$, we have \[\langle f,\mathbbm{1}_X\rangle_{\mathcal H_\mu} = \langle \Pi_\mu u, B_\mu\rangle_{\mathcal H_c}=0\] and thus $\|[f]\|_{\mathcal Q_\mu}=\|\Pi_\mu u\|_{\mathcal H_c}$. The reproducing property and the spectral bound $q_\mu<1$ used in the proof of \Cref{Qt existence} therefore yield
\begin{align*}
\langle\mathbf J_\mu u,u\rangle_{\mathcal H_c}
=\int_X f^2\,d\mu
=\langle K_\mu[f],[f]\rangle_{\mathcal Q_\mu} \leq q_\mu\|[f]\|_{\mathcal Q_\mu}^2
=q_\mu\|\Pi_\mu u\|_{\mathcal H_c}^2
\leq q_\mu\|u\|_{\mathcal H_c}^2.
\end{align*}
Since $\mathbf J_\mu$ is positive semidefinite and self-adjoint, $\|\mathbf J_\mu\|_{\operatorname{op}}\leq q_\mu<1$. Thus $B_\mu\in U_+$, completing the proof.
\end{proof}

\subsubsection{Inverting the change of variables}


\begin{lemma}[Inverse change of variables on $U_+$]
\label{ambient inverse change of vars}
\label{inverse change of vars}
For every $\beta\in U_+$, the map \[M:\mathcal{H}_c \to H^s(X)^* \quad \text{given by}\quad  M(\beta) = \beta H_c^{-1}(\beta)\] can be restricted to a $C^\infty$ diffeomorphism between an open neighborhood of $\beta$ contained in $U_+$ and an open neighborhood of $M(\beta)$ in $H^s(X)^*$. We intentionally denote the local inverse of $M$ by $\lambda\mapsto B_\lambda$, with $B_{M(\beta)}=\beta$; we choose this notation because it agrees with the original change of variables at probability measures in its domain. For every $h\in\mathcal H_c$, we have
\[
\bigl(\operatorname D_{(\cdot)}B_{M(\beta)}\bigr)^{-1}(h)
=\operatorname D_hM(\beta)
=hH_c^{-1}(\beta)+\beta H_c^{-1}(h).
\]
In particular, \[\left(\beta\mapsto\operatorname DM(\beta)\right) \in C^\infty(\mathcal{H}_c; \mathcal{L}(\mathcal{H}_c, H^s(X)^*)).\]
Finally, if $\beta\in\mathbb S(\mathcal H_c)$, then $M$ can be restricted to relatively open neighborhoods of $\beta$ in $\mathbb S(\mathcal H_c)$ and of $M(\beta)$ in $\{\lambda\in H^s(X)^*: \lambda(\mathbbm{1}_X)=1\}$, and $\operatorname D_hM(\beta)\in H^s(X)^*_0$ for every $h\in\beta^\perp$.
\end{lemma}

\begin{proof}
By \Cref{ambient basic smoothness}, $M\in C^\infty(\mathcal H_c;H^s(X)^*)$, and differentiating gives the displayed formula for $\operatorname D_hM(\beta)$. Decomposing $h=\Pi_\beta h+(\operatorname{id}_{\mathcal H_c}-\Pi_\beta)h$ and using the definition of $\mathbf J_\beta$ gives
\begin{align*}
H_c\bigl[a_\beta\operatorname D_hM(\beta)\bigr]
&=h+H_c\bigl[a_\beta hH_c^{-1}(\beta)\bigr] \\ &=h + H_c[a_\beta(\Pi_\beta h)H_c^{-1}(\beta)] + H_c[a_\beta(\operatorname{id}_{\mathcal{H}_c} - \Pi_\beta)(h)H_c^{-1}(\beta)] \\
&= h + \mathbf{J}_\beta h + H_c[a_\beta(\operatorname{id}_{\mathcal{H}_c} - \Pi_\beta)(h)H_c^{-1}(\beta)].
\end{align*}
Since $(\operatorname{id}_{\mathcal{H}_c} - \Pi_\beta)h = \frac{\langle h, \beta\rangle_{\mathcal{H}_c}}{\|\beta\|_{\mathcal{H}_c}^2}\beta$, the term on the right above becomes 
\begin{align*}
    H_c[a_\beta(\operatorname{id}_{\mathcal{H}_c} - \Pi_\beta)(h)H_c^{-1}(\beta)] = H_c\left[a_\beta \frac{\langle h, \beta\rangle_{\mathcal{H}_c}}{\|\beta\|_{\mathcal{H}_c}^2}\beta H_c^{-1}(\beta)\right] = \frac{\langle h, \beta\rangle_{\mathcal{H}_c}}{\|\beta\|_{\mathcal{H}_c}^2} H_c[H_c^{-1}(\beta)] = (\operatorname{id}_{\mathcal{H}_c} - \Pi_\beta)h
\end{align*}
and therefore $H_c[a_\beta\operatorname D_hM(\beta)] = (2\operatorname{id}_{\mathcal{H}_c} - \Pi_\beta + \mathbf{J}_\beta)h.$ Solving for $\operatorname{D}M(\beta)$ yields
\[\operatorname{D}_hM(\beta) = \beta H_c^{-1}\left((2\operatorname{id}_{\mathcal{H}_c} - \Pi_\beta + \mathbf{J}_\beta)(h)\right).\]
By the definition of $\mathbf{J}_\beta$, the operator $2\operatorname{id}_{\mathcal{H}_c} - \Pi_\beta + \mathbf{J}_\beta$ acts as multiplication by $2$ on $\operatorname{span}\{\beta\}$ and as $\operatorname{id}_{\beta^\perp}+\mathbf J_\beta|_{\beta^\perp}$ on $\beta^\perp$. Both restrictions are boundedly invertible because $\|\mathbf J_\beta\|_{\operatorname{op}}<1$. Thus $(2\operatorname{id}_{\mathcal{H}_c} - \Pi_\beta + \mathbf{J}_\beta)$ is boundedly invertible. Furthermore, by \Cref{sobolev reciprocal and logarithm,RKHS norm equivalence}, both $\beta$ and $a_\beta=\beta^{-1}$ belong to $H^s(X)$. Thus, by \Cref{rkhs algebra calculus}, multiplication by $a_\beta$ and multiplication by $\beta$ are bounded linear operators on $H^s(X)^*$. Since $a_\beta\beta=\mathbbm{1}_X$, these operators are mutual inverses, so multiplication by $a_\beta$ is a bounded linear isomorphism.
Finally, $H_c:H^s(X)^*\to\mathcal H_c$ is a bounded linear isomorphism since it is the Riesz map for the RKHS $\mathcal{H}_c$.  Thus, $h \mapsto \operatorname{D}_hM(\beta)$ is a bounded linear isomorphism. Applying the Banach-space Implicit Function Theorem \citep[Theorem 2.5.7]{abraham2012manifolds} to $M(\alpha)-\lambda=0$ gives open neighborhoods $W \subset H^s(X)^*$ of $M(\beta)$ and $L \subset U_+$ and a smooth local inverse $B: W \to L$. By the local uniqueness assertion of the Implicit Function Theorem, $M(\alpha)=\lambda$ for $(\lambda,\alpha)\in W\times L$ if and only if $\alpha=B_\lambda$. Setting $L'=L\cap M^{-1}(W)$, we obtain an open neighborhood of $\beta$ in $\mathcal H_c$, since $M$ is continuous. Moreover, $B(M(\alpha))=\alpha$ for every $\alpha\in L'$ and $M(B_\lambda)=\lambda$ for every $\lambda\in W$. Thus $M:L'\to W$ is a $C^\infty$ diffeomorphism with inverse $B$. Differentiating $B(M(\alpha))=\alpha$ gives
$\operatorname DB_{M(\beta)}\circ\operatorname DM(\beta)
=\operatorname{id}_{\mathcal H_c}$.
Since $\operatorname DM(\beta)$ is a bounded linear isomorphism, composing on the right with $(\operatorname DM(\beta))^{-1}$ yields
$\operatorname DB_{M(\beta)}=(\operatorname DM(\beta))^{-1}$, and hence
$\operatorname DM(\beta)\circ\operatorname DB_{M(\beta)}
=\operatorname{id}_{H^s(X)^*}$.
This is precisely the inverse-derivative identity associated with the Banach-space Inverse Function Theorem \citep[Theorem 2.5.7]{abraham2012manifolds}.

For agreement at probability measures, fix $\mu\in\mathcal P(X)$. Since $M(B_\mu)=\mu$, the preceding construction gives an open neighborhood $W$ of $\mu$ in $H^s(X)^*$ on which the smooth local inverse of $M$ is defined. Now let $\lambda\in W\cap\mathcal P(X)$ and define $\alpha$ as the inverse image of $\lambda$. Writing $B_\lambda$ temporarily for the \textit{original} change of variables, we have $H_c^{-1}(\alpha)= \alpha^{-1}\lambda$ by definition, and thus $H_c^{-1}(B_\lambda)=B_\lambda^{-1}\lambda$ by \Cref{a_t reciprocal}. Therefore, by Riesz representation and using the fact that $\alpha^{-1} - B_\lambda^{-1} = (B_\lambda - \alpha)/(\alpha B_\lambda)$, we have
\begin{align*}
\|\alpha-B_\lambda\|_{\mathcal H_c}^2
=\int_X(\alpha-B_\lambda)(\alpha^{-1}-B_\lambda^{-1})\,d\lambda =-\int_X\frac{(\alpha-B_\lambda)^2}{\alpha B_\lambda}\,d\lambda
\leq0.
\end{align*}
This implies $\alpha=B_\lambda$, proving agreement at probability measures. Finally, we have
\[M(\alpha)(\mathbbm 1_X)
=\bigl(\alpha H_c^{-1}(\alpha)\bigr)(\mathbbm 1_X)
 =H_c^{-1}(\alpha)(\alpha\mathbbm 1_X)
 =H_c^{-1}(\alpha)(\alpha)
=\langle\alpha,\alpha\rangle_{\mathcal H_c}
=\|\alpha\|_{\mathcal H_c}^2.\]

Now suppose $\beta\in U_+\cap\mathbb S(\mathcal H_c)$. The preceding identity shows that, for every $\alpha\in L'$, we have $\alpha\in\mathbb S(\mathcal H_c)$ if and only if $M(\alpha)\in \{\lambda\in H^s(X)^*:\lambda(\mathbbm{1}_X)=1\}$. Consequently,
\[
M(L'\cap\mathbb S(\mathcal H_c))=W\cap \{\lambda\in H^s(X)^*:\lambda(\mathbbm{1}_X)=1\}.
\]
Since $L'$ is open in $\mathcal H_c$ and $W$ is open in $H^s(X)^*$, these intersections are relatively open neighborhoods of $\beta$ and $M(\beta)$ in $U_+\cap\mathbb S(\mathcal H_c)$ and $\{\lambda\in H^s(X)^*:\lambda(\mathbbm{1}_X)=1\}$, respectively. The restrictions of $M$ and $B$ are smooth and inverses of each other, and hence define a $C^\infty$ diffeomorphism between these relatively open neighborhoods. Moreover, since $\operatorname D_hM(\beta)=hH_c^{-1}(\beta)+\beta H_c^{-1}(h)$, the definition of dual multiplication and the Riesz identity give
\begin{multline*}
(\operatorname D_hM(\beta))(\mathbbm{1}_X)
=\bigl(hH_c^{-1}(\beta)+\beta H_c^{-1}(h)\bigr)(\mathbbm{1}_X)
=H_c^{-1}(\beta)(h)+H_c^{-1}(h)(\beta)\\
=\langle\beta,h\rangle_{\mathcal H_c}
+\langle h,\beta\rangle_{\mathcal H_c}
=2\langle\beta,h\rangle_{\mathcal H_c}=0,
\end{multline*}
where the last equality follows from $h\in\beta^\perp$.
\end{proof}

\begin{lemma}[Smoothness of the relevant operators]
\label{ambient resolvent regularity}
For $\beta\in U_+$, define
\[\begin{aligned}
\mathbf{R}: U_+ &\to \mathcal{L}(\mathcal{H}_c, \mathcal{H}_c) \\
\beta &\mapsto (\operatorname{id}_{\mathcal H_c}-\mathbf K_\beta)^{-1}\Pi_\beta
\end{aligned}
\qquad \text{and}
\qquad
\begin{aligned}
    S: U_+ &\to \mathcal{L}(\mathcal{H}_c, \mathcal{H}_c) \\
    \beta &\mapsto \operatorname{id}_{\mathcal H_c}+2\mathbf J_\beta\mathbf R_\beta.
\end{aligned}
\]
These operators agree with the original ones at $\beta=B_\mu$, and the maps $(\beta\mapsto\mathbf R_\beta)$, $(\beta\mapsto S_\beta)$, and $(\beta\mapsto S_\beta^{-1})$ are elements of $C^\infty(U_+; \mathcal{L}(\mathcal{H}_c, \mathcal{H}_c)).$  Moreover, we have the explicit expressions
\begin{align*}
S_\beta =(\operatorname{id}_{\mathcal H_c}+\mathbf J_\beta)(\operatorname{id}_{\mathcal H_c}-\mathbf J_\beta)^{-1}, \quad \text{and} \quad 
S_\beta^{-1}=(\operatorname{id}_{\mathcal H_c}-\mathbf J_\beta)(\operatorname{id}_{\mathcal H_c}+\mathbf J_\beta)^{-1}.
\end{align*}
\end{lemma}

\begin{proof}
By \Cref{ambient coefficient domain}, $\mathbf K_\beta=\mathbf J_\beta$ and $\mathbf J_\beta\Pi_\beta=\Pi_\beta\mathbf J_\beta=\mathbf J_\beta$. Since $\|\mathbf J_\beta\|_{\operatorname{op}}<1$, both $\operatorname{id}_{\mathcal H_c}\pm\mathbf J_\beta$ are boundedly invertible. Moreover, $\Pi_\beta$ commutes with $\operatorname{id}_{\mathcal H_c}-\mathbf J_\beta$, and hence with its inverse \citep[Section 11.5, pp.~149--150]{robinson2020functional}. Using $\mathbf J_\beta\Pi_\beta=\mathbf J_\beta$, we obtain
\begin{align*}
S_\beta
=\operatorname{id}_{\mathcal H_c}
+2\mathbf J_\beta(\operatorname{id}_{\mathcal H_c}-\mathbf J_\beta)^{-1}\Pi_\beta =\operatorname{id}_{\mathcal H_c}
+2\mathbf J_\beta(\operatorname{id}_{\mathcal H_c}-\mathbf J_\beta)^{-1} =(\operatorname{id}_{\mathcal H_c}+\mathbf J_\beta)(\operatorname{id}_{\mathcal H_c}-\mathbf J_\beta)^{-1},
\end{align*}
which also yields the displayed formula for $S_\beta^{-1}$. Smoothness of $\beta\mapsto\mathbf R_\beta$ in operator norm follows from \Cref{smoothness of C beta}. By \Cref{smoothness of J beta}, $\beta\mapsto\mathbf J_\beta$ is $C^\infty$ in operator norm, so boundedness of operator composition gives the asserted smoothness of $S_\beta=\operatorname{id}_{\mathcal H_c}+2\mathbf J_\beta\mathbf R_\beta$. The formula above for $S_\beta^{-1}$, together with smoothness of operator inversion established in the proof of \Cref{smoothness of C beta}, yields smoothness of $\beta\mapsto S_\beta^{-1}$. Finally, at $\beta=B_\mu$, we have $M(\beta)=\mu$, $a_\beta=a_\mu$, and $\Pi_\beta=\Pi_\mu$. Thus \Cref{theta inverse,tilde K expression,Rbar expression,reparameterization} identify $\mathbf J_\beta$, $\mathbf K_\beta$, $\mathbf R_\beta$, and $S_\beta$ with their original counterparts.
\end{proof}

\begin{corollary}[Smoothness of extended operators]
\label{agreement of base derivatives}

For every $\mu\in\mathcal P(X)$, there exists an open neighborhood $W\subset H^s(X)^*$ of $\mu$ such that the maps
\[
\begin{aligned}
a:W &\to \mathcal H_c\\
\lambda &\mapsto a_\lambda
\end{aligned}
\qquad
\begin{aligned}
B:W &\to \mathcal H_c\\
\lambda &\mapsto B_\lambda
\end{aligned}
\qquad
\begin{aligned}
\Pi:W &\to \mathcal L(\mathcal H_c,\mathcal H_c)\\
\lambda &\mapsto \Pi_\lambda
\end{aligned}
\qquad
\begin{aligned}
\mathbf J:W &\to \mathcal L(\mathcal H_c,\mathcal H_c)\\
\lambda &\mapsto \mathbf J_\lambda
\end{aligned}
\]
and
\[
\begin{aligned}
\mathbf K:W &\to \mathcal L(\mathcal H_c,\mathcal H_c)\\
\lambda &\mapsto \mathbf K_\lambda
\end{aligned}
\qquad
\begin{aligned}
\mathbf R:W &\to \mathcal L(\mathcal H_c,\mathcal H_c)\\
\lambda &\mapsto \mathbf R_\lambda
\end{aligned}
\qquad
\begin{aligned}
S:W &\to \mathcal L(\mathcal H_c,\mathcal H_c)\\
\lambda &\mapsto S_\lambda
\end{aligned}
\]
are $C^\infty$ in their respective norm topologies. For every $\gamma\in\mathbb T_\mu^m$, the Fr\'echet derivative of each extension in direction $\gamma$ agrees with its previously derived Hadamard derivative from \Cref{sec: operator regularity}. If $\mu$ additionally satisfies \Cref{standing sinkhorn assumptions}.3, its restriction to $H^s(X)^*_0\cong T_\mu\mathcal P(X)$ is the unique continuous extension of the statistical Hadamard derivative.
\end{corollary}
\begin{proof}

Since $B_\mu\in U_+$, \Cref{ambient inverse change of vars} gives an open neighborhood $W\subset H^s(X)^*$ of $\mu$ and a $C^\infty$ local inverse $\lambda\mapsto B_\lambda\in U_+$ of $M$, which agrees with the original change of variables on $W\cap\mathcal P(X)$. For every $\lambda\in W$, define the extended operators by $a_\lambda=a_{B_\lambda}=(B_\lambda)^{-1}$, $\Pi_\lambda=\Pi_{B_\lambda}$, $\mathbf J_\lambda=\mathbf J_{B_\lambda}$, $\mathbf K_\lambda=\mathbf K_{B_\lambda}$, $\mathbf R_\lambda=\mathbf R_{B_\lambda}$, and $S_\lambda=S_{B_\lambda}$. By \Cref{ambient coefficient domain,ambient resolvent regularity}, each of the corresponding maps in $\beta$-coordinates is $C^\infty$ on $U_+$. Since $\lambda\mapsto B_\lambda$ is also $C^\infty$, the Banach-space chain rule implies that all the extended maps are $C^\infty$ on $W$ in their respective norm topologies. This proves the smoothness of the extended operator-valued maps on the neighborhood $W$ of $\mu$ in $H^s(X)^*.$

We now prove coincidence of the Fr\'echet derivatives with the previously derived Hadamard derivatives at $\lambda=\mu$. By \Cref{duality of continuous embedding}, the embeddings
\[
H^m(X)^*\hookrightarrow C^m(X)^*\hookrightarrow H^s(X)^*
\]
are continuous. In particular, convergence in $C^m(X)^*$ implies convergence in $H^s(X)^*$. Since the constructed extensions are Fr\'echet differentiable in the latter topology and agree with the original maps at probability measures, their restrictions to $\mathcal P(X)$ are Hadamard differentiable tangentially to $\mathbb T_\mu^m$ in the $C^m(X)^*$ topology, with derivatives given by the restrictions of their Fr\'echet derivatives. Note that this follows from the fact that the admissible sequences for Hadamard differentiation in the $C^m(X)^*$ topology are also admissible in $H^s(X)^*$ topology. Moreover, by definition, $\mathfrak M_\mu$ is dense in $\mathbb T_\mu^m$ in the $H^m(X)^*$ topology, and hence also in the $C^m(X)^*$ topology. Since both derivatives agree on the admissible perturbation directions $\mathfrak M_\mu$ (because the original maps agree on a neighborhood of $\mu$ due to \Cref{ambient inverse change of vars}), uniqueness of the Hadamard derivative \citep[Lemma 1]{gill1989nonparametric} implies that they coincide on all of $\mathbb T_\mu^m$. Finally, by \Cref{statistical riemannian tangent embedding}, if $\mu$ satisfies \Cref{standing sinkhorn assumptions}.3 then $\mathbb T_\mu^m=H^m(X)^*_0$ is dense in $H^s(X)^*_0$, so the restriction of the Fr\'echet derivative to $H^s(X)^*_0$ is the unique continuous extension of the statistical Hadamard derivative, which again follows from \citet[Lemma 1]{gill1989nonparametric}.
\end{proof}

\subsubsection{Smoothness of Coefficients}

\begin{lemma}[Smoothness and extension of the geodesic coefficients in $\beta$-coordinates]
\label{smoothness of geodesic coefficients}

The maps
\[
\begin{aligned}
\Pi:U_+ &\to \mathcal L(\mathcal H_c,\mathcal H_c)\\
\beta &\mapsto \Pi_\beta
\end{aligned}
\quad
\begin{aligned}
\mathbf R:U_+ &\to \mathcal L(\mathcal H_c,\mathcal H_c)\\
\beta &\mapsto \mathbf R_\beta
\end{aligned}
\quad
\begin{aligned}
S:U_+ &\to \mathcal L(\mathcal H_c,\mathcal H_c)\\
\beta &\mapsto S_\beta
\end{aligned}
\]
and
\[
\begin{aligned}
S^{-1}:U_+ &\to \mathcal L(\mathcal H_c,\mathcal H_c)\\
\beta &\mapsto S_\beta^{-1}
\end{aligned}
\qquad
\begin{aligned}
\Gamma:U_+\times\mathcal H_c\times\mathcal H_c &\to \mathcal H_c\\
(\beta,u,v) &\mapsto \Gamma_\beta(u,v)
\end{aligned}
\]
are $C^\infty$ in their respective norm topologies. Moreover, $\Gamma_\beta$ is bounded bilinear in its vector arguments and takes values in $\beta^\perp$. 
\end{lemma}

\begin{proof}
Smoothness of $\Pi_\beta$, $\mathbf R_\beta$, $S_\beta$, and $S_\beta^{-1}$ follows from \Cref{ambient coefficient domain,ambient resolvent regularity}. Since $\beta\mapsto S_\beta$ is $C^\infty$, the map $(\beta,u)\mapsto\operatorname D_uS_\beta$ is also $C^\infty$. By \Cref{smoothness of C beta}, the extended Riesz representative
\[
C_\beta(u,v)^\sharp
=2\left(
 a_\beta(\mathbf R_\beta u)(\mathbf R_\beta v)
-H_c\bigl[a_\beta^3(\mathbf R_\beta u)(\mathbf R_\beta v)M(\beta)\bigr]
\right)
\]
is $C^\infty$ in $(\beta,u,v)$. Define
\[
\Gamma_\beta(u,v)
=\frac12\Pi_\beta S_\beta^{-1}
\left[
 (\operatorname D_{\Pi_\beta u}S_\beta)\Pi_\beta v
 +(\operatorname D_{\Pi_\beta v}S_\beta)\Pi_\beta u
 -C_\beta(u,v)^\sharp
\right].
\]
Every constituent map is $C^\infty$, so boundedness of operator composition and evaluation proves smoothness of $\Gamma$. Its displayed formula also proves bounded bilinearity in $u,v$ and membership in $\beta^\perp$.
At $\beta=B_\mu$, the constituent operators agree with the original ones. For $u\in\mathfrak T_\mu\mathcal P(X)$, set $\sigma_u=\operatorname D_uM(\beta)$. By \Cref{ambient inverse change of vars,agreement of base derivatives}, the chain rule gives $\operatorname D_uS_\beta=\operatorname D_{\sigma_u}S_\mu$. Together with \Cref{Cmu riesz rep}, this identifies the extended connection with \Cref{gamma expression} on tangent inputs at probability-valued points.
\end{proof}

\begin{lemma}[Metric compatibility of the ambient connection]
\label{ambient metric compatibility}
For every $\beta\in U_+$ and $h,u,v\in\beta^\perp$, the extended operators satisfy
\[
\left\langle C_\beta(u,v)^\sharp,h\right\rangle_{\mathcal H_c}
=
\left\langle(\operatorname D_hS_\beta)u,v\right\rangle_{\mathcal H_c}.
\]
Consequently, the extended covariant derivative $\nabla_{\dot\beta_t}u_t=\Pi_{\beta_t}\dot u_t+\Gamma_{\beta_t}(\dot\beta_t,u_t)$ is metric-compatible along every $C^1$ curve $(\beta_t)\subset U_+\cap\mathbb S(\mathcal H_c)$. In particular, for any $C^1$ tangent fields $u_t,v_t\in\beta_t^\perp$,
\[
\frac{d}{dt}\tilde{\mathbf g}_{\beta_t}(u_t,v_t)
=
\tilde{\mathbf g}_{\beta_t}(\nabla_{\dot\beta_t}u_t,v_t)
+
\tilde{\mathbf g}_{\beta_t}(u_t,\nabla_{\dot\beta_t}v_t),
\]
where $\tilde{\mathbf g}_\beta(u,v)=\frac{\varepsilon}{2}\langle S_\beta u,v\rangle_{\mathcal H_c}$. Moreover, every ambient geodesic contained in $U_+\cap\mathbb S(\mathcal H_c)$ has constant metric speed.
\end{lemma}

\begin{proof}
Fix $\beta\in U_+$ and $h,u,v\in\beta^\perp$, and write $\xi=\mathbf R_\beta u$ and $\zeta=\mathbf R_\beta v$. By \Cref{ambient coefficient domain}, $\xi,\zeta\in\beta^\perp$. The definition of $\mathbf J_\beta$ gives
\begin{align*}
    \langle \mathbf{J}_\beta \xi, \zeta\rangle_{\mathcal{H}_c} &= \langle H_c[a_\beta^2(\Pi_\beta \xi)M(\beta)], \zeta \rangle_{\mathcal{H}_c}\\ 
    &= \left(a_\beta^2(\Pi_\beta \xi)M(\beta)\right)(\zeta)  \\ 
    &= M(\beta)\left(a_\beta^2 (\Pi_\beta\xi)\zeta\right) \tag{dual multiplication, \Cref{rkhs algebra calculus}} \\
    &= \beta H_c^{-1}(\beta)\left(a_\beta^2 (\Pi_\beta\xi)\zeta\right) \tag{$M(\beta) = \beta H_c^{-1}(\beta)$} \\
    &= H_c^{-1}(\beta)\left(a_\beta (\Pi_\beta\xi)(\Pi_\beta \zeta)\right).
\end{align*}
By \Cref{ambient basic smoothness}, the map $\beta\mapsto a_\beta=\beta^{-1}$ is $C^\infty$ on $V_+$. Differentiating $a_\beta\beta=\mathbbm{1}_X$ in direction $h$ and applying the product rule gives $(\operatorname D_h a_\beta)\beta+a_\beta h=0$, and hence $\operatorname D_h a_\beta=-a_\beta^2h$. 
Recall that $\xi,\zeta\in\beta^\perp$, so $\Pi_\beta\xi=\xi$ and $\Pi_\beta\zeta=\zeta$. Moreover, differentiating the identity
$\Pi_\beta w=w-\langle w,\beta\rangle_{\mathcal H_c}\beta/\|\beta\|_{\mathcal H_c}^2$
with respect to $\beta$, holding $w$ fixed, gives
$(\operatorname D_h\Pi_\beta)\xi=-\langle h,\xi\rangle_{\mathcal H_c}\beta/\|\beta\|_{\mathcal H_c}^2$,
and similarly for $\zeta$. Differentiating the identity in the preceding frame with respect to $\beta$ in direction $h$ (holding $\xi,\zeta$ fixed) therefore yields
\begin{align*}
\left\langle(\operatorname D_h\mathbf J_\beta)\xi,\zeta\right\rangle_{\mathcal H_c}
&=H_c^{-1}(h)(a_\beta\xi\zeta)
-H_c^{-1}(\beta)(a_\beta^2h\xi\zeta)\\
&\hspace{30mm}+H_c^{-1}(\beta)\bigl(a_\beta((\operatorname D_h\Pi_\beta)\xi)\zeta\bigr)
+H_c^{-1}(\beta)\bigl(a_\beta\xi((\operatorname D_h\Pi_\beta)\zeta)\bigr)\\
&=H_c^{-1}(h)(a_\beta\xi\zeta)
-H_c^{-1}(\beta)(a_\beta^2h\xi\zeta)\\
&\hspace{30mm}-\frac{\langle h,\xi\rangle_{\mathcal H_c}}{\|\beta\|_{\mathcal H_c}^2}H_c^{-1}(\beta)(\zeta)
-\frac{\langle h,\zeta\rangle_{\mathcal H_c}}{\|\beta\|_{\mathcal H_c}^2}H_c^{-1}(\beta)(\xi)\\
&=H_c^{-1}(h)(a_\beta\xi\zeta)
-H_c^{-1}(\beta)(a_\beta^2h\xi\zeta)\\
&=\langle a_\beta\xi\zeta,h\rangle_{\mathcal H_c}
-M(\beta)(a_\beta^3h\xi\zeta).
\end{align*}
Here, the third equality follows from the Riesz identity and the orthogonality relations $H_c^{-1}(\beta)(\xi)=\langle\beta,\xi\rangle_{\mathcal H_c}=0$ and $H_c^{-1}(\beta)(\zeta)=\langle\beta,\zeta\rangle_{\mathcal H_c}=0$. The final equality uses $M(\beta)=\beta H_c^{-1}(\beta)$ and $\beta a_\beta=\mathbbm{1}_X$. By \Cref{ambient resolvent regularity}, $S_\beta=2(\operatorname{id}-\mathbf J_\beta)^{-1}-\operatorname{id}$. Differentiating and using the definition of $\mathbf R_\beta$ therefore yields
\begin{align*}
\left\langle(\operatorname D_hS_\beta)u,v\right\rangle_{\mathcal H_c}
=2\left\langle(\operatorname D_h\mathbf J_\beta)\xi,\zeta\right\rangle_{\mathcal H_c} =2\left(\left\langle a_\beta\xi\zeta,h\right\rangle_{\mathcal H_c}
-M(\beta)(a_\beta^3h\xi\zeta)\right) =\left\langle C_\beta(u,v)^\sharp,h\right\rangle_{\mathcal H_c},
\end{align*}
where the final equality follows from the expression for $C_\beta(u,v)^\sharp$ in \Cref{smoothness of geodesic coefficients} and the Riesz identity.

We now establish metric compatibility. Since $S_\beta$ is self-adjoint and commutes with $\Pi_\beta$ by \Cref{ambient coefficient domain,ambient resolvent regularity}, the definition of $\Gamma_\beta$ in \Cref{smoothness of geodesic coefficients} gives
\begin{align*}
2\left\langle S_\beta\Gamma_\beta(h,u),v\right\rangle_{\mathcal H_c}
+2\left\langle S_\beta u,\Gamma_\beta(h,v)\right\rangle_{\mathcal H_c} &=2\left\langle(\operatorname D_hS_\beta)u,v\right\rangle_{\mathcal H_c}
+\left\langle(\operatorname D_uS_\beta)h,v\right\rangle_{\mathcal H_c}
-\left\langle C_\beta(h,v)^\sharp,u\right\rangle_{\mathcal H_c} \\&\hspace{33.5mm}+\left\langle(\operatorname D_vS_\beta)h,u\right\rangle_{\mathcal H_c}
-\left\langle C_\beta(h,u)^\sharp,v\right\rangle_{\mathcal H_c}\\ &=2\left\langle(\operatorname D_hS_\beta)u,v\right\rangle_{\mathcal H_c},
\end{align*}
where the final equality follows from the identity proved above. Applying this with $h=\dot\beta_t$ and differentiating $\tilde{\mathbf g}_{\beta_t}(u_t,v_t)$ gives the stated metric-compatibility identity, since $S_{\beta_t}$ preserves $\beta_t^\perp$. Finally, if $(\beta_t)$ satisfies the ambient geodesic equation, then $\nabla_{\dot\beta_t}\dot\beta_t=0$. Applying metric compatibility with $u_t=v_t=\dot\beta_t$ shows that $\frac{d}{dt}\tilde{\mathbf g}_{\beta_t}(\dot\beta_t,\dot\beta_t)=0$, completing the proof.
\end{proof}

All distribution (i.e. $H^s(X)^*$) indexed operators at a point $\mu=M(\beta)$ are henceforth interpreted through the operators associated with $\beta = B_\mu$. In particular, for $h\in\mathcal H_c$ and $\sigma_h=\operatorname D_hM(\beta)$, the chain rule gives $\operatorname D_{\sigma_h}\Pi_\mu=\operatorname D_h\Pi_\beta$ and $\operatorname D_{\sigma_h}S_\mu=\operatorname D_hS_\beta$. A distributional base derivative of $\Gamma_\beta$ is interpreted as the derivative of $\lambda\mapsto\Gamma_{B_\lambda}$, holding the vector arguments fixed. These are local identifications and do not require $M(\beta)$ to be a probability measure.

\begin{lemma}[Smoothness of the chart acceleration]
\label{smoothness of chart acceleration}
Fix $\beta_*\in U_+\cap\mathbb S(\mathcal H_c)$. Let $V_R=\{p\in\beta_*^\perp:\|p\|_{\mathcal H_c}<R\}$ and $\psi(p)=\beta_*\sqrt{1-\|p\|_{\mathcal H_c}^2}+p$, where $R\in(0,1)$ is chosen sufficiently small that $\psi(V_R)\subset U_+$. Identifying the coefficients with their extensions from \Cref{smoothness of geodesic coefficients}, define
\[
\mathscr A(p,u)
=-\|\operatorname D_u\psi(p)\|_{\mathcal H_c}^2p
-\Pi_{\beta_*}\Pi_{\psi(p)}\Gamma_{\psi(p)}
\bigl(\operatorname D_u\psi(p),\operatorname D_u\psi(p)\bigr).
\]
Then $\mathscr A:V_R\times\beta_*^\perp\to\beta_*^\perp$ is $C^\infty$ and is quadratic in $u$ for each fixed $p$.
\end{lemma}

\begin{proof}
The choice of $R$ is possible because $\psi(0)=\beta_*$ and $U_+$ is open by \Cref{ambient coefficient domain}. The chart is smooth, and $\operatorname D_u\psi(p)=u-\langle p,u\rangle_{\mathcal H_c}(1-\|p\|_{\mathcal H_c}^2)^{-1/2}\beta_*$. Thus $(p,u)\mapsto\operatorname D_u\psi(p)$ is $C^\infty$. By \Cref{smoothness of geodesic coefficients}, the maps $\beta\mapsto\Pi_\beta$ and $(\beta,u,v)\mapsto\Gamma_\beta(u,v)$ are $C^\infty$ on their respective domains over $U_+$. Boundedness of the inner product, operator evaluation, and the fixed projection $\Pi_{\beta_*}$ therefore gives smoothness of $\mathscr A$ by composition. Both terms take values in $\beta_*^\perp$, and their quadratic dependence on $u$ follows from linearity of $\operatorname D_{(\cdot)}\psi(p)$ and bilinearity of $\Gamma_{\psi(p)}$.
\end{proof}

\begin{prop}[Existence of ambient geodesics]
\label{small tangent full time existence}
Fix $\nu\in\mathcal P(X)$ and let $\beta_0=B_\nu$. For every $w\in\mathfrak T_\nu\mathcal P(X)$, the ambient geodesic initial-value problem with initial conditions $(\beta_0,w)$ admits a unique local solution in $U_+\cap\mathbb S(\mathcal H_c)$. Moreover, there exists $\delta_\nu>0$ such that, whenever $\|w\|_{\mathcal H_c}<\delta_\nu$, this solution exists throughout $[0,1]$ and remains in $U_+\cap\mathbb S(\mathcal H_c)$.
\end{prop}

\begin{proof}
By \Cref{ambient coefficient domain}, $\beta_0\in U_+$, so choose a sphere chart $\psi$ centered at $\beta_0$ with image in $U_+$ as in \Cref{smoothness of chart acceleration}. By \Cref{chart geodesic ODE}, the coordinate equation is $\dot p_t=u_t$, $\dot u_t=\mathscr A(p_t,u_t)$, with $(p_0,u_0)=(0,w)$. Its vector field is $C^\infty$, so \Cref{local smoothness theorem} gives a unique local solution. Pulling it back through $\psi$ proves the first assertion.

Apply \Cref{local smoothness theorem} at $(p,u)=(0,0)$. Choose $\tau>0$ with $[0,\tau]$ contained in its common open time interval, and $r>0$ such that every $(0,v)$ with $\|v\|_{\mathcal H_c}<r$ belongs to its neighborhood of initial conditions. The resulting solutions $(p_t^v,u_t^v)$ are defined on $[0,\tau]$ and satisfy $\psi(p_t^v)\in U_+$. Set $\delta_\nu=\tau r$. For $\|w\|_{\mathcal H_c}<\delta_\nu$, take $v=w/\tau$ and define $p_t=p_{\tau t}^v$, $u_t=\tau u_{\tau t}^v$ for $t\in[0,1]$. Quadratic dependence of $\mathscr A$ on its velocity argument gives
\[
\dot p_t=u_t,
\qquad
\dot u_t=\tau^2\mathscr A(p_{\tau t}^v,u_{\tau t}^v)
=\mathscr A(p_t,u_t).
\]
Since $(p_0,u_0)=(0,w)$, uniqueness identifies this curve with the desired coordinate geodesic. Its image under $\psi$ lies in $U_+\cap\mathbb S(\mathcal H_c)$ for every $t\in[0,1]$.
\end{proof}

\begin{lemma}[Uniqueness of the linearized geodesic equation]
\label{uniqueness linearized geodesic}
For every tangent-bundle chart $\Psi$ induced by a sphere chart as in \Cref{smoothness of chart acceleration}, $\operatorname D\Psi_{(\beta,u)}$ is a bounded linear isomorphism with inverse $\operatorname D\Psi^{-1}_{\Psi(\beta,u)}$. Consequently, the linear initial value problem in \Cref{linearizing expmap}, subject to its tangency constraint, admits at most one solution.
\end{lemma}

\begin{proof}
The tangent-bundle chart is $\Psi(\beta,u)=(\Pi_{\beta_*}\beta,\Pi_{\beta_*}u)$, with smooth inverse $\Psi^{-1}(p,u)=(\psi(p),\operatorname D_u\psi(p))$. Differentiating the two inverse relations shows that their differentials are bounded linear inverses on the corresponding tangent spaces.

Suppose $h_t$ and $\widetilde h_t$ solve the linear initial value problem with the same initial data. The tangency condition and its time derivative give $\langle h_t,\beta_t\rangle_{\mathcal H_c}=0$ and $\langle\dot h_t,\beta_t\rangle_{\mathcal H_c}+\langle h_t,\dot\beta_t\rangle_{\mathcal H_c}=0$, and similarly for $\widetilde h_t$. Thus $(h_t,\dot h_t)$ and $(\widetilde h_t,\dot{\widetilde h}_t)$ belong to the tangent space of $T\mathbb S(\mathcal H_c)$ at $(\beta_t,\dot\beta_t)$. On each chart interval, applying $\operatorname D\Psi_{(\beta_t,\dot\beta_t)}$ to these pairs converts the linearized geodesic equation into \Cref{state space ivp}. The resulting coordinate variations have the same initial data, so uniqueness of the linear variational equation in \Cref{local smoothness theorem} makes them equal. Applying $\operatorname D\Psi^{-1}$ gives equality of the original pairs. Repeating across a finite sequence of chart intervals covering the compact reference trajectory proves $h_t=\widetilde h_t$ throughout $[0,1]$.
\end{proof}

\begin{corollary}[Smoothness of the chart parallel transport field]
\label{smoothness of chart parallel transport}
Under the hypotheses and notation of \Cref{smoothness of chart acceleration}, and identifying the coefficients with the same extensions on $U_+$, define
\begin{align*}
\mathscr R(p,u,r)
= -\left\langle\operatorname D_r\psi(p),\operatorname D_u\psi(p)\right\rangle_{\mathcal H_c}p -\Pi_{\beta_*}\Pi_{\psi(p)}\Gamma_{\psi(p)}
\bigl(\operatorname D_u\psi(p),\operatorname D_r\psi(p)\bigr).
\end{align*}
Then $\mathscr R:V_R\times\beta_*^\perp\times\beta_*^\perp\to\beta_*^\perp$ is $C^\infty$. Moreover, for each fixed $(p,u)\in V_R\times\beta_*^\perp$, the map $r\mapsto\mathscr R(p,u,r)$ is bounded linear.
\end{corollary}

\begin{proof}
By the proof of \Cref{smoothness of chart acceleration}, both $(p,u)\mapsto\operatorname D_u\psi(p)$ and $(p,r)\mapsto\operatorname D_r\psi(p)$ are $C^\infty$. Together with \Cref{smoothness of geodesic coefficients}, boundedness of the inner product and operator evaluation, and boundedness of $\Pi_{\beta_*}$, this proves that both terms defining $\mathscr R$ are $C^\infty$ by composition. Both take values in $\beta_*^\perp$. Finally, for fixed $(p,u)$, $r\mapsto\operatorname D_r\psi(p)$ is bounded linear and $\Gamma_{\psi(p)}$ is bounded bilinear in its vector arguments. Hence both terms defining $\mathscr R(p,u,r)$ are bounded linear in $r$.
\end{proof}

\subsection{Proof of \Cref{three coordinate ode}} \label{proof of three coordinate ode}

We seek an expression for $\dot r_t^i$, which we will do through $w_t$. Note that $\dot w_t = \Pi_{\mu_t} \dot w_t + (\operatorname{id} - \Pi_{\mu_t})\dot w_t$, and by the parallel transport equations, 
\[\Pi_{\mu_t}\dot w_t = -\Pi_{\mu_t}\Gamma_{\beta_t}(\dot \beta_t, w_t).\]
To obtain the normal component of $\dot w_t$, we can differentiate the identity $\langle w_t, \beta_t\rangle_{\mathcal{H}_c} = 0$ to say $\langle \dot w_t, \beta_t\rangle_{\mathcal{H}_c} = - \langle w_t, \dot \beta_t \rangle_{\mathcal{H}_c}$, and thus $\dot w_t = -\langle w_t, \dot \beta_t\rangle_{\mathcal{H}_c}\beta_t - \Pi_{\mu_t}\Gamma_{\beta_t}(\dot \beta_t, w_t).$ It therefore follows that 
\begin{align*}
    \dot r_t^i = - \langle w_t, \dot \beta_t\rangle_{\mathcal{H}_c} \Pi_{\mu_{t_i}} \beta_t - \Pi_{\mu_{t_i}} \Pi_{\mu_t}\Gamma_{\beta_t}(\dot \beta_t, w_t) \quad \text{for $t \in I_{t_i}$.}
\end{align*}
Converting to state-space coordinates $(p^i, u^i, r^i)$ and identifying $\Pi_{\mu_t}$ with $\Pi_{\beta_t}$ implies
\begin{align*}
    \dot r_t^i = - \langle \operatorname{D}_{r_t^i}\psi_{t_i}(p_t^i), \operatorname{D}_{u_t^i}\psi_{t_i}(p_t^i)\rangle_{\mathcal{H}_c}  p_t^i - \Pi_{\mu_{t_i}} \Pi_{\psi_{t_i}(p_t^i)} \Gamma_{\psi_{t_i}(p_t^i)}(\operatorname{D}_{u_t^i}\psi_{t_i}(p_t^i), \operatorname{D}_{r_t^i}\psi_{t_i}(p_t^i)).
\end{align*}
We see that $\dot r_t^i = \mathscr{R}_i(p_t^i, u_t^i, r_t^i)$ indeed. Recalling the definition of $\mathscr{A}$ from \Cref{chart geodesic ODE} and specializing it to the chart $\psi_{t_i}$ concludes the proof.
\qed{}

\subsection{Proof of \Cref{chart perturbation boundary condition}} \label{proof of chart perturbation boundary condition}

Recall that $\dot r_t^i = \mathscr{R}_i(p_t^i, u_t^i, r_t^i)$. Thus, we can readily compute the time derivative of the first variation of $r_t^i$ by taking the spatial derivative of $(p_t^i, u_t^i, r_t^i) \mapsto \mathscr{R}_i(p_t^i, u_t^i, r_t^i)$ as 
\begin{align*}
    \dot{\mathfrak r}_t^i = \operatorname{D}_{1, \mathfrak{p}_t^i}\mathscr{R}_i(p_t^i, u_t^i, r_t^i) + \operatorname{D}_{2, \mathfrak u_t^i}\mathscr{R}_i(p_t^i, u_t^i, r_t^i) + \operatorname{D}_{3, \mathfrak r_t^i}\mathscr{R}_i(p_t^i, u_t^i, r_t^i) \quad \text{for $t \in I_{t_i}$.}
\end{align*}
For $i > 1$, the initial condition is obtained from the solution in the previous chart. In particular, it is obtained via the differential of the chart transition as
\begin{equation*}
    \begin{pmatrix}
        \mathfrak{p}_{s_{i-1}}^i \\
        \mathfrak u_{s_{i - 1}}^i \\
        \mathfrak r_{s_{i - 1}}^i
    \end{pmatrix} = \operatorname{D}_{(\mathfrak{p}_{s_{i-1}}^{i-1},\mathfrak u_{s_{i - 1}}^{i - 1}, \mathfrak{r}_{s_{i - 1}}^{i - 1})}(\Psi_i \circ \Psi^{-1}_{i - 1})(p^{i-1}_{s_{i-1}}, u^{i-1}_{s_{i-1}}, r^{i-1}_{s_{i-1}}).
\end{equation*}
Otherwise, if $i = 1$, then $(\mathfrak{p}^1_0, \mathfrak{u}_0^1, \mathfrak{r}_0^1)$ is obtained by converting the initial condition of the $\mathcal{H}_c$ perturbation variables to chart perturbation variables (\Cref{fig:imp-linearization-notation}). In particular, recall that our initial position in $\mathcal{H}_c$ coordinates and our initial perturbation in $\mathcal{H}_c$ coordinates are given by

\[(\beta_0, \dot \beta_0, w_0) = (B_\mu, \log_\mu(\nu), \log_\mu(\rho)), \quad \text{and} \quad (h_0, \dot h_0, k_0) = (\operatorname{D}_\gamma B_\mu, \operatorname{D}_{(\gamma, \eta)}\log_\mu(\nu), \operatorname{D}_{(\gamma, \zeta)}\log_\mu(\rho)), \]
respectively. To convert this to chart perturbation variables we simply push through $\operatorname{D}_{(\cdot)}\Psi_1$. In particular, we obtain the initial condition
\begin{align*}
    \begin{pmatrix}
        \mathfrak{p}^1_0 \\ \mathfrak{u}_0^1 \\\mathfrak{r}_0^1
    \end{pmatrix} = \operatorname{D}_{(h_0, \dot h_0, k_0)}\Psi_1(\beta_0, \dot \beta_0, w_0).
\end{align*}
Summarizing, we have 
\begin{equation}
    \begin{pmatrix}
        \mathfrak{p}_{s_{i-1}}^i \\
        \mathfrak u_{s_{i - 1}}^i \\
        \mathfrak r_{s_{i - 1}}^i
    \end{pmatrix} = \begin{cases}\operatorname{D}_{(\mathfrak{p}_{s_{i-1}}^{i-1},\mathfrak u_{s_{i - 1}}^{i - 1}, \mathfrak{r}_{s_{i - 1}}^{i - 1})}(\Psi_i \circ \Psi^{-1}_{i - 1})(p^{i-1}_{s_{i-1}}, u^{i-1}_{s_{i-1}}, r^{i-1}_{s_{i-1}}) & \text{if $i > 1$} \\
    \operatorname{D}_{(h_0, \dot h_0, k_0)}\Psi_1(\beta_0, \dot \beta_0, w_0) & \text{otherwise}.
    \end{cases}
\end{equation}

\qed{}

\subsection{Proof of \Cref{first variation sinkhorn distance}}\label{proof of first variation sinkhorn distance}

Set $u_t=(1-t)u_0+tu_1$ and, for any probability-valued curve $\boldsymbol\eta$, define $(T_r\boldsymbol\eta)_t=e^{ru_t}\eta_t/\eta_t(e^{ru_t})$. Let $\boldsymbol\mu^r$ be a variational minimizer joining $\mu^r$ and $\nu^r$. Since $T_r\boldsymbol\mu$ is an admissible competitor for the variational problem with endpoints $(\mu^r, \nu^r)$ and $T_{-r}\boldsymbol\mu^r$ is an admissible competitor for the variational problem with endpoints $(\mu, \nu)$, minimality gives
\begin{align}
\mathscr E(\boldsymbol\mu^r)-\mathscr E(T_{-r}\boldsymbol\mu^r)
\leq d_S^2(\mu^r,\nu^r)-d_S^2(\mu,\nu) \leq \mathscr E(T_r\boldsymbol\mu)-\mathscr E(\boldsymbol\mu). \label{energy bound}
\end{align}
By \Cref{strong stability variational geodesics}, $(B_{\mu_t^r})_t\to(B_{\mu_t})_t$ strongly in $H^1([0,1];\mathcal H_c)$. Since $M$ is smooth near the population curve by \Cref{ambient inverse change of vars}, it follows that $\boldsymbol\mu^r\to\boldsymbol\mu$ strongly in $H^1([0,1];H^s(X)^*)$.

Recall that the proof of \Cref{geodesic coincidence} establishes the local $C^\infty$ regularity of $\lambda\mapsto A_\lambda$ in operator norm and differentiates the energy $\mathscr E$. 
The proof of \Cref{geodesic coincidence} establishes that the energy $\mathscr E$ admits a continuously Fr\'echet differentiable extension to an $H^1([0,1];H^s(X)^*)$-neighborhood of $\boldsymbol\mu$. It therefore suffices to establish the first-order expansion of the reweighting. For any probability-valued curve $\boldsymbol\eta$, we have
\[
\frac{(T_r\boldsymbol\eta)_t-\eta_t}{r}
=
\frac{b_{r,t}\eta_t-\eta_t(b_{r,t})\eta_t}
{1 + r\eta_t(b_{r,t})},
\quad \text{where}\quad
b_{r,t}=\frac{e^{ru_t}-1}{r}.
\]
By the smoothness of pointwise exponentiation established in \Cref{rkhs algebra calculus}, we have $(b_{r,t})_t\to (u_t)_t$ in $W^{1,\infty}([0,1];H^s(X))$. Moreover, the boundedness of multiplication and dual pairing on $H^s(X)$ and its dual, together with the Banach-valued product rule, imply that these operations extend to bounded bilinear maps from $W^{1,\infty}([0,1];H^s(X))\times H^1([0,1];H^s(X)^*)$ into $H^1([0,1];H^s(X)^*)$ and $H^1([0,1])$, respectively. Thus, whenever $\boldsymbol\eta^r\to\boldsymbol\mu$ strongly in the space $H^1([0,1];H^s(X)^*)$, we obtain $b_r\boldsymbol\eta^r\to u\boldsymbol\mu$ and $\boldsymbol\eta^r(b_r)\to\boldsymbol\mu(u)$ in their respective $H^1$ spaces. Since multiplication $H^1([0,1])\times H^1([0,1];H^s(X)^*)\to H^1([0,1];H^s(X)^*)$ is also bounded bilinear, it follows that
\[
b_r\boldsymbol\eta^r-\boldsymbol\eta^r(b_r)\boldsymbol\eta^r
\to (u-\boldsymbol\mu(u))\boldsymbol\mu
=:\boldsymbol\sigma
\quad\text{in }H^1([0,1];H^s(X)^*).
\]
On the other hand, $\eta_t^r(e^{ru_t})=1+r\eta_t^r(b_{r,t})$, since $\eta_t^r(\mathbbm{1}_X)=1$. Hence the denominators converge to one in $H^1([0,1])$. By the Sobolev embedding $H^1([0,1])\hookrightarrow C([0,1])$ \citep[Theorem 4.12]{adams2003sobolev}, the denominators remain uniformly positive for sufficiently small $|r|$. Continuity of scalar inversion in $H^1([0,1])$ therefore gives $\eta^r(e^{ru})^{-1}\to1$ in $H^1([0,1])$. Combining the convergence of the numerator and denominator yields
\[
\frac{T_r\boldsymbol\eta^r-\boldsymbol\eta^r}{r}
\to\boldsymbol\sigma
\quad\text{in }H^1([0,1];H^s(X)^*),
\qquad
\sigma_t=(u_t-\mu_t(u_t))\mu_t.
\]
Applying this both to $\boldsymbol\eta^r=\boldsymbol\mu$ and to $\boldsymbol\eta^r=\boldsymbol\mu^r$, with $-r$ in the latter case, continuous Fr\'echet differentiability of $\mathscr E$ yields
\begin{align*}
\mathscr E(T_r\boldsymbol\mu)-\mathscr E(\boldsymbol\mu)
=r\operatorname D_{\boldsymbol\sigma}\mathscr E(\boldsymbol\mu)+o(|r|), \quad \text{and}\quad
\mathscr E(\boldsymbol\mu^r)-\mathscr E(T_{-r}\boldsymbol\mu^r)
=r\operatorname D_{\boldsymbol\sigma}\mathscr E(\boldsymbol\mu)+o(|r|).
\end{align*}
Consequently, \Cref{energy bound} gives
\[
d_S^2(\mu^r,\nu^r)-d_S^2(\mu,\nu)
=
r\operatorname D_{\boldsymbol\sigma}\mathscr E(\boldsymbol\mu)+o(|r|).
\]
It remains to identify the energy variation. Recall the operator $A_\mu$ from the proof of \Cref{geodesic coincidence}. Differentiating the energy and applying the Euler--Lagrange identity \eqref{pulled back distributional identity}, together with the regularity established in \Cref{geodesic coincidence}, gives
\begin{align}
\operatorname D_{\boldsymbol\sigma}\mathscr E(\boldsymbol\mu)
&=\frac{\varepsilon}{2}\int_0^1
\left[
\left\langle(\operatorname D_{\sigma_t}A_{\mu_t})\dot\mu_t,\dot\mu_t\right\rangle_{H^s(X)^*}
+2\left\langle A_{\mu_t}\dot\mu_t,\dot\sigma_t\right\rangle_{H^s(X)^*}
\right]dt \notag\\
&=\varepsilon\left[
\left\langle A_{\mu_t}\dot\mu_t,\sigma_t\right\rangle_{H^s(X)^*}
\right]_{t=0}^{t=1} \notag\\
&=2\tilde{\mathbf g}_\nu(\dot\beta_1,\operatorname D_kB_\nu)
-2\tilde{\mathbf g}_\mu(v,\operatorname D_hB_\mu).
\label{energy first variation endpoint identity}
\end{align}
Here the second equality follows by integration by parts, while the final equality follows from the definition of $A_\mu$ and $\tilde{\mathbf g}_\mu$. Substitution into the preceding expansion proves the claim.
\qed{}

\subsection{Proof of \Cref{sinkhorn distance hadamard diff}} \label{proof of sinkhorn distance hadamard diff}
We first establish Hadamard differentiability. By \citet[Theorem~5.2]{lavenant2024riemannian}, $d_S$ is uniformly equivalent to the $\mathcal H_c$ distance in $B$-coordinates. Since $B$ admits a local $C^1$ extension by \Cref{agreement of base derivatives}, there exist neighborhoods of $\mu,\nu$ in $H^s(X)^*$ and $C>0$ such that
\begin{equation}
|d_S^2(\alpha,\zeta)-d_S^2(\alpha',\zeta')|
\leq C\left(
\|\alpha-\alpha'\|_{H^s(X)^*}
+\|\zeta-\zeta'\|_{H^s(X)^*}
\right) \label{sinkhorn distance lipschitz bound}
\end{equation}
for probability measures in these neighborhoods. Next, we note that the reweighting directions considered in \Cref{first variation sinkhorn distance} are dense in $H^s(X)^*_0$. Indeed, the proof of \Cref{statistical riemannian tangent embedding} shows that smooth zero-mass densities $h\,dx$ are dense in this space. Since $\mu=\rho\,dx$ with $\rho\geq\rho_{\min}>0$, we can write $h\,dx=(h/\rho)\mu$, where $\mu(h/\rho)=0$. 
Since $h/\rho\in L^1(\mu)$, the density of $C(X)$ in $L^1(\mu)$ \citep[Theorem~3.14]{rudin1987realcomplex}, together with the density of polynomials in $C(X)$ \citep[Theorem 7.32, Stone-Weierstrauss Theorem]{rudin1976principles}, yields $u_j\in C^\infty(X)$ such that $u_j\to h/\rho$ in $L^1(\mu)$. Since $\int_Xh\,dx=0$ and $H^s(X)\hookrightarrow C(X)$ \citep[Theorem 4.12]{adams2003sobolev}, we have
\begin{align*}
\|(u_j-\mu(u_j))\mu-h\,dx\|_{H^s(X)^*}
&=\sup_{\|\varphi\|_{H^s(X)}\leq 1}
\left|
\mu\left(\left(u_j-\frac h\rho\right)\varphi\right)
-\mu\left(u_j-\frac h\rho\right)\mu(\varphi)
\right|\\
&\leq \sup_{\|\varphi\|_{H^s(X)}\leq 1}
\left[
\left|\mu\left(\left(u_j-\frac h\rho\right)\varphi\right)\right|
+\left|\mu\left(u_j-\frac h\rho\right)\right||\mu(\varphi)|
\right]\\
&\leq \sup_{\|\varphi\|_{H^s(X)}\leq 1}
\left(\|\varphi\|_\infty+|\mu(\varphi)|\right)
\left\|u_j-\frac h\rho\right\|_{L^1(\mu)}\\
&\leq 2\sup_{\|\varphi\|_{H^s(X)}\leq 1}
\|\varphi\|_\infty
\left\|u_j-\frac h\rho\right\|_{L^1(\mu)}\\
&\lesssim 2\left\|u_j-\frac h\rho\right\|_{L^1(\mu)}
\to 0,
\end{align*}
The same argument applies at $\nu$. Now write $\Lambda(\gamma,\eta):=-2\tilde{\mathbf g}_\mu(v,\operatorname D_\gamma B_\mu)+2\tilde{\mathbf g}_\nu(\dot\beta_1,\operatorname D_\eta B_\nu)$ for the expression in the lemma statement, which is bounded and linear. Let $t_n\downarrow0$ and let $(\alpha_n,\zeta_n)\in\mathcal P(X)\times \mathcal P(X)$ satisfy
\[
\frac{\alpha_n-\mu}{t_n}\to\gamma,
\qquad
\frac{\zeta_n-\nu}{t_n}\to\eta
\]
in $H^s(X)^*$. Given a $\delta>0$, choose reweighting directions $h=(u_0-\mu(u_0))\mu$ and $k=(u_1-\nu(u_1))\nu$ such that $\|\gamma-h\|_{H^s(X)^*}+\|\eta-k\|_{H^s(X)^*}<\delta$. Define $\alpha_n'=e^{t_nu_0}\mu/\mu(e^{t_nu_0})$ and $\zeta_n'=e^{t_nu_1}\nu/\nu(e^{t_nu_1})$. 
Since $(e^{t_nu_0}-1)/t_n\to u_0$ uniformly on $X$ and $\mu(e^{t_nu_0})\to1$, we have
\[\frac{\alpha_n'-\mu}{t_n}
=\frac{(e^{t_nu_0}-1)/t_n-\mu((e^{t_nu_0}-1)/t_n)}{\mu(e^{t_nu_0})}\mu\to h\]
in $H^s(X)^*$, and the same argument gives $(\zeta_n'-\nu)/t_n\to k$.
 By \Cref{first variation sinkhorn distance},
\[
\frac{d_S^2(\alpha_n',\zeta_n')-d_S^2(\mu,\nu)}{t_n}
\to\Lambda(h,k).
\]
The preceding Lipschitz bound \eqref{sinkhorn distance lipschitz bound} and boundedness of $\Lambda$ therefore imply
\begin{multline*}
    \left|
\frac{d_S^2(\alpha_n,\zeta_n)-d_S^2(\mu,\nu)}{t_n}
-\Lambda(\gamma,\eta)
\right|\leq
\frac{|d_S^2(\alpha_n,\zeta_n)-d_S^2(\alpha_n',\zeta_n')|}{t_n} \\+
\left|
\frac{d_S^2(\alpha_n',\zeta_n')-d^2_S(\mu,\nu)}{t_n}
-\Lambda(h,k)
\right|
+
|\Lambda(h-\gamma,k-\eta)|.
\end{multline*}
which then gives
\[
\limsup_{n\to\infty}
\left|
\frac{d_S^2(\alpha_n,\zeta_n)-d_S^2(\mu,\nu)}{t_n}
-\Lambda(\gamma,\eta)
\right|
\leq(C+\|\Lambda\|_{\operatorname{op}})\delta.
\]
Since $\delta>0$ is arbitrary, this proves full Hadamard differentiability. The restriction to $\mathbb T_\mu^m\times\mathbb T_\nu^m$ follows from \Cref{statistical riemannian tangent embedding}.

It remains to identify $\Lambda$ with the derivative of the norm of the logarithmic map. Fix $(\gamma,\eta)\in H^s(X)^*_0\times H^s(X)^*_0$ and consider the mass-one perturbations $\alpha^r=\mu+r\gamma$ and $\zeta^r=\nu+r\eta$ in the smooth extensions of \Cref{agreement of base derivatives}. By the implicit-function construction in the proof of \Cref{linearizing logmap}, these determine a $C^1$ family of $\beta$-variable geodesics $\beta^r$ with initial velocities $v^r=\log_{\alpha^r}(\zeta^r)$. Set $\lambda_t^r=M(\beta_t^r)$ and $\sigma_t=\partial_r\lambda_t^r|_{r=0}$. Then $\sigma_0=\gamma$ and $\sigma_1=\eta$. The extended geodesic equations and metric compatibility established in \Cref{ambient metric compatibility} therefore give

\[
\frac{d}{dt}
\left\langle
S_{\lambda_t^r}\dot\beta_t^r,
\dot\beta_t^r
\right\rangle_{\mathcal H_c}
=
-\left\langle
(\operatorname D_{\dot\lambda_t^r}S_{\lambda_t^r})
\dot\beta_t^r,\dot\beta_t^r
\right\rangle_{\mathcal H_c}
+
C_{\lambda_t^r}(\dot\beta_t^r,\dot\beta_t^r)
(\dot\beta_t^r)
=0.
\]
Thus energy satisfies
$\mathscr E(\boldsymbol\lambda^r)
=\tilde{\mathbf g}_{\alpha^r}(v^r,v^r)$.
Differentiating at $r=0$ and applying the Euler--Lagrange identity \eqref{pulled back distributional identity}, as in the proof of \Cref{first variation sinkhorn distance}, yields
\[
\left.\frac{d}{dr}\right|_{r=0}
\tilde{\mathbf g}_{\alpha^r}(v^r,v^r)
=
2\tilde{\mathbf g}_\nu(\dot\beta_1,\operatorname D_\eta B_\nu)
-2\tilde{\mathbf g}_\mu(v,\operatorname D_\gamma B_\mu)
=
\Lambda(\gamma,\eta).
\]
On the other hand, differentiating the initial metric speed using \Cref{operator metric equivalence} and \Cref{linearizing logmap} gives
\[
\left.\frac{d}{dr}\right|_{r=0}
\tilde{\mathbf g}_{\alpha^r}(v^r,v^r)
=
\frac{\varepsilon}{2}
\left\langle
(\operatorname D_\gamma S_\mu)v,v
\right\rangle_{\mathcal H_c}
+
\varepsilon
\left\langle
S_\mu v,
\operatorname D_{(\gamma,\eta)}\log_\mu(\nu)
\right\rangle_{\mathcal H_c}.
\]
Combining these identities establishes the second representation of the derivative and completes the proof.
\qed{}

\subsection{Additional Results for \Cref{sec: clts}}

\begin{prop}\label{norm and distance coincide}
    Under the assumptions of \Cref{geometric clt}, we have $d_S^2(\mu, \nu) = \tilde{\mathbf{g}}_{\mu}(v,v).$
\end{prop}
\begin{proof}
    By definition and by our assumption that $\beta_t^{\mu, \nu}$ is the variational minimizer,
    \[d_S^2(\mu, \nu) = \int_{0}^1 \tilde{\mathbf{g}}_{\mu_t}(\dot \beta_t^{\mu, \nu}, \dot \beta_t^{\mu, \nu})\, dt.\]
    Since $\beta_t^{\mu, \nu}$ satisfies the geodesic equations (\Cref{geodesic coincidence}) we know that $\nabla^{\operatorname{LC}}_{\dot \beta_t}  \dot \beta_t = 0$ for a.e. $t \in [0,1]$, where we write $ \beta_t = \beta_t^{\mu, \nu}$ for shorthand. Then, since $\nabla^{\operatorname{LC}}$ is Levi-Civita (\Cref{levi civita}), metric compatibility reads
    \[\frac{d}{dt}\tilde{\mathbf{g}}_{\mu_t}(\dot \beta_t,\dot \beta_t) = \tilde{\mathbf{g}}_{\mu_t}(\nabla^{\operatorname{LC}}_{\dot \beta_t} \dot \beta_t, \dot \beta_t) + \tilde{\mathbf{g}}_{\mu_t}(\dot \beta_t, \nabla^{\operatorname{LC}}_{\dot \beta_t}\dot \beta_t)  = 0.\]
    Thus, $\tilde{\mathbf{g}}_{\mu_t}(\dot \beta_t,\dot \beta_t) = \tilde{\mathbf{g}}_{\mu}(v, v)$ since $v = \dot \beta_0.$
\end{proof}

\begin{lemma}[Brownian bridge in the statistical tangent space]
\label{brownian bridge concentration}
Let $m>d/2$ and $\mu\in\mathcal P(X)$. The $\mu$-Brownian bridge $\mathbb G_\mu$ in $\ell^\infty(B^m)$ admits a tight, Borel measurable realization in $\mathbb T_\mu^m$, equipped with the $H^m(X)^*$ norm. In particular, under the canonical embedding into $\ell^\infty(B^m)$, $\mathbb G_\mu$ takes values almost surely in a separable subset of $\mathbb T_\mu^m$.
\end{lemma}
\begin{proof}
We proceed with a modified version of the proof of Lemma 17 from \citet{goldfeld2024limit}. Since \(m>d/2\), the continuous Sobolev embedding \(H^m(X)\hookrightarrow C^0(X)\) implies
\[
|\delta_x(f)|=|f(x)|\le \|f\|_{C^0(X)}\lesssim \|f\|_{H^m(X)}
\]
uniformly in \(x\in X\), and hence \(\sup_{x\in X}\|\delta_x\|_{H^m(X)^*}<\infty\). Thus, since $H^m(X)$ is a separable Hilbert space, it admits an orthonormal basis $(e_j)_{j\geq1}$ of $H^m(X)$. Parseval's identity \citep[Theorem~5.9]{brezis2011functional} and the Tonelli--Fubini theorems \citep[Theorems 4.4--4.5]{brezis2011functional} then give
\[
    \sum_{j=1}^{\infty}\operatorname{Var}_\mu(e_j)
    \leq \sum_{j = 1}^\infty\int_X |e_j(x)|^2\, d\mu(x)\leq
    \int_X\sum_{j=1}^{\infty}|e_j(x)|^2\,d\mu(x)
    =
    \int_X\|\delta_x\|_{H^m(X)^*}^2\,d\mu(x)
    <\infty.
\]
Now let $\mathcal F_H=\{f\in H^m(X):\|f\|_{H^m(X)}\leq1\}$ and consider the covariance bilinear form
\[
    \mathcal C_\mu(f,g)
    :=
    \operatorname{Cov}_\mu(f,g),
    \qquad f,g\in H^m(X).
\]
By Cauchy--Schwarz and the continuous embedding $H^m(X)\hookrightarrow C^0(X)$,
\[
    |\mathcal C_\mu(f,g)|
    \leq
    \sqrt{\operatorname{Var}_\mu(f)}
    \sqrt{\operatorname{Var}_\mu(g)}
    \lesssim
    \|f\|_{H^m(X)}\|g\|_{H^m(X)}.
\]
Hence, by the Riesz representation theorem, there exists a bounded self-adjoint operator
$Q_\mu:H^m(X)\to H^m(X)$ satisfying $\langle Q_\mu f,g\rangle_{H^m(X)}
=
    \operatorname{Cov}_\mu(f,g).$ Moreover, the operator $Q_\mu$ is positive since
$\langle Q_\mu f,f\rangle_{H^m(X)}
=\operatorname{Var}_\mu(f)\geq0$, and the preceding estimate gives
\[
    \operatorname{tr}(Q_\mu)
    =
    \sum_{j=1}^\infty
    \langle Q_\mu e_j,e_j\rangle_{H^m(X)}
    =
    \sum_{j=1}^\infty
    \operatorname{Var}_\mu(e_j)
    <\infty.
\]
Thus $Q_\mu$ is positive and trace class. By \citet[Lemma~23]{dashti2017bayesian}, there exists a centered Gaussian random element $Z_\mu$ in $H^m(X)$ with covariance operator $Q_\mu$. Consequently,
\[
    \widetilde{\mathbb G}_\mu(f)
    :=
    \langle Z_\mu,f\rangle_{H^m(X)},
    \qquad f\in\mathcal F_H,
\]
defines a tight $\mu$-Brownian bridge in $\ell^\infty(\mathcal F_H)$, since $\mathbb E[
        \widetilde{\mathbb G}_\mu(f)
        \widetilde{\mathbb G}_\mu(g)
    ]
    =
    \operatorname{Cov}_\mu(f,g).$ In particular, $\mathcal F_H$ is $\mu$-pre-Gaussian. We may therefore apply \citet[Lemma~17]{goldfeld2024limit} with
$\mathcal F=\mathcal F_H$. Writing
\[
    \mathcal M_\mu
    :=
    \left\{
        g\,\mu:
        g \text{ is bounded and measurable and }
        \mu(g)=0
    \right\},
\]
their result gives $\operatorname{spt}(\widetilde{\mathbb G}_\mu) \subseteq
    \overline{\mathcal M_\mu}^{\,\ell^\infty(\mathcal F_H)}.$ For every $g\,\mu\in\mathcal M_\mu$, choose $t>\|g\|_\infty$ and set
$\nu=(1+g/t)\mu$. Then $\nu\in\mathcal P_\mu$ and $g\,\mu=t(\nu-\mu)\in\mathfrak M_\mu.$ Hence $\mathcal M_\mu\subseteq\mathfrak M_\mu$. Moreover, on linear functionals on $H^m(X)$,
\[
    \|\sigma\|_{\ell^\infty(\mathcal F_H)}
    =
    \sup_{\|f\|_{H^m(X)}\leq1}|\sigma(f)|
    =
    \|\sigma\|_{H^m(X)^*}.
\]
It follows that $\operatorname{spt}(\widetilde{\mathbb G}_\mu)\subseteq
    \overline{\mathfrak M_\mu}^{\,H^m(X)^*}
    =
    \mathbb T_\mu^m.$ Thus $\widetilde{\mathbb G}_\mu$ takes values almost surely in $\mathbb T_\mu^m$.

Finally, since $C^m(X)\hookrightarrow H^m(X)$ continuously, restriction to $B^m$ defines a continuous map
$H^m(X)^*\to\ell^\infty(B^m)$. The restriction of
$\widetilde{\mathbb G}_\mu$ to $B^m$ is a centered tight Gaussian process with covariance
$\operatorname{Cov}_\mu(f,g)$, and therefore gives a realization of the
$\mu$-Brownian bridge $\mathbb G_\mu$ in $\ell^\infty(B^m)$. Since
$\mathbb T_\mu^m$ is a closed subspace of the separable Hilbert space
$H^m(X)^*$, it is separable, and so its image under the canonical embedding into
$\ell^\infty(B^m)$ is separable as well.
\end{proof}

\begin{lemma}[Stability of variational geodesics]
\label{stability variational geodesics}
Suppose $(\mu_t)_{t\in[0,1]}$ is the unique constant-speed minimizer of the variational problem in \Cref{variational geodesic} among admissible probability-measure-valued curves joining $\mu$ and $\nu$. Let $\mu_n\rightharpoonup\mu$ and $\nu_n\rightharpoonup\nu$, and let $(\mu_t^n)_{t\in[0,1]}$ be any constant-speed minimizer of the same variational problem joining $\mu_n$ and $\nu_n$. Then
\[
    \sup_{t\in[0,1]}d_S(\mu_t^n,\mu_t)\to0.
\]
\end{lemma}

\begin{proof}
Since $d_S$ metrizes weak convergence, the triangle inequality gives
\[
|d_S(\mu_n,\nu_n)-d_S(\mu,\nu)|
\leq d_S(\mu_n,\mu)+d_S(\nu_n,\nu)\to0.
\]
Since $(\mu_t^n)_{t\in[0,1]}$ is a variational minimizing geodesic,
$\mathscr E(\boldsymbol\mu^n)=d_S^2(\mu_n,\nu_n)\to d_S^2(\mu,\nu)<\infty$.
Consequently, $\sup_n\mathscr E(\boldsymbol\mu^n)<\infty$. By \citet[Theorem~2.11]{lavenant2026approximation}, every subsequence admits a further subsequence converging uniformly in $d_S$ to a limiting curve $(\tilde\mu_t)_{t\in[0,1]}$. Since
$d_S(\mu_s^n,\mu_t^n)=|t-s|d_S(\mu_n,\nu_n)$,
passing to the limit gives
$d_S(\tilde\mu_s,\tilde\mu_t)=|t-s|d_S(\mu,\nu)$
for every $s,t\in[0,1]$. Moreover, $\tilde\mu_0=\mu$ and $\tilde\mu_1=\nu$, so $(\tilde\mu_t)_{t\in[0,1]}$ is a constant-speed minimizing metric geodesic joining $\mu$ and $\nu$. 
As noted by \citet[Remark 5.5]{lavenant2024riemannian}, a metric geodesic is not necessarily known to minimize the variational energy, so we must establish this property before invoking uniqueness of the variational minimizer. In particular, by the weak $H^1([0,1];\mathcal H_c)$ convergence in \citet[Theorem 2.11]{lavenant2026approximation} and the weak lower semicontinuity of the energy \citep[Theorem 3.4]{lavenant2026approximation}, we have
\[
d_S^2(\mu,\nu)
\leq\mathscr E(\tilde{\boldsymbol\mu})
\leq\liminf_{k\to\infty}\mathscr E(\boldsymbol\mu^{n_k})
=d_S^2(\mu,\nu).
\]
Hence $(\tilde\mu_t)_{t\in[0,1]}$ is a variational minimizer joining $\mu$ and $\nu$. Uniqueness identifies it with $(\mu_t)_{t\in[0,1]}$, and the conclusion follows.
\end{proof}

\begin{lemma}[Strong stability of variational geodesics]
\label{strong stability variational geodesics}
Under the assumptions of \Cref{stability variational geodesics}, write $\beta_t^n=B_{\mu_t^n}$ and $\beta_t=B_{\mu_t}$. Then $\beta^n\to\beta$ strongly in $H^1([0,1];\mathcal H_c)$.
\end{lemma}
\begin{proof}
By \Cref{stability variational geodesics} and \citet[Theorem~2.11]{lavenant2026approximation}, we have $\beta^n\to\beta$ strongly in $C([0,1];\mathcal H_c)$ and $\dot\beta^n\overset{w}{\rightharpoonup}\dot\beta$ weakly in $L^2([0,1];\mathcal H_c)$. Moreover, \citet[Proposition~4.7]{lavenant2026approximation} and uniform convergence in $d_S$ imply
\[
    \sup_{t\in[0,1]}
    \|S_{\mu_t^n}-S_{\mu_t}\|_{\operatorname{op}}
    \to0.
\]
Since the minimizing energies converge and the metric tensor is uniformly coercive by \citet[Equation~(2.11)]{lavenant2026approximation}, the sequence $(\dot\beta^n)$ is bounded in $L^2([0,1];\mathcal H_c)$. Consequently,
\begin{align*}
    \int_0^1\langle S_{\mu_t}\dot\beta_t^n,\dot\beta_t^n\rangle_{\mathcal H_c}\,dt
    &=
    \int_0^1\langle S_{\mu_t^n}\dot\beta_t^n,\dot\beta_t^n\rangle_{\mathcal H_c}\,dt+o(1)\\
    &=\frac{2}{\varepsilon}\mathscr E(\boldsymbol\mu^n)+o(1)\\
    &\to\frac{2}{\varepsilon}\mathscr E(\boldsymbol\mu)
    =\int_0^1\langle S_{\mu_t}\dot\beta_t,\dot\beta_t\rangle_{\mathcal H_c}\,dt.
\end{align*}
Since $S_{\mu_t}$ is bounded by construction, weak convergence of $\dot\beta^n$ and self adjointness of $S_{\mu_t}$ (\Cref{operator metric equivalence}) give
\[
    \int_0^1\langle S_{\mu_t}\dot\beta_t^n,\dot\beta_t\rangle_{\mathcal H_c}\,dt
    \to
    \int_0^1\langle S_{\mu_t}\dot\beta_t,\dot\beta_t\rangle_{\mathcal H_c}\,dt.
\]
Combining these identities yields
\[
    \int_0^1
    \langle S_{\mu_t}(\dot\beta_t^n-\dot\beta_t),
    \dot\beta_t^n-\dot\beta_t\rangle_{\mathcal H_c}\,dt
    \to0.
\]
By \Cref{S tangent invariance,ambient self adjointness}, $S_{\mu_t}$ is self-adjoint, preserves $B_{\mu_t}^{\perp}$, and acts as the identity on $\operatorname{span}\{B_{\mu_t}\}$. Consequently, the tangent coercivity bound of \citet[Proposition 4.15]{lavenant2024riemannian} extends to all of $\mathcal H_c$ by orthogonal decomposition. Together with the strong convergence of $\beta^n$ in $C([0,1];\mathcal H_c)$, this proves the claim.
\end{proof}

\subsection{Supplemental Results}

\begin{lemma}[Smoothness of $\mathbf J_\beta$]
\label{smoothness of J beta}
The map $\beta\mapsto\mathbf J_\beta$ is $C^\infty$ from $V_+$ into $\mathcal L(\mathcal H_c,\mathcal H_c)$.
\end{lemma}

\begin{proof}
For $g\in\mathcal H_c$, define the multiplication operator $\mathsf L_g\in\mathcal L(\mathcal H_c,\mathcal H_c)$ by $\mathsf L_g u=gu$. Since $\mathcal H_c$ is a Banach algebra (\Cref{rkhs algebra calculus}), we have $\|\mathsf L_g u\|_{\mathcal H_c}\lesssim\|g\|_{\mathcal H_c}\|u\|_{\mathcal H_c}$, so $g\mapsto\mathsf L_g$ is bounded linear. Similarly, for $\lambda\in\mathcal H_c^*$, define $\mathsf Q_\lambda\in\mathcal L(\mathcal H_c,\mathcal H_c^*)$ by $\mathsf Q_\lambda(g)=g\lambda$. By the boundedness of dual multiplication established in the proof of \Cref{rkhs algebra calculus}, we have
\[
\|\mathsf Q_\lambda g\|_{\mathcal H_c^*}
\lesssim\|g\|_{\mathcal H_c}\|\lambda\|_{\mathcal H_c^*},
\]
so $\lambda\mapsto\mathsf Q_\lambda$ is bounded linear. By definition,
\[
\mathbf J_\beta
=
H_c\circ\mathsf Q_{M(\beta)}
\circ\mathsf L_{a_\beta^2}\circ\Pi_\beta.
\]
By \Cref{ambient basic smoothness}, the maps $\beta\mapsto a_\beta=\beta^{-1}$ and $\beta\mapsto\Pi_\beta$ are $C^\infty$ on $V_+$, and $\beta\mapsto M(\beta)$ is a smooth quadratic map. Since pointwise multiplication is bounded bilinear on $\mathcal H_c$, the map $\beta\mapsto a_\beta^2$ is also $C^\infty$. Consequently, $\beta\mapsto\mathsf L_{a_\beta^2}$ and $\beta\mapsto\mathsf Q_{M(\beta)}$ are $C^\infty$ in operator norm. Finally, $H_c:\mathcal H_c^*\to\mathcal H_c$ is bounded linear and operator composition is bounded multilinear. Therefore $\beta\mapsto\mathbf J_\beta$ is $C^\infty$ in operator norm.
\end{proof}

\begin{lemma}[Smoothness of $\mathbf K_\beta$]
\label{smoothness of K beta}
The map $\beta\mapsto\mathbf K_\beta$ is $C^\infty$ from $V_+$ into $\mathcal L(\mathcal H_c,\mathcal H_c)$. Moreover, $\mathbf K_\beta=\mathbf J_\beta$ for every $\beta\in V_+$.
\end{lemma}

\begin{proof}
For $\beta\in V_+$ and $u\in\mathcal H_c$, the reproducing property gives
\[
\langle\mathbf J_\beta u,\beta\rangle_{\mathcal H_c}
=M(\beta)(a_\beta\Pi_\beta u)
=\langle\Pi_\beta u,\beta\rangle_{\mathcal H_c}=0.
\]
Thus $\Pi_\beta\mathbf J_\beta=\mathbf J_\beta$, and hence $\mathbf K_\beta=\Pi_\beta\mathbf J_\beta=\mathbf J_\beta$. The asserted smoothness therefore follows directly from \Cref{smoothness of J beta}.
\end{proof}

\begin{lemma}[Smoothness of $\mathbf R_\beta$ and $C_\beta(u,v)^\sharp$]
\label{smoothness of C beta}
The map $\beta\mapsto\mathbf R_\beta$ is $C^\infty$ from $U_+$ into $\mathcal L(\mathcal H_c,\mathcal H_c)$, and the map $(\beta,u,v)\mapsto C_\beta(u,v)^\sharp$ is $C^\infty$ from $U_+\times\mathcal H_c\times\mathcal H_c$ into $\mathcal H_c$.
\end{lemma}

\begin{proof}
By \Cref{smoothness of K beta,ambient basic smoothness}, the maps $\beta\mapsto\mathbf K_\beta$ and $\beta\mapsto\Pi_\beta$ are $C^\infty$ in operator norm. Since $\mathbf K_\beta=\mathbf J_\beta$ and $\|\mathbf J_\beta\|_{\operatorname{op}}<1$ for every $\beta\in U_+$, the operator $\operatorname{id}_{\mathcal H_c}-\mathbf K_\beta$ is boundedly invertible. Moreover, operator inversion is $C^\infty$ on the open set of boundedly invertible operators: its Fr\'echet derivative is $\operatorname D_H\operatorname{inv}(A)=-A^{-1}HA^{-1}$ by the argument in \Cref{tilde R hadamard diff}, and bounded bilinearity of operator composition allows this formula to be differentiated repeatedly. Therefore
\[
\mathbf R_\beta
=(\operatorname{id}_{\mathcal H_c}-\mathbf K_\beta)^{-1}\Pi_\beta
\]
depends $C^\infty$-smoothly on $\beta\in U_+$ in operator norm. Recall from \Cref{Cmu riesz rep} that the Riesz representative admits the extension
\[
C_\beta(u,v)^\sharp
=
2\left(
a_\beta(\mathbf R_\beta u)(\mathbf R_\beta v)
-
H_c\left[
a_\beta^3(\mathbf R_\beta u)(\mathbf R_\beta v)M(\beta)
\right]
\right).
\]
Since $\beta\mapsto\mathbf R_\beta$ is $C^\infty$ in operator norm, and the evaluation map $(A,u)\mapsto Au$ is bounded bilinear, both $(\beta,u)\mapsto\mathbf R_\beta u$ and $(\beta,v)\mapsto\mathbf R_\beta v$ are $C^\infty$. Moreover, $\beta\mapsto a_\beta$ is $C^\infty$ on $V_+$ by \Cref{ambient basic smoothness}. By the Banach algebra property of $\mathcal H_c$, pointwise multiplication is bounded multilinear. Consequently, the maps
\[
(\beta,u,v)\mapsto
a_\beta(\mathbf R_\beta u)(\mathbf R_\beta v),
\qquad
(\beta,u,v)\mapsto
a_\beta^3(\mathbf R_\beta u)(\mathbf R_\beta v)
\]
are $C^\infty$ into $\mathcal H_c$. Since $\beta\mapsto M(\beta)$ is $C^\infty$ into $\mathcal H_c^*$ and multiplication $\mathcal H_c\times\mathcal H_c^*\to\mathcal H_c^*$ is bounded bilinear, the map
\[
(\beta,u,v)\mapsto
a_\beta^3(\mathbf R_\beta u)(\mathbf R_\beta v)M(\beta)
\]
is $C^\infty$ into $\mathcal H_c^*$. Finally, $H_c:\mathcal H_c^*\to\mathcal H_c$ is bounded linear, so both terms defining $C_\beta(u,v)^\sharp$ are $C^\infty$ into $\mathcal H_c$, as desired.
\end{proof}

\begin{lemma} \label{coordinate chart regularity}
    The change of variables maps $\psi$ and $\Psi$ are diffeomorphisms onto their images. Moreover, their Fr\'echet derivatives are bounded linear isomorphisms at every point of their respective domains.
\end{lemma}

\begin{proof}
    Recall that $V_R=
        \left\{
        q\in\mathfrak{T}_\mu\mathcal{P}(X):
        \|q\|_{\mathcal{H}_c}<R<1
        \right\}$ and the sphere chart $\psi$ is given by
    \[
        \psi(q)
        =
        B_\mu\sqrt{1-\|q\|_{\mathcal{H}_c}^2}+q.
    \]
    Since $R<1$, the map
    $q\mapsto(1-\|q\|_{\mathcal{H}_c}^2)^{1/2}$ is smooth on $V_R$, rendering $\psi$ a smooth map. Moreover, since
    $\Pi_\mu B_\mu=0$ and $\Pi_\mu q=q$ for
    $q\in\mathfrak{T}_\mu\mathcal{P}(X)$, $\Pi_\mu\psi(q)=q.$ Thus the inverse of $\psi$ on $\psi(V_R)$ is simply
    $\psi^{-1}(\beta)=\Pi_\mu\beta$, which is the restriction of a bounded
    linear operator and is therefore smooth. Hence $\psi$ is a
    diffeomorphism onto its image.

    Now we will compute the Fr\'echet derivative of $\psi$. Fix $q\in V_R$ and $\gamma\in\mathfrak{T}_\mu\mathcal{P}(X)$. Since $B_\mu$ is fixed with respect to $q$, it suffices to differentiate the scalar-valued map $q\mapsto (1-\|q\|_{\mathcal{H}_c}^2)^{1/2}$. First, by the bilinearity of the inner product,
    \[
        \operatorname{D}_\gamma\|q\|_{\mathcal{H}_c}^2
        =
        \operatorname{D}_\gamma\langle q,q\rangle_{\mathcal{H}_c}
        =
        2\langle q,\gamma\rangle_{\mathcal{H}_c}.
    \]
    Applying the chain rule to the map $s\mapsto\sqrt{s}$ therefore gives
    \[
        \operatorname{D}_\gamma
        \sqrt{1-\|q\|_{\mathcal{H}_c}^2}
        =
        -\frac{\langle q,\gamma\rangle_{\mathcal{H}_c}}
        {\sqrt{1-\|q\|_{\mathcal{H}_c}^2}}.
    \]
    Since the Fr\'echet derivative of $q\mapsto q$ in the direction $\gamma$ is simply $\gamma$, we conclude that
    \begin{equation}\label{chart differential}
        \operatorname{D}_\gamma\psi(q)
        =
        \gamma
        -
        \frac{\langle q,\gamma\rangle_{\mathcal{H}_c}}
        {\sqrt{1-\|q\|_{\mathcal{H}_c}^2}}
        B_\mu.
    \end{equation}
    The denominator is strictly positive since $q\in V_R$ and $R<1$. In particular, at the center of the chart, $\operatorname{D}_\gamma\psi(0)=\gamma$. Observe that the Fr\'echet derivative is a bounded linear map from
    $\mathfrak{T}_\mu\mathcal{P}(X)$ to
    $T_{\psi(q)}\mathbb S(\mathcal H_c)=\psi(q)^\perp$, and its inverse is simply the restriction $\Pi_\mu|_{\psi(q)^\perp}$. To see this, firstly note that we have $\Pi_\mu\operatorname{D}_\gamma\psi(q)=\gamma$. Conversely, if $h\in\psi(q)^\perp$, write
    $h=\Pi_\mu h+\lambda B_\mu$; taking the inner product of this expression with $\psi(q)$ and rearranging to solve for $\lambda$ yields
    \begin{equation*}
        \lambda
        =-\frac{\langle \Pi_\mu h, \psi(q)\rangle_{\mathcal{H}_c}}{\langle B_\mu, \psi(q)\rangle_{\mathcal{H}_c}} = -\frac{\langle \Pi_\mu h, B_\mu(1-\|q\|_{\mathcal{H}_c}^2)^{1/2}+q\rangle_{\mathcal{H}_c}}{\langle B_\mu, B_\mu(1-\|q\|_{\mathcal{H}_c}^2)^{1/2}+q\rangle_{\mathcal{H}_c}}=
        -
        \frac{
        \langle q,\Pi_\mu h\rangle_{\mathcal{H}_c}
        }{
        \sqrt{1-\|q\|_{\mathcal{H}_c}^2}
        }
    \end{equation*}
    Then, using the formula of \Cref{chart differential} and taking $\gamma = \Pi_\mu h$ yields
    $\operatorname{D}_{\Pi_\mu h}\psi(q)=\Pi_{\mu}h + \lambda B_\mu = h$. Thus, since $\gamma \mapsto \operatorname{D}_\gamma\psi(q)$ is a bounded linear bijection between two Banach spaces, the bounded inverse theorem \citep[Corollary 2.12]{rudin1991functional} implies that its inverse is continuous and thus bounded. Thus the Fr\'echet derivative is a bounded linear isomorphism. Now recall that the induced tangent-bundle chart is $\Psi(\beta,u) = (\Pi_\mu\beta,\Pi_\mu u).$ If $(p,w)\in
    V_R\times\mathfrak{T}_\mu\mathcal{P}(X)$, then
    \[
        \Psi^{-1}(p,w)
        =
        \left(
        \psi(p),
        \operatorname{D}_w\psi(p)
        \right).
    \]
    Indeed,
    $\operatorname{D}_w\psi(p)\in
    T_{\psi(p)}\mathbb S(\mathcal H_c)=\psi(p)^\perp$ and
    $\Pi_\mu\operatorname{D}_w\psi(p)=w$.
    Since $\psi$ is smooth, the map
    $(p,w)\mapsto\operatorname{D}_w\psi(p)$ is smooth, and hence
    $\Psi^{-1}$ is smooth. The map $\Psi$ itself is the restriction of
    the bounded linear map $(\beta,u)\mapsto (\Pi_\mu\beta,\Pi_\mu u),$ and is therefore smooth. Consequently, $\Psi$ is a diffeomorphism
    onto its image.

    Since \(\psi\) and \(\Psi\) are diffeomorphisms, their inverses are Fréchet differentiable. Differentiating both identities \(\psi^{-1}\circ\psi=\operatorname{id}\) and \(\psi\circ\psi^{-1}=\operatorname{id}\) gives
    \[ \operatorname{D}_{(\cdot)}\psi^{-1}(\psi(q))\circ \operatorname{D}_{(\cdot)}\psi(q)=\operatorname{id}, \quad \text{and} \quad \operatorname{D}_{(\cdot)}\psi(q)\circ \operatorname{D}_{(\cdot)}\psi^{-1}(\psi(q))=\operatorname{id}. \]
    Hence \(\gamma \mapsto \operatorname{D}_\gamma\psi(q)\) is invertible with inverse \(\gamma \mapsto \operatorname{D}_{\gamma}\psi^{-1}(\psi(q))\). Since Fréchet derivatives are bounded linear operators, \(\gamma \mapsto \operatorname{D}_\gamma\psi(q)\) is a bounded linear isomorphism. The same argument applied to \(\Psi\) yields the corresponding conclusion for \(\operatorname{D}_{(\cdot)}\Psi(\beta,u)\).
\end{proof}

\begin{lemma} \label{bounded linearity}
    The map $(\gamma, a) \mapsto \operatorname{D}_{(\gamma, a)}\exp_{\mu}(v)$ is a bounded linear map in $(\gamma, a).$
\end{lemma}
\begin{proof}
     On each interval $[s_{i-1},s_i]$, the map $
(\mathfrak p_{s_{i-1}}^{i},\mathfrak u_{s_{i-1}}^{\,i})
\mapsto
(\mathfrak p_t^{i},\mathfrak u_t^{i})$ given by $
\operatorname D_{2}\alpha_i(t-s_{i-1},x_i)$ is bounded and linear since $\alpha_i$ is $C^1$. Likewise, the differentials of the transition maps $\Psi_{i+1}\circ\Psi_i^{-1}$ and the chart inverses $\Psi_i^{-1}$ are bounded linear. Since only finitely many charts are required, their composition shows that $(h_0,\dot h_0)
\mapsto 
((h_t,\dot h_t))_{t\in[0,1]}$ is bounded and linear as a map into $C^0([0,1];\mathcal H_c\times\mathcal H_c)$. Equivalently, $(h_0,\dot h_0)\mapsto (h_t)_{t\in[0,1]}$ is bounded and linear into $C^1([0,1];\mathcal H_c)$. Finally, since $(\gamma,a)\mapsto(\operatorname D_\gamma B_\mu,a)$ is bounded linear, the derivative $
(\gamma,a)\mapsto
\operatorname D_{(\gamma,a)}\exp_\mu(v)
=
(h_t)_{t\in[0,1]}$ is a bounded linear map on $\mathbb T_{(\mu,v)}\mathfrak D_{\exp}$, and hence the exponential map is fully Hadamard differentiable.
\end{proof}

\begin{lemma}[Linearity of the parallel imputation Hadamard derivative]
\label{linearity of parallel imputation variation}
Suppose that the assumptions of \Cref{linearizing imp} hold, and let
$k_t$ be the variation of the parallel field induced by
$(\gamma,\eta,\zeta)\in
\mathbb T_\mu^m\times\mathbb T_\nu^m\times\mathbb T_\rho^m$.
Then the map
\[
    (\gamma,\eta,\zeta)
    \to
    (k_t)_{t\in[0,1]}
\]
is bounded and linear as a map into
$C^1([0,1];\mathcal H_c)$. In particular,
$(\gamma,\eta,\zeta)\mapsto k_1$ is bounded and linear. Consequently,
writing $w=\operatorname{PT}_{\mu\to\nu}(\log_\mu(\rho))$, the map
\[
    (\gamma,\eta,\zeta)
    \mapsto
    \operatorname D_{(\eta,k_1)}\exp_\nu(w)
\]
is bounded and linear.
\end{lemma}

\begin{proof}
By \Cref{hadamard deriv of change of var} and
\Cref{linearizing logmap}, the map
\[
    (\gamma,\eta,\zeta)
    \to
    (h_0,\dot h_0,k_0)
    =
    \left(
        \operatorname D_\gamma B_\mu,
        \operatorname D_{(\gamma,\eta)}\log_\mu(\nu),
        \operatorname D_{(\gamma,\zeta)}\log_\mu(\rho)
    \right)
\]
is bounded and linear. Since $\Psi_1$ and
$G_{\operatorname{PT}}$ are $C^1$, their differentials are bounded
linear operators. Projecting the differential of
$G_{\operatorname{PT}}$ onto its third component therefore shows that
$(\gamma,\eta,\zeta)\mapsto(k_t)_{t\in[0,1]}$ is bounded and linear.
Since evaluation at $t=1$ is bounded and linear, the same holds for
$(\gamma,\eta,\zeta)\mapsto k_1$.

Finally, by \Cref{linearizing expmap}, the differential of the
exponential map is bounded and linear in its joint perturbation.
Since $(\gamma,\eta,\zeta)\mapsto(\eta,k_1)$ is bounded and linear, the
claimed bounded linearity of the parallel imputation variation follows
by composition.
\end{proof}

\section{Miscellaneous Results}

\begin{lemma}[Duality of $\hookrightarrow$ for Sobolev spaces] \label{duality of continuous embedding}
    Let $X\subset\mathbb R^d$ be a closed Euclidean ball of positive radius and let $m\in\mathbb N$. Then $C^m(X) \hookrightarrow H^m(X)$ densely, and $H^m(X)^* \hookrightarrow C^m(X)^*.$
    Moreover, if $s>m+d/2$, then
    $H^s(X)\hookrightarrow C^m(X)$ continuously
    and densely. Consequently,
    \[
        H^m(X)^*\hookrightarrow C^m(X)^*
        \hookrightarrow H^s(X)^*.
    \]
\end{lemma}
\begin{proof}
    Define the inclusion operator $i: C^m(X) \to H^m(X)$ given by $i(f) = f.$ Definitionally, we have
    \[\|f\|_{H^m(X)} = \left(\sum_{|\alpha| \leq m}\|\partial^\alpha f\|_{L^2(X)}^2\right)^{1/2} \leq \sum_{|\alpha| \leq m}\|\partial^\alpha f\|_{L^2(X)} \lesssim_X \sum_{|\alpha| \leq m}\|\partial^\alpha f\|_{L^\infty(X)} = \|f\|_{C^m(X)}.\]
    Thus, $i$ is clearly a bounded -- hence, continuous -- linear operator. Since $i$ is linear, establishing injectivity can be achieved by showing that $i(f) = 0 \implies f = 0$. Since $i(f) = f \in H^m(X)$ and equality in $H^m(X)$ is defined a.e., we know that $i(f) = 0$ implies $f = 0$ Lebesgue a.e. on $X$. Since $f$ is continuous and is Lebesgue a.e. $0$, we know $f$ must also be zero everywhere. This proves injectivity of $i$. Thus, $C^m(X) \hookrightarrow H^m(X).$ For the density of the image of $i$, see \citet[Theorem 3.17]{adams2003sobolev}. 
    
    Now for the reverse embedding, we can define the dual inclusion map $i^*: H^m(X)^* \to C^m(X)^*$ as $i^*(\sigma) = \sigma$. Definitionally, for any $\sigma \in H^m(X)^*$ we have
    \begin{align*}
        \|\sigma\|_{C^m(X)^*} = \sup_{\|f\|_{C^m(X)} \leq 1}|\sigma(f)| \lesssim_X \sup_{\|f\|_{H^m(X)} \leq 1}|\sigma(f)| = \|\sigma\|_{H^m(X)^*}.
    \end{align*}
    Thus $i^*$ is a bounded linear operator. Again, one can show injectivity by showing that $i^*(\sigma) = 0 \implies \sigma = 0$. To see why this holds, note that $0 = i^*(\sigma) = \sigma$ in $C^m(X)^*.$ We will now show that $\sigma = 0$ in $C^m(X)^*$ implies $\sigma = 0$ in $H^m(X)^*$. Let $f \in H^m(X)$ be arbitrary and let $\{f_n\} \subset C^m(X)$ be a sequence of functions converging to $f$ in $H^m(X)$ (which we know exists by density). Then $\sigma(f_n) = i^*(\sigma)(f_n) = 0$ for all $n$ since $\sigma = 0$ in $C^m(X)^*.$ Now by boundedness (and thus continuity) of $\sigma$,
    \[ 0 = \lim_{n \to \infty}\sigma(f_n) =  \sigma(\lim_{n \to \infty}f_n) = \sigma(f).\]
    Since $f \in H^m(X)$ was arbitrary, we conclude $\sigma = 0$ in $H^m(X)^*$. This completes the proof of injectivity. 
    
    Finally, since $s>m+d/2$, the Sobolev embedding
    theorem \citep[Theorem 4.12]{adams2003sobolev} gives $H^s(X)\hookrightarrow C^m(X)$
    continuously. 
    Moreover, since $X$ is a closed Euclidean ball, Whitney's extension and approximation results \citep[Theorem I and Lemma 5]{whitney1934analytic} imply that restrictions of $C^\infty(\mathbb R^d)$ functions are dense in $C^m(X)$ in the $C^m$ norm. Since these restrictions belong to $H^s(X)$, the inclusion $H^s(X)\hookrightarrow C^m(X)$ has dense image. Duality then yields the
    continuous injection
    $C^m(X)^*\hookrightarrow H^s(X)^*$.

\end{proof}

\begin{lemma}[Multiplication and pointwise exponentiation]
\label{rkhs algebra calculus}
Pointwise multiplication $\mathcal H_c\times\mathcal H_c\to\mathcal H_c$ and dual multiplication $\mathcal H_c\times\mathcal H_c^*\to\mathcal H_c^*$, defined by $(g\lambda)(f)=\lambda(gf)$, are bounded bilinear maps. Moreover, pointwise exponentiation $f\mapsto e^{f/\varepsilon}$ is $C^\infty$ from $\mathcal H_c$ to itself, i.e. $(f \mapsto e^{f/\varepsilon}) \in C^{\infty}(\mathcal{H}_c; \mathcal{H}_c)$; its $k$th derivative at $f$ maps $(h_1,\dots,h_k)$ to $\varepsilon^{-k}e^{f/\varepsilon}\prod_{i=1}^k h_i$ for every $k\geq1$. The same assertions hold with $H^s(X)$ and $H^s(X)^*$ in place of $\mathcal H_c$ and $\mathcal H_c^*$.
\end{lemma}

\begin{proof}
By \Cref{RKHS norm equivalence} and the Sobolev multiplication theorem \citep[Theorem 4.39]{adams2003sobolev}, there exists $C_A\geq1$ such that $\|fg\|_{\mathcal H_c}\leq C_A\|f\|_{\mathcal H_c}\|g\|_{\mathcal H_c}$. Consequently,
\[
\|g\lambda\|_{\mathcal H_c^*}
=\sup_{\|f\|_{\mathcal H_c}\leq1}|\lambda(gf)|
\leq C_A\|g\|_{\mathcal H_c}\|\lambda\|_{\mathcal H_c^*}.
\]
This proves the bounded bilinearity of dual multiplication. For pointwise exponentiation, since $\mathbbm{1}_X\in H^s(X)\simeq\mathcal H_c$ and $\|f^j\|_{\mathcal H_c}\leq C_A^{j-1}\|f\|_{\mathcal H_c}^j$ for $j\geq1$, the exponential series $e^{f/\varepsilon}=\sum_{j=0}^\infty f^j/(\varepsilon^j j!)$ converges absolutely in $\mathcal H_c$. Indeed, for every $R>0$ and $\|f\|_{\mathcal H_c}\leq R$, we have 
\[
\left\|\frac{f^j}{\varepsilon^j j!}\right\|_{\mathcal H_c} =\frac{1}{\varepsilon^j j!}\|f^j\|_{\mathcal H_c} \le\frac{C_A^{j-1}}{\varepsilon^j j!}
\|f\|_{\mathcal H_c}^j \le\frac{C_A^{j-1}R^j}{\varepsilon^j j!} =\frac{1}{C_A}\frac{(C_AR/\varepsilon)^j}{j!}.
\]
Summing then gives
\[
\sum_{j=1}^\infty
\left\|\frac{f^j}{\varepsilon^j j!}\right\|_{\mathcal H_c}
\le \sum_{j=1}^\infty
\frac{1}{C_A}\frac{(C_AR/\varepsilon)^j}{j!}
=\frac{1}{C_A}
\sum_{j=1}^\infty\frac{(C_AR/\varepsilon)^j}{j!}
=\frac{1}{C_A}
\left(e^{C_AR/\varepsilon}-1\right)<\infty.
\]
Moreover, for every $N\geq1$, the triangle inequality gives
\[
\sup_{\|f\|_{\mathcal H_c}\leq R}
\left\|\sum_{j=N}^\infty\frac{f^j}{\varepsilon^j j!}\right\|_{\mathcal H_c}
\leq\frac{1}{C_A}\sum_{j=N}^\infty\frac{(C_AR/\varepsilon)^j}{j!}\to0
\]
as $N\to\infty$, since the right-hand side is the tail of a convergent exponential series. Thus, by the Weierstrass $M$-test for Banach-space-valued series \citep[Proposition 28.1.18]{erdman2014problemtext}, the exponential series converges uniformly on every bounded subset of $\mathcal H_c$.
Moreover, for every $k\geq1$ and $j\geq k$, repeated differentiation gives
\[
\operatorname D^k\left(\frac{f^j}{\varepsilon^j j!}\right)[h_1,\ldots,h_k]
=\frac{f^{j-k}}{\varepsilon^j(j-k)!}\prod_{i=1}^k h_i.
\]
By the same multiplication bound, whenever $\|f\|_{\mathcal H_c}\leq R$,
\[
\left\|\operatorname D^k\left(\frac{f^j}{\varepsilon^j j!}\right)\right\|_{\operatorname{op}}
\leq \frac{C_A^{k-1}}{\varepsilon^k}
\frac{(C_AR/\varepsilon)^{j-k}}{(j-k)!}.
\]
Consequently, the series of $k$th derivatives converges uniformly on bounded subsets in the norm of bounded $k$-linear operators, since its terms are dominated by a convergent exponential series. As this holds for every $k\geq1$, termwise differentiation is justified to all orders, proving that pointwise exponentiation is $C^\infty$ with
$\operatorname D^k(e^{f/\varepsilon})[h_1,\ldots,h_k]
=\varepsilon^{-k}e^{f/\varepsilon}\prod_{i=1}^k h_i$. Hence pointwise exponentiation is $C^\infty$ with the stated derivatives. The assertions for $H^s(X)$ and its dual follow from norm equivalence. 
\end{proof}

\begin{lemma}[Sobolev regularity of reciprocals and logarithms]
\label{sobolev reciprocal and logarithm}
    Let $X\subset\mathbb R^d$ be a closed Euclidean ball and let $s\in\mathbb N$ satisfy $s>d/2$. If $u\in H^s(X)$ satisfies $u\geq\delta$ on $X$ for some $\delta>0$, then $1/u,\log u\in H^s(X)$.
\end{lemma}
\begin{proof}
    Since $H^s(X)\hookrightarrow C(X)$ continuously, the range of $u$ is contained in a compact interval $[\delta,M]\subset(0,\infty)$. The conclusion follows from the Sobolev composition theorem \citep[Lemma 2.1]{avalos2026sobolev}, applied to the smooth functions $t\mapsto 1/t$ and $t\mapsto\log t$ on an open interval containing $[\delta,M]$.
\end{proof}

\subsection{Proof of \Cref{continuous embedding}} \label{proof of continuous embedding}

The assumption $c\in C^{(m,m)}(X\times X)$ implies that
$k_c=\exp(-c/\varepsilon)\in C^{(m,m)}(X\times X)$. Therefore,
\citet[Lemma~B.1]{lavenant2024riemannian} gives, for every
$f\in\mathcal H_c$,
\[
    \|f\|_{C^m(X)}
    \leq
    \|f\|_{\mathcal H_c}
    \|k_c\|_{C^{(m,m)}(X\times X)}^{1/2}.
\]
Since $\mathcal H_c$ is an RKHS of functions on $X$, the canonical
inclusion $\mathcal H_c\to C^m(X)$ is injective, and hence
$\mathcal H_c\hookrightarrow C^m(X)$ continuously. For the self-transport kernel, \citet[Proposition~3.7]{lavenant2024riemannian}
shows that $f_{\mu,\mu}\in C^m(X)$. Consequently,
$a_\mu=\exp(f_{\mu,\mu}/\varepsilon)\in C^m(X)$, and since
\[
    k_\mu(x,y)
    =
    a_\mu(x)a_\mu(y)k_c(x,y),
\]
it follows that $k_\mu\in C^{(m,m)}(X\times X)$. Applying
\citet[Lemma~B.1]{lavenant2024riemannian} once more yields
\[
    \|f\|_{C^m(X)}
    \leq
    \|f\|_{\mathcal H_\mu}
    \|k_\mu\|_{C^{(m,m)}(X\times X)}^{1/2}
\]
for every $f\in\mathcal H_\mu$. Again, the canonical inclusion is
injective because $\mathcal H_\mu$ is an RKHS of functions on $X$.
Therefore $\mathcal H_\mu\hookrightarrow C^m(X)$ continuously. 
\qed{}


\end{appendices}

{\bibliography{main.bib}}

\end{document}